\documentclass[10pt,twocolumn,twoside]{IEEEtran}
\IEEEoverridecommandlockouts

\pdfoutput=1

\usepackage{epsfig}
\usepackage{graphicx}
\usepackage{amsmath}
\usepackage{amssymb}
\usepackage{url}
\usepackage{array}
\graphicspath{{images/}}
\usepackage{setspace}
\usepackage{booktabs}
\usepackage{multirow}
\usepackage{ifpdf}
\usepackage{grffile}

\def\etal{\emph{et al.}}
\def\trainset{\textit{S}}
\def\image{\mathbf{I}}
\def\groundtruth{\mathbf{G}}
\def\side{\textrm{side}}
\def\allweights{\mathbf{W}}
\def\weights{\mathbf{w}}
\def\Pr{\textrm{Pr}}
\def\pred{\textrm{pred}}
\def\over3{\textrm{over3}}
\def\all{\textrm{all}}

\makeatletter
\def\hlinew#1{%
	\noalign{\ifnum0=`}\fi\hrule \@height #1 \futurelet
	\reserved@a\@xhline}
\makeatother%

\begin{document}

\title{Learning to Refine Object Contours with a Top-Down Fully Convolutional
Encoder-Decoder Network}

\author{Yahui~Liu,
	~Jian Yao$^{\dagger}$,
	~Li~Li, ~Xiaohu Lu and~Jing~Han
    \IEEEcompsocitemizethanks{\IEEEcompsocthanksitem All authors are with Computer Vision and Remote Sensing (CVRS) Lab., School of Remote Sensing and Information Engineering, Wuhan University, Wuhan, Hubei, China.
      $^{\dagger} $ E-mail: {jian.yao@whu.edu.cn} (Jian Yao), 	Web: {http://cvrs.whu.edu.cn/}.}
}

\markboth{
	}%
{Liu \MakeLowercase{\textit{et al.}}: Learning to Refine Object Contours with a Top-Down Fully Convolutional Encoder-Decoder Network}

\maketitle

\begin{abstract}
   We develop a novel deep contour detection algorithm with a top-down fully convolutional encoder-decoder network. Our proposed method, named TD-CEDN, solves two important issues in this low-level vision problem: (1) learning multi-scale and multi-level features; and (2) applying an effective top-down refined approach in the networks. TD-CEDN performs the pixel-wise prediction by means of leveraging features at all layers of the net. Unlike skip connections and previous encoder-decoder methods, we first learn a coarse feature map after the encoder stage in a feedforward pass, and then refine this feature map in a top-down strategy during the decoder stage utilizing features at successively lower layers. Therefore, the deconvolutional process is conducted stepwise, which is guided by Deeply-Supervision Net providing the integrated direct supervision. The above proposed technologies lead to a more precise and clearer prediction. Our proposed algorithm achieved the state-of-the-art on the BSDS500 dataset (ODS F-score of 0.788), the PASCAL VOC2012 dataset (ODS F-score of 0.588), and and the NYU Depth dataset (ODS F-score of 0.735).
\end{abstract}


\IEEEpeerreviewmaketitle

\section{Introduction}
\label{Sec:Introduction}

\IEEEPARstart{O}{bject} contour detection is a classical and fundamental task in computer vision, which is of great significance to numerous computer vision applications, including segmentation~\cite{arbelaez2011contour, arbelaez2014multiscale}, object proposals~\cite{uijlings2013selective, zitnick2014edge}, object detection/recognition~\cite{ferrari2008groups, xiaofeng2012discriminatively}, optical flow~\cite{revaud2015epicflow}, and  occlusion and depth reasoning~\cite{amer2015monocular, hoiem2007recovering}. In general, contour detectors offer no guarantee that they will generate closed contours and hence don't necessarily provide a partition of the image into regions~\cite{arbelaez2011contour}. Since visually salient edges correspond to  variety of visual patterns, designing a universal approach to solve such tasks is difficult~\cite{dollar2015fast}. It makes sense that precisely extracting edges/contours from natural images involves visual perception of various ``levels"~\cite{hubel1962receptive, marr1980theory}, which makes it to be a challenging problem.

\begin{figure}[th]
	\small
	\centering
	\renewcommand{\tabcolsep}{1pt}
	\begin{tabular}{ccc}
		\includegraphics[width=0.32\linewidth]{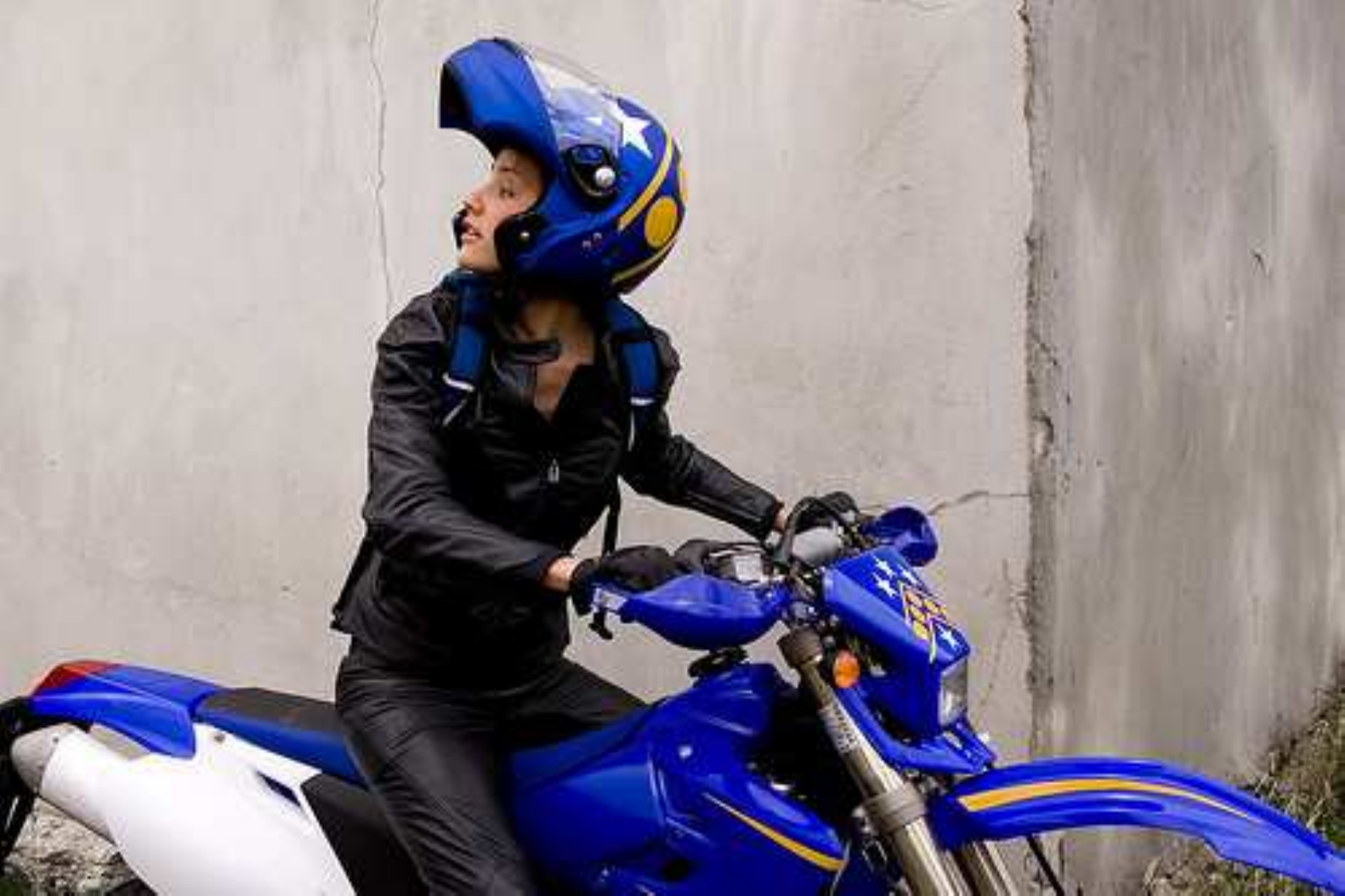} &
		\includegraphics[width=0.32\linewidth]{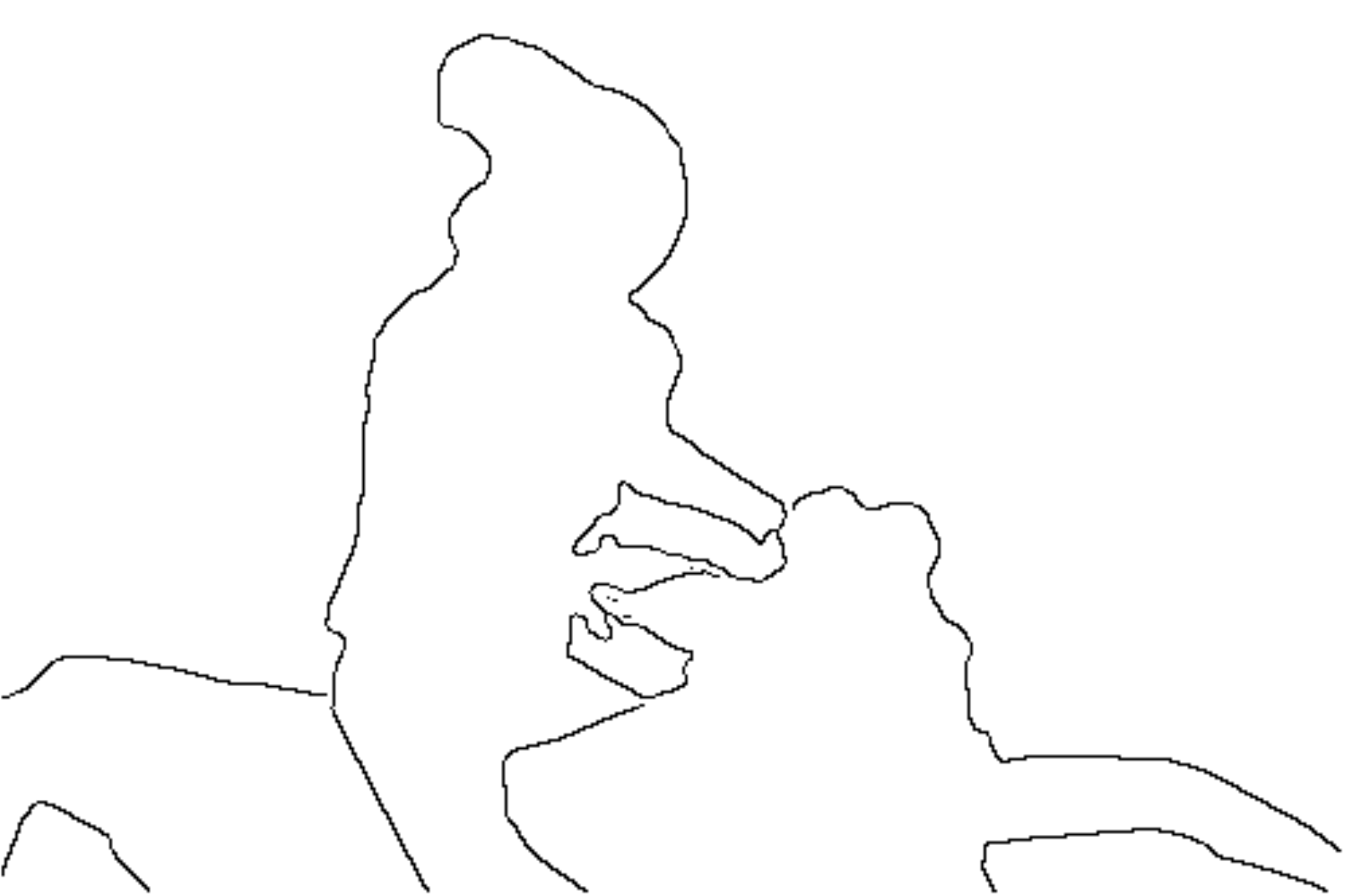} &
		\includegraphics[width=0.32\linewidth]{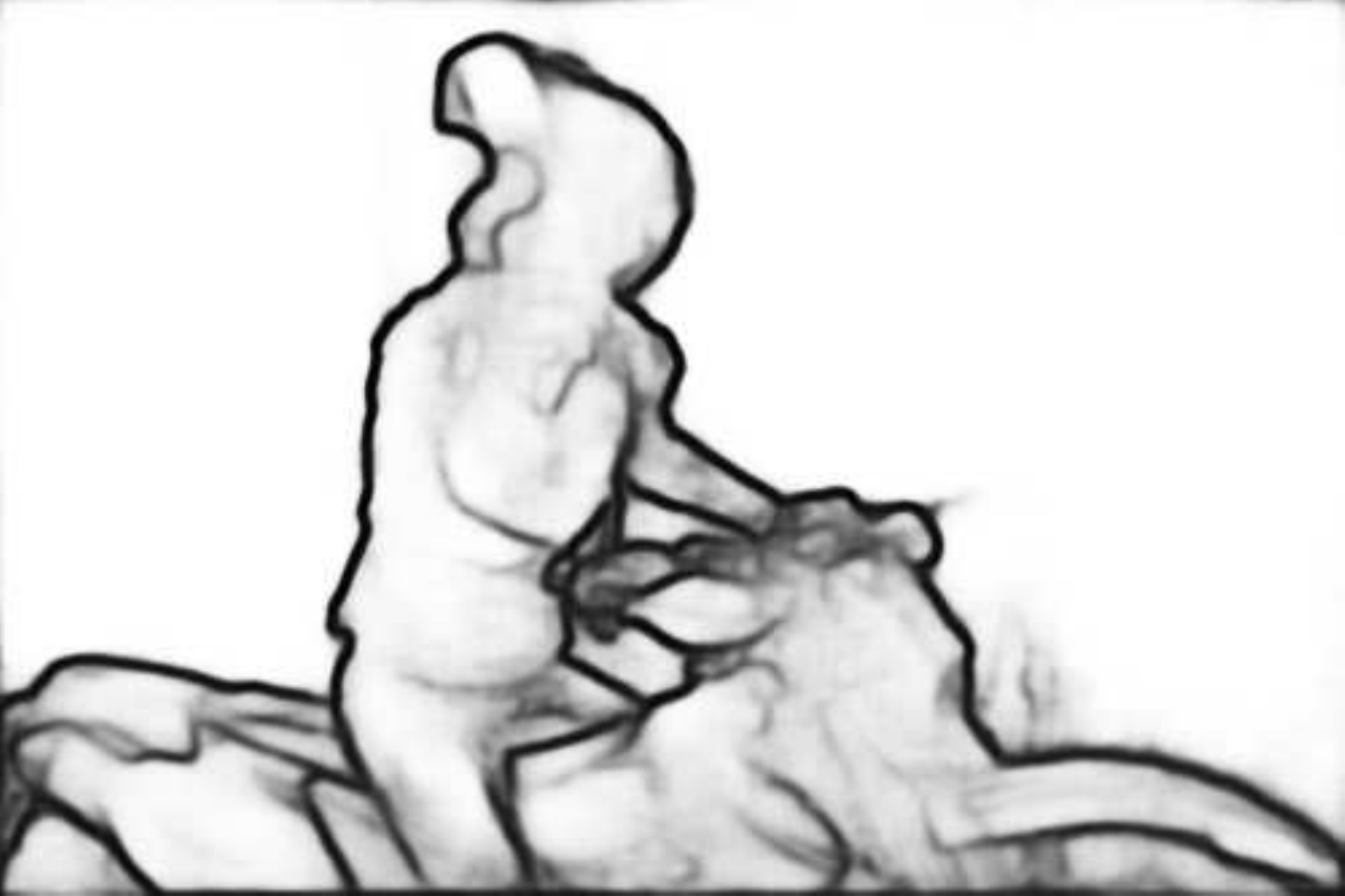} \\
		Raw image & Ground truth & CEDN \\
		\includegraphics[width=0.32\linewidth]{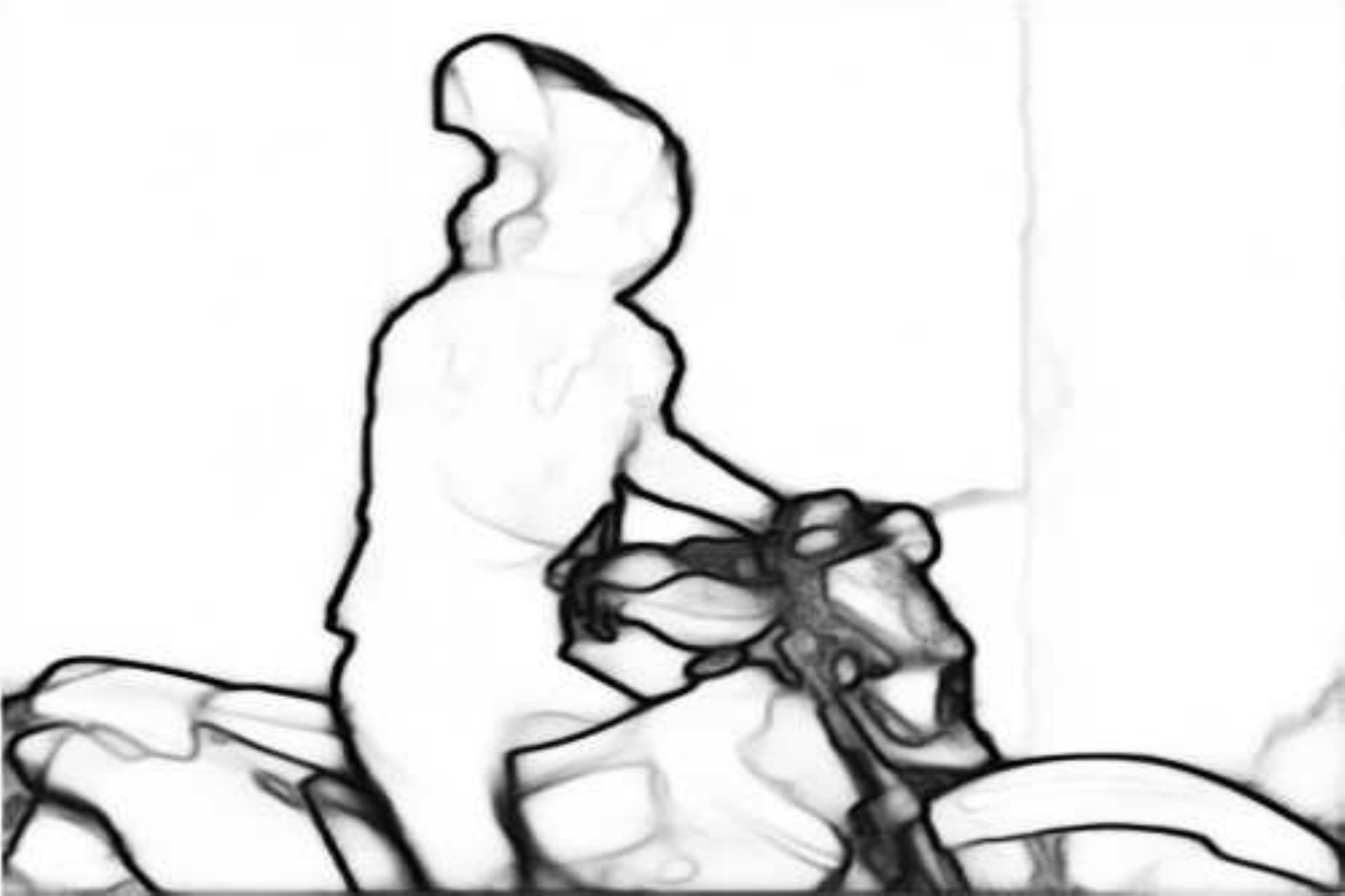} &
		\includegraphics[width=0.32\linewidth]{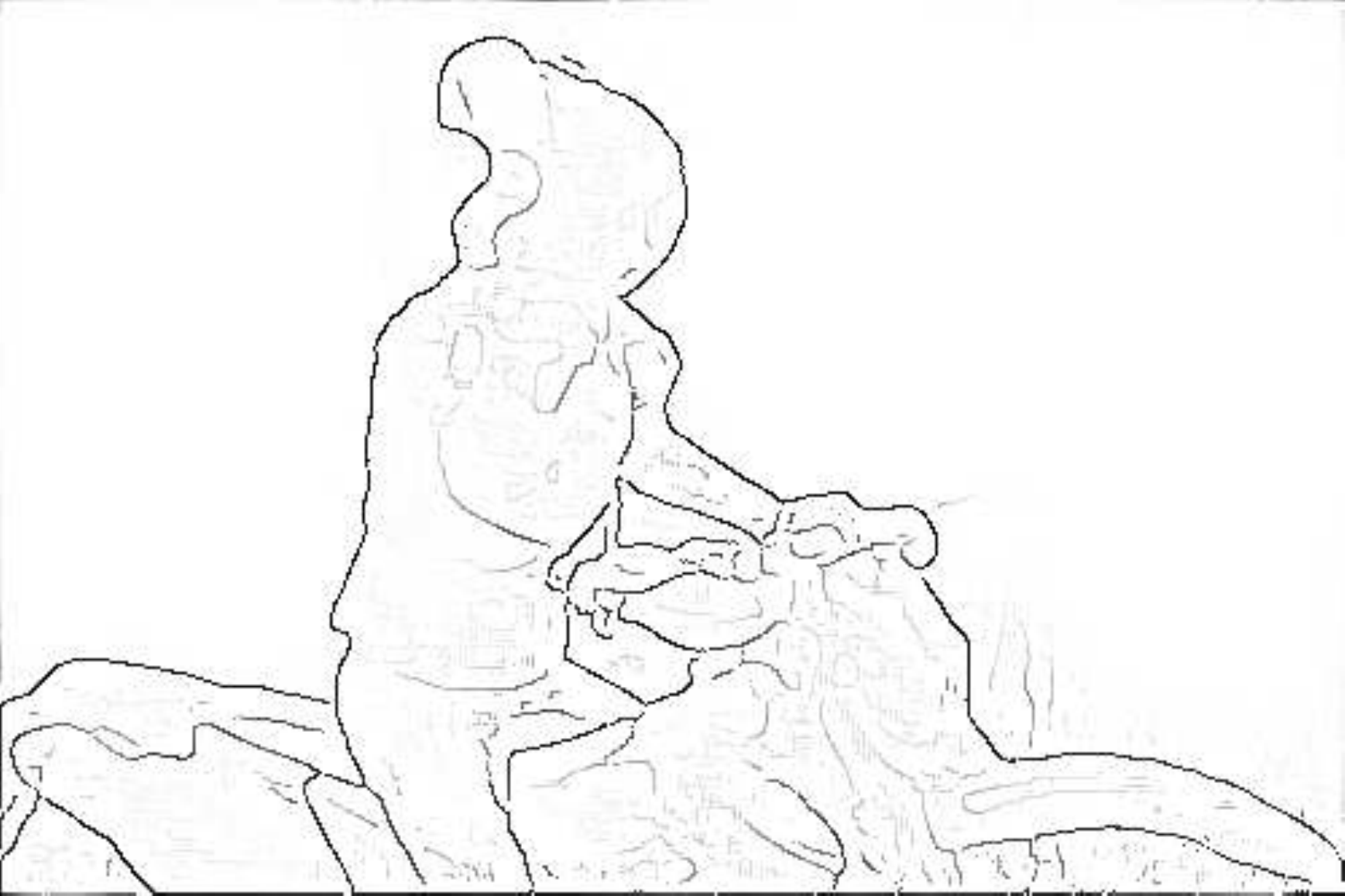} &
		\includegraphics[width=0.32\linewidth]{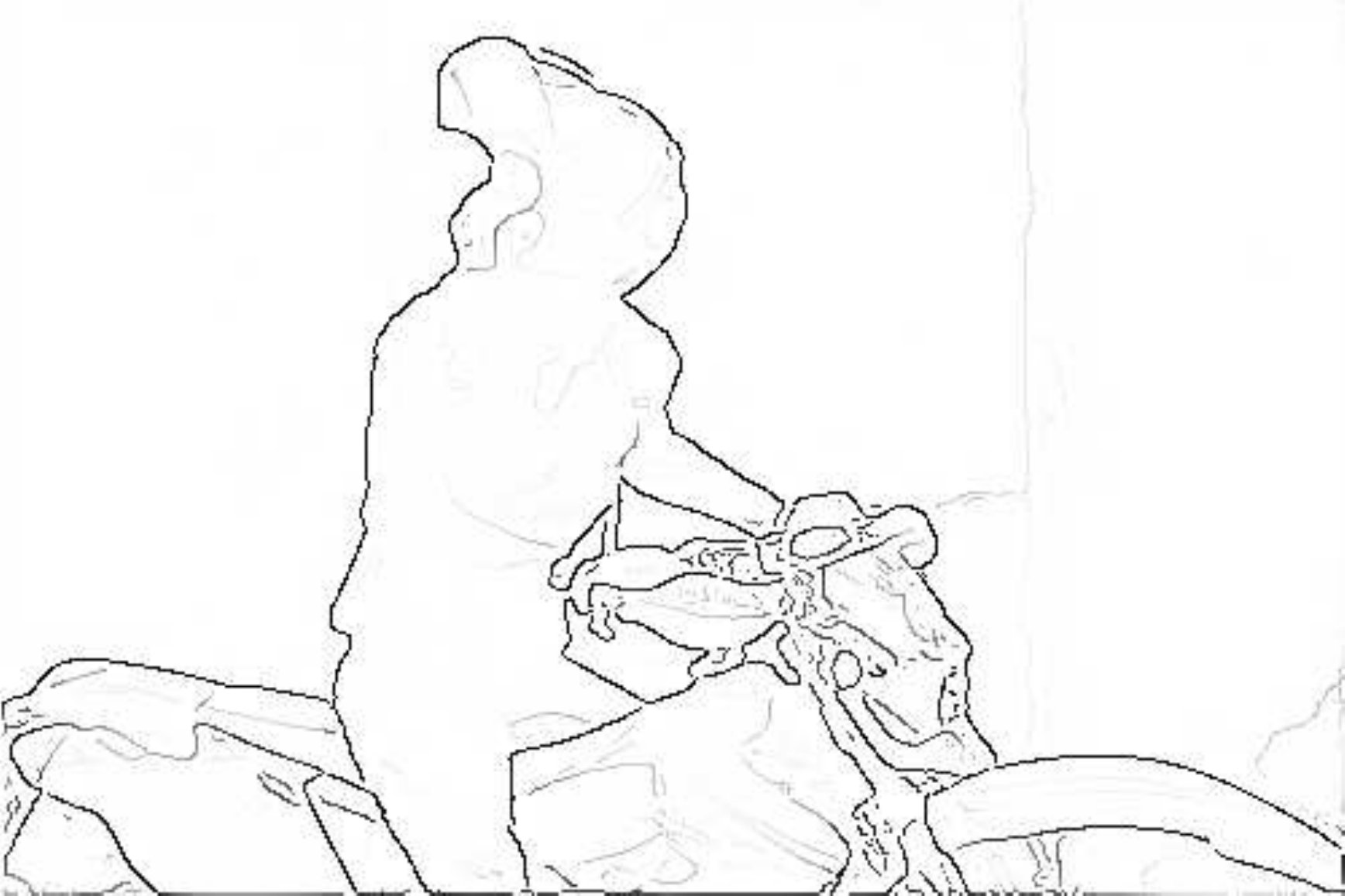} \\
		TD-CEDN (ours) & thinned CEDN & thinned TD-CEDN \\
		& & (ours)\\
	\end{tabular}
	\caption{An illustration of object contour detection. For each input raw image, our model not only can effectively learn to detect contours of foreground objects but also predict results with less noises in the local variational patches of the raw image, compared with the latest CEDN method~\cite{yang2016object}. The thinned contours are obtained by applying a standard non-maximal suppression technique to the probability map of contour.}
	\label{Fig:Demo}
\end{figure}

Edge detection has experienced an extremely rich history. A variety of approaches have been developed in the past decades. Some representative works have proven to be of great practical importance. Especially, the establishment of a few standard benchmarks, BSDS500~\cite{martin2001database}, NYUDv2~\cite{silberman2012indoor} and PASCAL VOC~\cite{everingham2010pascal}, provides a critical baseline to evaluate the performance of each algorithm. The state-of-the-art edge/contour detectors~\cite{arbelaez2011contour, gupta2013perceptual, lim2013sketch, xie2015holistically} explore multiple features as input, including brightness, color, texture, local variance and depth computed over multiple scales. It indicates that multi-scale and multi-level features improve the capacities of the detectors. 
Recently, the supervised deep learning methods, such as deep Convolutional Neural Networks (CNNs), have achieved the state-of-the-art performances in such field, including $N^{4}$-Fields~\cite{ganin2014n}, DeepContour~\cite{shen2015deepcontour}, DeepEdge~\cite{bertasius2015deepedge}, HED~\cite{xie2015holistically}, and CEDN~\cite{yang2016object}. The CNNs-based methods are powerful visual models that yield hierarchical features, which can provide an ideal method to aggregate multiple ``levels". 

\begin{figure*}
	\centering
	\includegraphics[width=\linewidth]{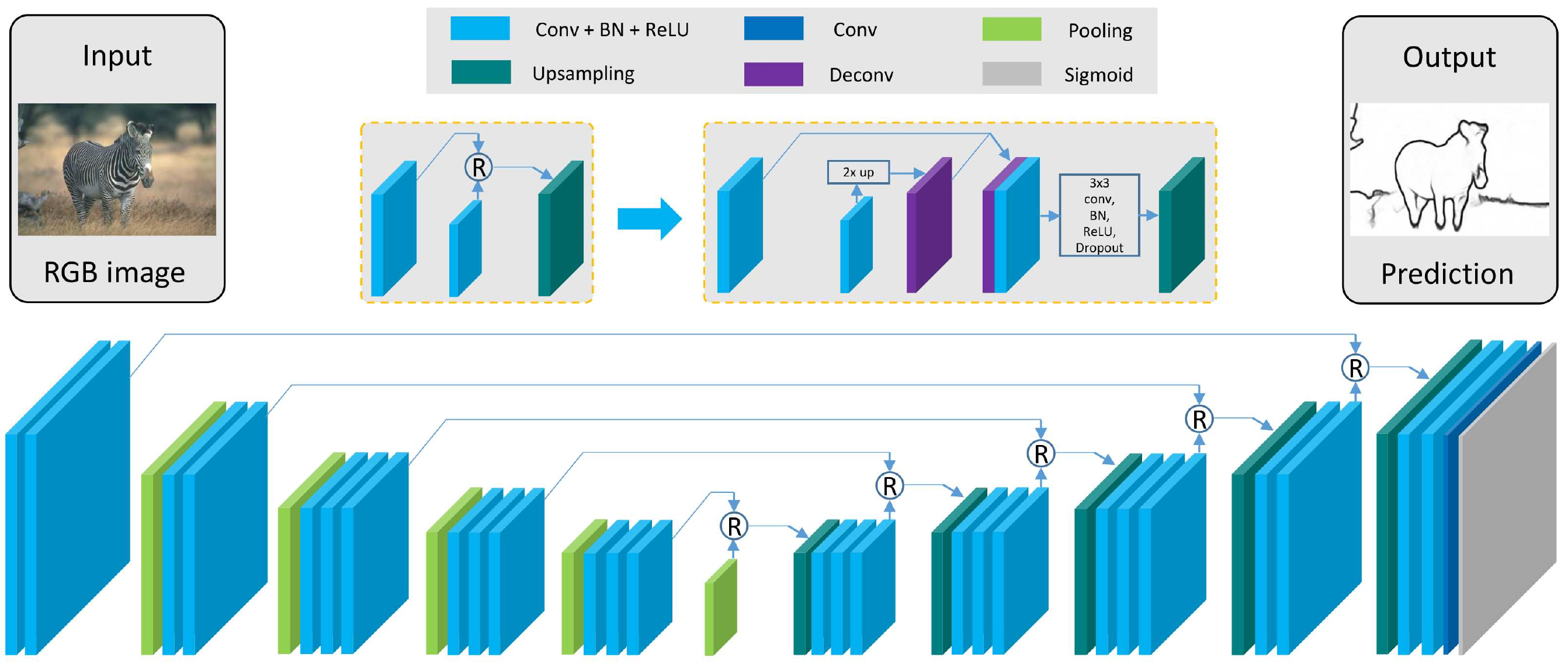}\\
	\centering
	\caption{An illustration of our proposed TD-CEDN architecture. In this architecture, there are no fully connected layers, the side-output layers are inserted after the convolutional layers, the deep supervision is applied at each side-output layer and then all of them are concatenated to form a fused final output. In this way, the final output layer learns multi-scale and multi-level features as the plane size of the input of side-output layers becomes smaller and the receptive field size becomes larger.}
	\label{Fig:Architecture}
\end{figure*}

In this paper, we develop a pixel-wise and end-to-end contour detection system, Top-Down Convolutional Encoder-Decoder Network (TD-CEDN), which is inspired by the success of Fully Convolutional Networks (FCN)~\cite{long2015fully}, HED,  Encoder-Decoder networks~\cite{noh2015learning, badrinarayanan2015segnet, yang2016object} and the bottom-up/top-down architecture~\cite{pinheiro2016learning}. Being fully convolutional, the developed TD-CEDN can operate on an arbitrary image size and the encoder-decoder network emphasizes its symmetric structure which is similar to the SegNet~\cite{badrinarayanan2015segnet} and DeconvNet~\cite{noh2015learning} but not the same, as shown in Fig.~\ref{Fig:Architecture}. In TD-CEDN, we initialize our encoder stage with VGG-16 net~\cite{simonyan2014very} (up to the ``pool5" layer) and apply Bath Normalization (BN)~\cite{ioffe2015batch} to reduce the internal covariate shift between each convolutional layer and the Rectified Linear Unit (ReLU)~\cite{nair2010rectified} layer. Our decoder network can be regarded as a mirrored version of the encoder network, and has multiple series of upsampling, convolutional, BN and ReLU layers. The upsampling with a refined module differs from previous unpooling/deconvolution~\cite{noh2015learning} and max-pooling indices~\cite{badrinarayanan2015segnet} technologies. We first concatenate the output of the deconvolutional layer and features at the successively lower layer of the encoder stage. Then a convolutional layer is trained to learn to generate refined results with the concatenated feature map. With such upsampling strategy, the proposed network learns and optimizes the meaningful features for contour detection from the high level to the low level visual perception in a top-down manner. To learn multi-scale and multi-level features efficiently, the Deeply-Supervision Net (DSN)~\cite{lee2015deeply} is applied to provide the integrated direct supervision by supervising each output of upsampling. Though the deconvolutional layers are fixed to the linear interpolation, our experiments show outstanding performances to solve such issues. 
Fig.~\ref{Fig:Demo} presents several samples of object contour detection, in which our method obtains more precise and clear predictions.

The remainder of this paper is organized as follows. In Section~\ref{Sec:RelatedWork}, we review related work on the pixel-wise semantic prediction networks. The main idea and details of the proposed network are explained in Section~\ref{Sec:td-cedn-netwrok}.  A quantitative comparison of our method to the two state-of-the-art contour detection methods is presented in Section~\ref{Sec:Results} followed by the conclusion drawn in Section~\ref{Sec:Conclusion}.

\section{Related Work}
\label{Sec:RelatedWork}

Semantic pixel-wise prediction is an active research task, which is fueled by the open datasets~\cite{martin2001database, everingham2010pascal, silberman2012indoor}. With the development of deep networks, the best performances of contour detection have been continuously improved. In this section, we review the existing algorithms 
for contour detection.

Early approaches to contour detection~\cite{canny1986computational, morrone1987feature, freeman1991design, lindeberg1998edge} aim at quantifying the presence of boundaries through local measurements, which is the key stage of designing detectors.  The Canny detector~\cite{canny1986computational}, which is perhaps the most widely used method up to now, models edges as a sharp discontinuities in the local gradient space, adding non-maximum suppression and hysteresis thresholding steps. The oriented energy methods~\cite{morrone1987feature, freeman1991design} tried to obtain a richer description via using a family of quadrature pairs of even and odd symmetric filters. Lindeberg~\cite{lindeberg1998edge}, who realized that the extracted descriptors of an image may be strongly dependent on the scales which the operators are applied, proposed a filter-based method with an automatic scale selection algorithm. 

The local approaches took into account more feature spaces, such as color and texture, and applied learning methods for cue combination~\cite{martin2002learning, martin2004learning, dollar2006supervised, mairal2008discriminative, xiaofeng2012discriminatively, arbelaez2011contour, arbelaez2014multiscale}. The \textit{Pb} work of Martin \etal~\cite{martin2002learning, martin2004learning} formulated features that responded to gradients in brightness, color and texture, and made use of them as input of a logistic regression classifier to predict the probability of boundaries. Ren \etal~\cite{ren2008multi} combined features extracted from multi-scale local operators based on the \textit{Pb} work~\cite{martin2004learning}, which provided additional localization and relative contrast cues for the boundary classifier. Dollar \etal~\cite{dollar2006supervised} proposed a Boosted Edge Learning (BEL) algorithm which attempted to learn an edge classifier in the form of a probabilistic boosting tree from a rich set of simple local features. This algorithm was extended by Zheng \etal~\cite{zheng2010detecting} who used contextual and shape features to refine the edge maps.  Mairal \etal~\cite{mairal2008discriminative} and Ren \etal~\cite{xiaofeng2012discriminatively} used discriminative sparse models to represent local image patches. A major difference of their works was the use of oriented gradients: the former used K-SVD to represent multi-scale patches and classified a patch directly while the latter used Sparse Code Gradients (SCG) measuring contrast between oriented half-discs. Arbel{\'a}ez \etal~\cite{arbelaez2011contour} combined multiple local cues into a globalization framework based on spectral clustering for contour detection, called \textit{gPb}. Furthermore, Arbel{\'a}ez \etal~\cite{arbelaez2014multiscale} developed a normalized cuts algorithm, which provided a faster speed to the eigenvector computation required for contour globalization~\cite{arbelaez2011contour, xiaofeng2012discriminatively}.

Some researches focused on the mid-level structures of local patches, such as straight lines, parallel lines, T-junctions, Y-junctions and so on~\cite{wu2007compositional, kontschieder2011structured, lim2013sketch, dollar2015fast}, which are termed as \textit{structure learning}~\cite{nowozin2011structured}. Wu \etal~\cite{wu2007compositional} presented a compositional boosting method to detect 17 unique local edge structures. Kontschieder \etal~\cite{kontschieder2011structured} incorporated structural information in the random forests~\cite{breiman2001random} for the task of semantic labeling for image patches. Lim \etal~\cite{lim2013sketch} typically utilized a few hundred tokens, which captured a majority of common edge structures, to learn to classify local image patches into \textit{sketch tokens}. The Structured Forests method~\cite{dollar2015fast} took advantage of a large number of manually designed multi-scale features in a structured learning framework applied to random decision forests.

Some other methods~\cite{ren2005scale, felzenszwalb2006min, zhu2007untangling} tried to solve this issue with different strategies. Ren \etal~\cite{ren2005scale} presented a model of curvilinear grouping taking advantage of piecewise linear representation of contours and a conditional random field to capture continuity and the frequency of different junction types. They computed a constrained Delaunay triangulation (CDT), which was scale-invariant and tended to fill gaps in the detected contours, over the set of found local contours. Felzenszwalb \etal~\cite{felzenszwalb2006min} generated a global interpretation of an image in term of a small set of salient smooth curves. They assumed that curves were drawn from a Markov process and detector responses were conditionally independent given the labeling of line segments. The curve finding algorithm searched for optimal curves by starting from short curves and iteratively expanding ones, which was translated into a general weighted min-cover problem. Zhu \etal~\cite{zhu2007untangling} proposed to first threshold the output of ~\cite{martin2004learning} and then create a weighted edgels graph, where the weights measured directed collinearity between neighboring edgels.

Recently, deep learning methods have achieved great successes for various applications in computer vision, including contour detection~\cite{ganin2014n, kivinen2014visual, shen2015deepcontour, bertasius2015deepedge, xie2015holistically, yang2016object}. Ganin \etal~\cite{ganin2014n} proposed a $N^{4}$-Fields method to process an image in a patch-by-patch manner. An input patch was first passed through a pretrained CNN and then the output features were mapped to an annotation edge map using the nearest-neighbor search. Kivinen \etal~\cite{kivinen2014visual} used a traditional CNN architecture, which applied multiple streams to integrate multi-scale and multi-level features, to achieve contour detection. Shen \etal~\cite{shen2015deepcontour} developed a method, called DeepContour, 
in which a contour patch was an input of a CNN model and the output was treated as a compact cluster which was assigned by a shape label. The final contours were fitted with the various shapes by different model parameters by a divide-and-conquer strategy. Bertasius \etal~\cite{bertasius2015deepedge} designed a multi-scale deep network which consists of five convolutional layers and a bifurcated fully-connected sub-networks. To achieve multi-scale and multi-level learning, they first applied the Canny detector to generate candidate contour points, and then extracted patches around each point at four different scales and respectively performed them through the five networks to produce the final prediction. The above mentioned four methods~\cite{ganin2014n, kivinen2014visual, shen2015deepcontour, bertasius2015deepedge} are all patch-based but not end-to-end training and holistic image prediction networks. Xie \etal~\cite{xie2015holistically} and Yang \etal~\cite{yang2016object} developed two end-to-end and pixel-wise prediction fully convolutional networks. In the work of Xie \etal~\cite{xie2015holistically}, a number of properties, which are key and likely to play a role in a successful system in such field, are summarized: (1) carefully designed detector and/or learned features~\cite{martin2004learning, dollar2006supervised}, (2) multi-scale response fusion~\cite{ren2008multi, arbelaez2014multiscale}, (3) engagement of multiple levels of visual perception~\cite{hubel1962receptive, marr1980theory, hou2013boundary}, (4) structural information~\cite{lim2013sketch, dollar2015fast}, etc. Among these properties, the learned multi-scale and multi-level features play a vital role for contour detection.
To achieve this goal, deep architectures have developed three main strategies: (1) inputing images at several scales into one or multiple streams~\cite{kivinen2014visual, bertasius2015deepedge, eigen2015predicting}; (2) combining feature maps from different layers of a deep architecture~\cite{xie2015holistically, hariharan2015hypercolumns, liu2015parsenet}; (3) improving the decoder/deconvolution networks~\cite{yang2016object, badrinarayanan2015segnet, noh2015learning}. HED~\cite{xie2015holistically} and CEDN~\cite{yang2016object}, which achieved the state-of-the-art performances, are representative works of the above-mentioned second and third strategies. HED integrated FCN~\cite{long2015fully} and DSN~\cite{lee2015deeply} to learn meaningful features from multiple level layers in a single trimmed VGG-16 net. Their integrated learning of hierarchical features was in distinction to previous multi-scale approaches. CEDN focused on applying a more complicated deconvolution network, which was inspired by DeconvNet~\cite{noh2015learning} and was composed of deconvolution, unpooling and ReLU layers, to improve upsampling results. Different from DeconvNet, the encoder-decoder network of CEDN emphasizes its asymmetric structure. Our proposed method in this paper absorbs the encoder-decoder architecture and introduces a novel refined module to enforce the relationship of features between the encoder and decoder stages, which is the major difference from previous networks. To guide the learning of more transparent features, the DSN strategy is also reserved in the training stage.

\section{The TD-CEDN Network}
\label{Sec:td-cedn-netwrok}

In this section, we describe our contour detection method with the proposed top-down fully convolutional encoder-decoder network.

\subsection{Architecture}
\label{Sec:Architecture}

Similar to CEDN~\cite{yang2016object}, we formulate contour detection as a binary image labeling problem where ``0" and ``1" refer to ``non-contour" and ``contour", respectively. Fig.~\ref{Fig:Architecture} illustrates the entire architecture of our proposed network for contour detection. The encoder-decoder network is composed of two parts: encoder/convolution and decoder/deconvolution networks. We use the layers up to ``pool5" from the VGG-16 net~\cite{simonyan2014very} as the encoder network. Different from the original network, we apply the BN~\cite{ioffe2015batch} layer to reduce the internal covariate shift between each convolutional layer and the  ReLU~\cite{nair2010rectified} layer. With such adjustment, we can still initialize the training process from weights trained for classification on the large dataset~\cite{russakovsky2015imagenet}. Owing to discarding the fully connected layers after ``pool5", higher resolution feature maps are retained while reducing the parameters of the encoder network significantly (from 134M to 14.7M).

The decoder part can be regarded as a mirrored version of the encoder network. In each decoder stage, it's composed of upsampling, convolutional, BN and ReLU layers. The upsampling process is conducted stepwise with a refined module which differs from previous unpooling/deconvolution~\cite{noh2015learning} and max-pooling indices~\cite{badrinarayanan2015segnet} technologies, which will be described in details in Section~\ref{Sec:RefinedModule}. The complete configurations of our network are outlined in Table~\ref{tab:ConvNetConfig}. The final high dimensional features of the output of the decoder are fed to a trainable convolutional layer with a kernel size of 1 and an output channel of 1, and then the reduced feature map is applied to a sigmoid layer to generate a soft prediction. Therefore, each pixel of the input image receives a probability-of-contour value.

\begin{table}[tb]
	\begin{center}
		\caption{The configurations of our proposed TD-CEDN network for contour detection.}
		\renewcommand{\tabcolsep}{3pt}
		\renewcommand{\arraystretch}{1.1}
		\begin{tabular}{c|c}\hline
			 \textbf{Encoder} &  \textbf{Decoder}  \\
			\hlinew{1pt} 
			conv1\_1-3-64, BN, ReLU  & upsample5  \\
			conv1\_2-3-64, BN, ReLU  & deconv5\_3-3-512, BN, ReLU  \\
			pool1 & deconv5\_2-3-512, BN, ReLU  \\
			conv2\_1-3-128, BN, ReLU & deconv5\_1-3-512, BN, ReLU\\
			conv2\_2-3-128, BN, ReLU & upsample4 \\
			pool2 & deconv4\_3-3-512, BN, ReLU\\
			conv3\_1-3-256, BN, ReLU & deconv4\_2-3-512, BN, ReLU \\
			conv3\_2-3-256, BN, ReLU & deconv4\_1-3-512, BN, ReLU \\
			conv3\_3-3-256, BN, ReLU & upsample3 \\
			pool3 &  deconv3\_3-3-256, BN, ReLU\\
			conv4\_1-3-512, BN, ReLU & deconv3\_2-3-256, BN, ReLU \\
			conv4\_2-3-512, BN, ReLU & deconv3\_1-3-256, BN, ReLU \\
			conv4\_3-3-512, BN, ReLU & upsample2 \\
			pool4 & deconv2\_2-3-128, BN, ReLU \\
			conv5\_1-3-512, BN, ReLU & deconv2\_1-3-128, BN, ReLU\\
			conv5\_2-3-512, BN, ReLU & upsample1 \\
			conv5\_3-3-512, BN, ReLU & deconv1\_2-3-64, BN, ReLU\\
			pool5 & deconv1\_1-3-64, BN, ReLU\\
			& pred-1-1, Sigmoid\\ \hline
		\end{tabular}
		\label{tab:ConvNetConfig}
	\end{center}
	\begin{spacing}{0.85}
	{\footnotesize
		\textbf{Note:} In the encoder part, all of the pooling layers are max-pooling with a 2$\times$2 window and a stride 2 (non-overlapping window). The convolutional layer parameters are denoted as ``conv/deconv$\langle$stage\_index$\rangle$-$\langle$receptive field size$\rangle$-$\langle$number of channels$\rangle$". BN and ReLU represent the batch normalization and the activation function, respectively.
	}
	\end{spacing}
\end{table}

\subsection{Refined Module}
\label{Sec:RefinedModule}

\begin{figure}[tbh]
	\small
	\centering
	\renewcommand{\tabcolsep}{0pt}
	\begin{tabular}{c}
		\includegraphics[width=\linewidth]{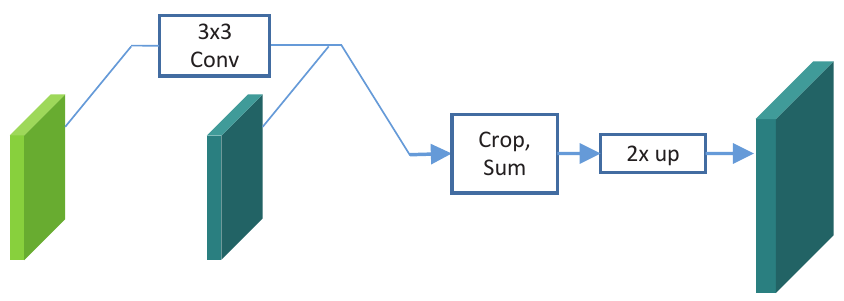} \\
		(a) The refined module of FCN \\
		\includegraphics[width=\linewidth]{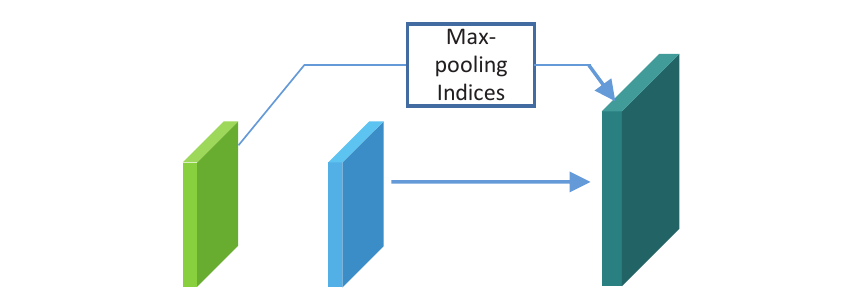} \\
		(b) The refined module of SegNet \\
		\includegraphics[width=\linewidth]{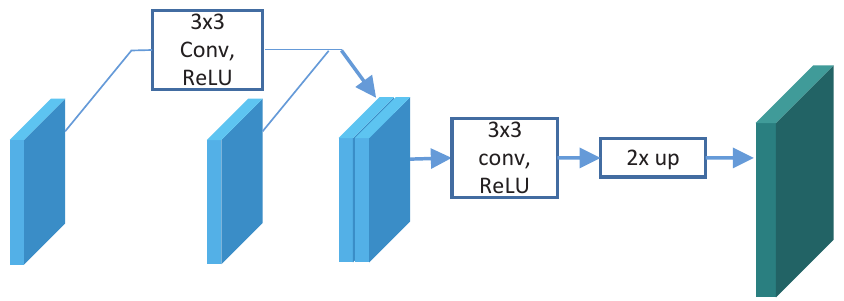} \\
		(c) The refined module of SharpMask \\
		\includegraphics[width=\linewidth]{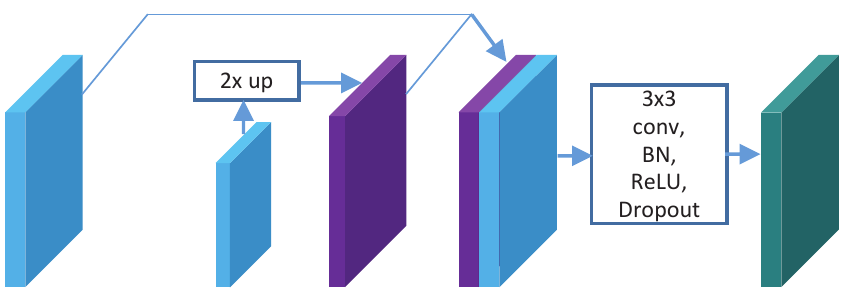} \\
		(d) The used refined module for our proposed TD-CEDN\\
	\end{tabular}
	\caption{An illustration of differences of the refined modules for FCN, SegNet, SharpMask and our proposed TD-CEDN.}
	\label{Fig:RefinedModule}
\end{figure}

Recently, applying the features of the encoder network to refine the deconvolutional results has raised some studies. FCN~\cite{long2015fully} combined the lower pooling layer with the current upsampling layer following by summing the cropped results and the output feature map was upsampled.  SegNet~\cite{badrinarayanan2015segnet} used the max pooling indices to upsample (without learning) the feature maps and convolved with a trainable decoder network. SharpMask~\cite{pinheiro2016learning} concatenated the current feature map of the decoder network with the output of the convolutional layer in the encoder network, which had the same plane size. Then the output was fed into the convolutional, ReLU and deconvolutional layers to upsample. Our refined module differs from the above mentioned methods. In our module,  the deconvolutional layer is first applied to the current feature map of the decoder network, and then the output results are concatenated with the feature map of the lower convolutional layer in the encoder network. The final upsampling results are obtained through the convolutional, BN, ReLU and dropout~\cite{srivastava2014dropout} layers.

Fig.~\ref{Fig:RefinedModule} shows the refined modules of FCN~\cite{long2015fully}, SegNet~\cite{badrinarayanan2015segnet}, SharpMask~\cite{pinheiro2016learning} and our proposed TD-CEDN. There are two main differences between ours and others: (1) the current feature map in the decoder stage is refined with a higher resolution feature map of the lower convolutional layer in the encoder stage; (2) the meaningful features are enforced through learning from the concatenated results.


\subsection{Formulation}
\label{Sec:Formulation}

The most of the notations and formulations of the proposed method follow those of HED~\cite{xie2015holistically}. The training set is denoted by $\trainset = \{ (\image_i, \groundtruth_i)\}_{i=1}^N$, where the image sample $\image_i$ refers to the $i$-th raw input image and $\groundtruth_i$ refers to the corresponding ground truth edge map of $\image_i$. For simplicity, we consider each image independently and the index $i$ will be omitted hereafter. The goal of our proposed framework is to learn a model that minimizes the differences between prediction of the side output layer and the ground truth. Each side-output layer is regarded as a pixel-wise classifier with the corresponding weights $\weights$. Note that there are $M$ side-output layers, in which DSN~\cite{lee2015deeply} is applied to provide supervision for learning meaningful features. Therefore, the weights are denoted as $\weights=\{(\weights^{(1)}, \ldots, \weights^{(M)})\}$. The objective function is defined as the following loss:
\begin{equation}\label{Eq:side-loss}
\begin{aligned}
	\mathcal{L}_{\side}(\image, \groundtruth, \allweights, \weights)  & =  \sum_{m=1}^{M}\alpha_{m} \ell_{\side}(\image, \groundtruth, \allweights, \weights^{(m)}) \\
	& = \sum_{m=1}^{M}{\alpha_{m} \Delta(\hat{\groundtruth}^{(m)}, \groundtruth, \allweights, \weights^{(m)})} ,
\end{aligned}
\end{equation}
where $\allweights$ denotes the collection of all standard network layer parameters, $\ell_{\side}$ refers to the image-level loss function for the side-output, $\hat{\groundtruth}^{(m)}$ is the prediction results of the $m$-th side-output layer, which is upsampled to the raw image size if necessary, $\Delta$ is a cross-entropy classification loss function, and $\alpha_{m}$ is a hyper-parameter referring to the loss weight for each side-output layer.

Given that over 90\% of the ground truth is non-contour. A cost-sensitive loss function, which balances the loss between contour and non-contour classes and differs from the CEDN~\cite{yang2016object} fixing the balancing weight for the entire dataset, is applied. Therefore, the traditional cross-entropy loss function is redesigned as follows:
\begin{equation}\label{Eq:cross-entropy}
\begin{aligned}
\Delta = & -\beta\sum_{k=1}^{|\image|} \log\Pr(\groundtruth(k)=1|\image(k); \allweights, \weights^{(m)}) \\
 &- (1-\beta)\sum_{k=1}^{|\image|} \log\Pr(\groundtruth(k)=0|\image(k); \allweights, \weights^{(m)}),
\end{aligned}
\end{equation}
where $\beta$ refers to a class-balancing weight, and $\image(k)$ and $\groundtruth(k)$ denote the values of the $k$-th pixel in $\image$ and $\groundtruth$, respectively. For a training image $\image$, $\beta = \frac{|\image|_{-}}{|\image|}$ and $1-\beta = \frac{|\image|_{+}}{|\image|}$ where $|\image|$, $|\image|_{-}$ and $|\image|_{+}$ refer to total number of all pixels, non-contour (negative) pixels and contour (positive) pixels, respectively.

\subsection{Supervision}
\label{Sec:Supervision}

\begin{figure}
	\centering
	\includegraphics[width=\linewidth]{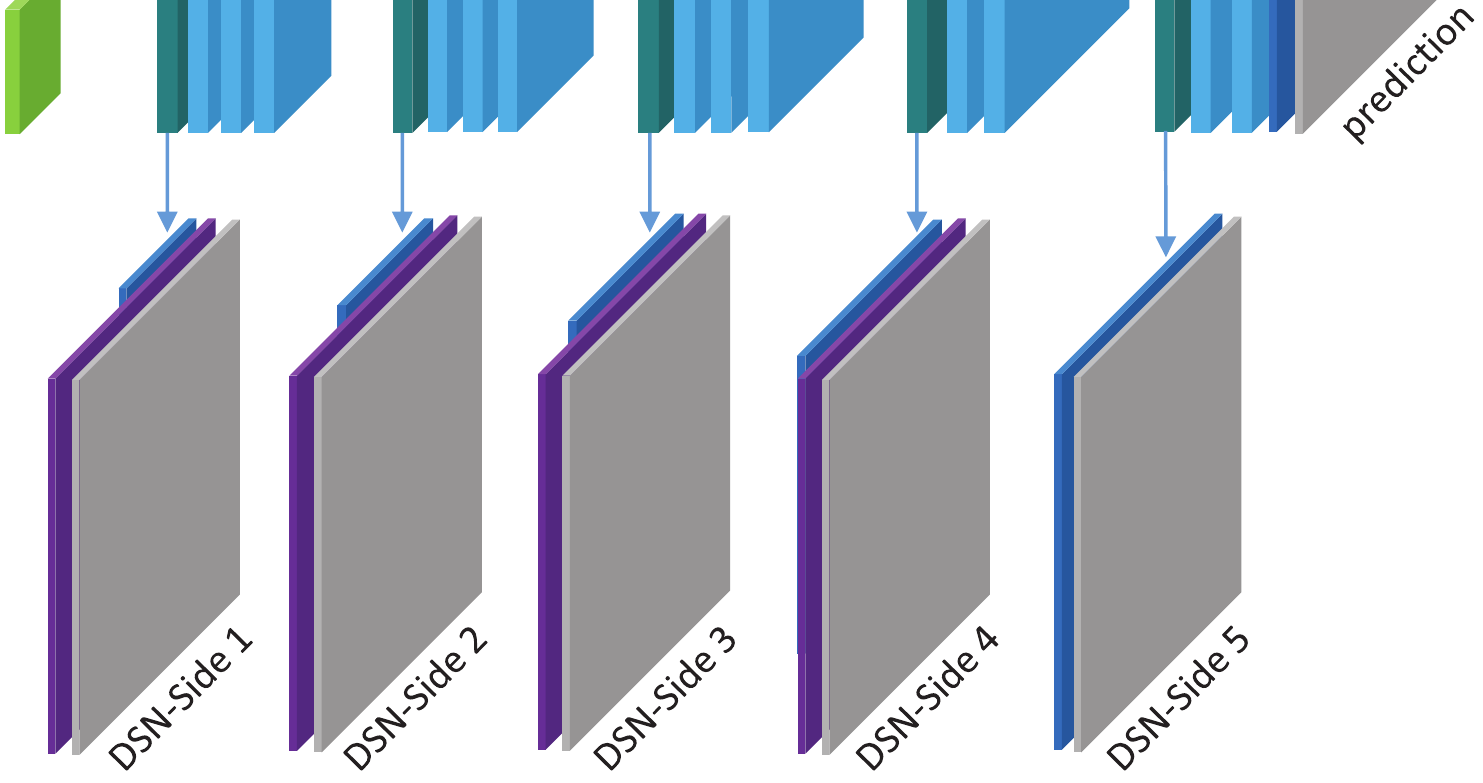}\\
	\caption{An illustration of the DSN supervision method. All the outputs of the upsampling layers, from ``upsample5" to ``upsample1", are resized to the same size as the original input image by the convolutional and deconvolutional methods. Then, the sigmoid layers are applied to calculate the cross-entropy loss. The loss of ``DSN-Side $m$" ($m=1, ..., 5$)   corresponds to $\alpha_{m}\Delta(\hat{\groundtruth}^{(m)}, \groundtruth, \allweights, \weights^{(m)})$ and the loss of ``prediction" corresponds to $\mathcal{L}_{\pred}$.}
	\label{Fig:side-ouput-loss}
\end{figure}

We use the DSN~\cite{lee2015deeply} to supervise each upsampling stage, as shown in Fig.~\ref{Fig:side-ouput-loss}. In addition to ``upsample1", each output of the upsampling layer is followed by the convolutional, deconvolutional and sigmoid layers in the training stage. Each side-output can produce a loss termed $\mathcal{L}_{\side}$. The final prediction also produces a loss term $\mathcal{L}_{\pred}$, which is similar to Eq.~(\ref{Eq:cross-entropy}):
\begin{equation}\label{Eq:pred-loss}
\begin{aligned}
\mathcal{L}_{\pred}(\image, \groundtruth, \allweights)  = & -\beta\sum_{k=1}^{|\image|} \log\Pr(\groundtruth(k)=1|\image(k); \allweights) \\
&- (1-\beta)\sum_{k=1}^{|\image|} \log\Pr(\groundtruth(k)=0|\image(k); \allweights),
\end{aligned}
\end{equation}
where $\image(k)$, $\groundtruth(k)$, $|\image|$ and $\beta$ have the same meanings with those in Eq.~(\ref{Eq:cross-entropy}).
Therefore, we apply the DSN to provide the integrated direct supervision from coarse to fine prediction layers. The overall loss function is formulated as:
\begin{equation}\label{Eq:final-loss}
\mathcal{L}(\image, \groundtruth, \allweights)  = \mathcal{L}_{\side}(\image, \groundtruth, \allweights, \weights)
 + \mathcal{L}_{\pred}(\image, \groundtruth, \allweights).
\end{equation}
In our testing stage, the DSN side-output layers will be discarded, which differs from the HED network. HED fused the output of side-output layers to obtain a final prediction, while we just output the final prediction layer. Even so, the results show a pretty good performances on several datasets, which will be presented in Section~\ref{Sec:Results}.

\subsection{Training}
\label{Sec:Training}

 We trained our network using the publicly available \textit{Caffe}~\cite{jia2014caffe} library and built it on the top of the implementations of FCN~\cite{long2015fully}, HED~\cite{xie2015holistically}, SegNet~\cite{badrinarayanan2015segnet} and CEDN~\cite{yang2016object}. The Stochastic Gradient Descent (SGD)~\cite{bottou2010large} method was used to optimize our network. The model parameters we tuned are listed as follows: the size of mini-batch (1), the learning rate (1e-6), the loss weight $\alpha_m$ for each side-output layer (1.0), the momentum (0.9), the weight decay (2e-4), the training iteration number (2e4, and the learning rate follows a polynomial decay with a power of 0.8). During the training stage, all the images were resized to 400$\times$400 for memory efficiency. In later experiments, the values of all parameters mentioned above were fixed. Traditional data augmentation methods, like rotating and flipping, were also applied to augment the training dataset. All models were trained and tested on a single NVIDIA TITAN X GPU.

\section{Results}
\label{Sec:Results}

In this section, we comprehensively evaluated our method on three popularly used contour detection datasets: BSDS500, PASCAL VOC 2012 and NYU Depth, by comparing with two state-of-the-art contour detection methods: HED~\cite{xie2015holistically} and CEDN~\cite{yang2016object}.

\noindent \textbf{BSDS500:} The majority of our experiments were performed on the BSDS500 dataset. BDSD500~\cite{martin2001database} is a standard benchmark for contour detection. It is composed of 200 training, 100 validation and 200 testing images. Each image has 4-8 hand annotated ground truth contours. Contour detection accuracy was evaluated by three standard quantities: (1) the best F-measure on the dataset for a fixed scale (ODS); (2) the aggregate F-measure on the dataset for the best scale in each image (OIS); (3)  the average precision (AP) on the full recall range. Recent works, HED~\cite{xie2015holistically} and CEDN~\cite{yang2016object}, which have achieved the best performances on the BSDS500 dataset, are two baselines which our method was compared to. Like other methods, a standard non-maximal suppression technique was applied to obtain thinned contours before evaluation.

\begin{figure}[t]
	\centering
	\renewcommand{\tabcolsep}{1pt}
	\begin{tabular}{cccc}
		Raw image & Ground truth & Raw image & Ground truth \\
		\includegraphics[width=0.24\linewidth]{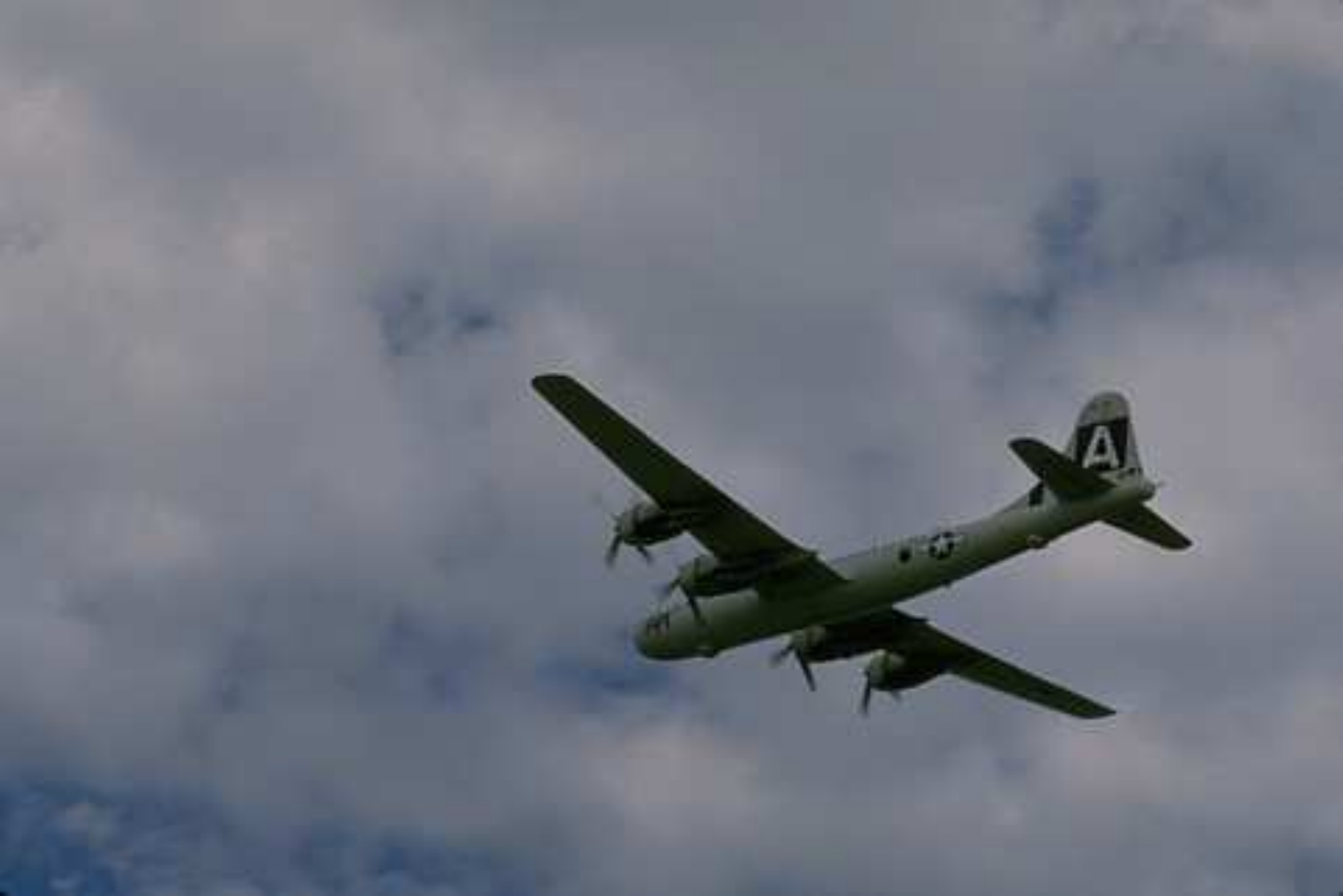} &
		\includegraphics[width=0.24\linewidth]{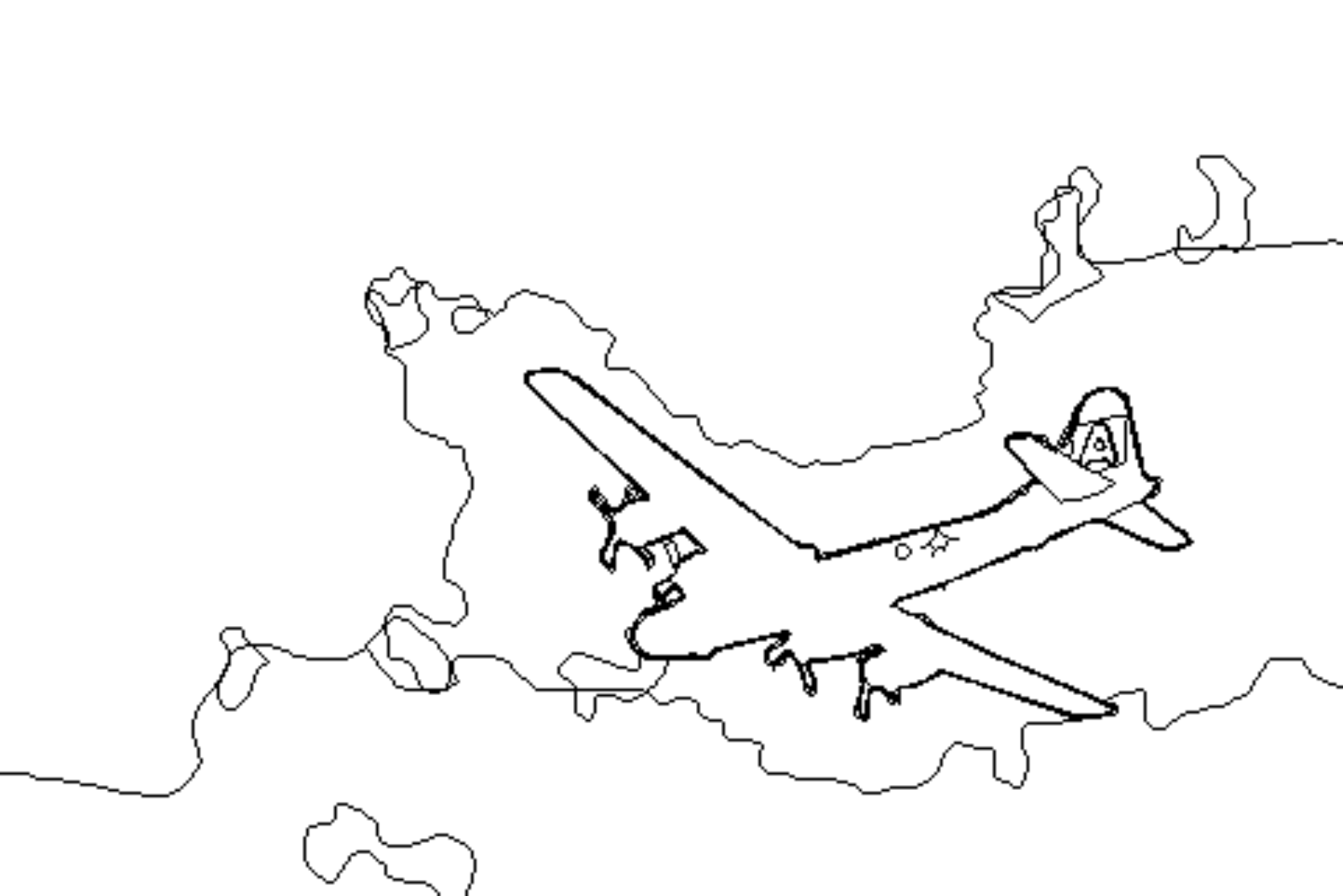} &
		\includegraphics[width=0.24\linewidth]{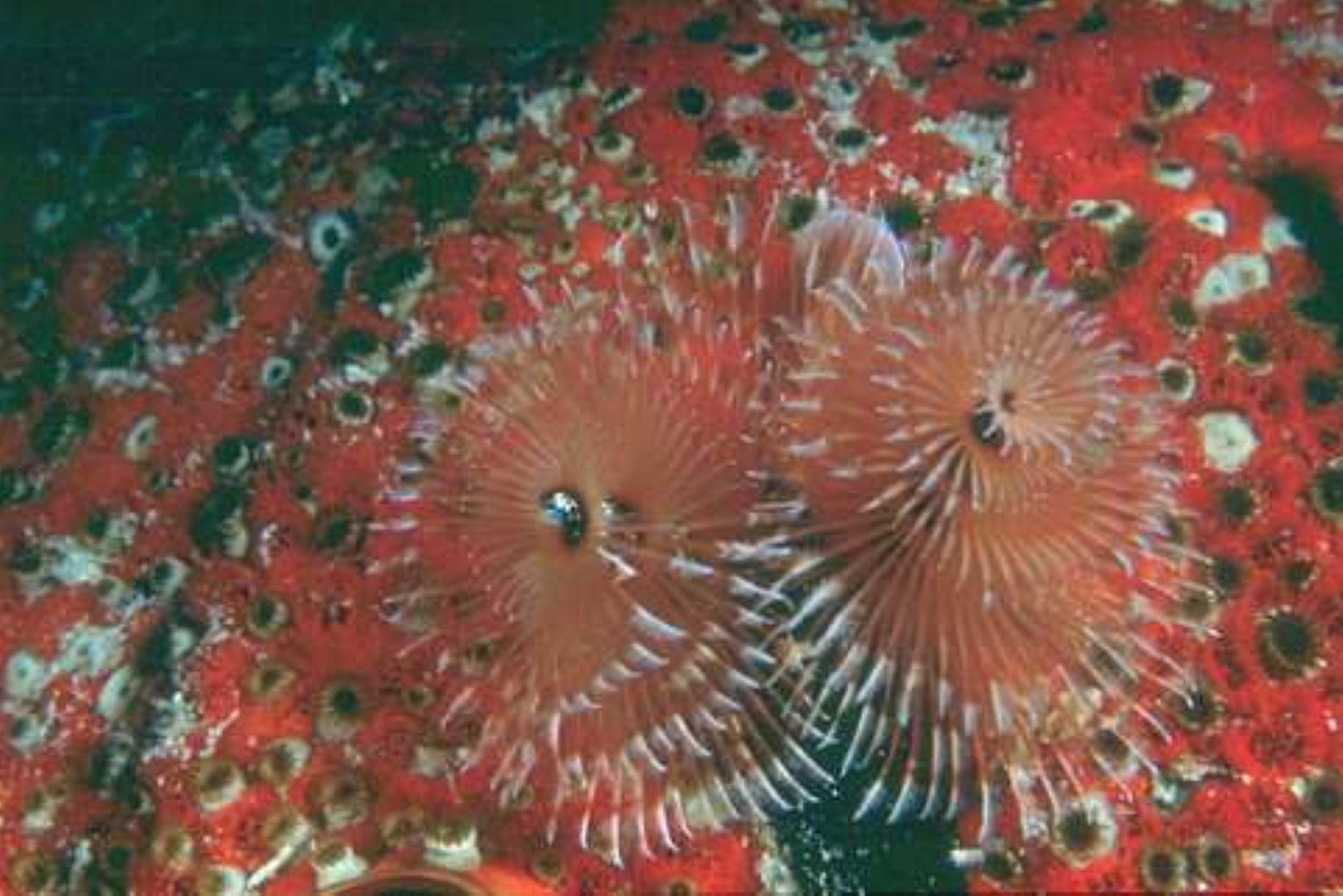} &
		\includegraphics[width=0.24\linewidth]{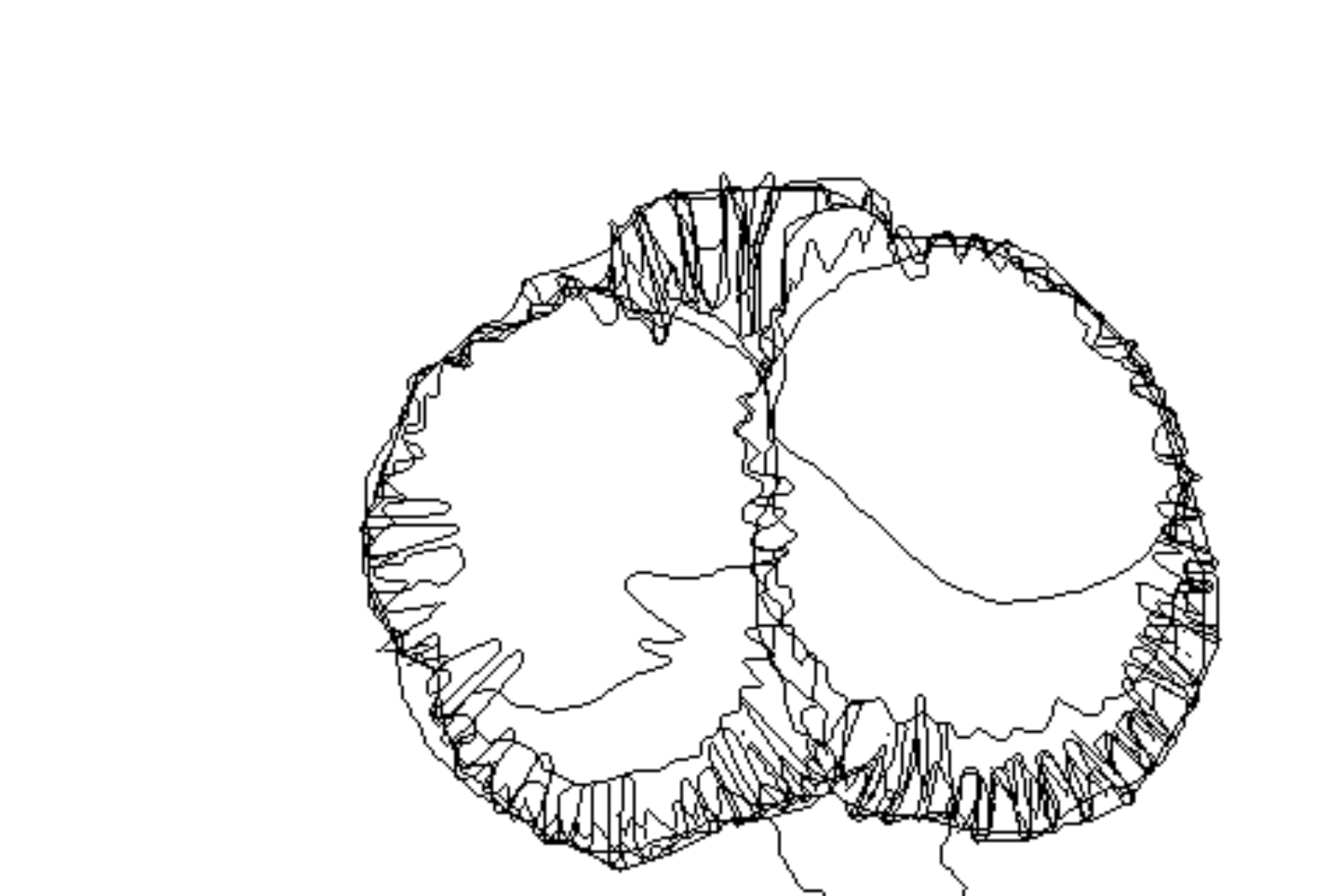} \\
		\includegraphics[width=0.24\linewidth]{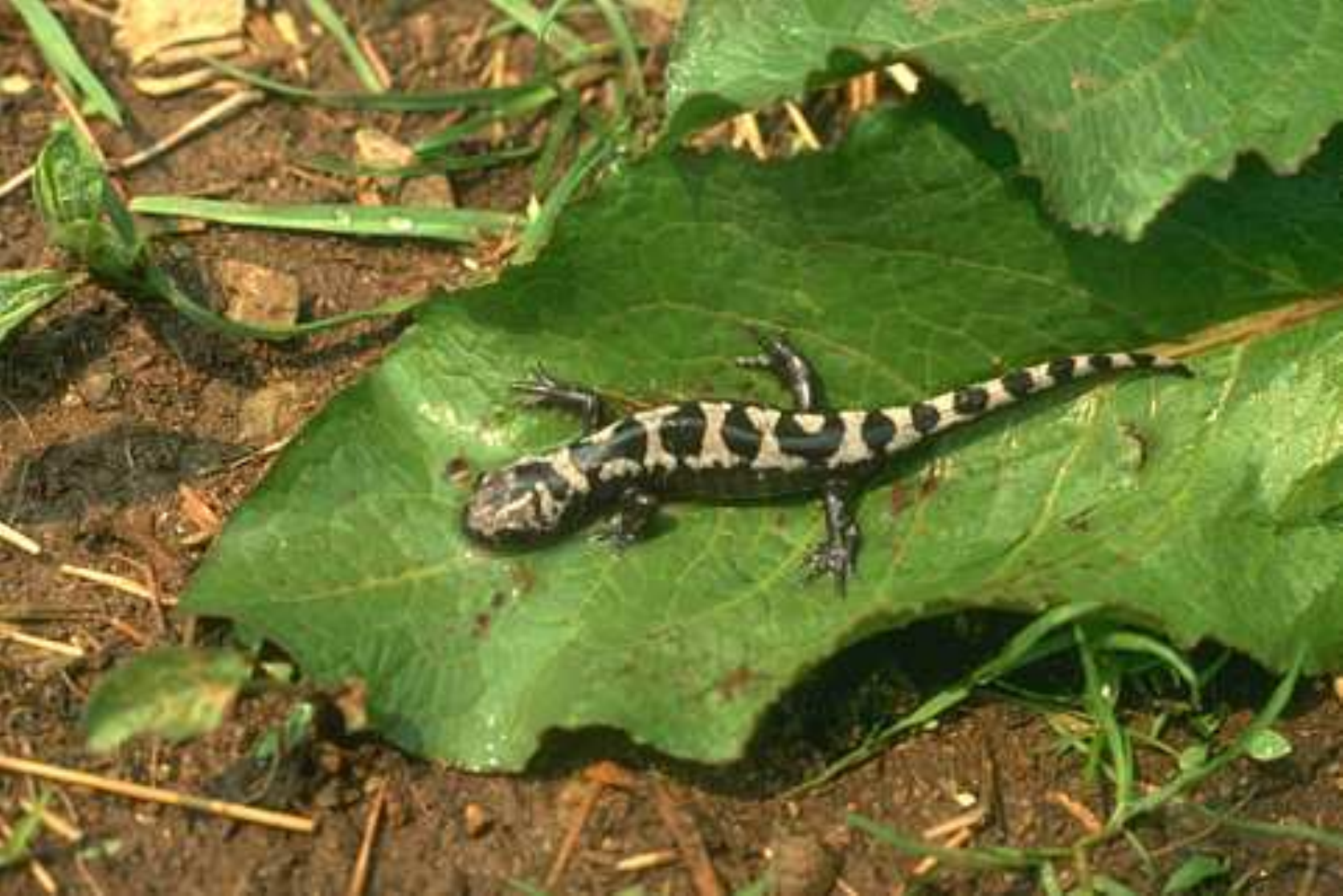} &
		\includegraphics[width=0.24\linewidth]{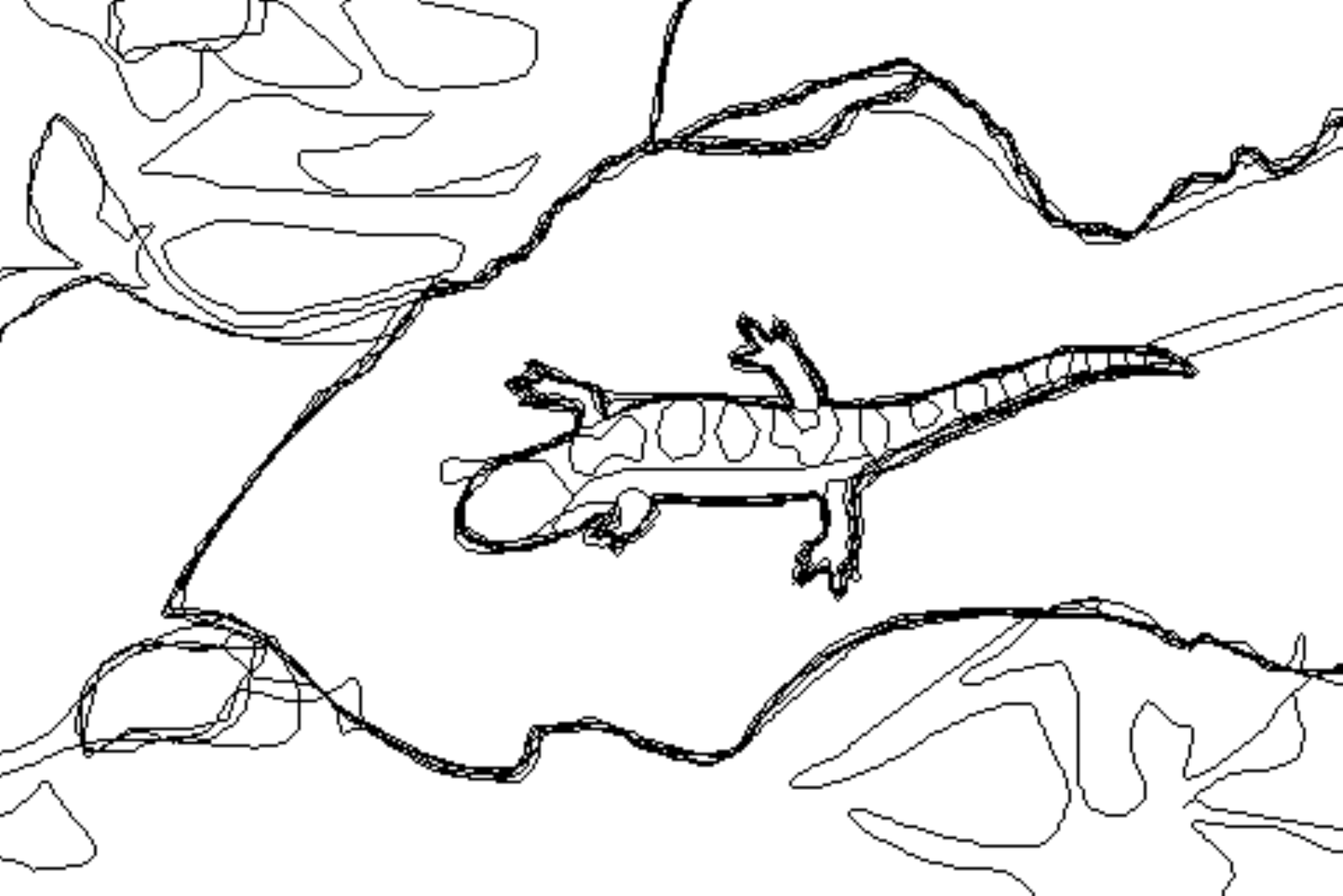} &
		\includegraphics[width=0.24\linewidth]{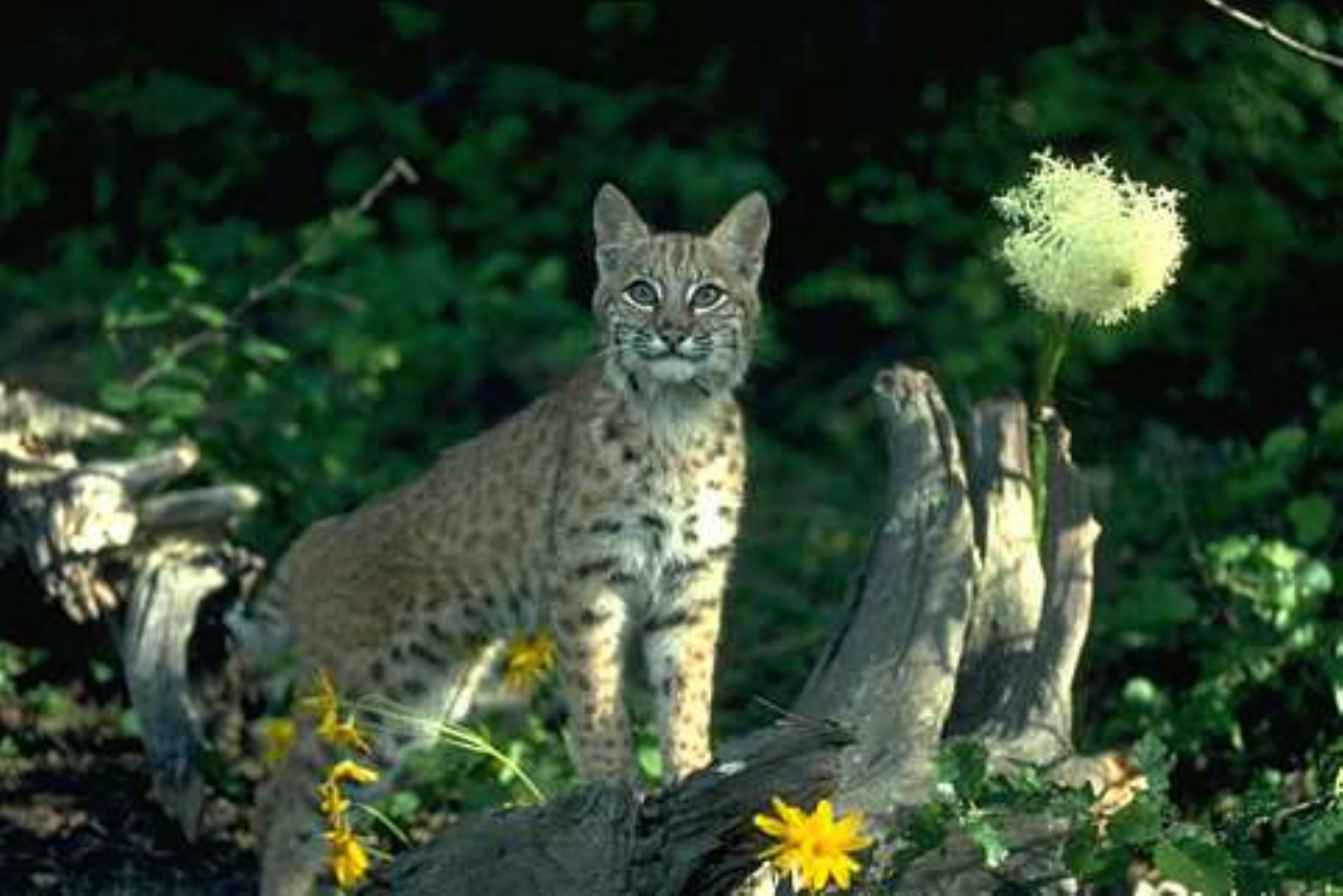} &
		\includegraphics[width=0.24\linewidth]{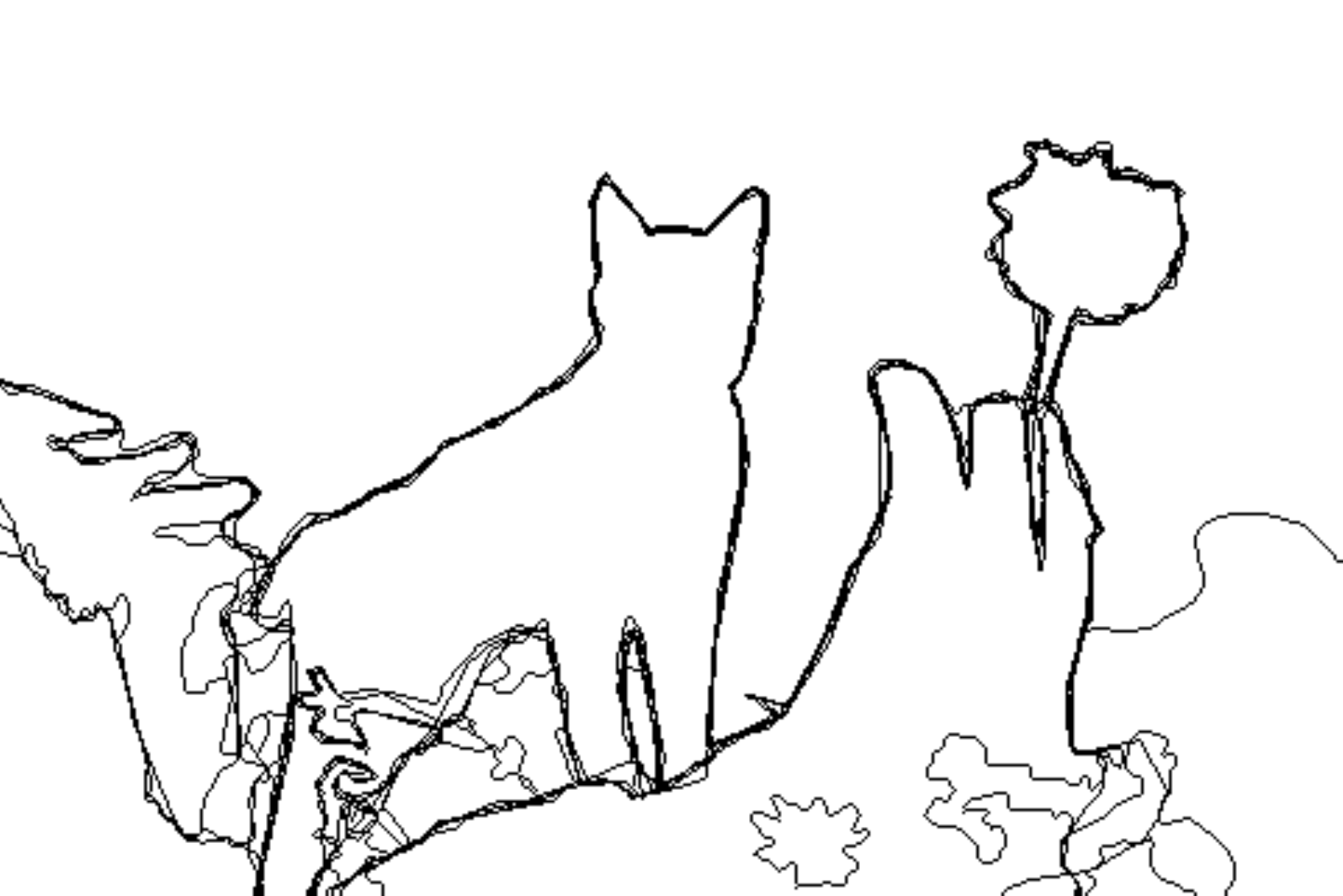} \\
	\end{tabular}
	\caption{Four samples in the BSDS500 dataset.}
	\label{Fig:bsds-dataset}
	\vspace{-0.5em}
\end{figure}

\begin{figure}[tbh]
	\centering
	\includegraphics[width=\linewidth]{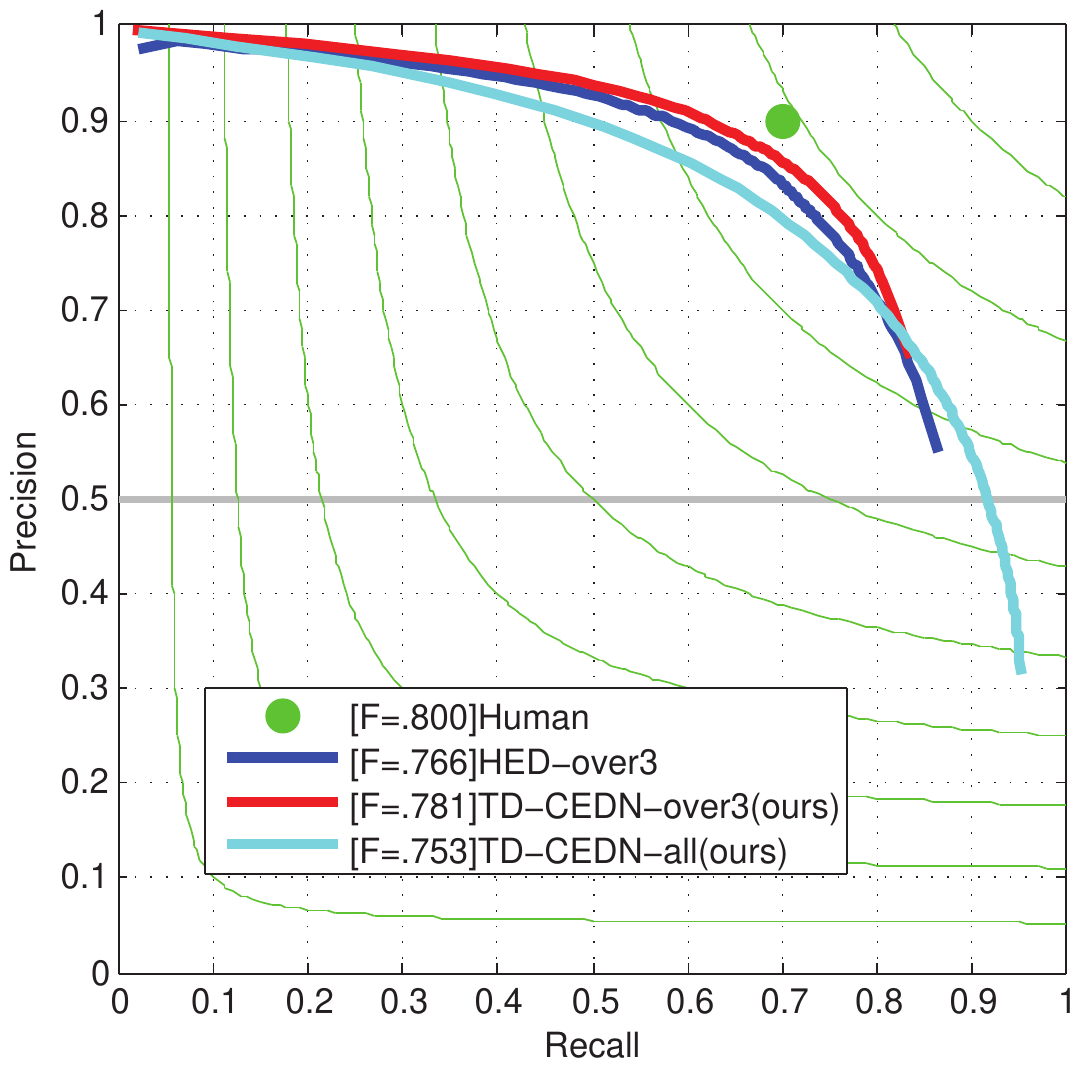}\\
	\centering
	\caption{The PR curve for contour detection with different assumptions of the ground truth on the BSDS500 dataset.}
	\label{Fig:bsds-a}
\end{figure}

Considering that the dataset was annotated by multiple individuals independently, as samples illustrated in Fig.~\ref{Fig:bsds-dataset},  we trained the dataset with two strategies: (1) assigning a pixel a positive label if only if it's labeled as positive by at least three annotators, otherwise this pixel was labeled as negative; (2) treating all annotated contour labels as positives. Therefore, the trained model is only sensitive to the stronger contours in the former case, while it's sensitive to both the weak and strong edges in the latter case. Fig.~\ref{Fig:bsds-a} shows the results of HED and our method, where the ``HED-over3" denotes the HED network trained with the above-mentioned first training strategy which was provided by Xie~\etal~\footnotemark[1]\footnotetext[1]{HED pretrained model:~\url{http://vcl.ucsd.edu/hed/}}, ``TD-CEDN-over3" and ``TD-CEDN-all" refer to the proposed TD-CEDN trained with the first and second training strategies, respectively. According to the results, the performances show a big difference with these two training strategies. When the trained model is sensitive to the stronger contours, it shows a better performance on precision but a poor performance on recall in the PR curve. When the trained model is sensitive to both the weak and strong contours, it shows an inverted results.

\begin{figure}[tbh]
	\centering
	\includegraphics[width=\linewidth]{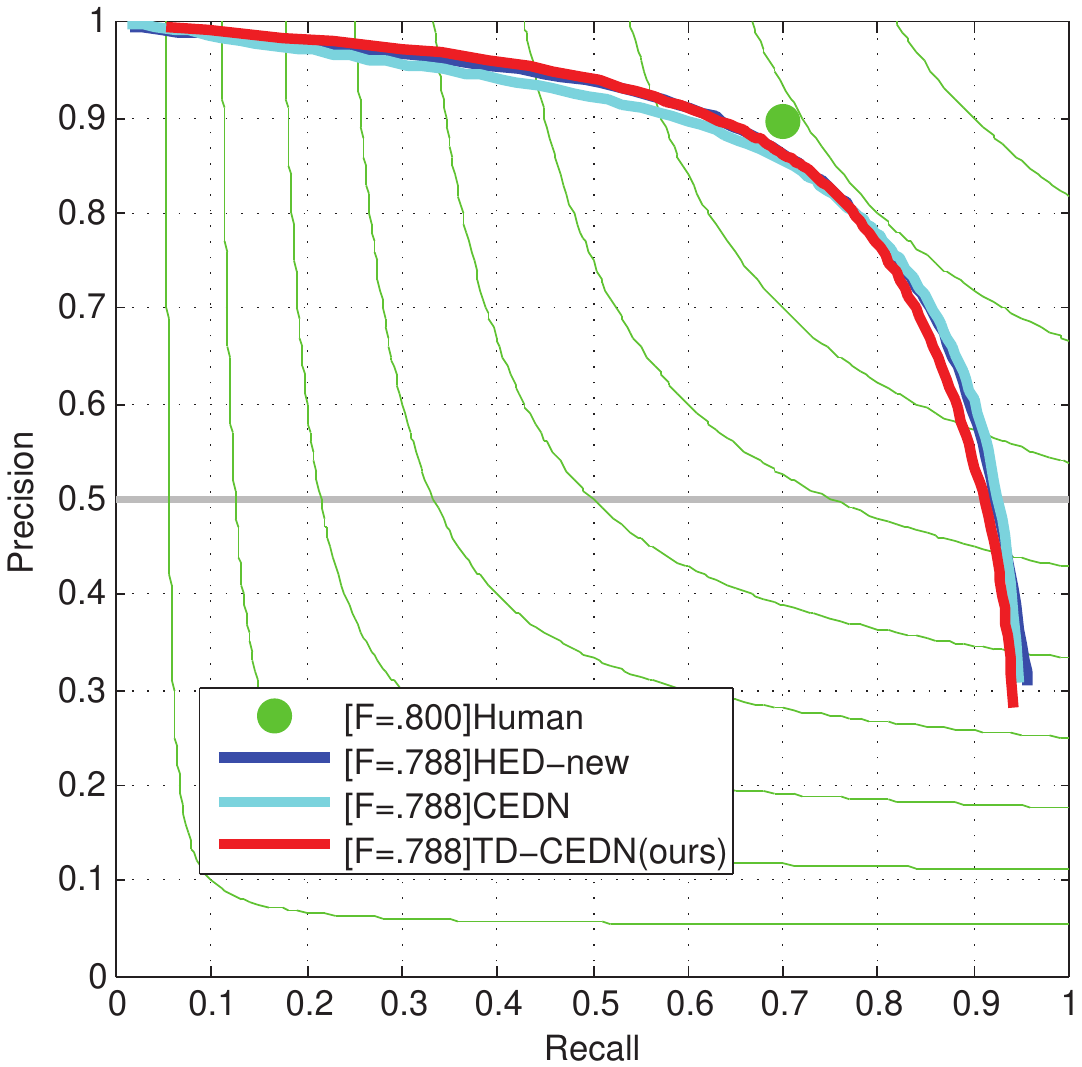}\\
	\centering
	\caption{The results on the BSDS500 dataset. Our proposed TD-CEDN network achieved the best ODS F-score of 0.788. }
	\label{Fig:bsds-b}
\end{figure}

With the observation, we applied a simple method to solve such problem. For an image, the predictions of two trained models are denoted as $\hat{\groundtruth}_{\over3}$ and $\hat{\groundtruth}_{\all}$, respectively. A simple fusion strategy is defined as:
\begin{equation}\label{Eq:fusion}
\hat{\groundtruth} = \gamma\hat{\groundtruth}_{\over3} + (1-\gamma)\hat{\groundtruth}_{\all},
\end{equation}
where $\gamma$ is a hyper-parameter controlling the weight of the prediction of the two trained models. For simplicity, we set $\gamma$ as a constant value of 0.5. For simplicity, the ``TD-CEDN-over3", ``TD-CEDN-all" and ``TD-CEDN" refer to the results of $\hat{\groundtruth}_{\over3}$, $\hat{\groundtruth}_{\all}$ and $\hat{\groundtruth}$, respectively. Fig.~\ref{Fig:bsds-b} shows the fused performances compared with HED and CEDN, in which our method achieved the state-of-the-art performances. Table~\ref{tab:BSDS_PRF_measures} shows the detailed statistics on the BSDS500 dataset, in which our method achieved the best performances in ODS=0.788 and OIS=0.809. 

Fig.~\ref{Fig:ours-hed} presents several predictions which were generated by the ``HED-over3" and ``TD-CEDN-over3" models. With the same training strategy, our method achieved the best ODS=0.781 which is higher than the performance of ODS=0.766 for HED, as shown in Fig.~\ref{Fig:bsds-a}. Observing the predicted maps, our method predicted the contours more precisely and clearly, which seems to be a refined version. Fig.~\ref{Fig:ours-cedn} presents our fused results and the CEDN published predictions. Our results present both the weak and strong edges better than CEDN on visual effect. Moreover, to suppress the image-border contours appeared in the results of CEDN, we applied a simple image boundary region extension method to enlarge the input image 10 pixels around the image during the testing stage. The enlarged regions were cropped to get the final results.

\begin{table}
	\begin{center}
		\caption{{Contour detection results on the BSDS500 dataset.}}
		\renewcommand{\tabcolsep}{6pt}
		\renewcommand{\arraystretch}{1.1}
		\begin{tabular}{l|l|l|l}\hline
			&   ODS & OIS & AP   \\
			\hlinew{1pt} 
			Human & .80 & .80 & - \\ \hline
			HED~\cite{xie2015holistically} & .782 & .804 & .833 \\
			HED-new~\footnotemark[1] & \textbf{.788} & .808 & \textbf{.840} \\
			CEDN~\cite{yang2016object} & \textbf{.788} & .804 & .821 \\ \hline
			TD-CEDN-all (ours) & .753 & .780 & .807\\
			TD-CEDN-over3 (ours) & .781 & .796 & .757\\
			TD-CEDN (ours) & \textbf{.788} & \textbf{.809} & .833 \\ \hline
		\end{tabular}
		\label{tab:BSDS_PRF_measures}
	\end{center}
\end{table}

\begin{figure}[tbh]
	\small
	\centering
	\renewcommand{\tabcolsep}{1pt}
	\begin{tabular}{cccc}
		\includegraphics[width=0.24\linewidth]{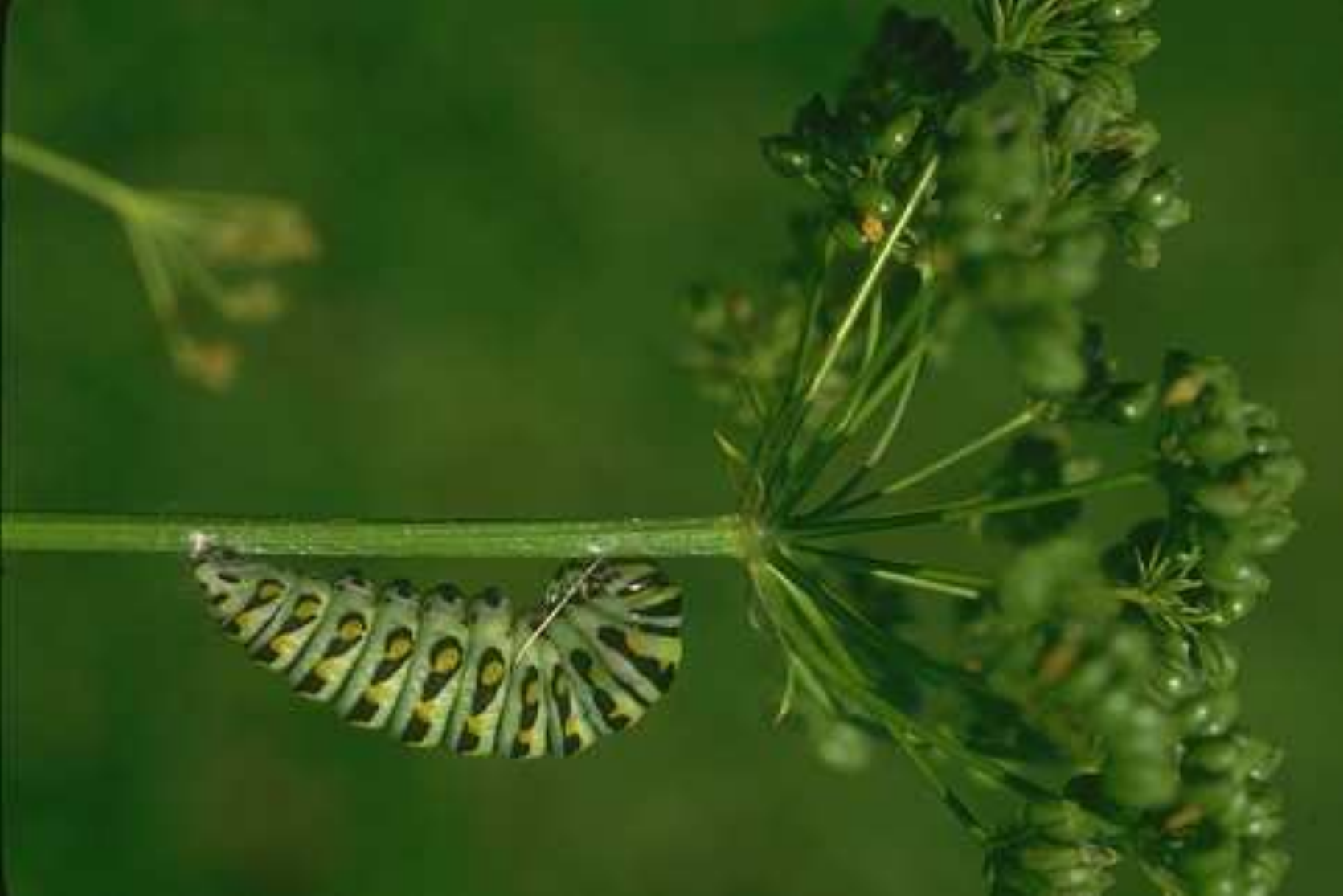} &
		\includegraphics[width=0.24\linewidth]{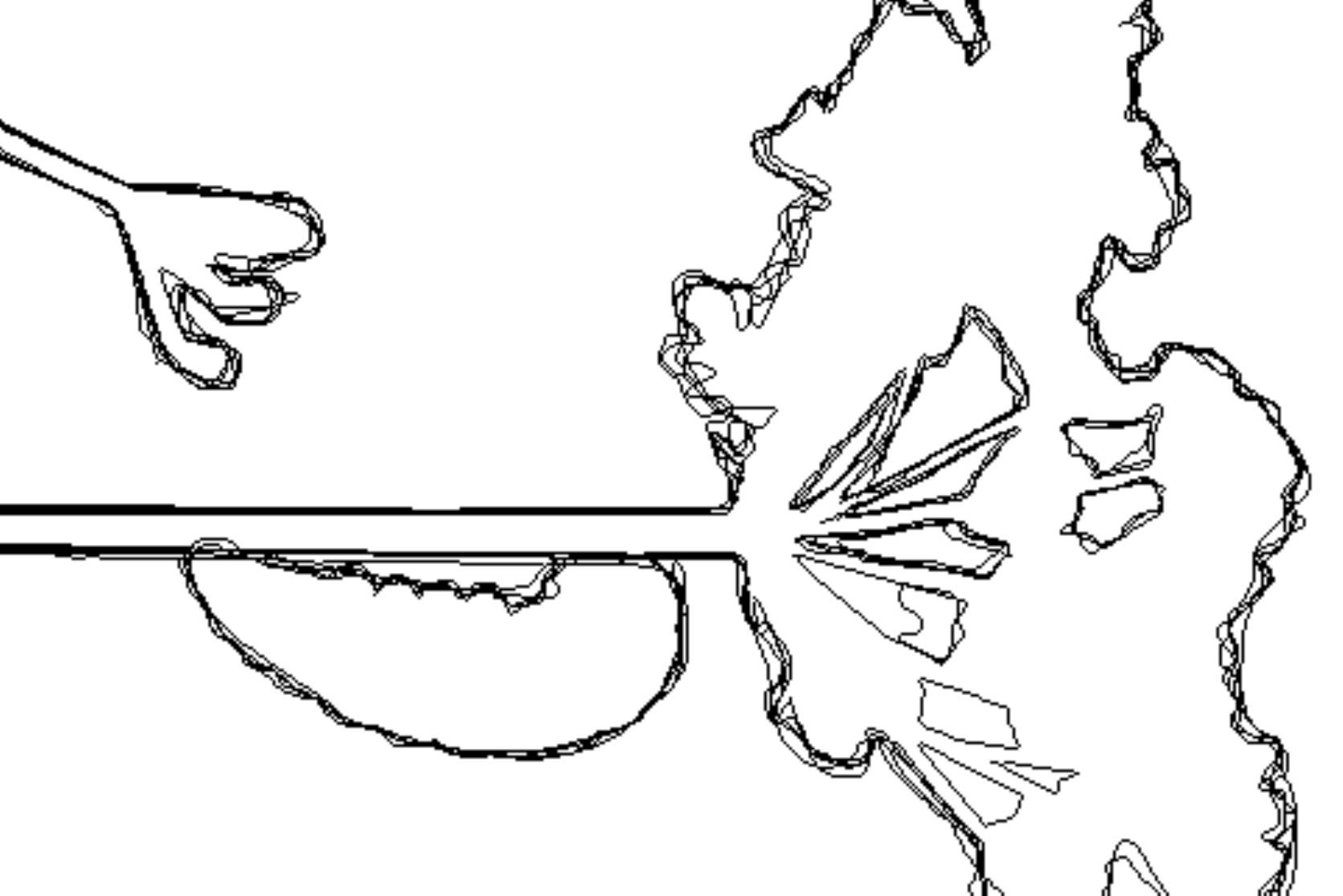} &
		\includegraphics[width=0.24\linewidth]{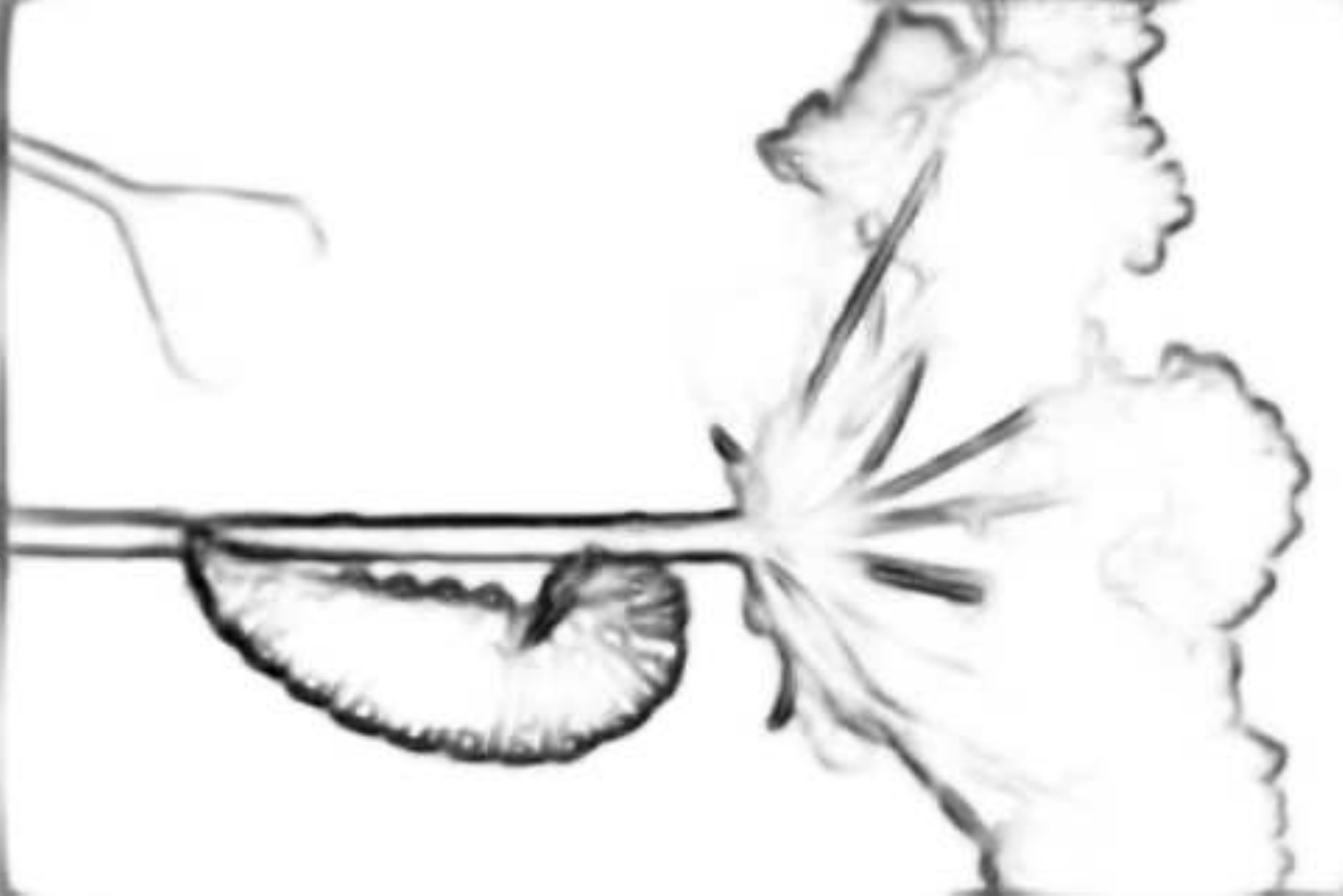} &
		\includegraphics[width=0.24\linewidth]{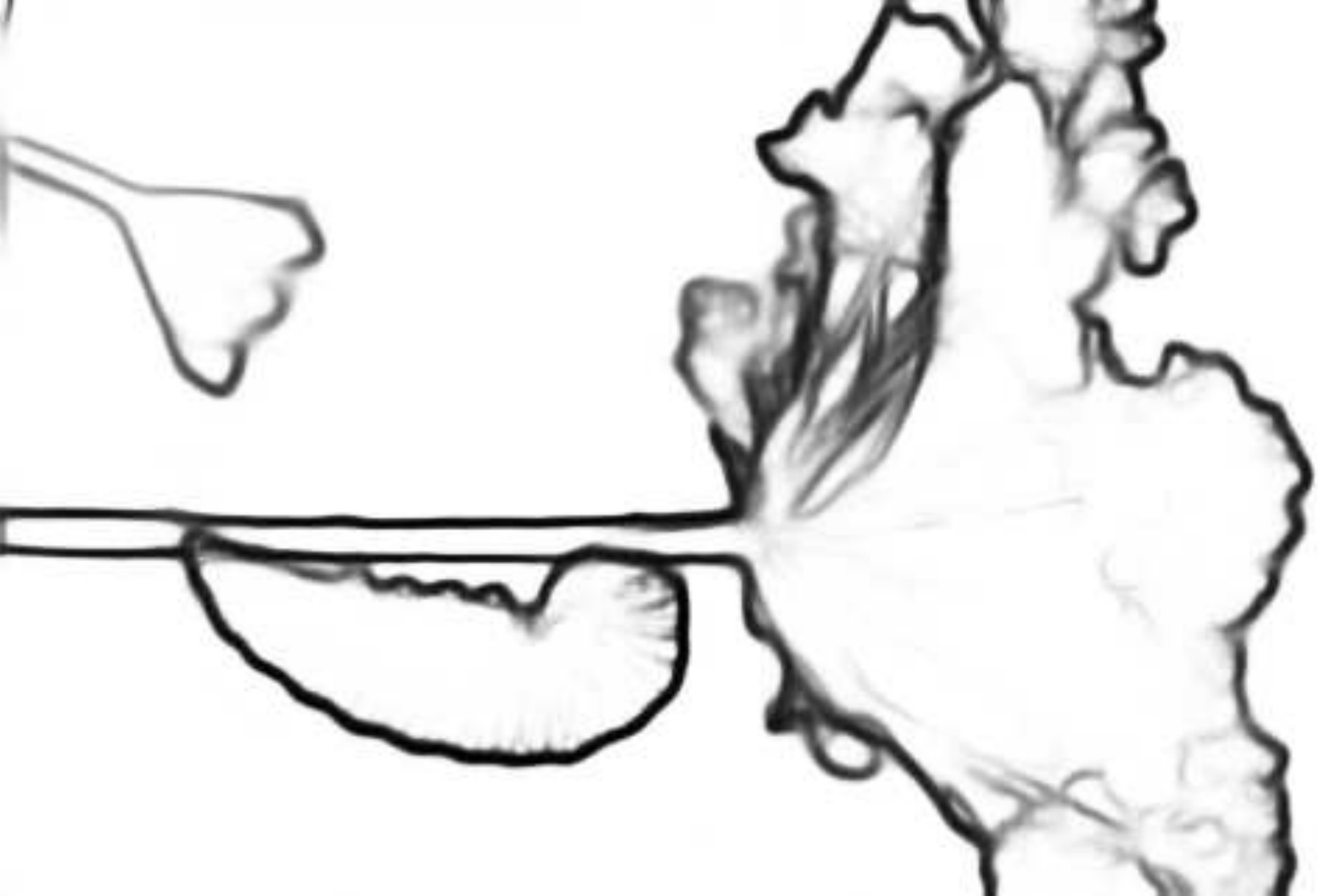} \\
		\includegraphics[width=0.24\linewidth]{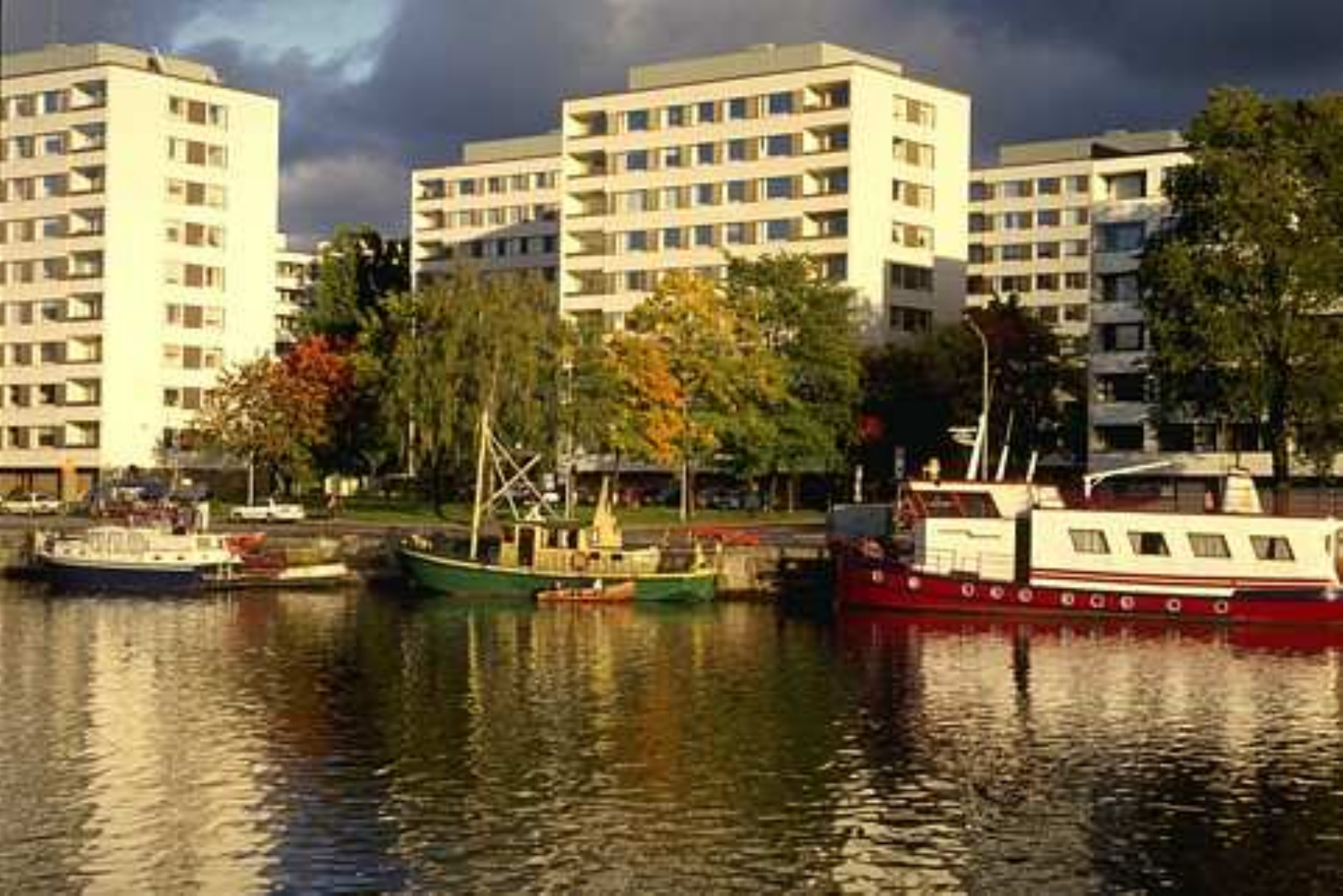} &
		\includegraphics[width=0.24\linewidth]{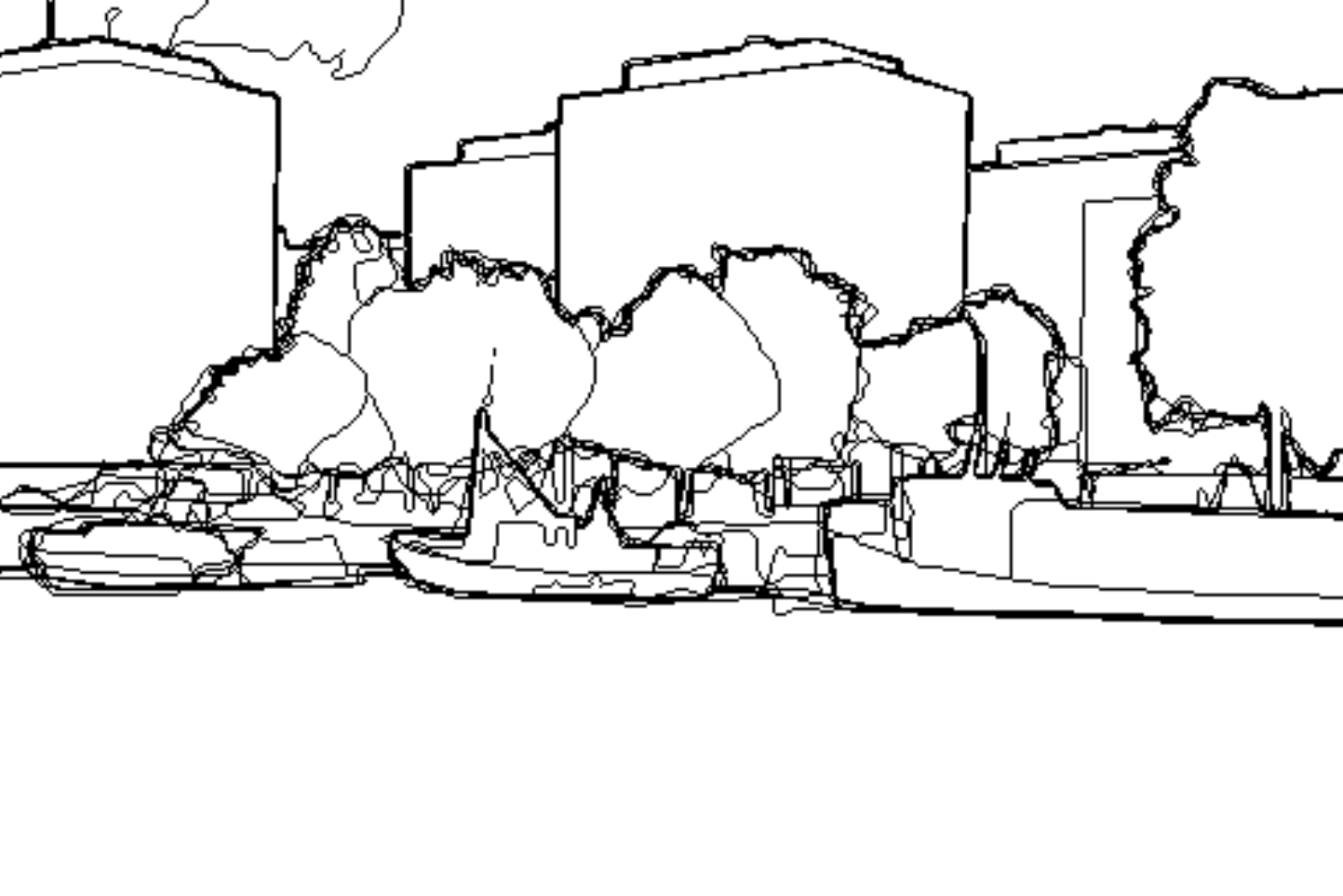} &
		\includegraphics[width=0.24\linewidth]{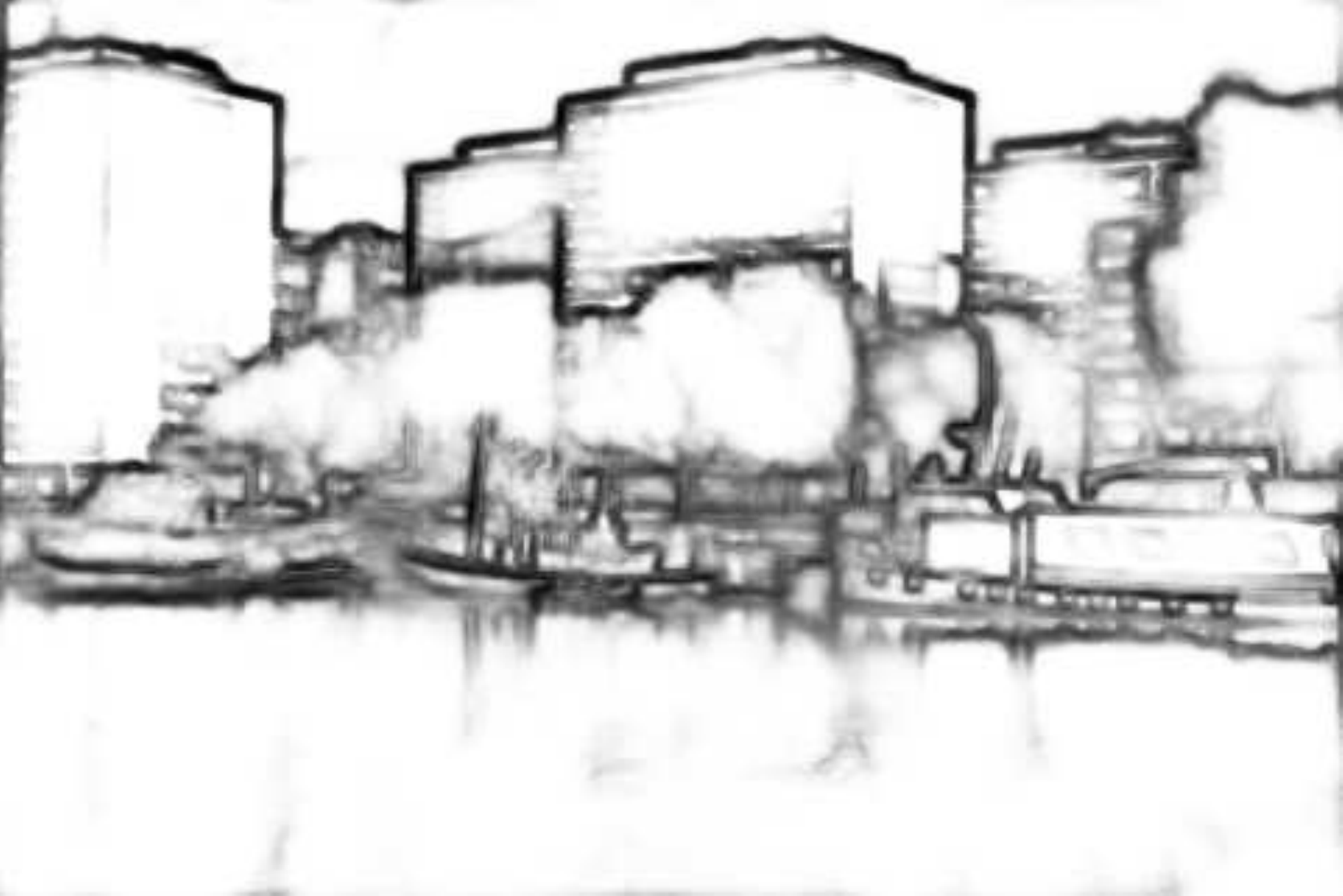} &
		\includegraphics[width=0.24\linewidth]{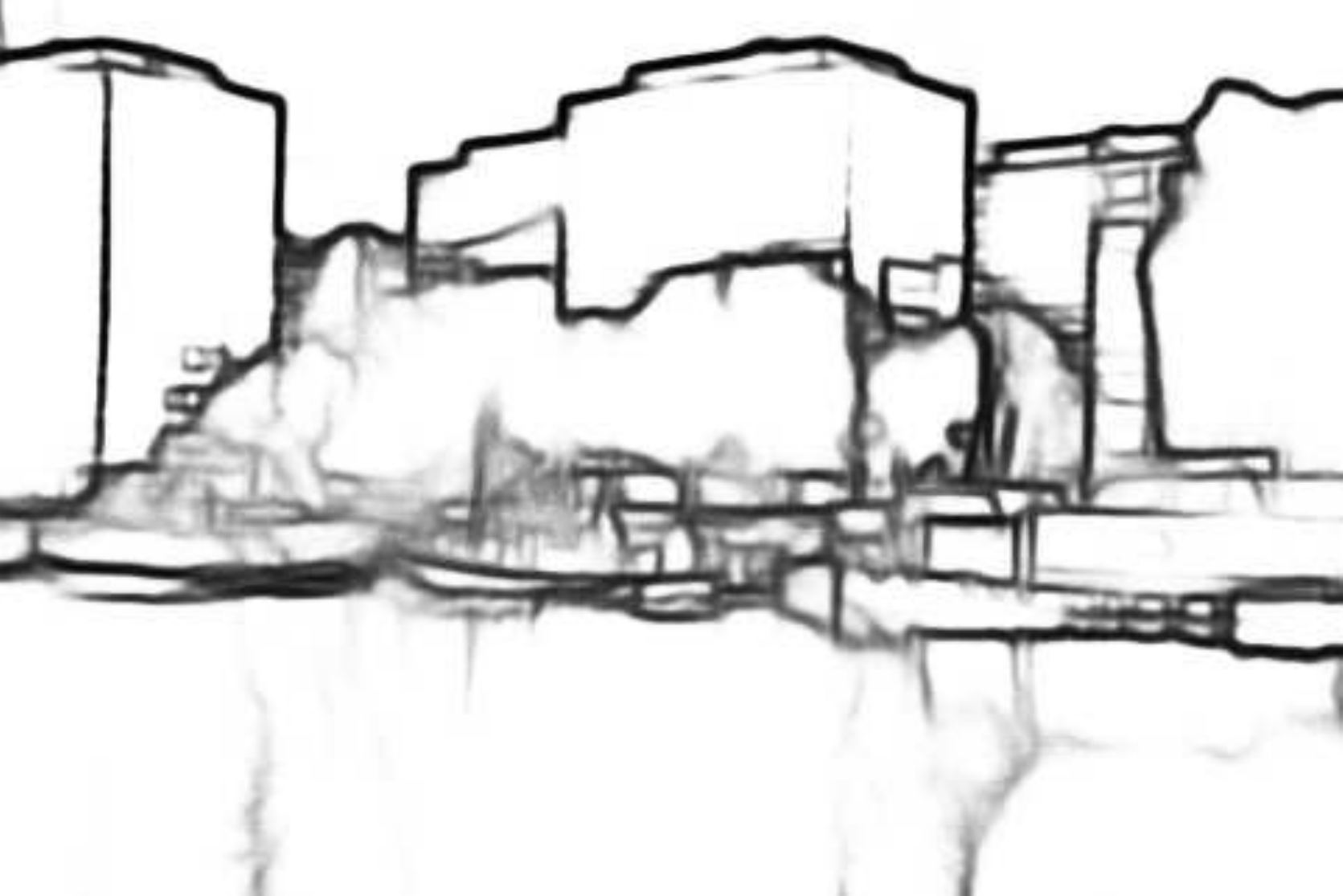} \\
		\includegraphics[width=0.24\linewidth]{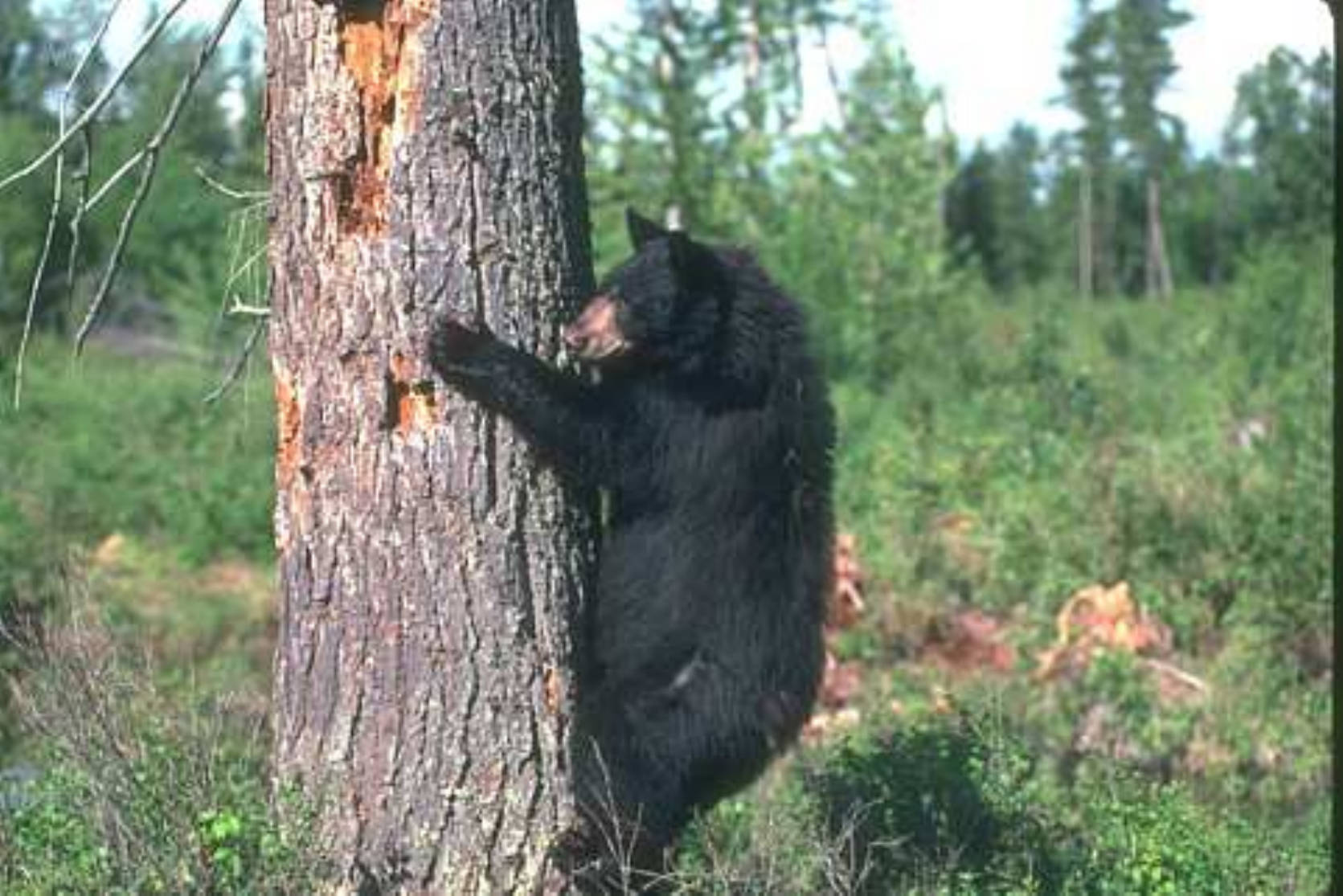} &
		\includegraphics[width=0.24\linewidth]{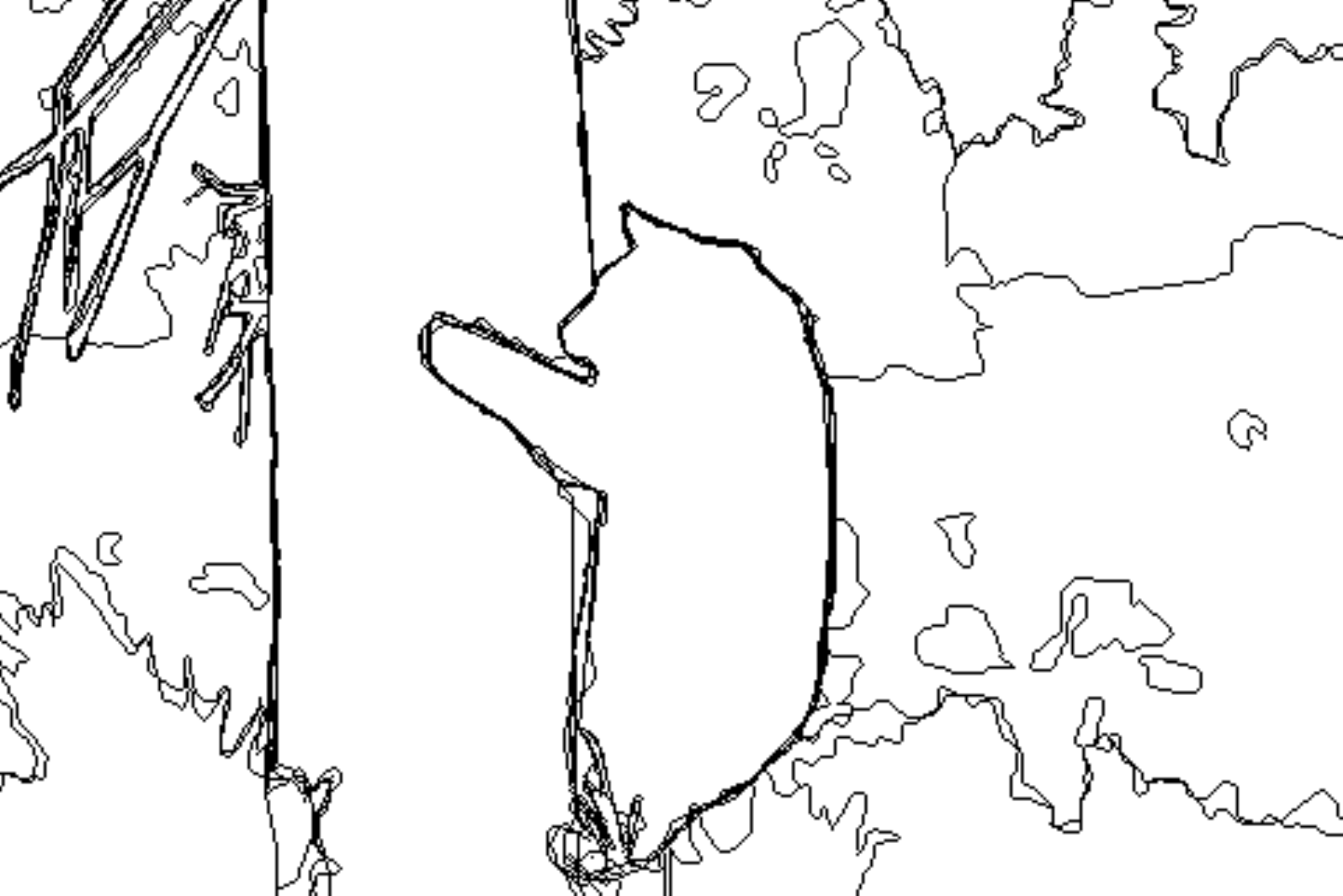} &
		\includegraphics[width=0.24\linewidth]{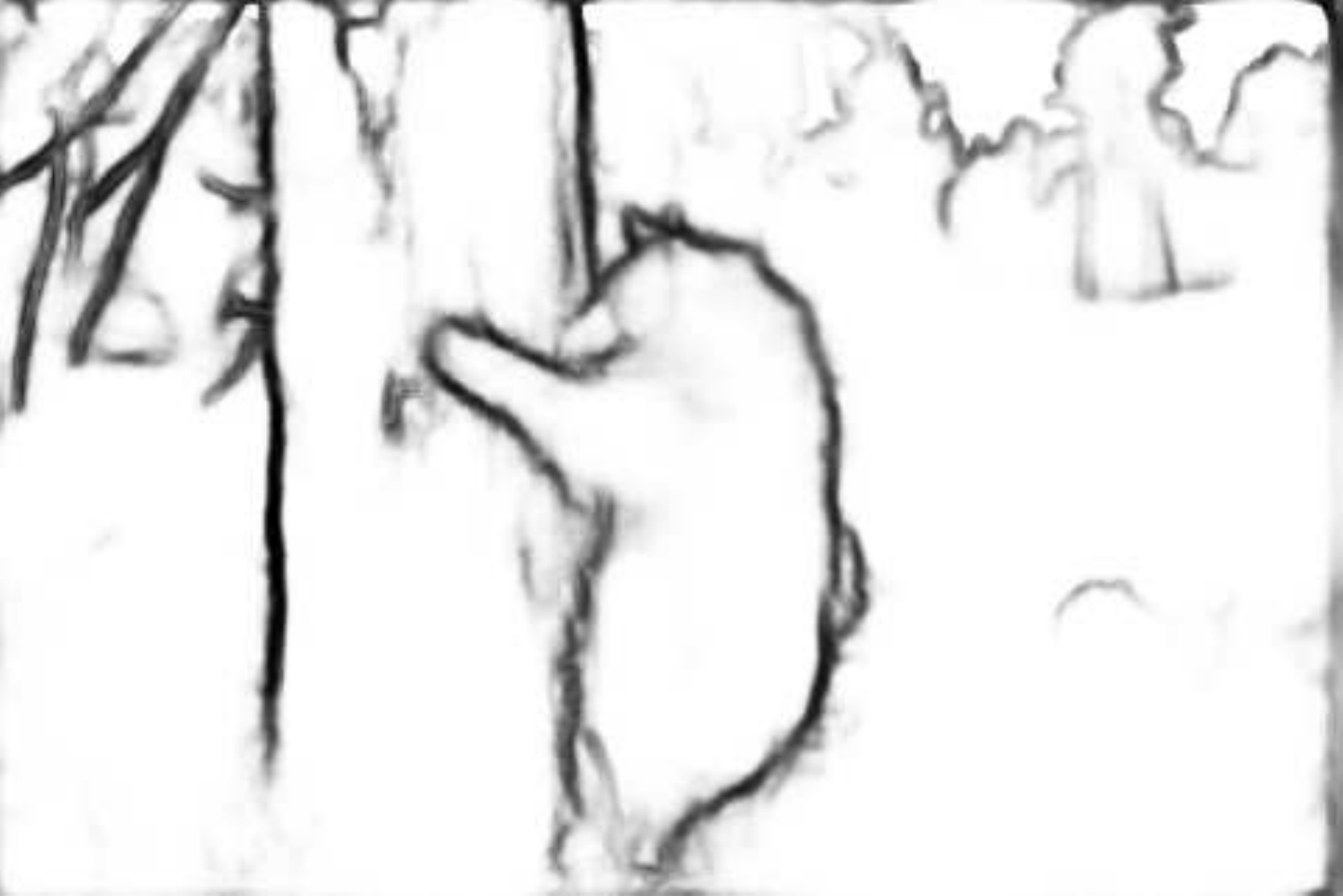} &
		\includegraphics[width=0.24\linewidth]{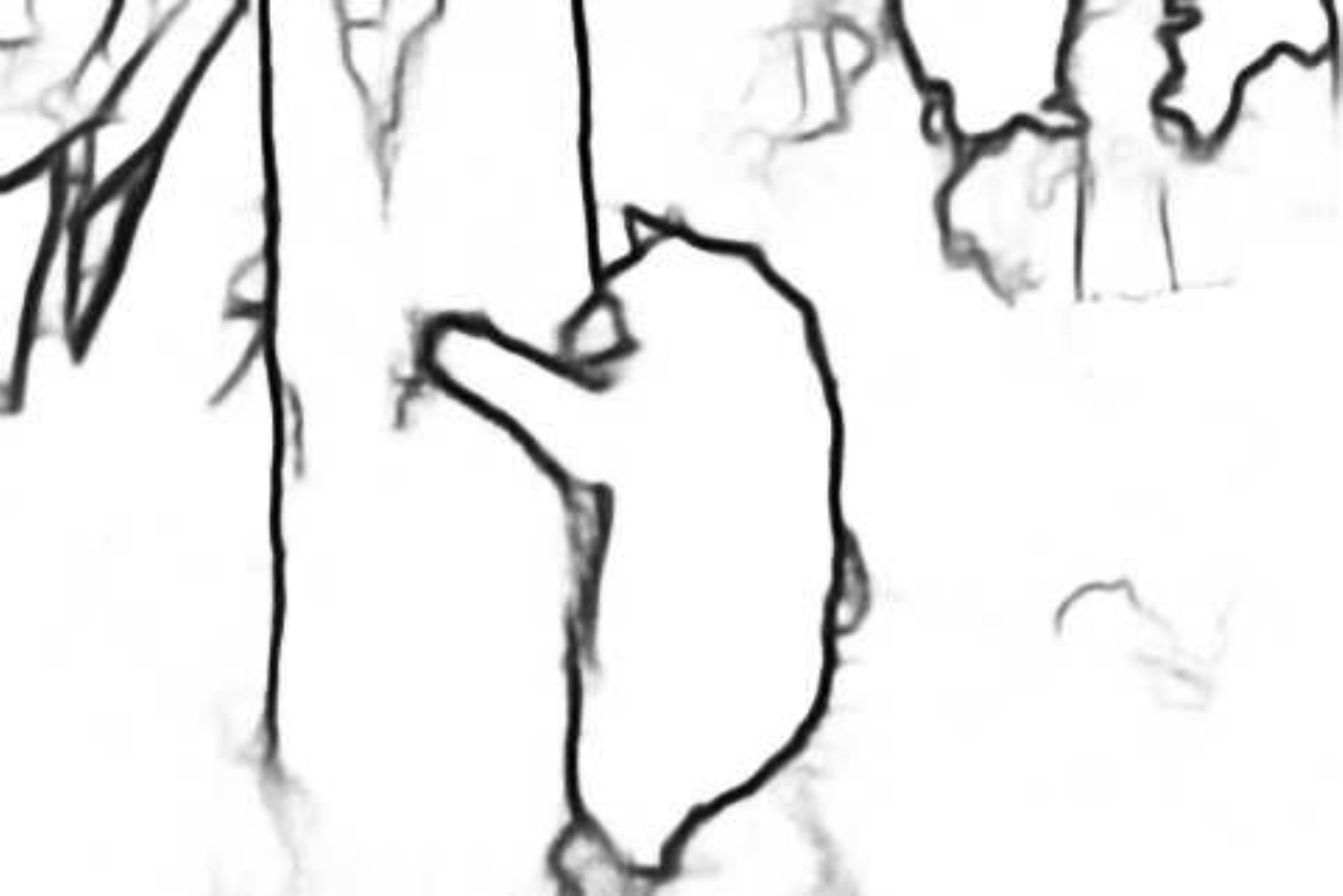} \\
		\includegraphics[width=0.24\linewidth]{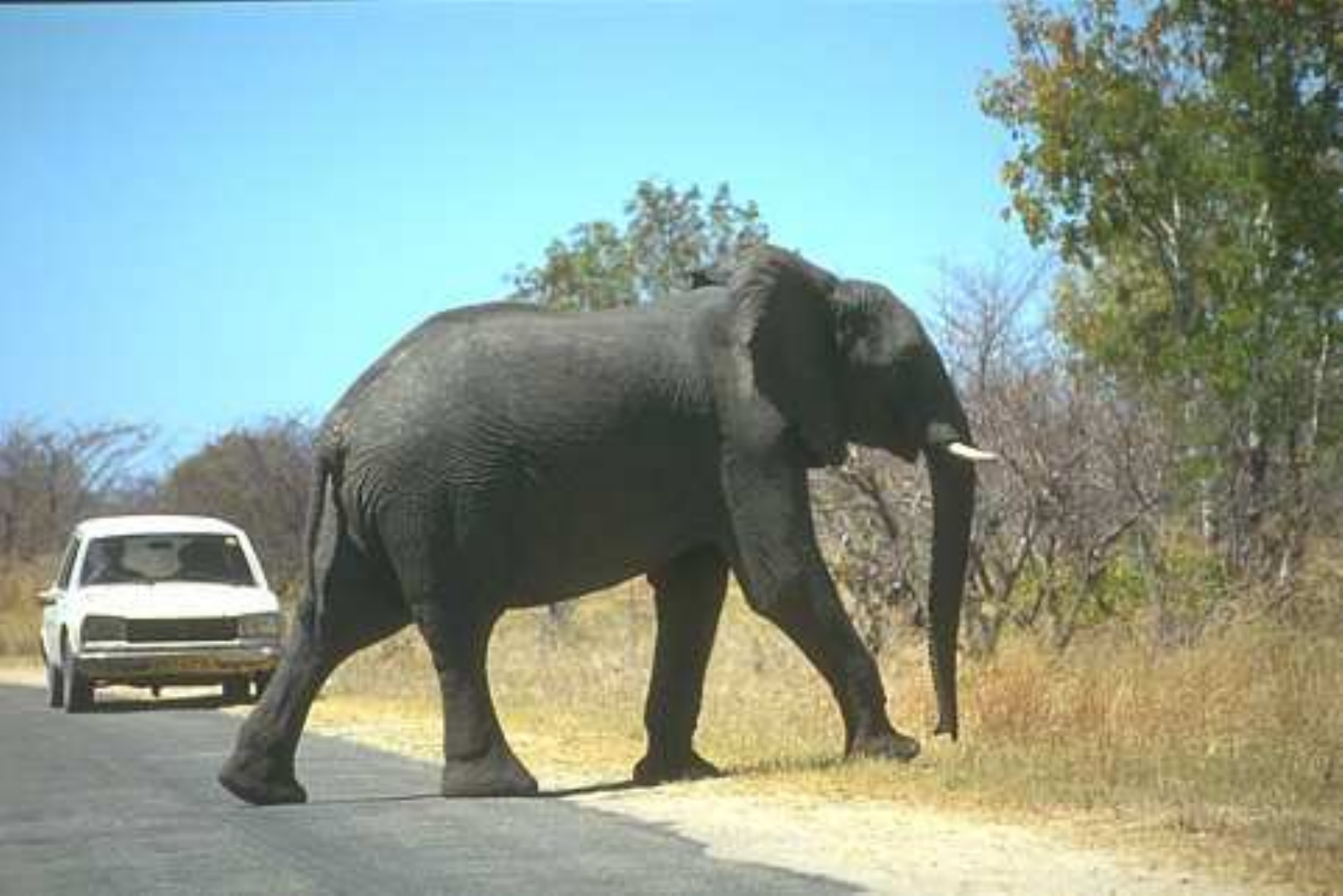} &
		\includegraphics[width=0.24\linewidth]{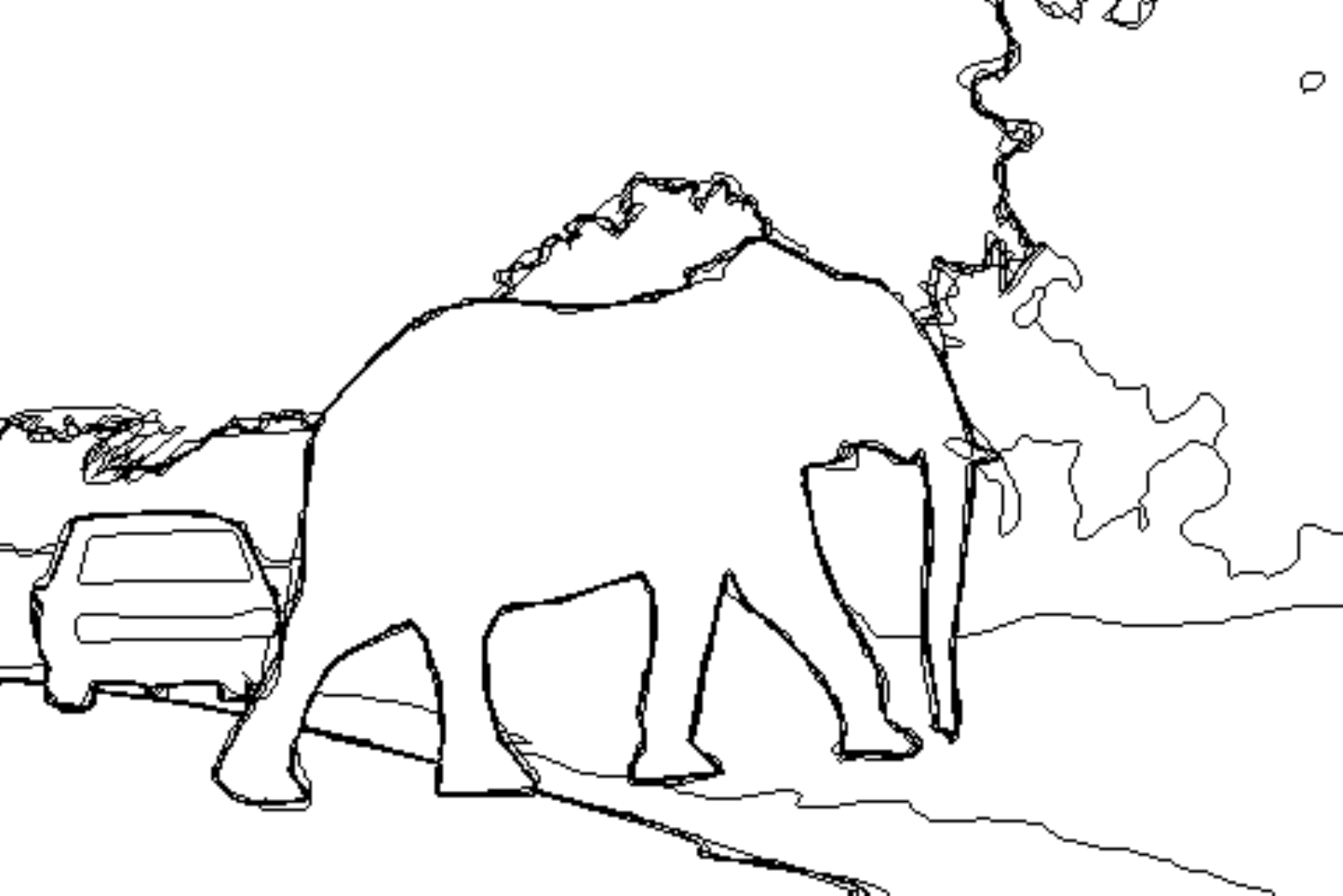} &
		\includegraphics[width=0.24\linewidth]{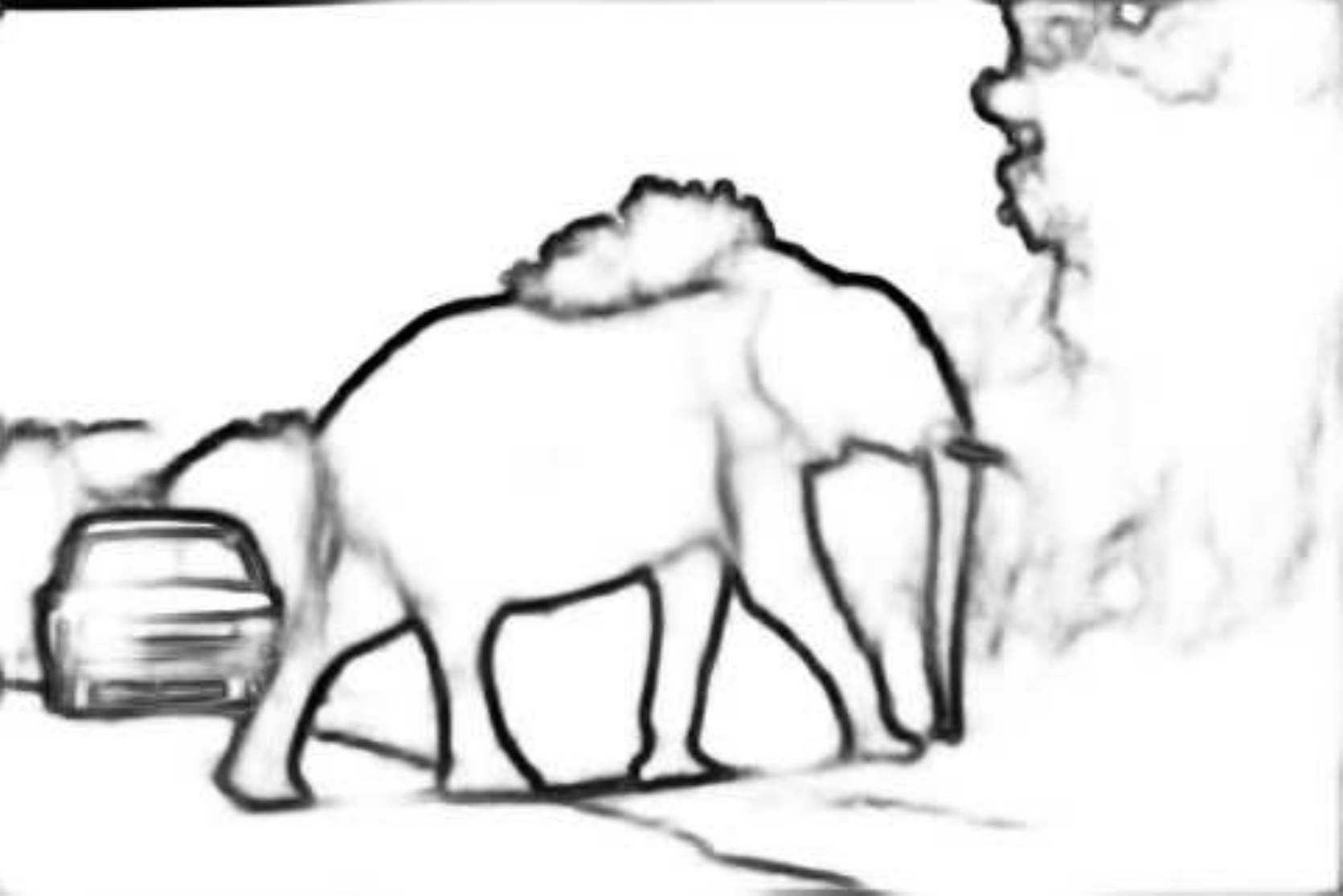} &
		\includegraphics[width=0.24\linewidth]{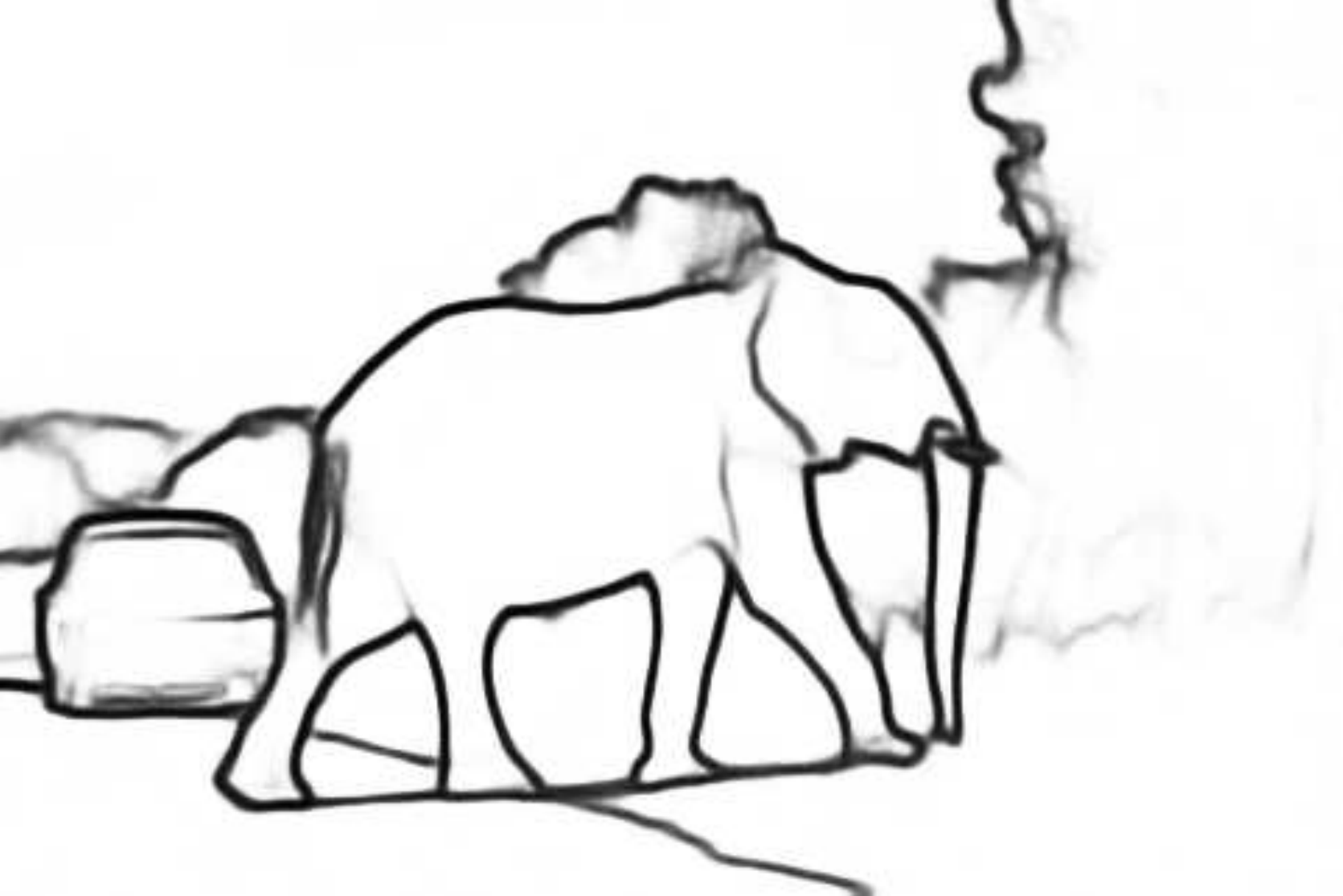} \\
		\includegraphics[width=0.24\linewidth]{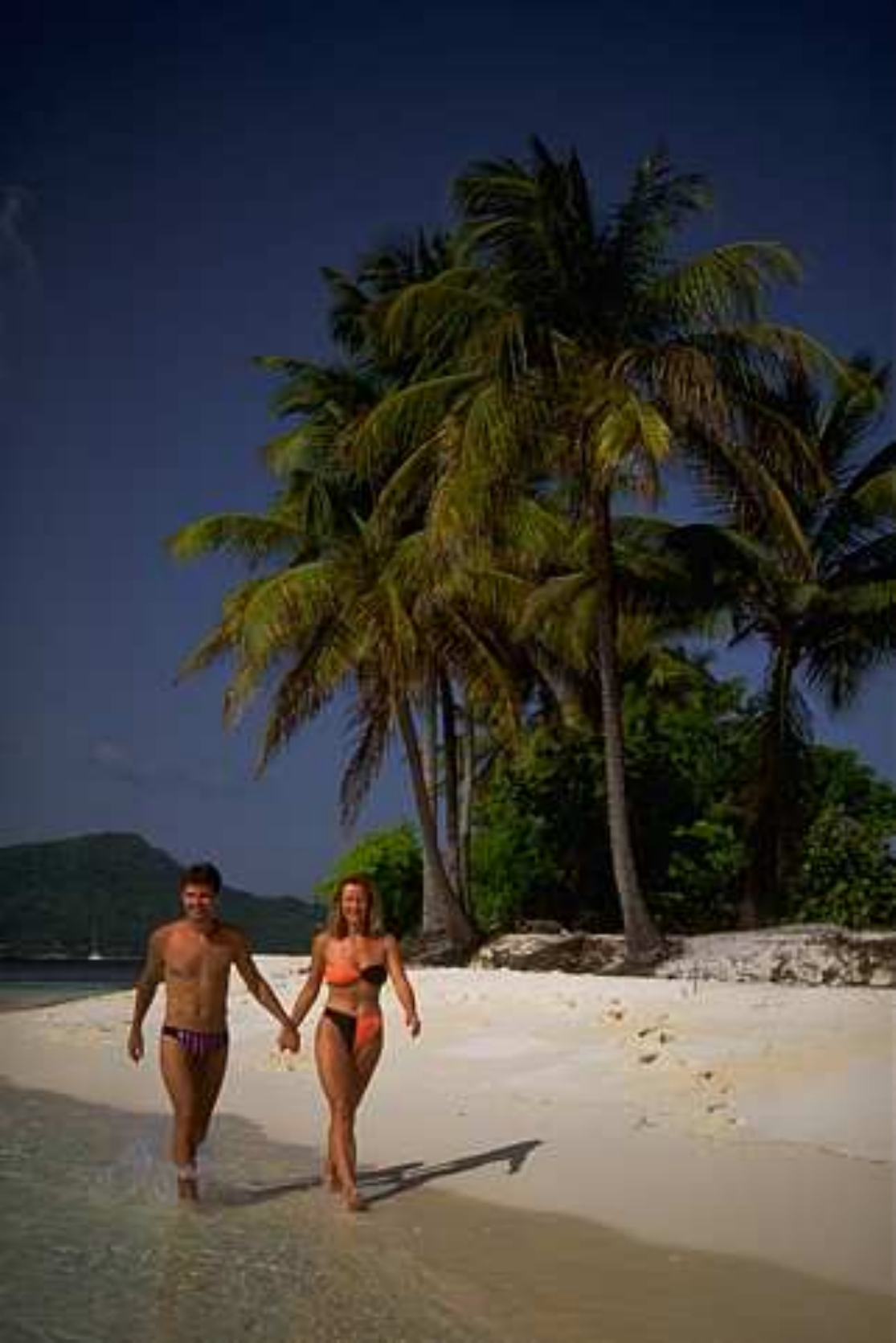} &
		\includegraphics[width=0.24\linewidth]{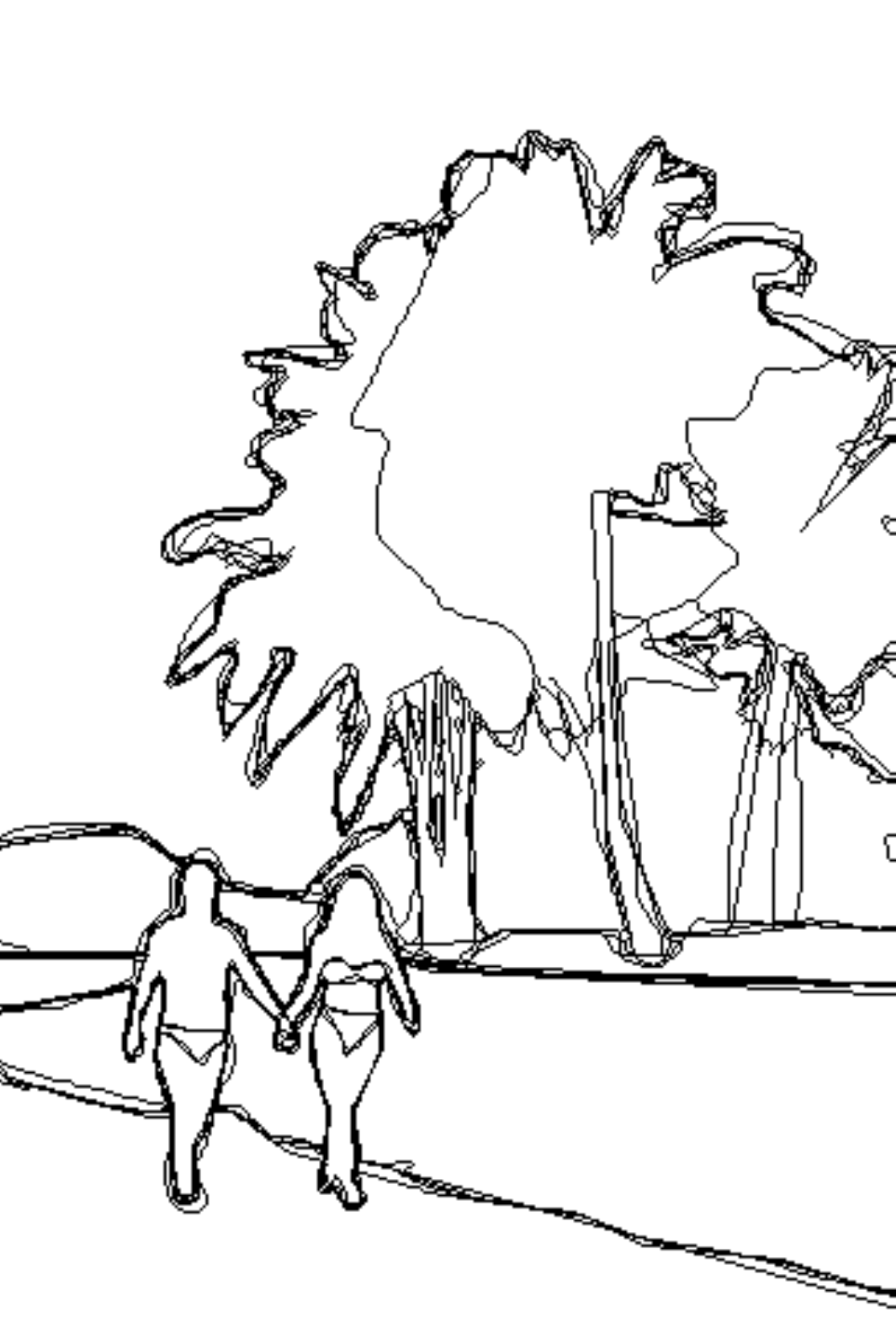} &
		\includegraphics[width=0.24\linewidth]{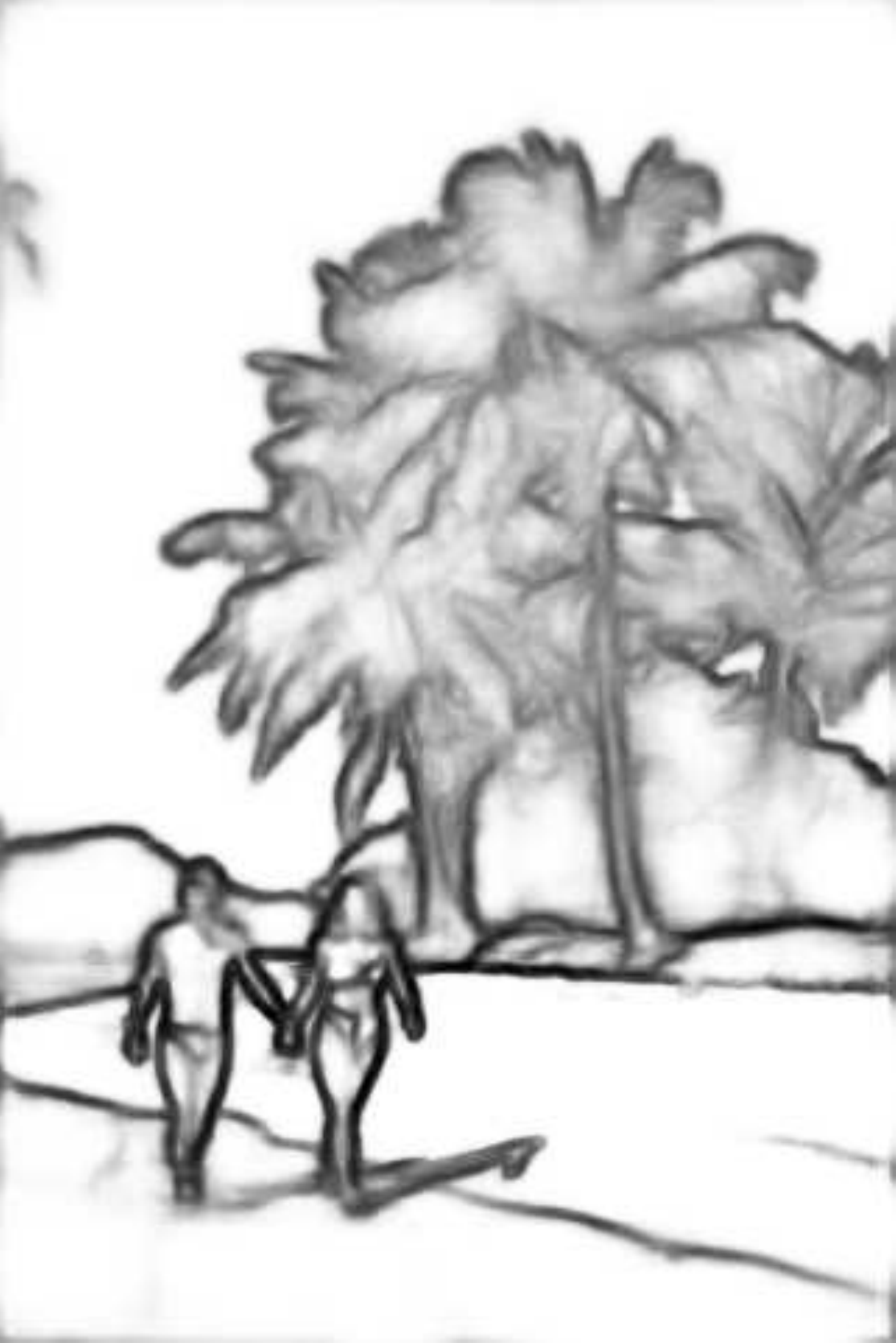} &
		\includegraphics[width=0.24\linewidth]{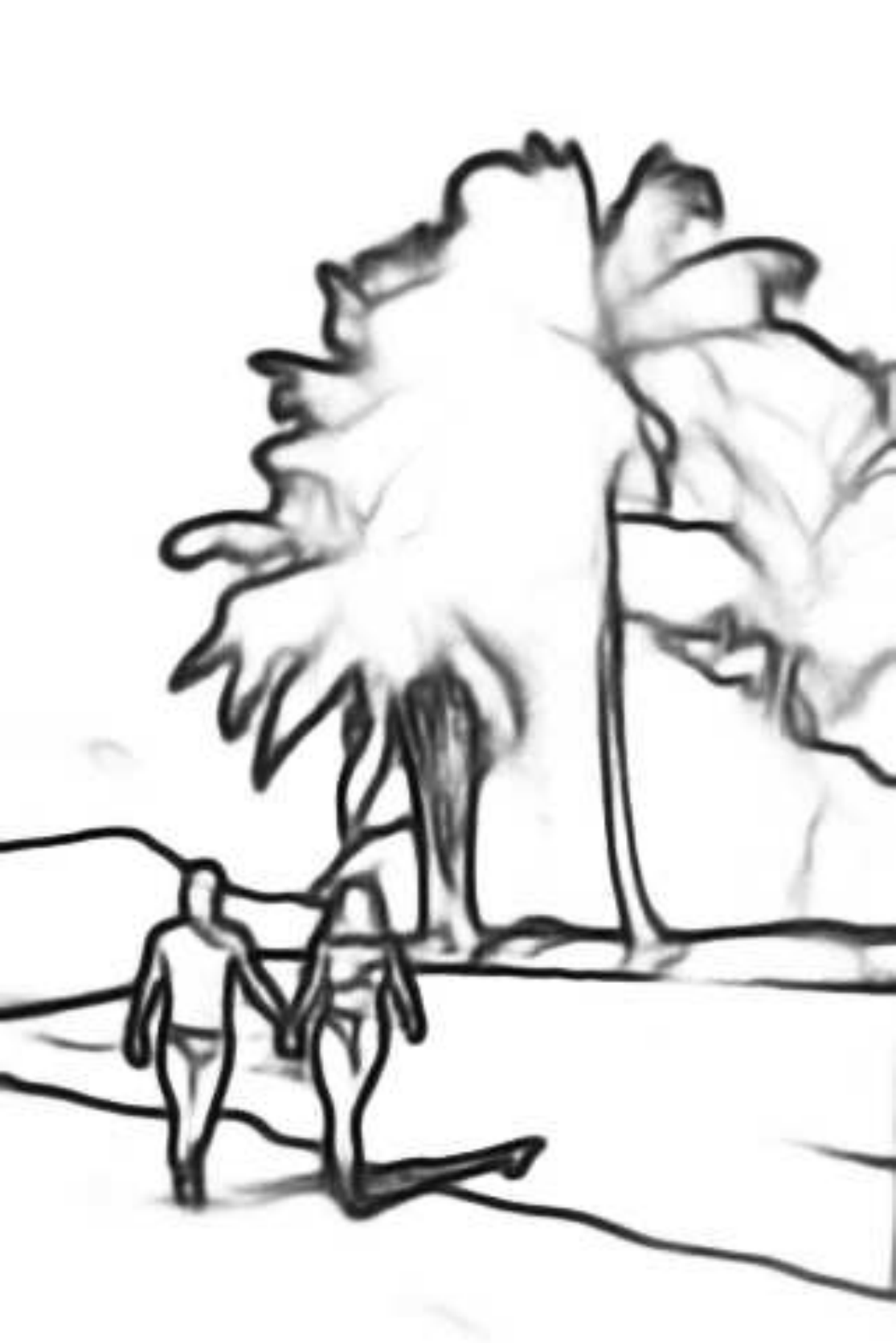} \\
		\includegraphics[width=0.24\linewidth]{bsds/326025-a} &
		\includegraphics[width=0.24\linewidth]{bsds/326025-b} &
		\includegraphics[width=0.24\linewidth]{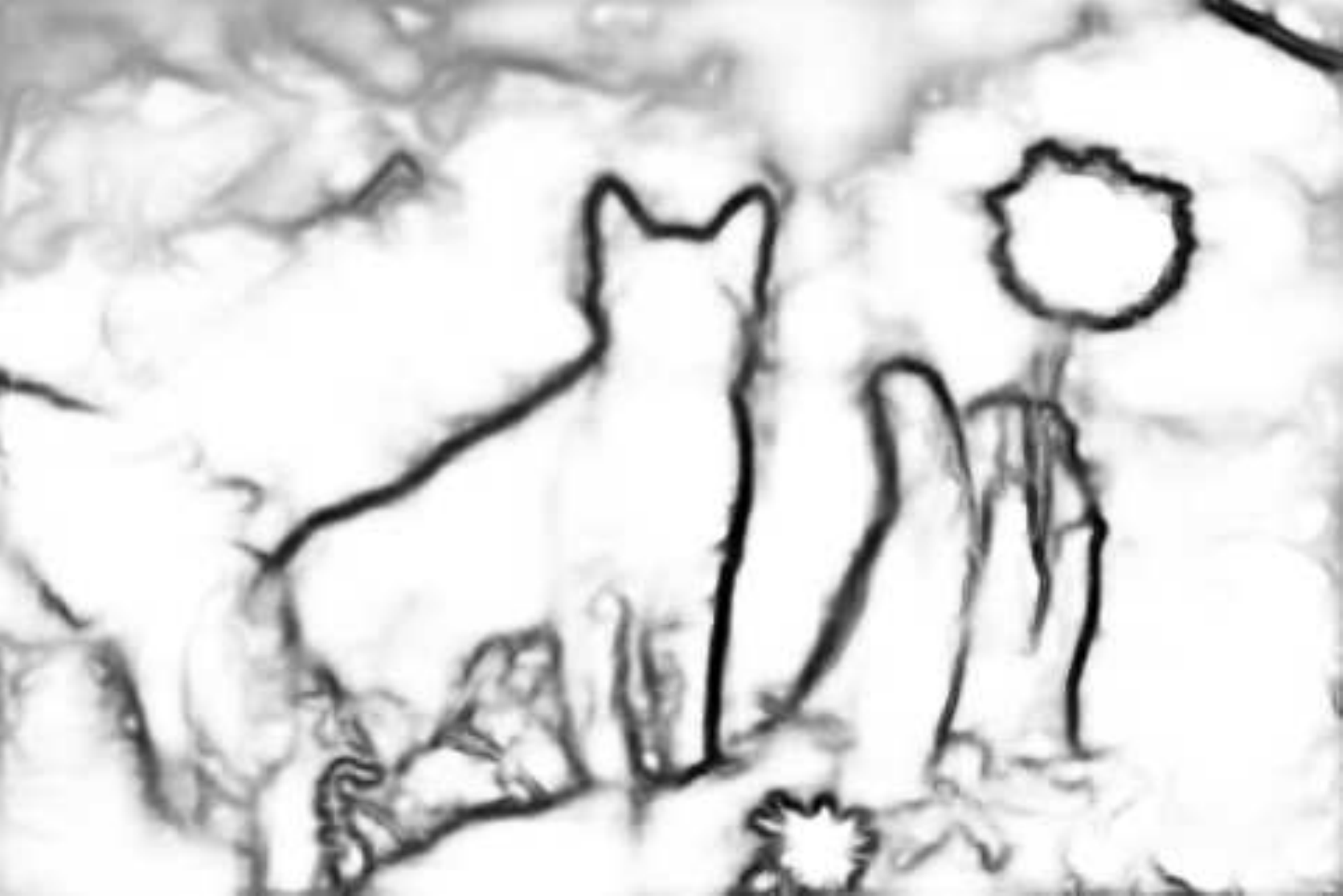} &
		\includegraphics[width=0.24\linewidth]{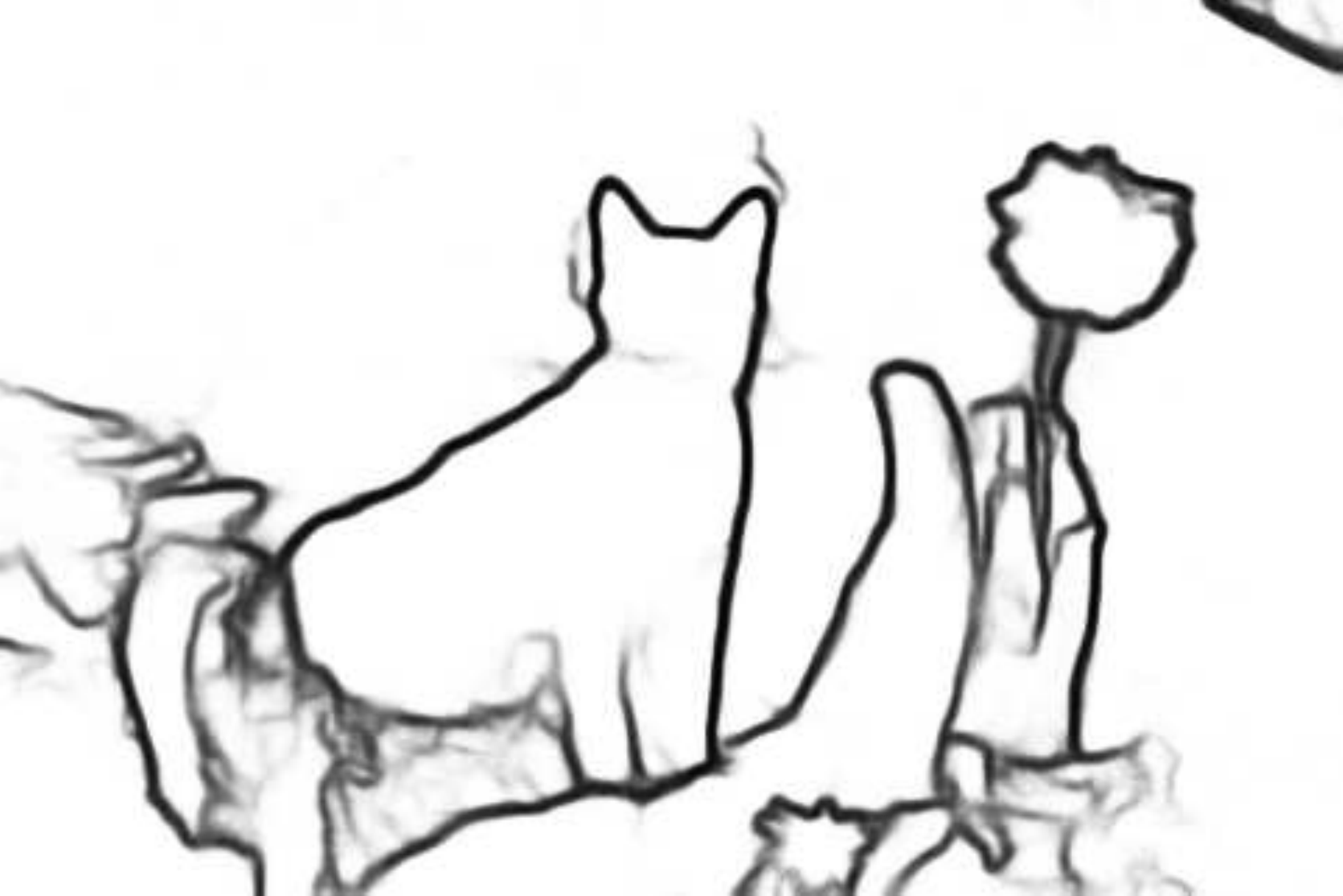} \\
		Raw image & Ground truth & HED-over3 & TD-CEDN-over3 \\[-0.25em]
		& & & (ours)\\
	\end{tabular}
	\caption{Several predictions obtained by ``HED-over3" and ``TD-CEDN-over3 (ours)" models tested on seven samples in the BSDS500 dataset.}
	\label{Fig:ours-hed}
\end{figure}

\begin{figure}[tbh]
	\small
	\centering
	\renewcommand{\tabcolsep}{1pt}
	\begin{tabular}{cccc}
		\includegraphics[width=0.24\linewidth]{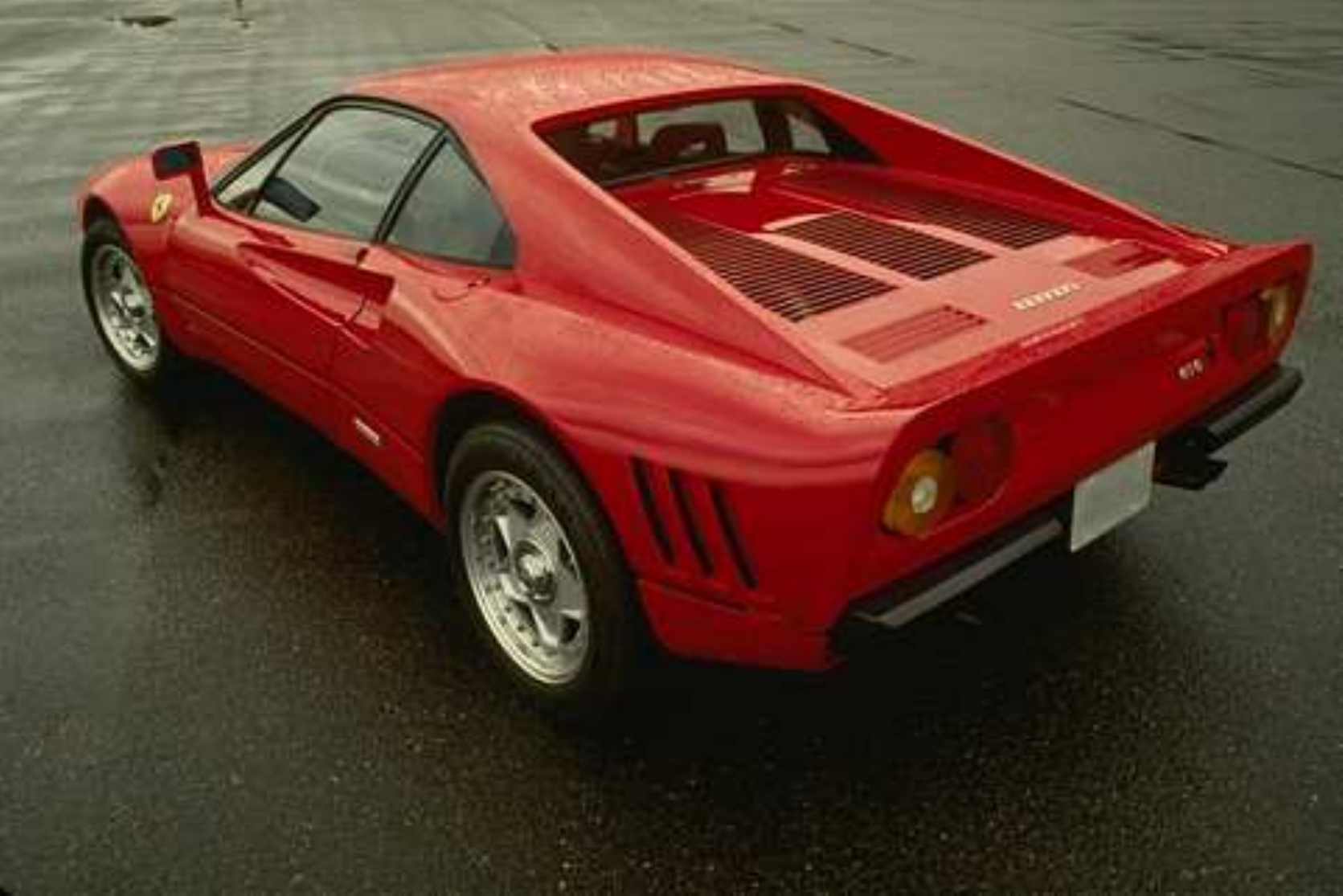} &
		\includegraphics[width=0.24\linewidth]{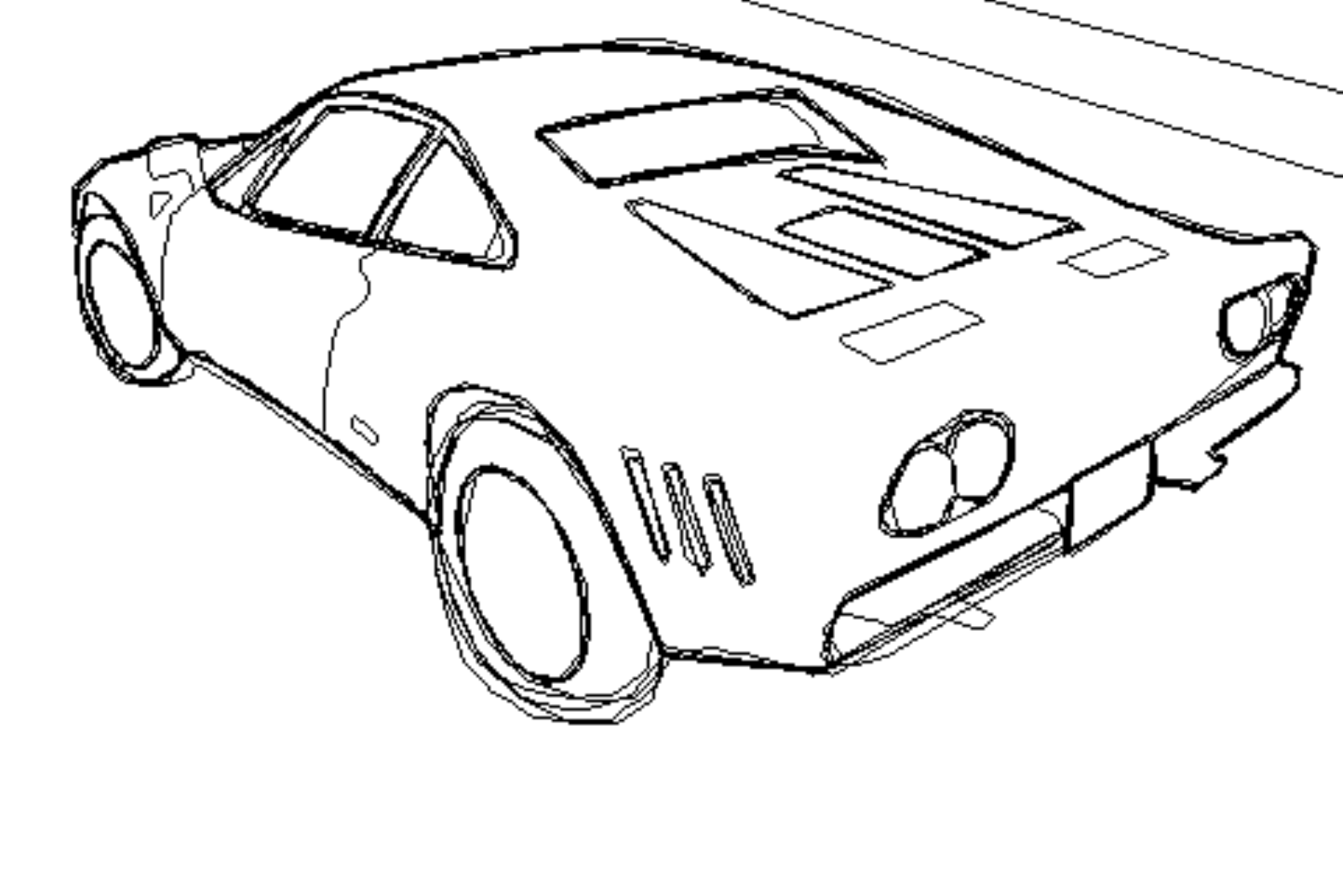} &
		\includegraphics[width=0.24\linewidth]{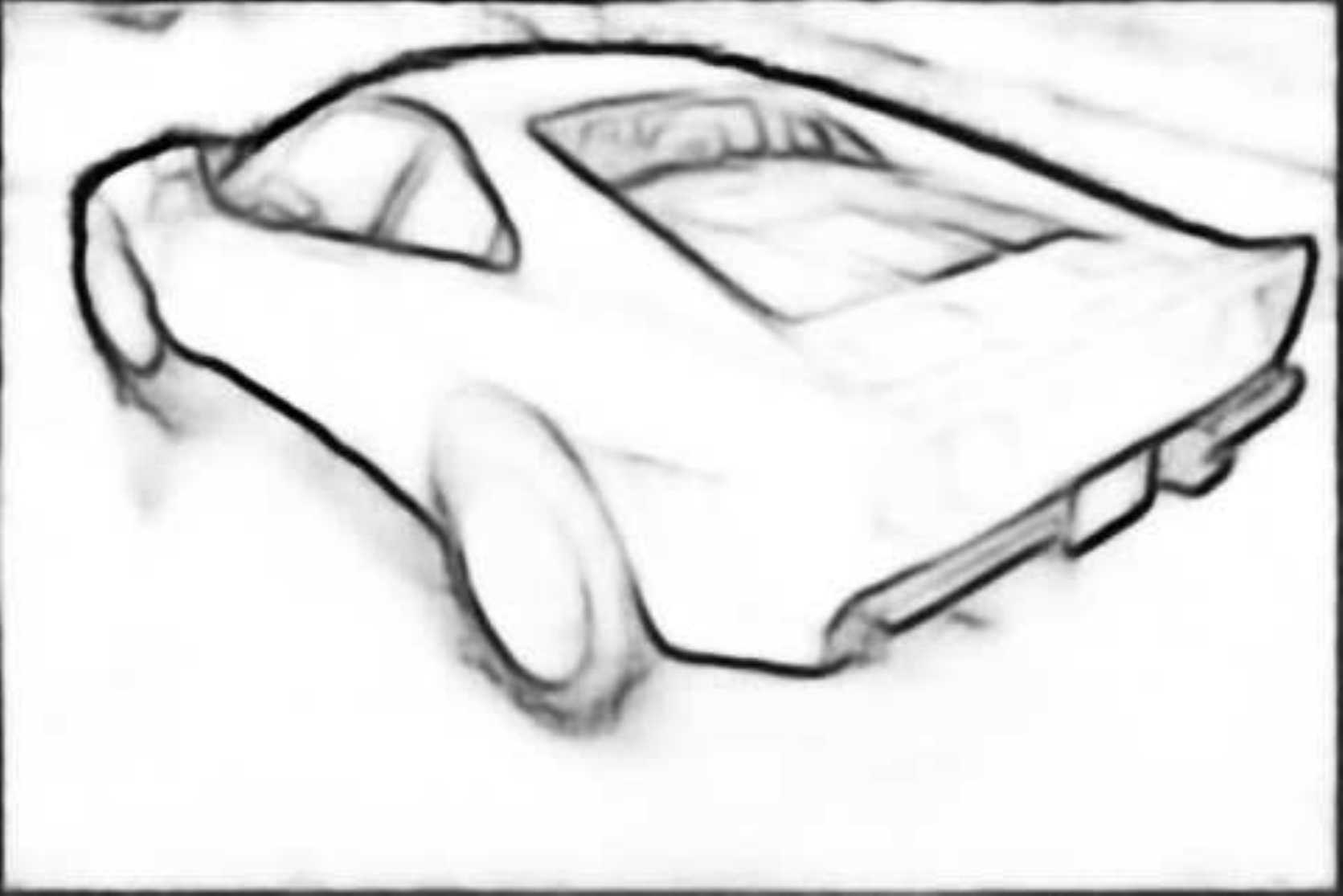} &
		\includegraphics[width=0.24\linewidth]{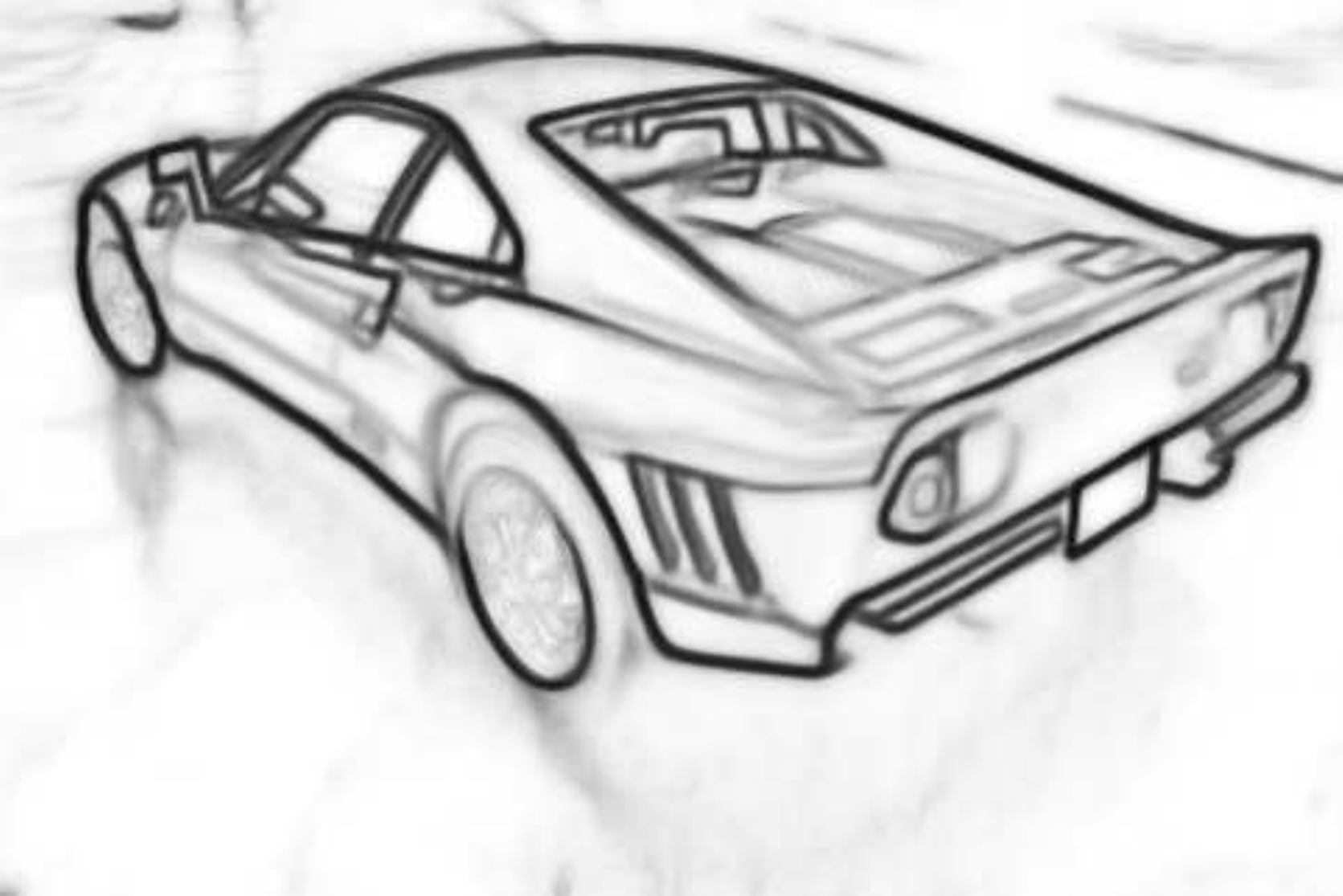} \\
		\includegraphics[width=0.24\linewidth]{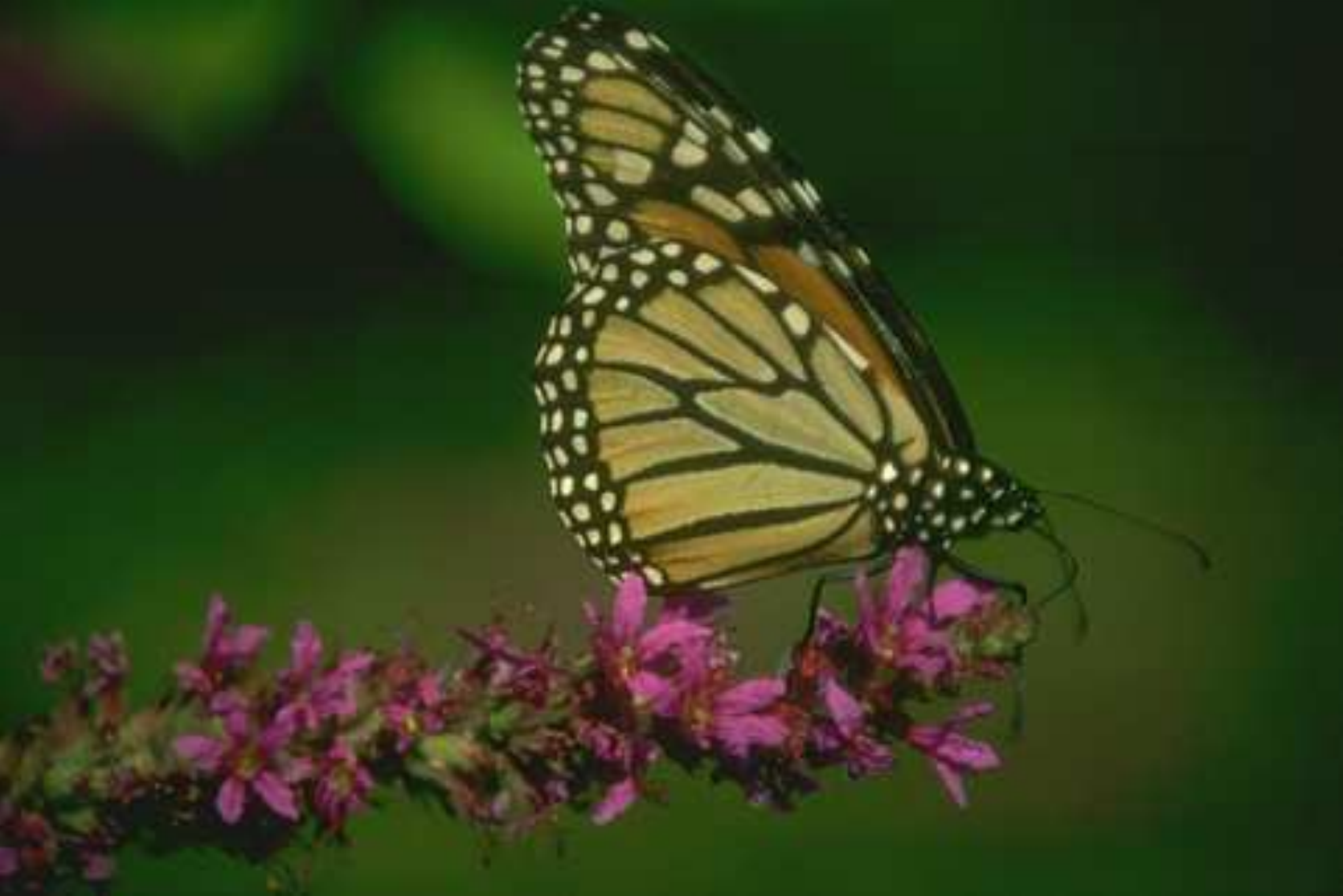} &
		\includegraphics[width=0.24\linewidth]{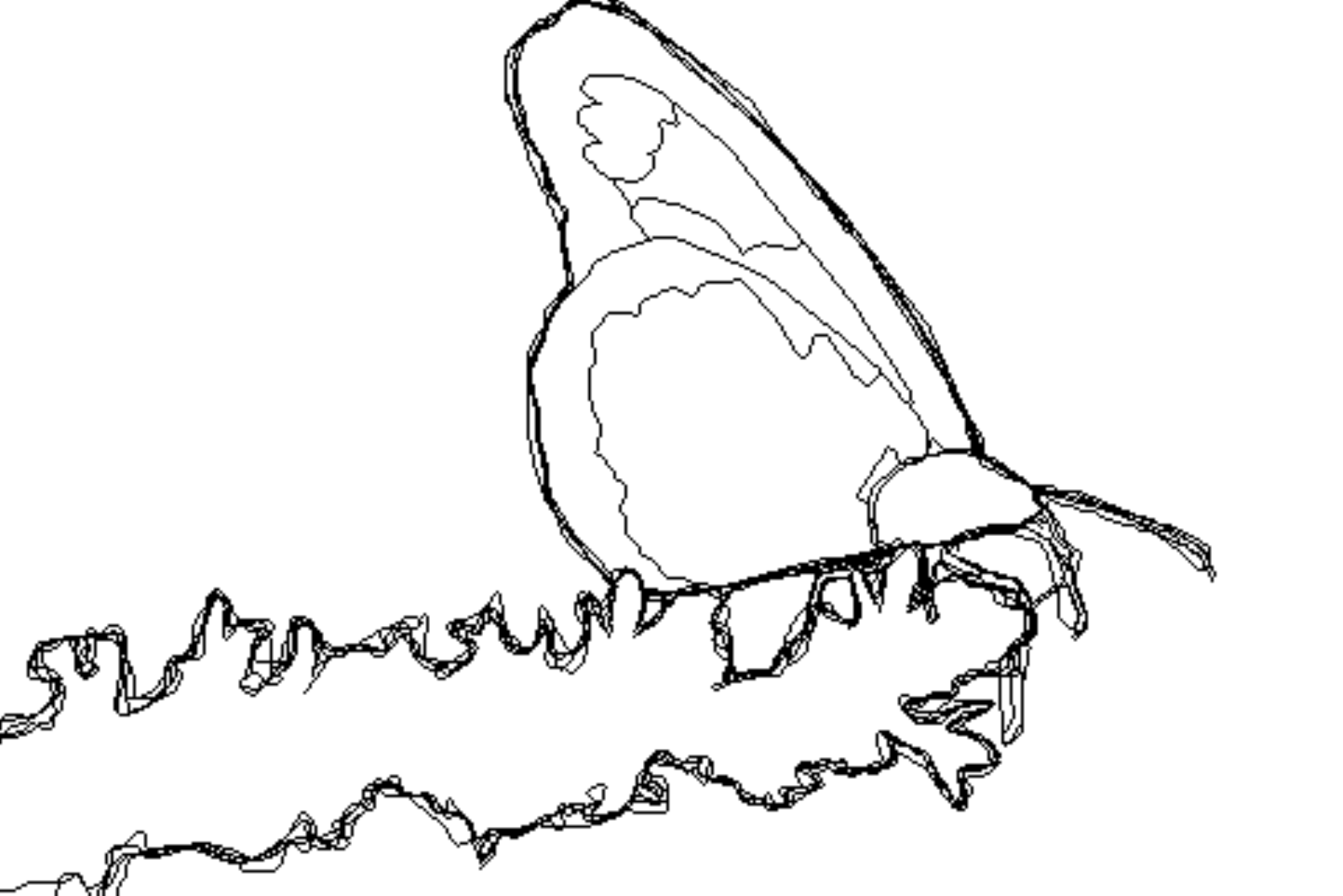} &
		\includegraphics[width=0.24\linewidth]{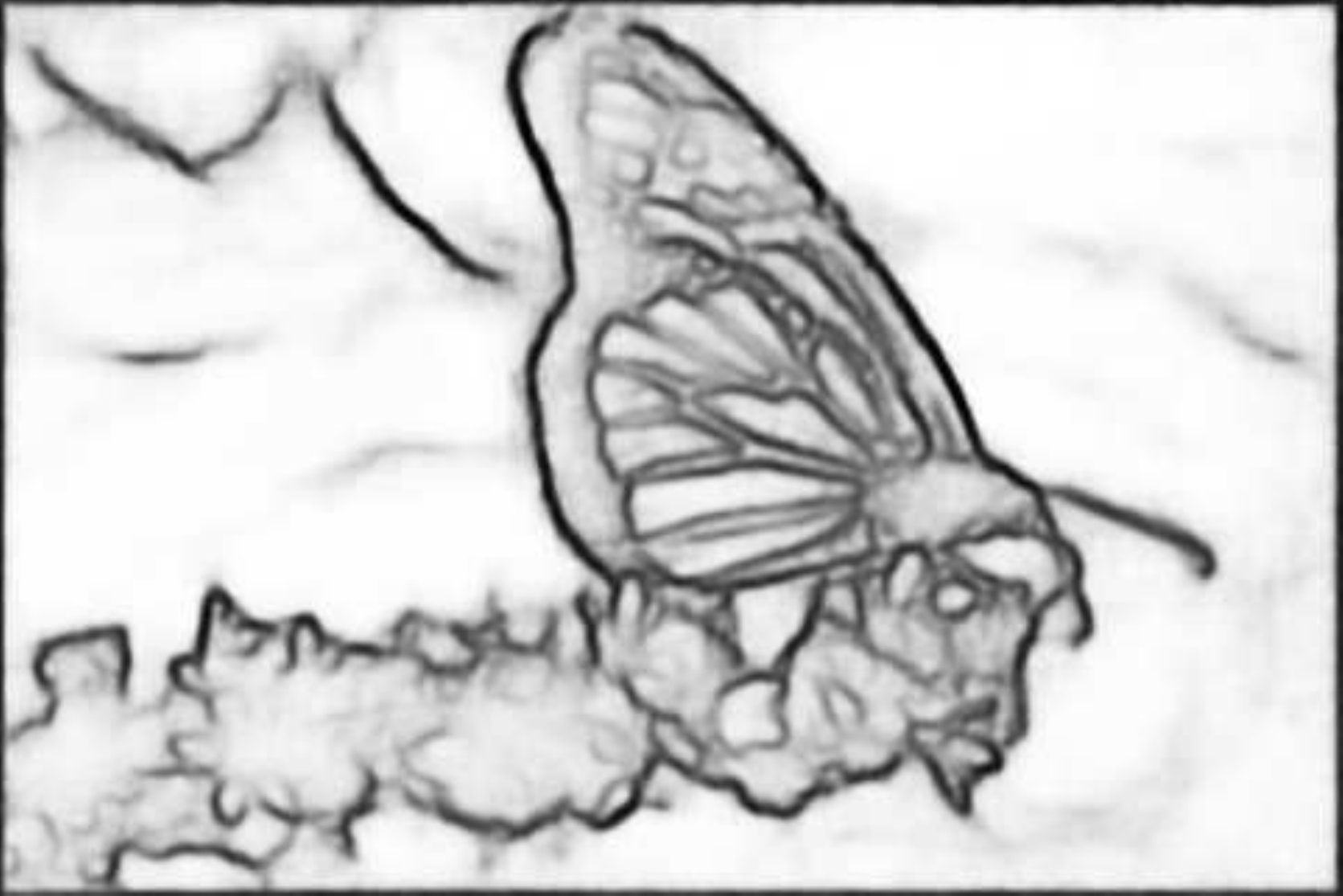} &
		\includegraphics[width=0.24\linewidth]{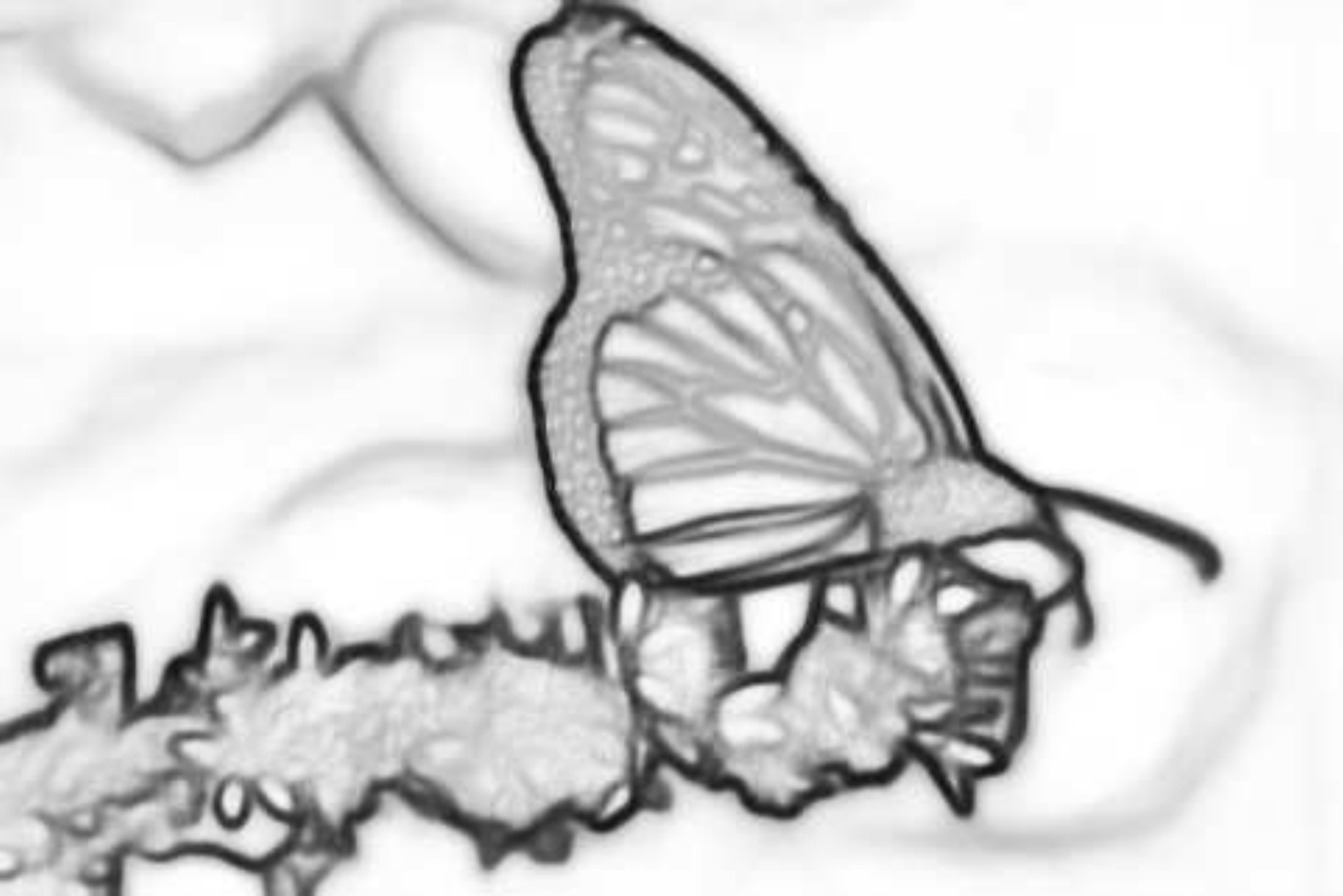} \\
		\includegraphics[width=0.24\linewidth]{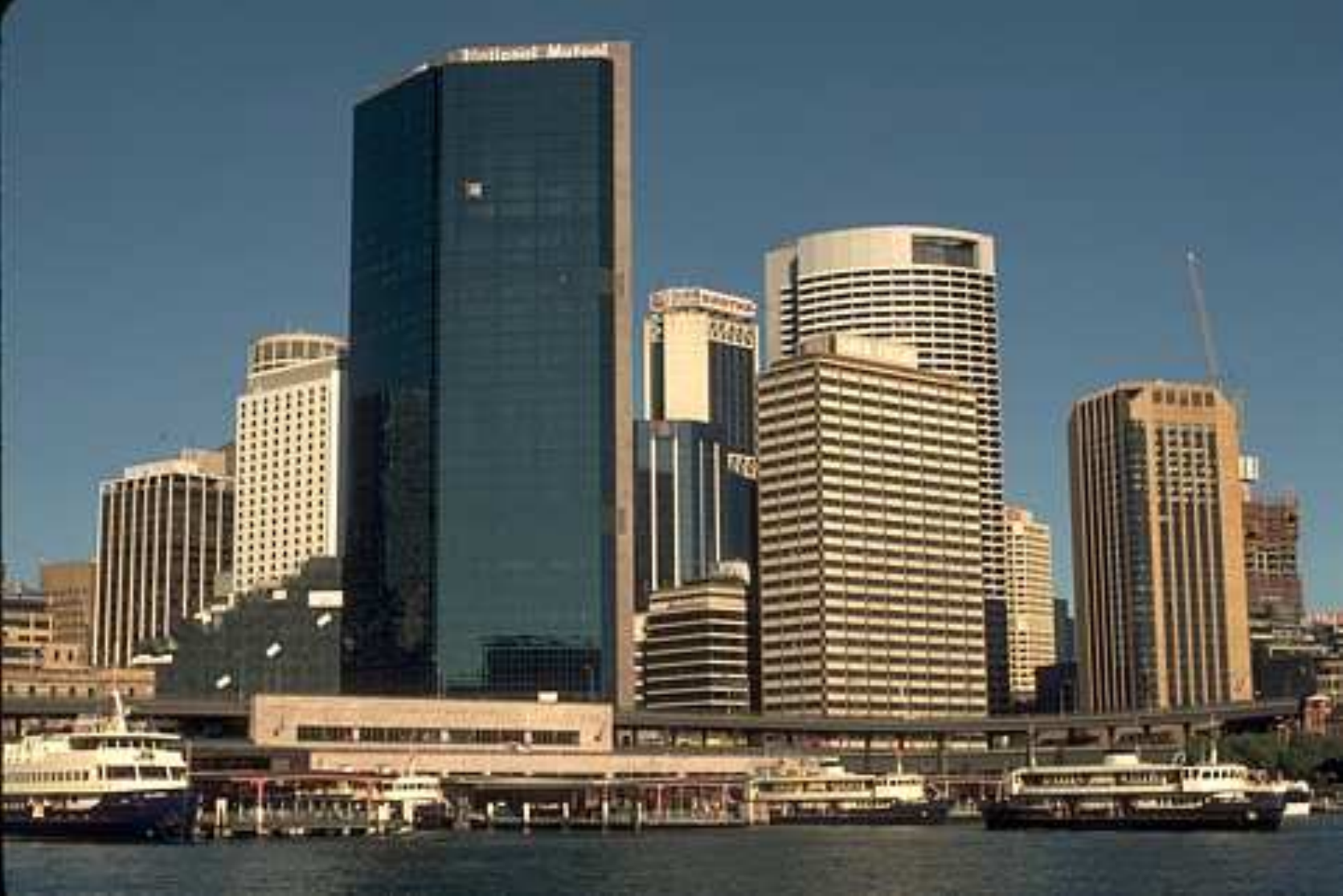} &
		\includegraphics[width=0.24\linewidth]{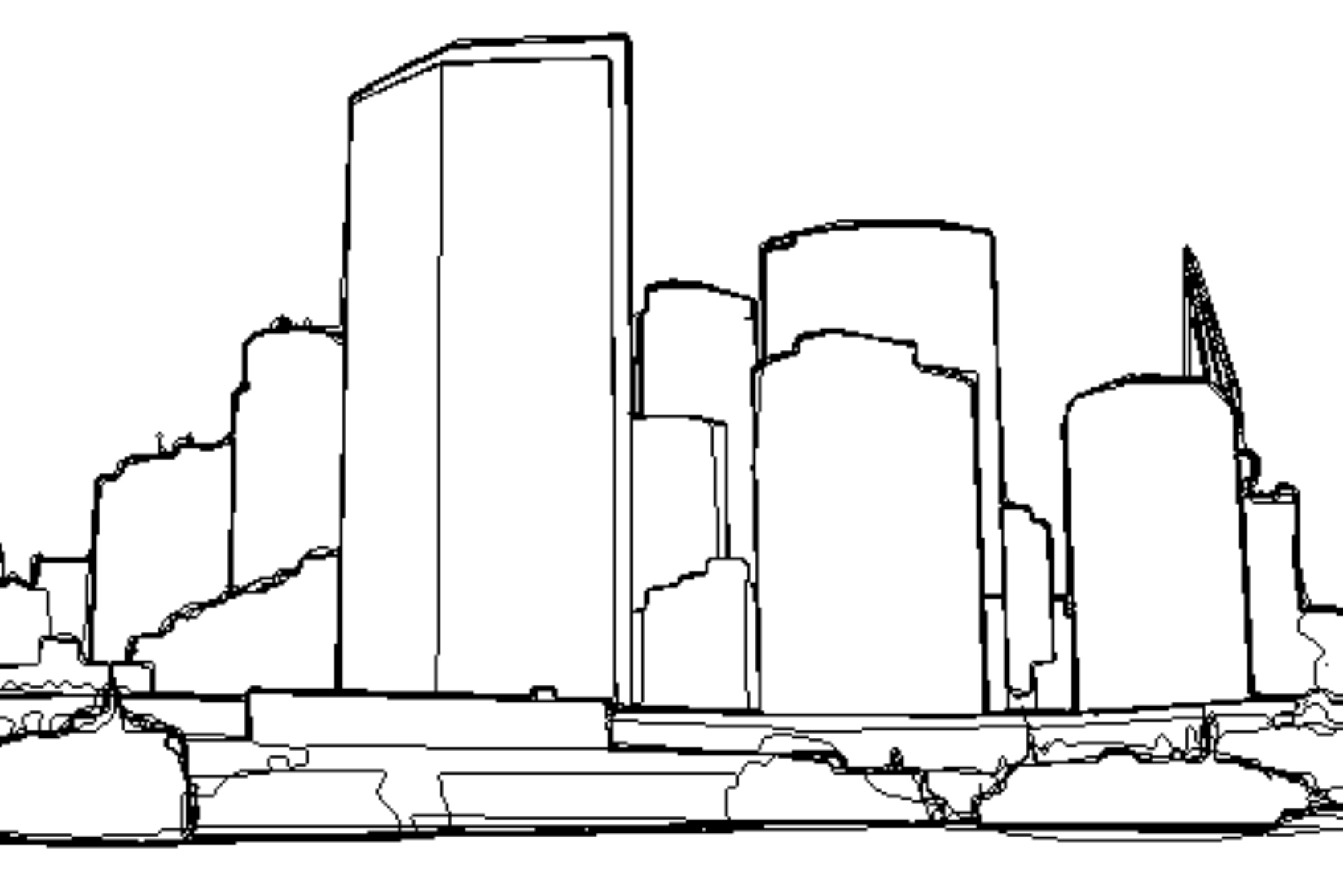} &
		\includegraphics[width=0.24\linewidth]{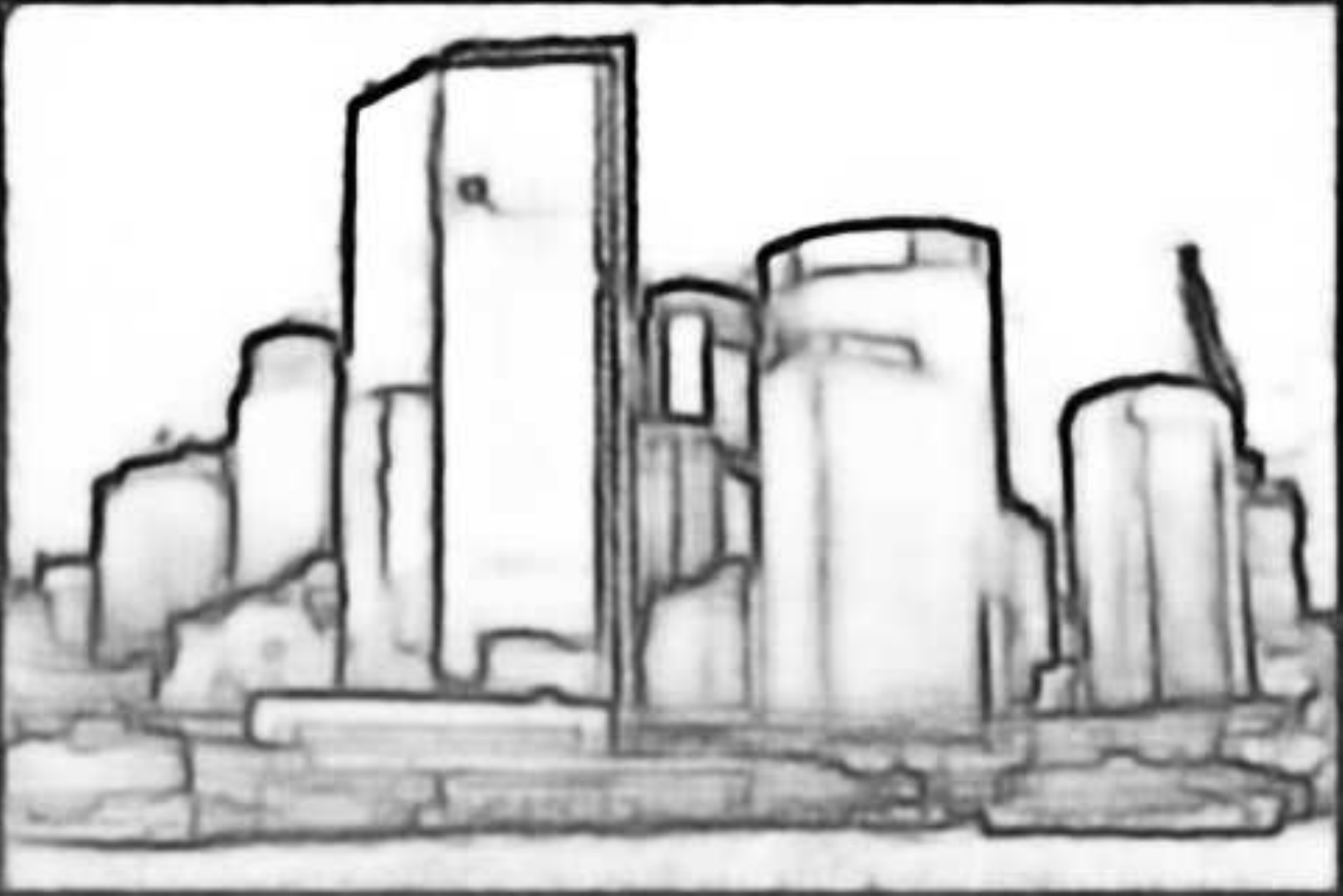} &
		\includegraphics[width=0.24\linewidth]{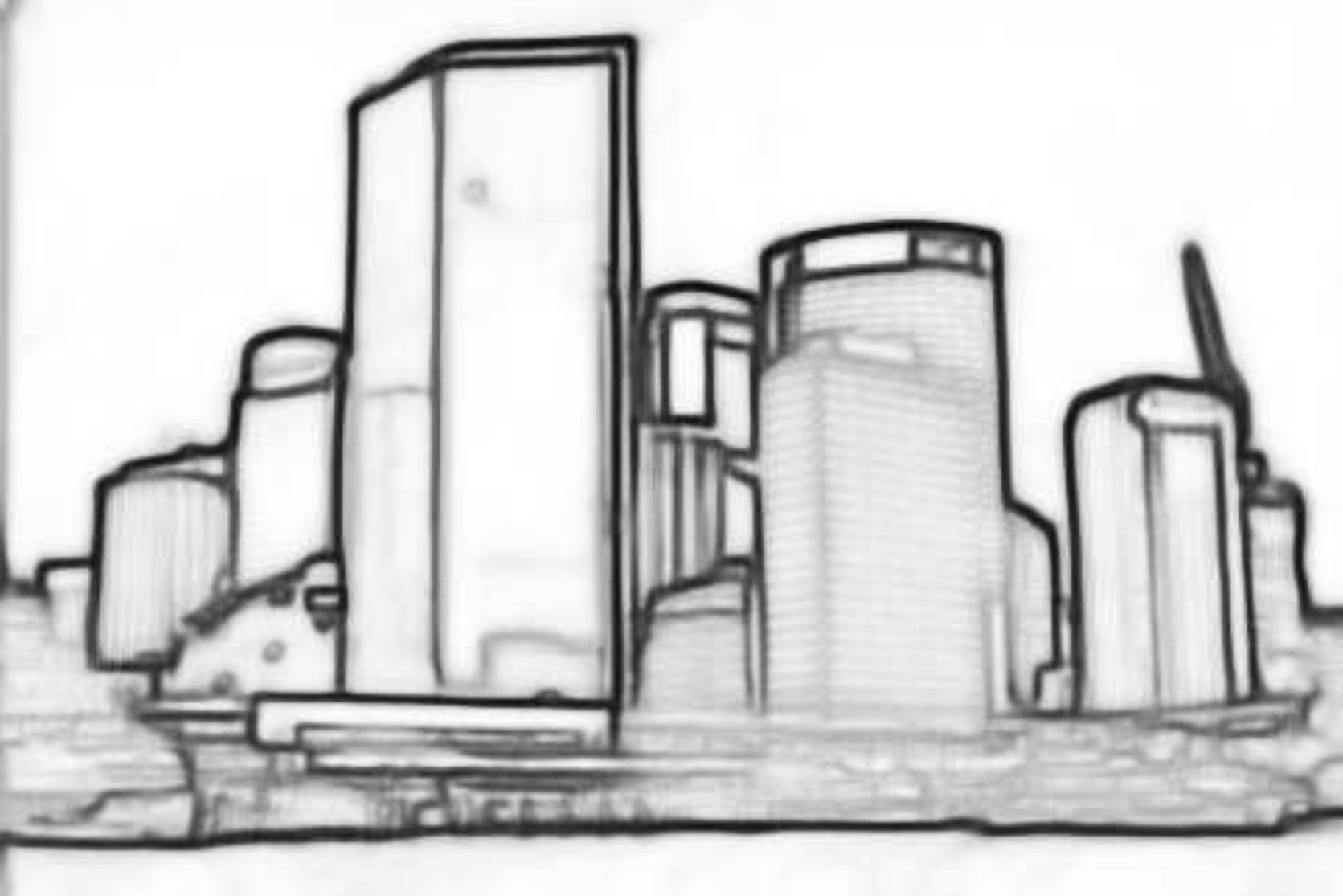} \\
		\includegraphics[width=0.24\linewidth]{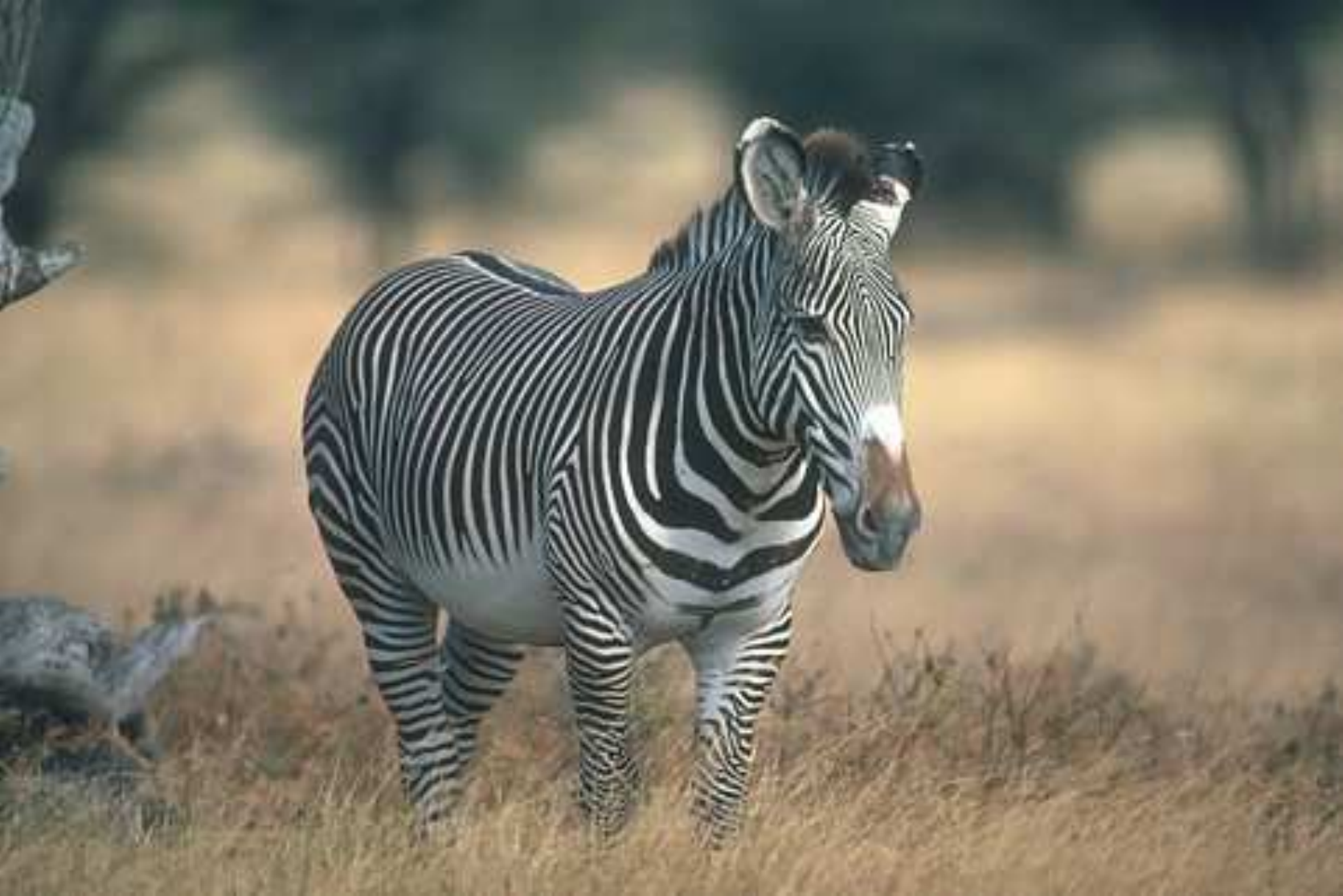} &
		\includegraphics[width=0.24\linewidth]{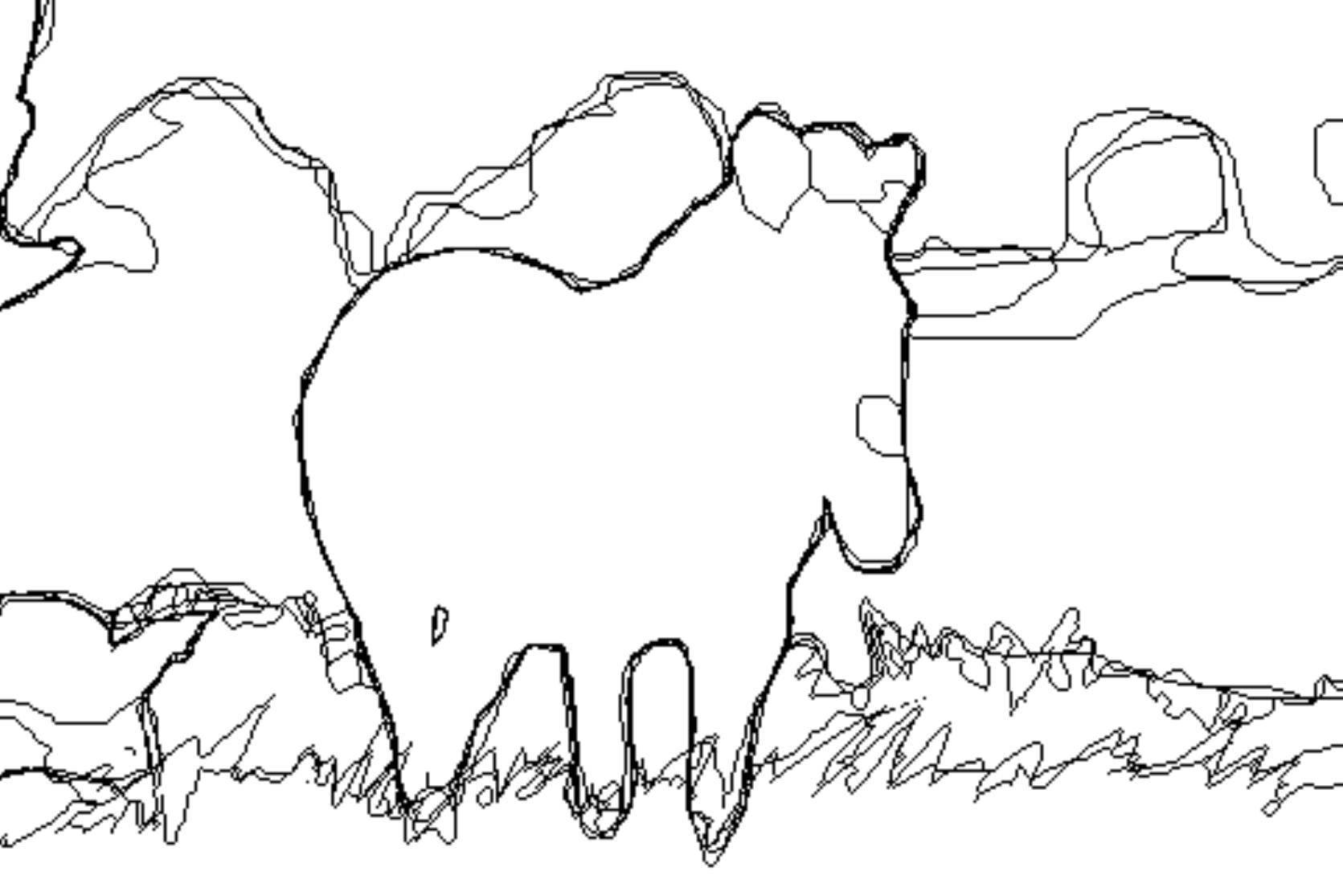} &
		\includegraphics[width=0.24\linewidth]{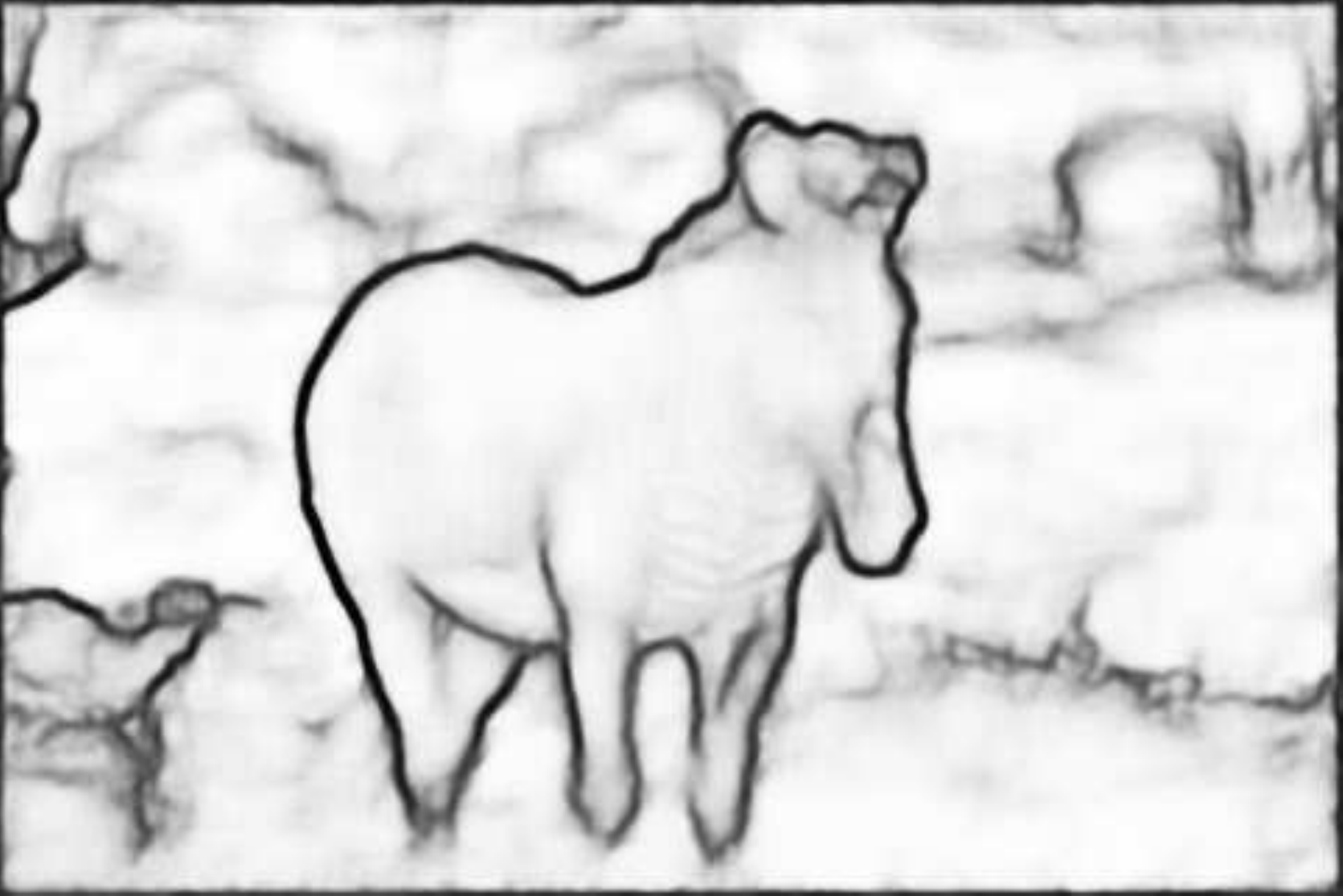} &
		\includegraphics[width=0.24\linewidth]{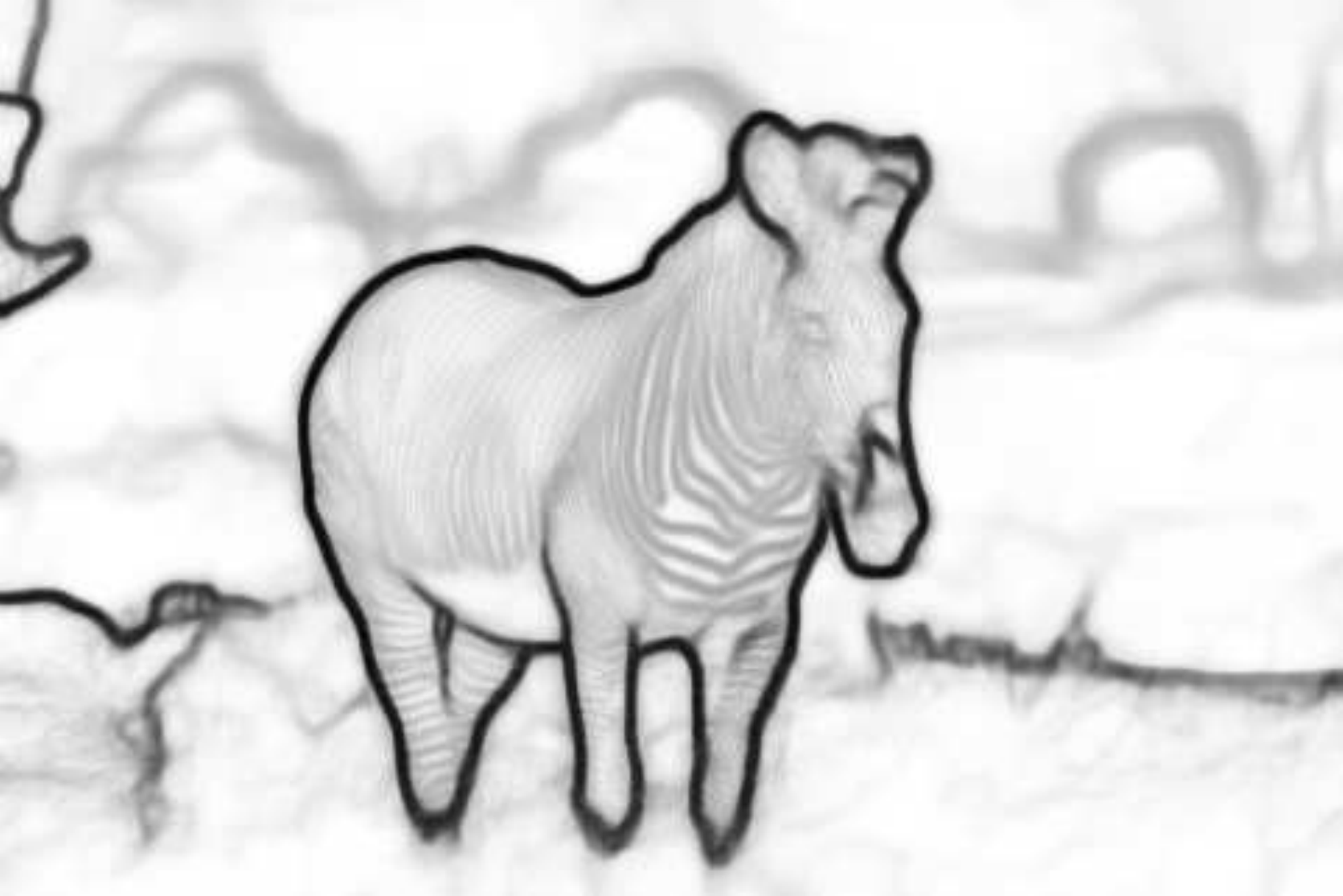} \\
		\includegraphics[width=0.24\linewidth]{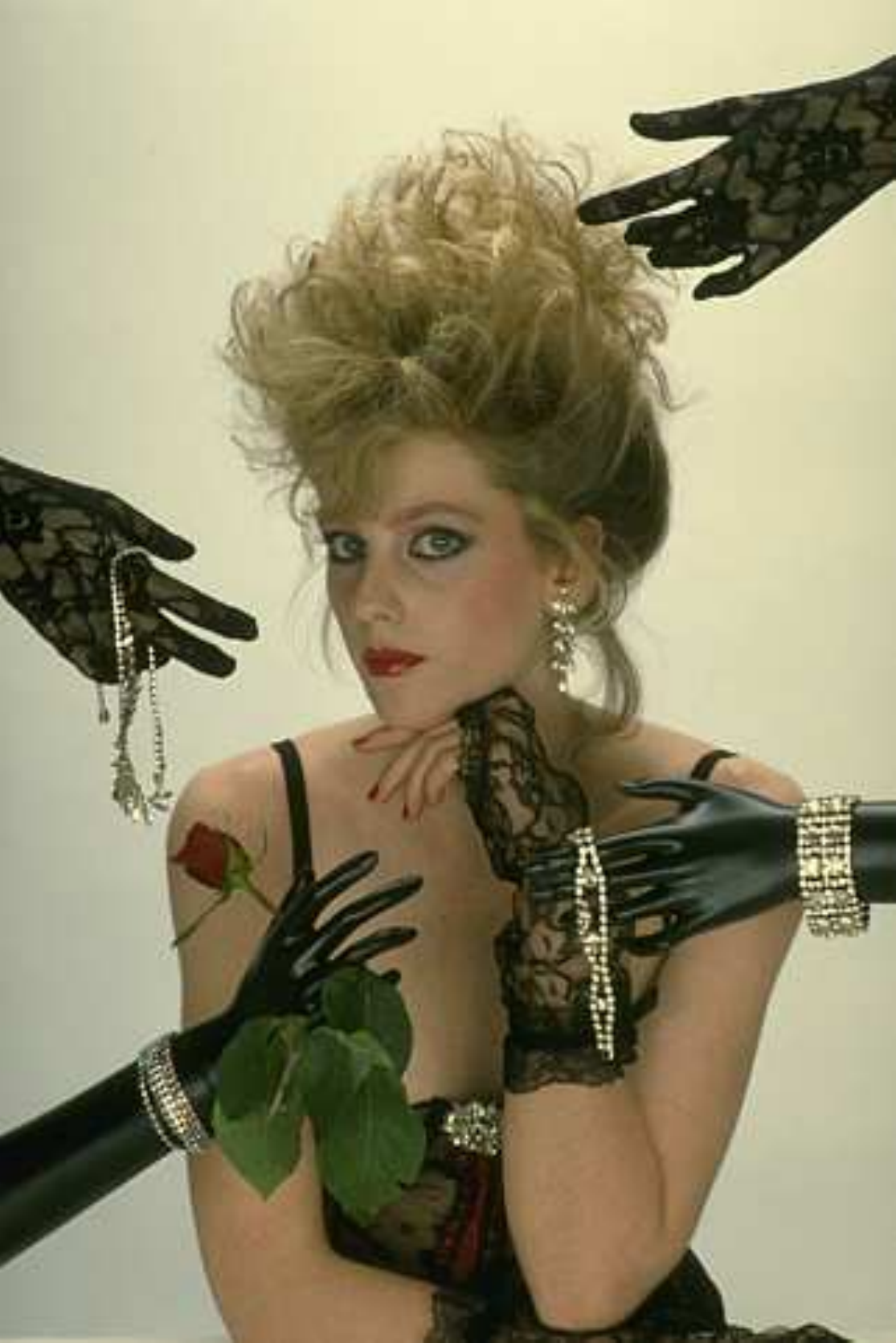} &
		\includegraphics[width=0.24\linewidth]{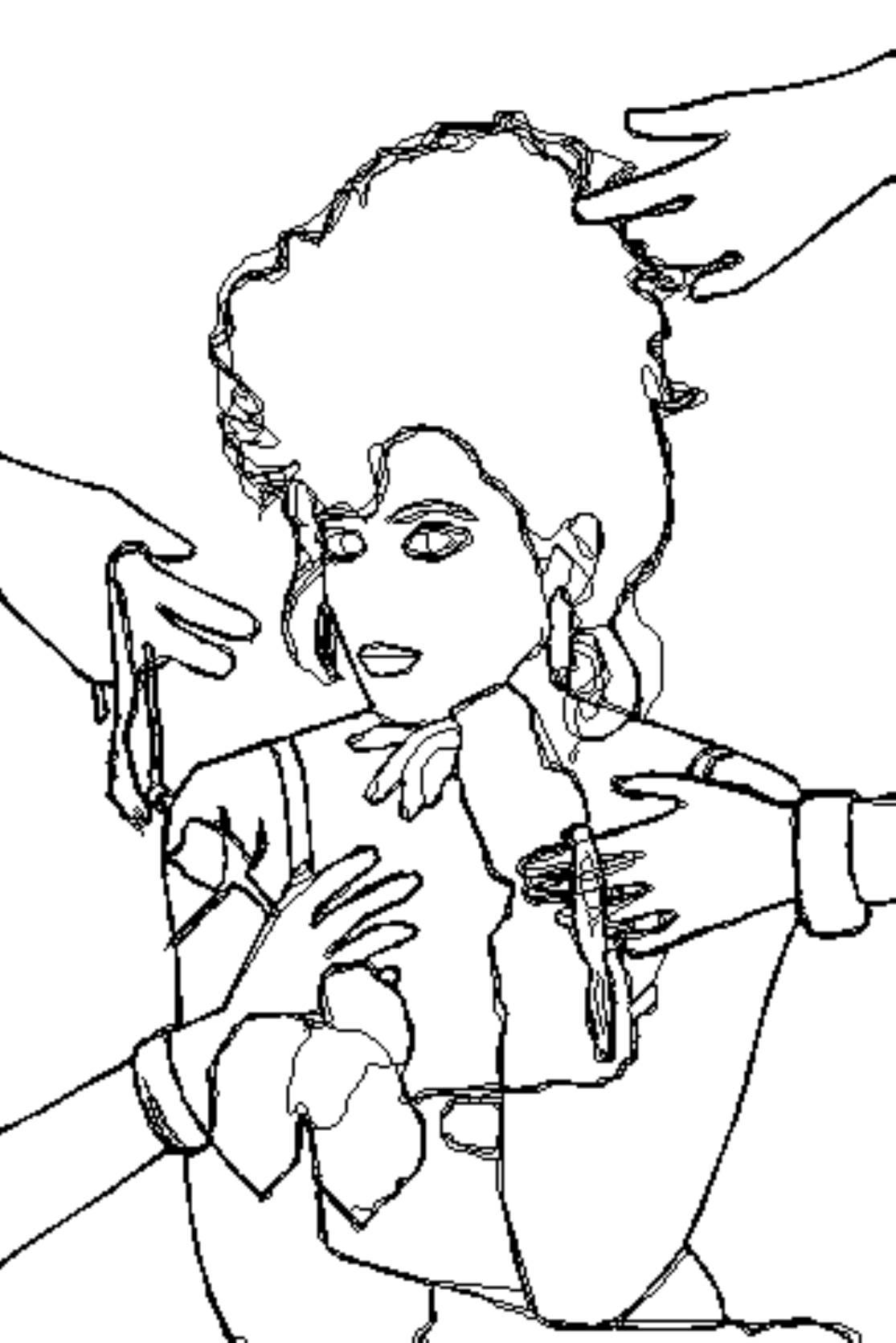} &
		\includegraphics[width=0.24\linewidth]{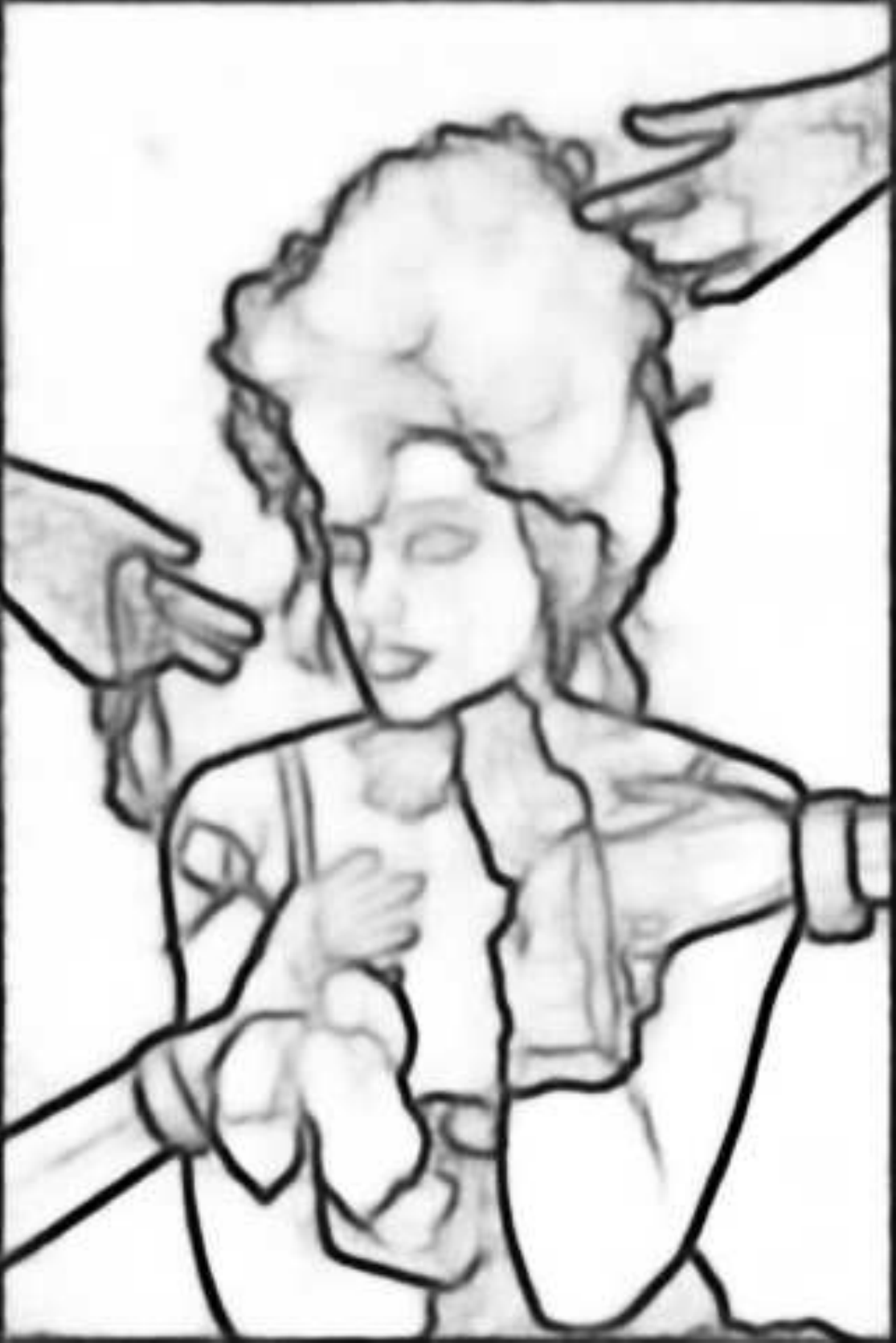} &
		\includegraphics[width=0.24\linewidth]{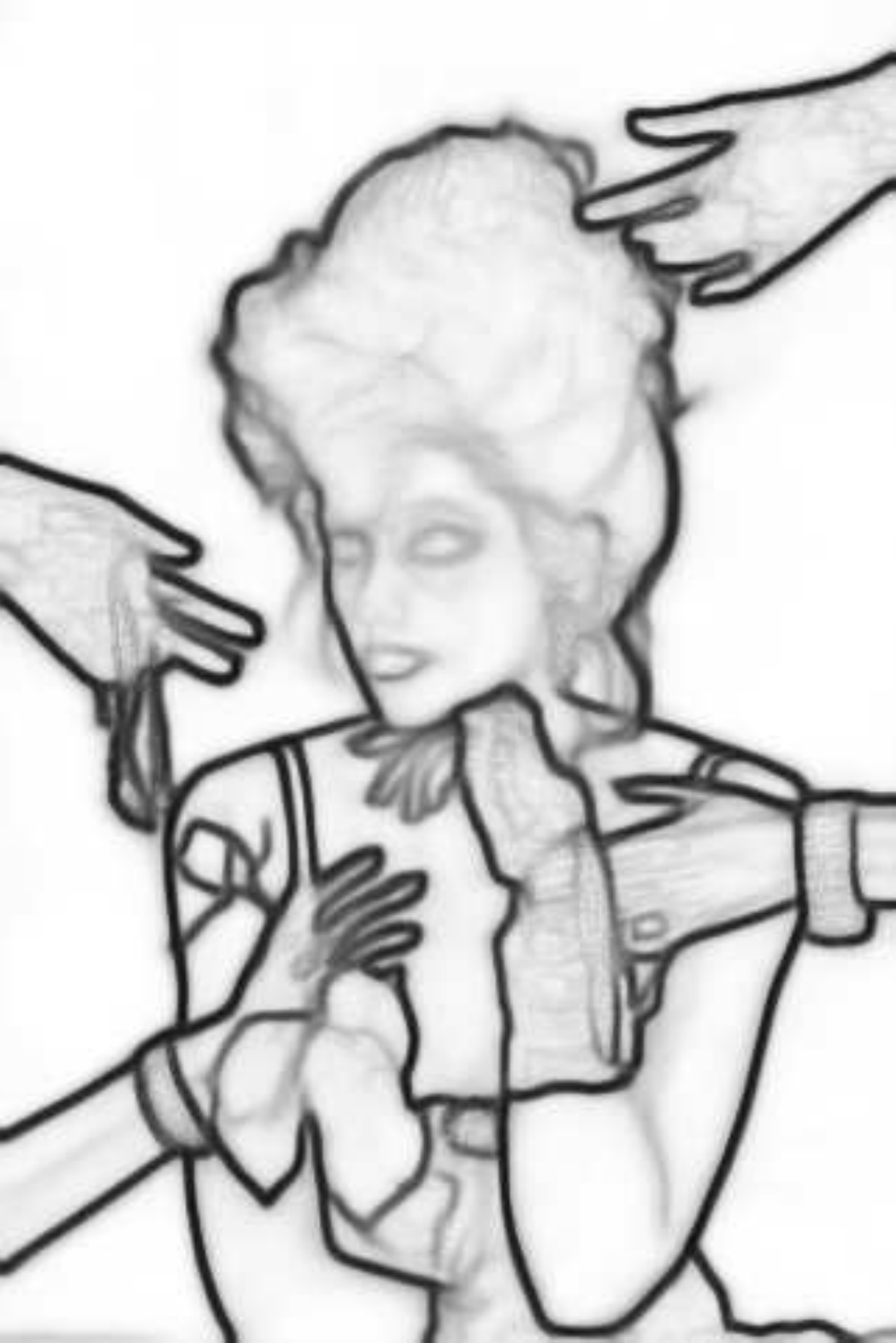} \\
		\includegraphics[width=0.24\linewidth]{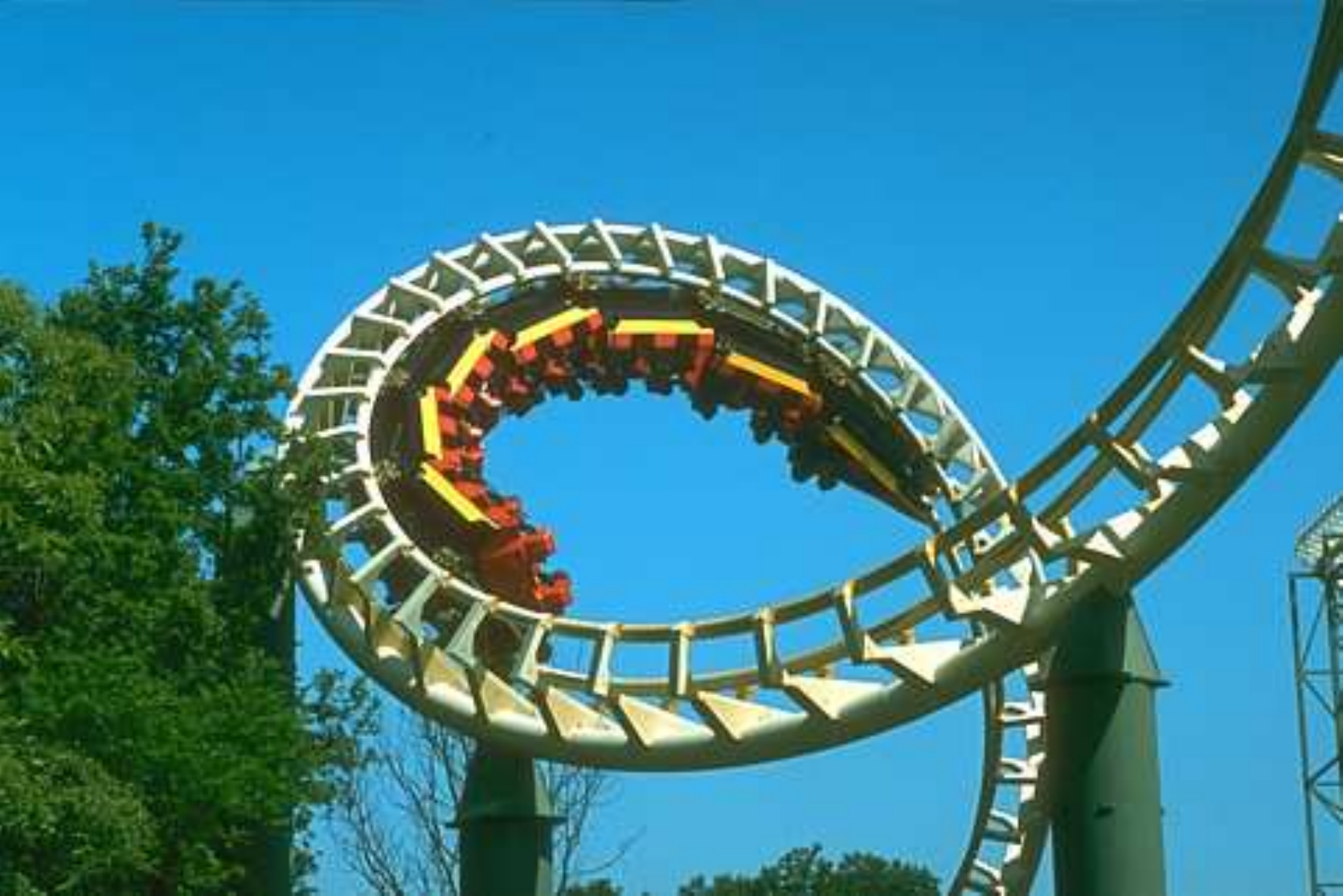} &
    	\includegraphics[width=0.24\linewidth]{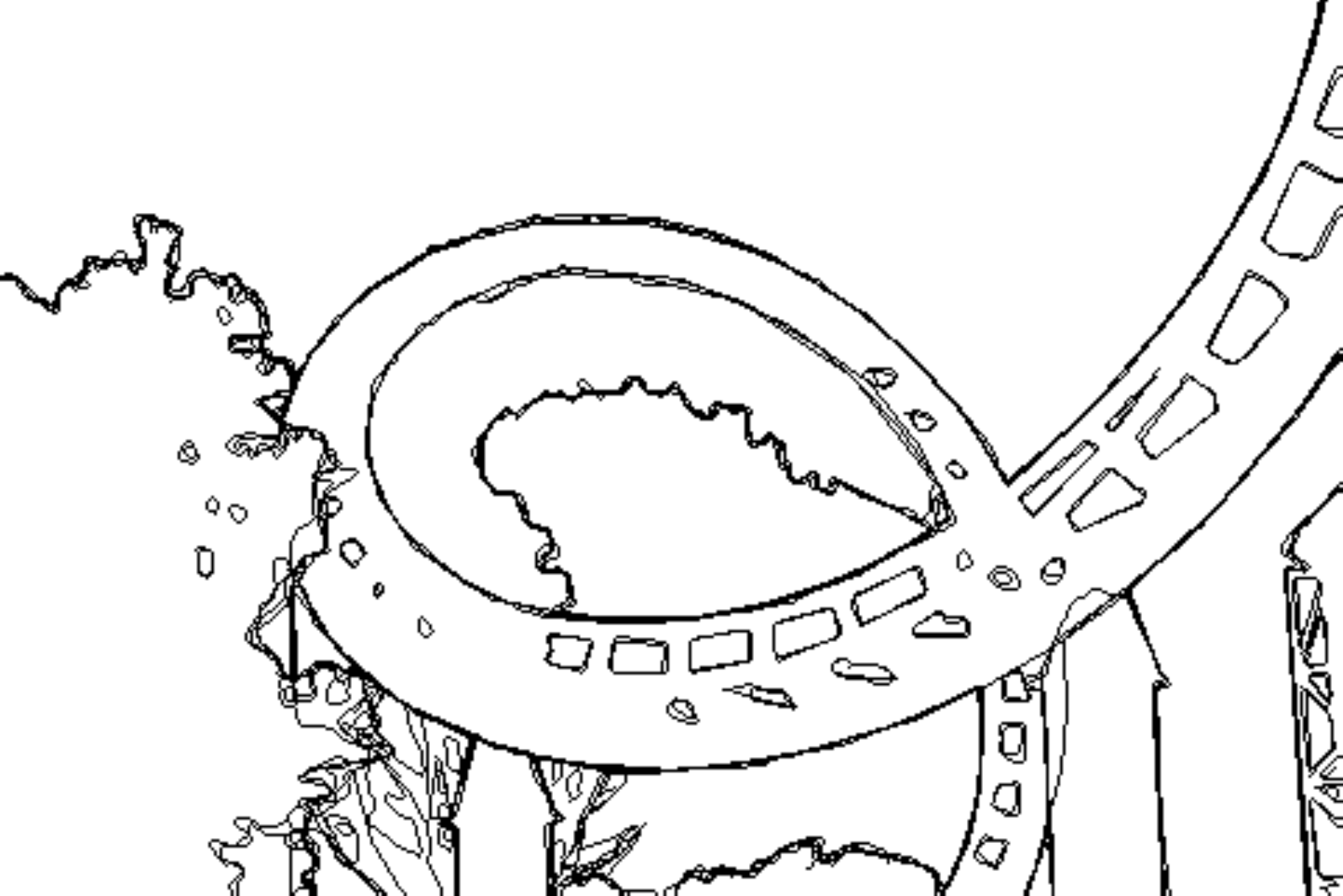} &
		\includegraphics[width=0.24\linewidth]{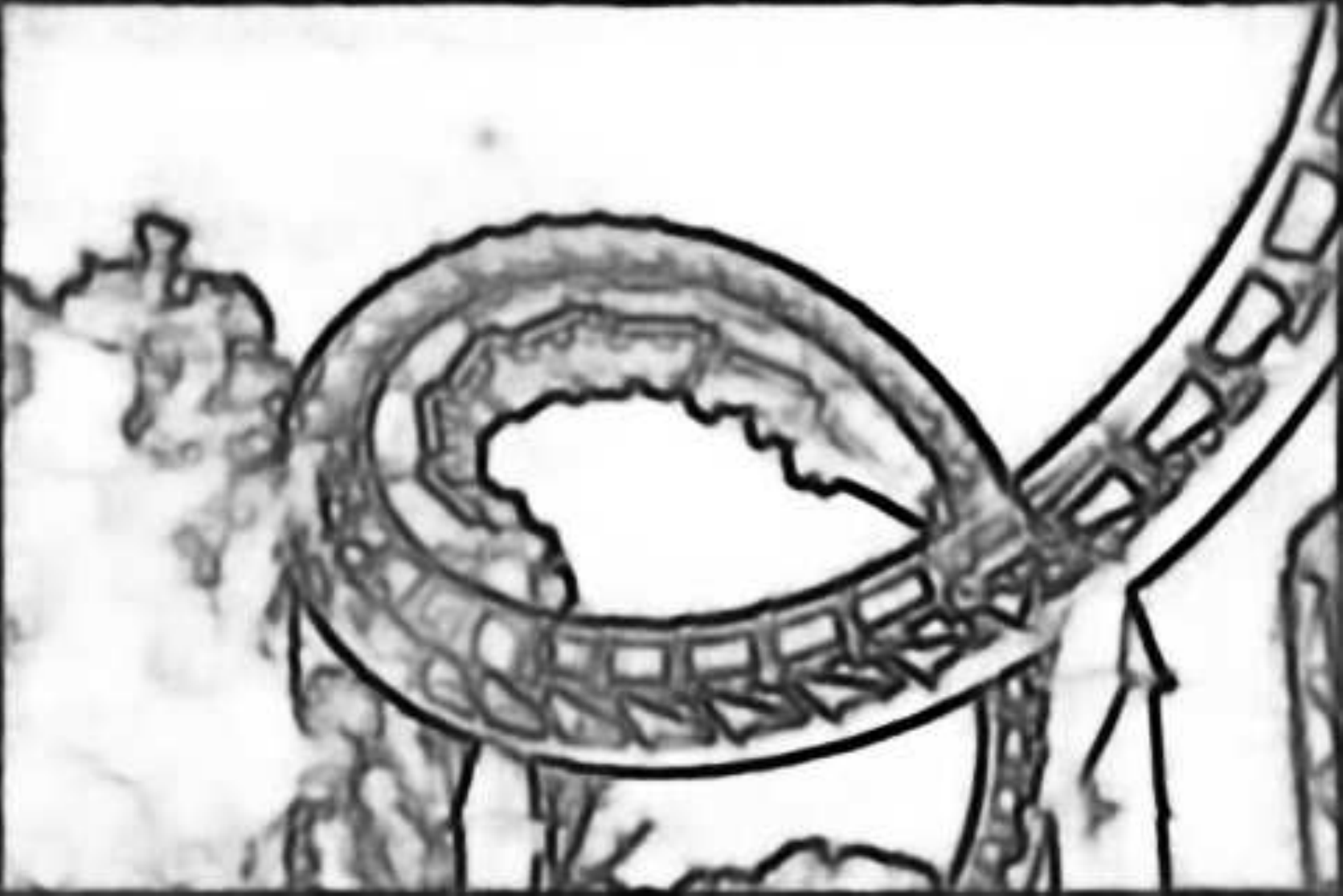} &
		\includegraphics[width=0.24\linewidth]{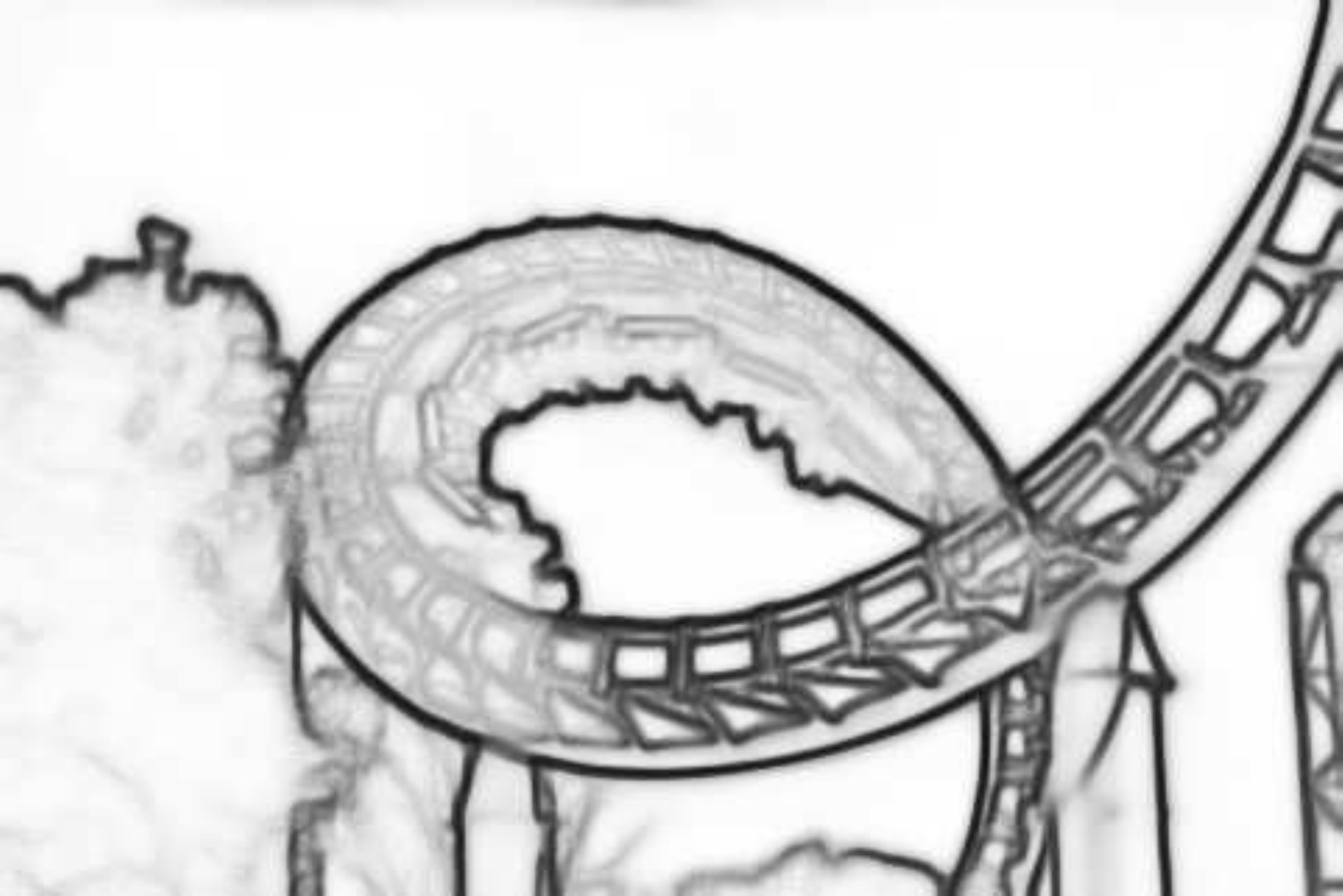} \\
		Raw image & Ground truth & CEDN & TD-CEDN (ours) \\
	\end{tabular}
	\caption{Several predictions obtained by ``CEDN" and ``TD-CEDN (ours)" models tested on seven samples in the BSDS500 dataset.}
	\label{Fig:ours-cedn}
\end{figure}

\noindent \textbf{PASCAL VOC 2012:} The PASCAL VOC dataset~\cite{everingham2010pascal} is a widely-used benchmark with high-quality annotations for object detection and segmentation. The VOC 2012 release includes 11530 images for 20 classes covering a series of common object categories, such as ``person", ``animal", ``vehicle" and ``indoor".  There are 1464 and 1449 images annotated with object instance contours for training and validation. With the further contribution of Hariharan \etal~\cite{hariharan2011semantic}, we can get 10528 and 1449 images for training and validation. Yang \etal~\cite{yang2016object} has cleaned up the dataset and applied it to evaluate the performances of object contour detection. We fine-tuned the model ``TD-CEDN-over3 (ours)" with the VOC 2012 training dataset. The same measurements applied on the BSDS500 dataset were evaluated. 

\begin{figure}[tbh]
	\centering
	\includegraphics[width=\linewidth]{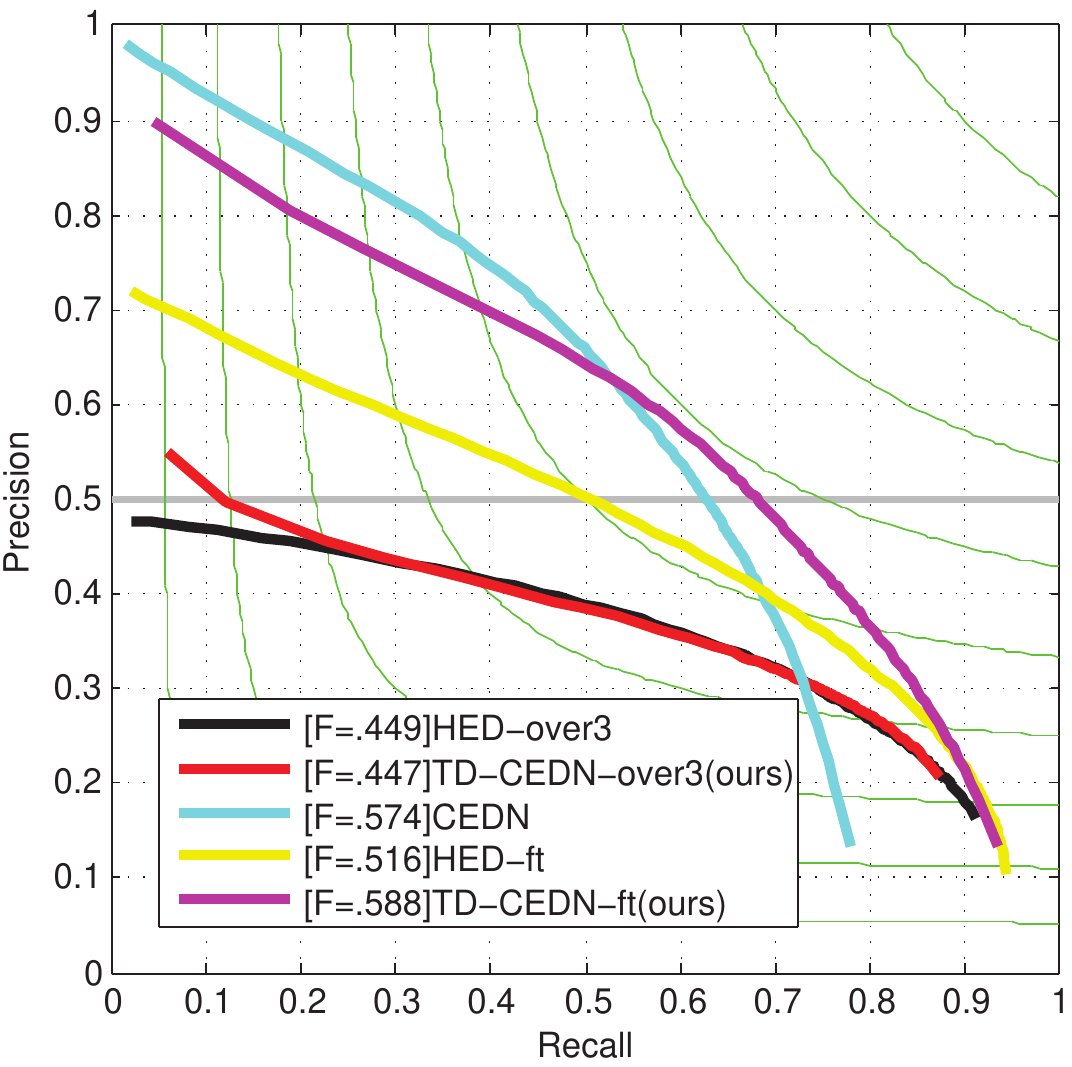}\\
	\centering
	\caption{The PR curve for contour detection on the PASCAL VOC 2012 validation dataset. Our fine-tuned model achieved the best ODS F-score of 0.588.}
	\label{Fig:voc12}
\end{figure}

\begin{figure*}[!t]
	\small
	\centering
	\renewcommand{\tabcolsep}{2pt}
	\begin{tabular}{ccccc}
		\includegraphics[width=0.19\linewidth]{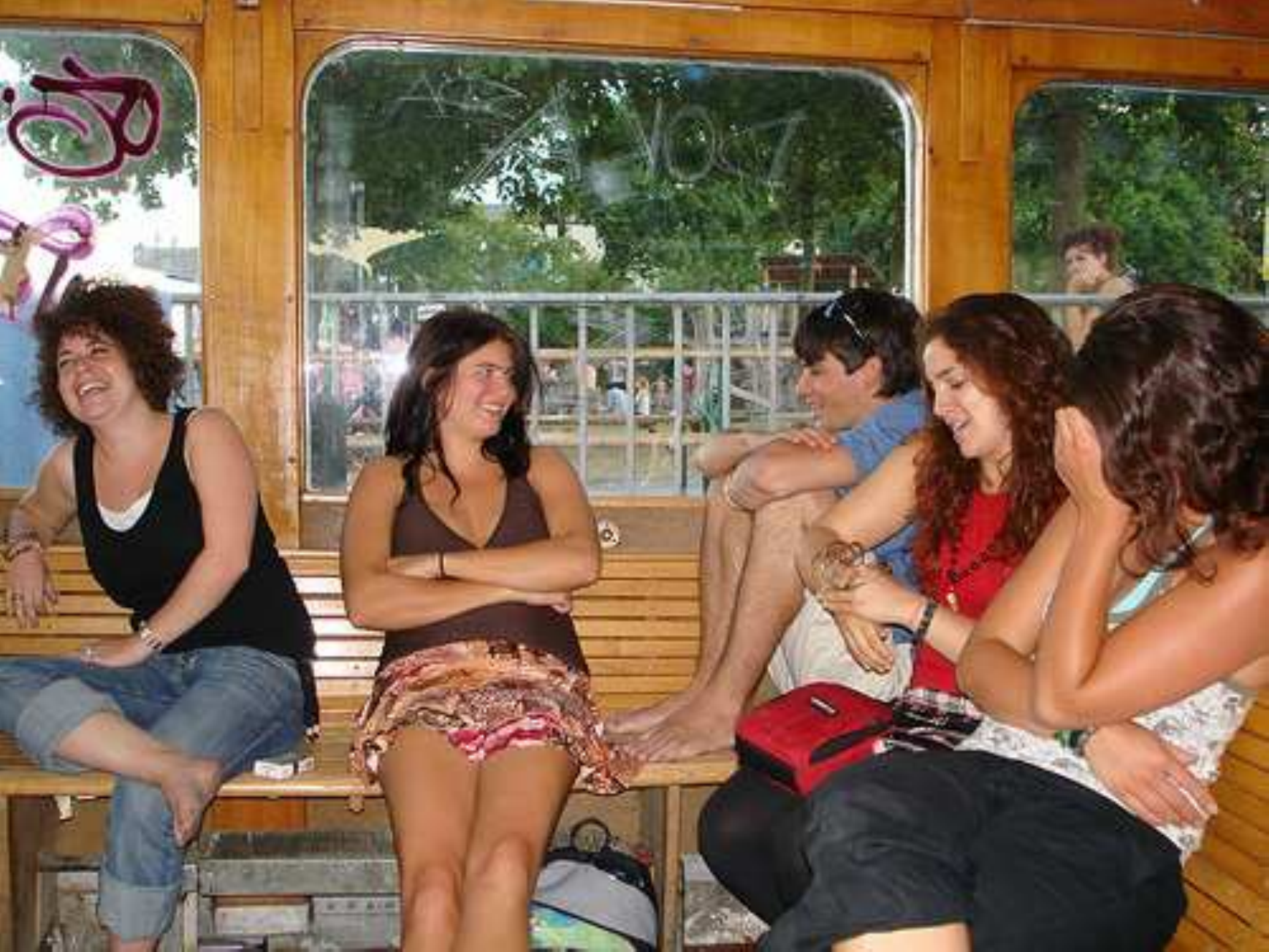} &
		\includegraphics[width=0.19\linewidth]{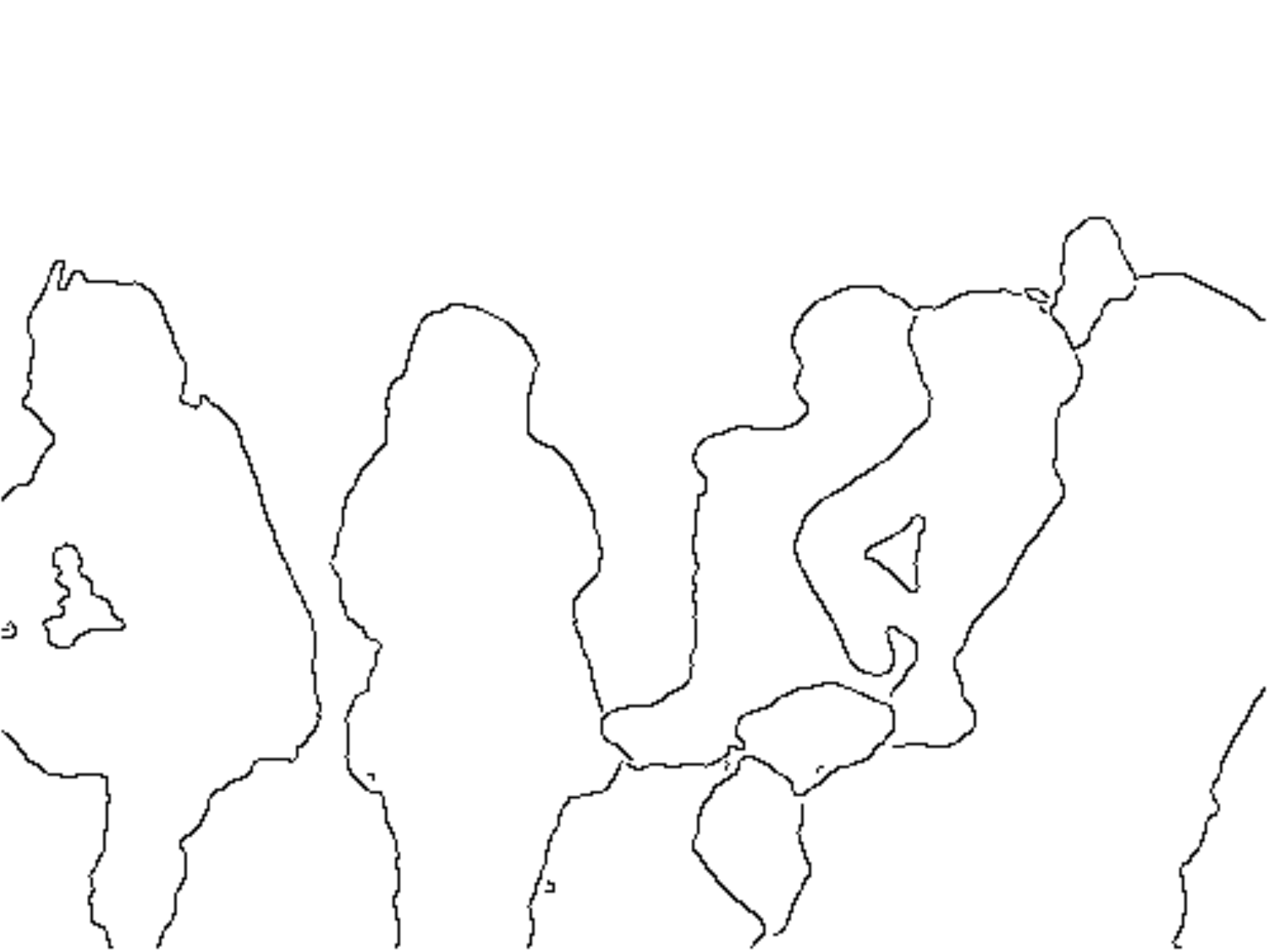} &
		\includegraphics[width=0.19\linewidth]{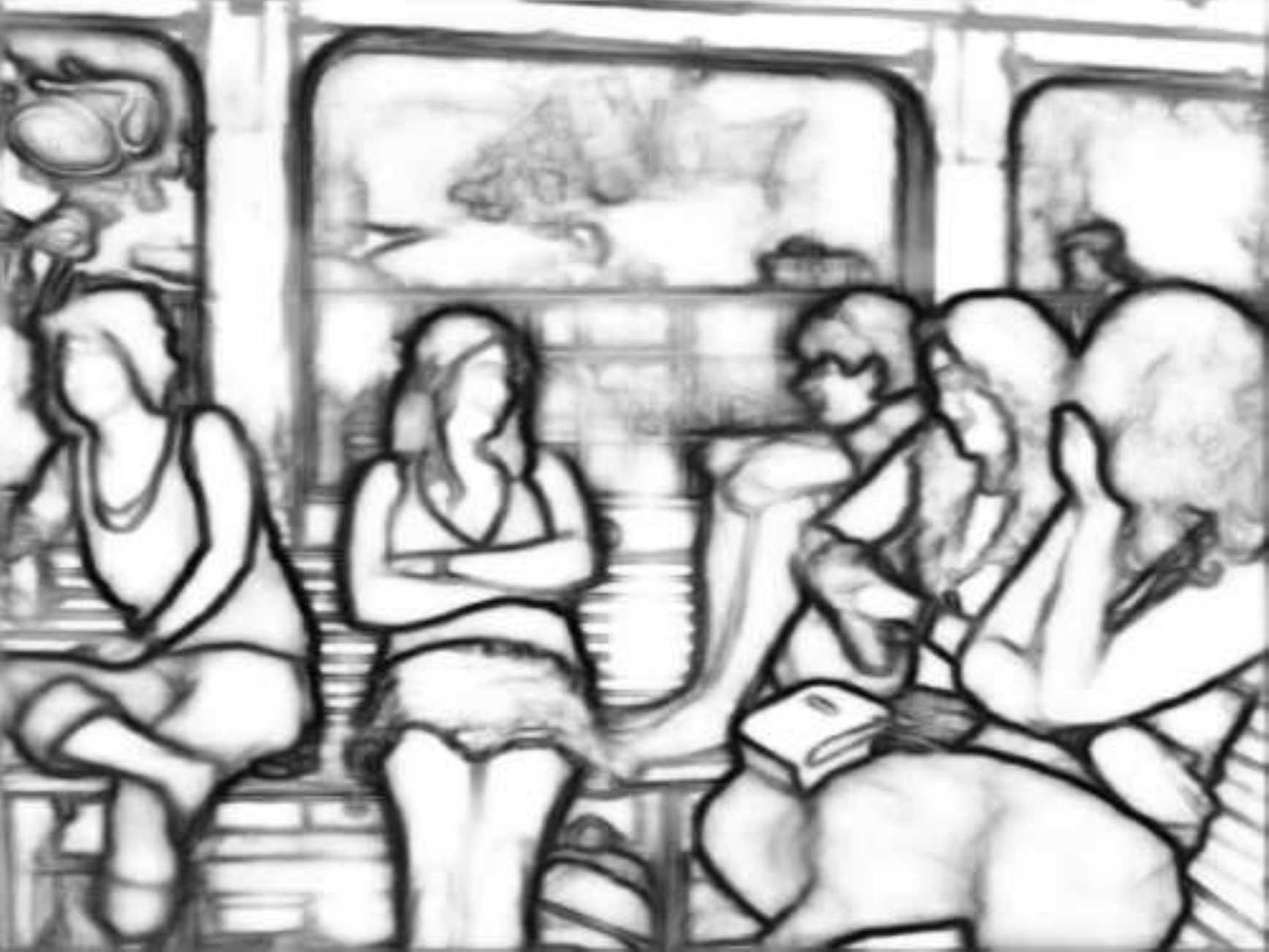} &
		\includegraphics[width=0.19\linewidth]{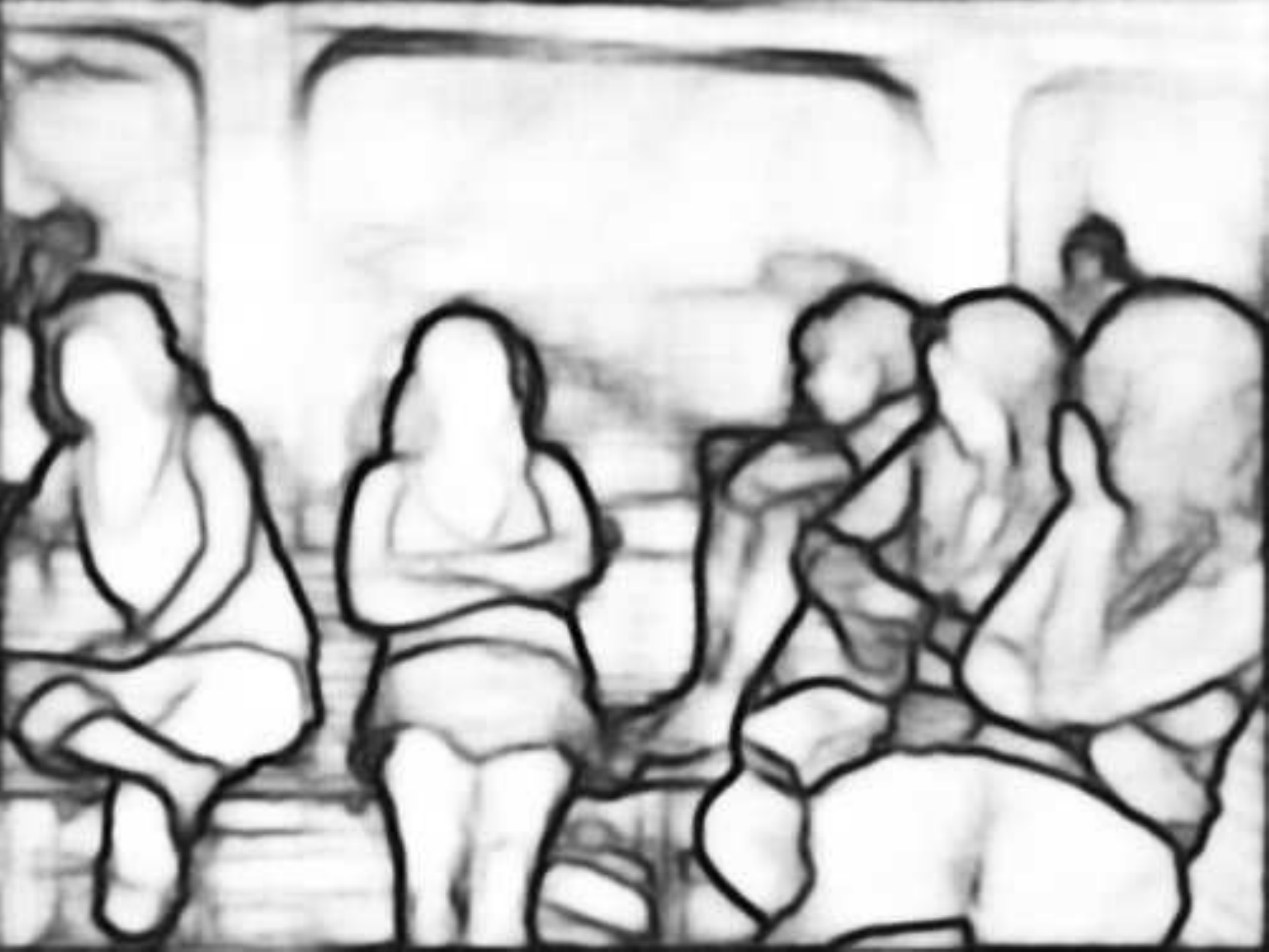} &
		\includegraphics[width=0.19\linewidth]{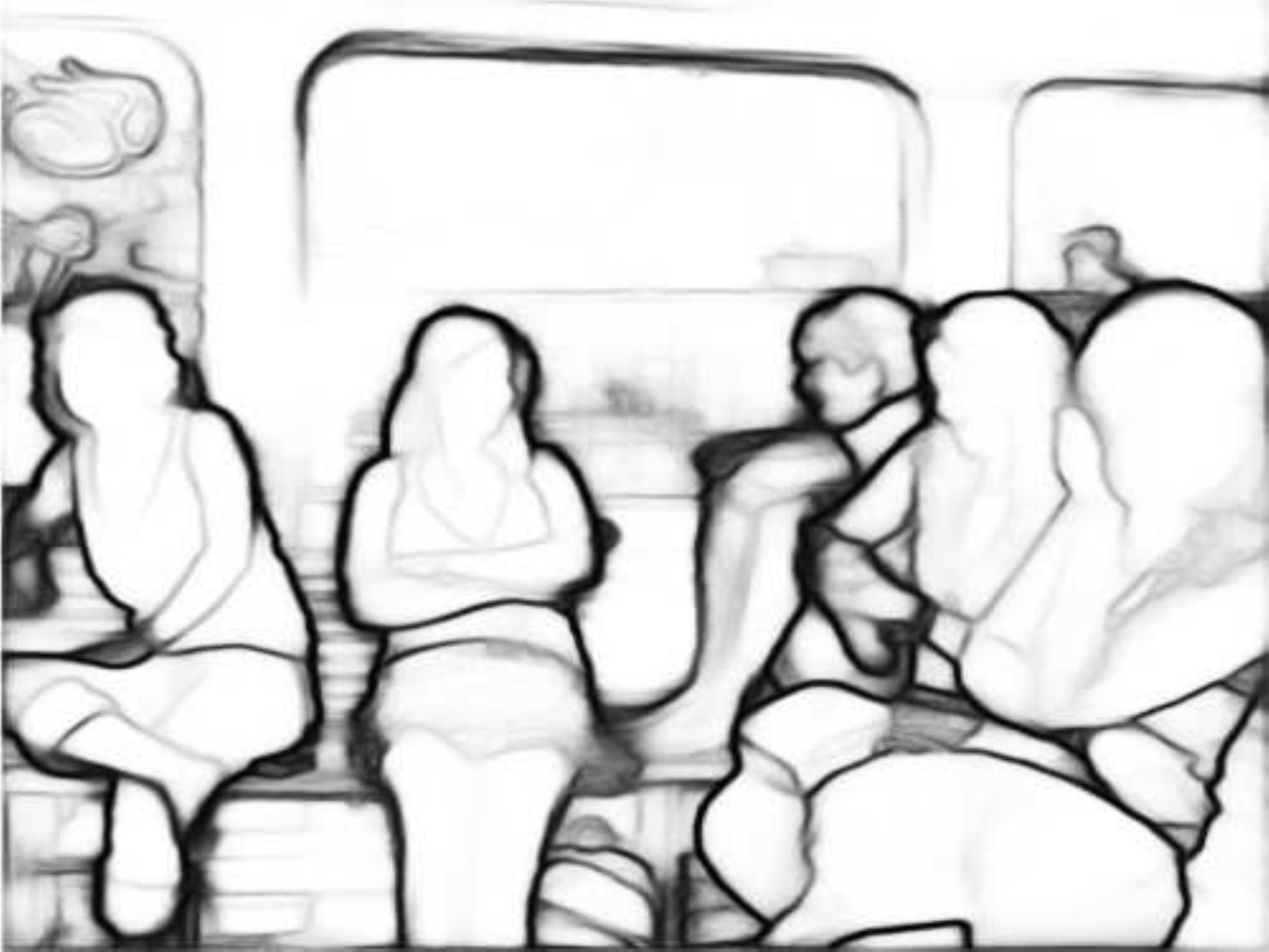} \\
		\includegraphics[width=0.19\linewidth]{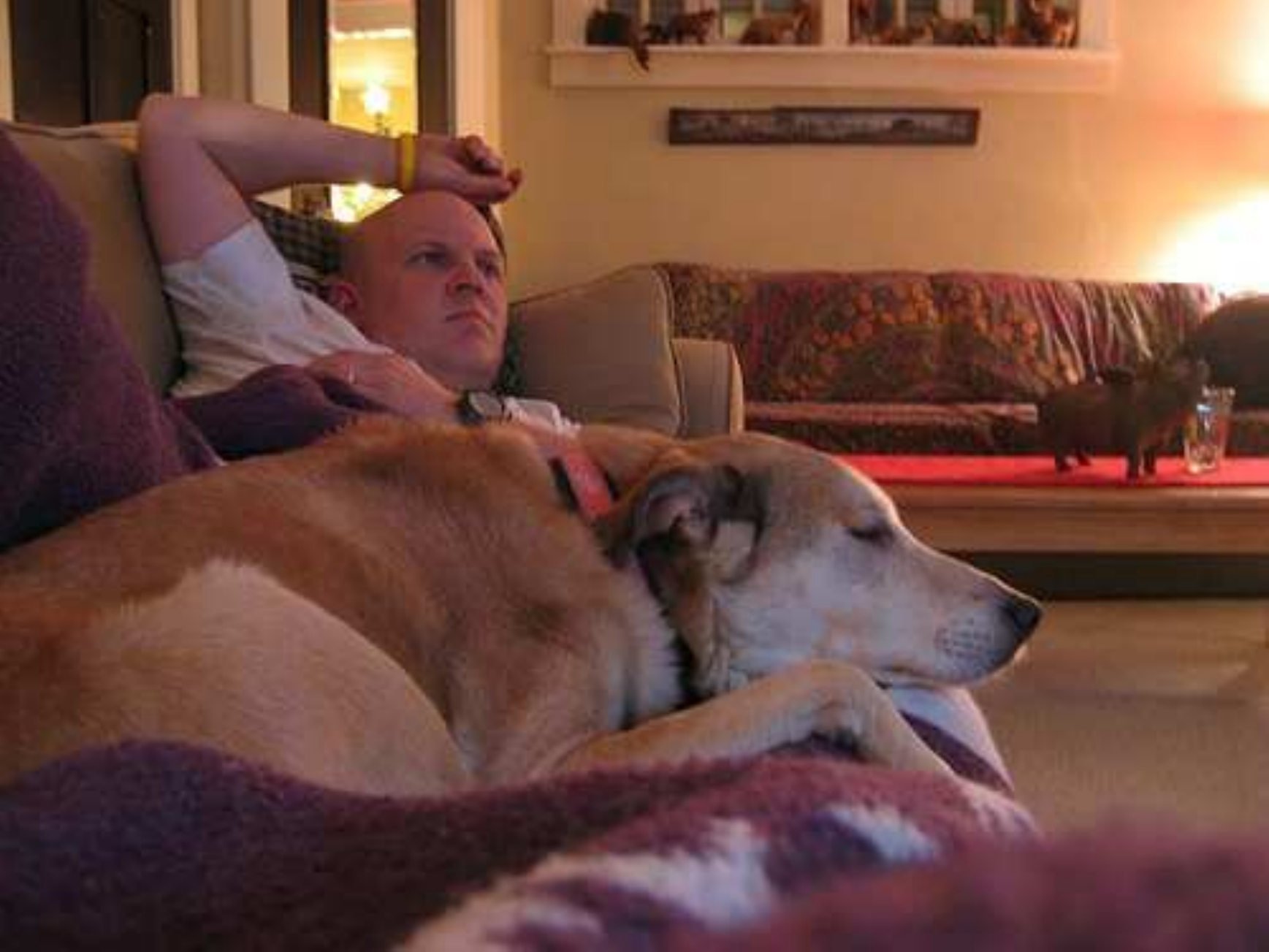} &
		\includegraphics[width=0.19\linewidth]{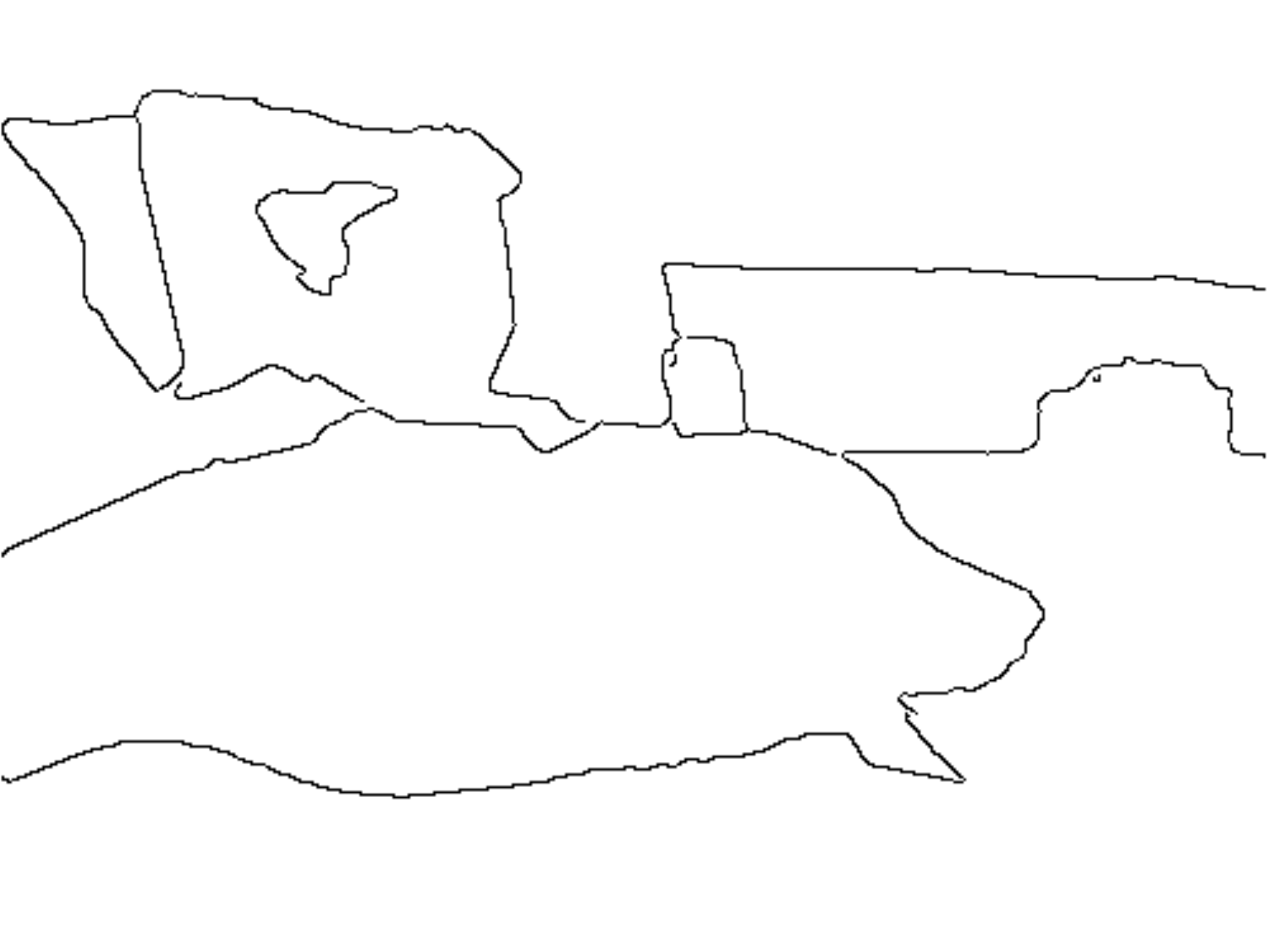} &
		\includegraphics[width=0.19\linewidth]{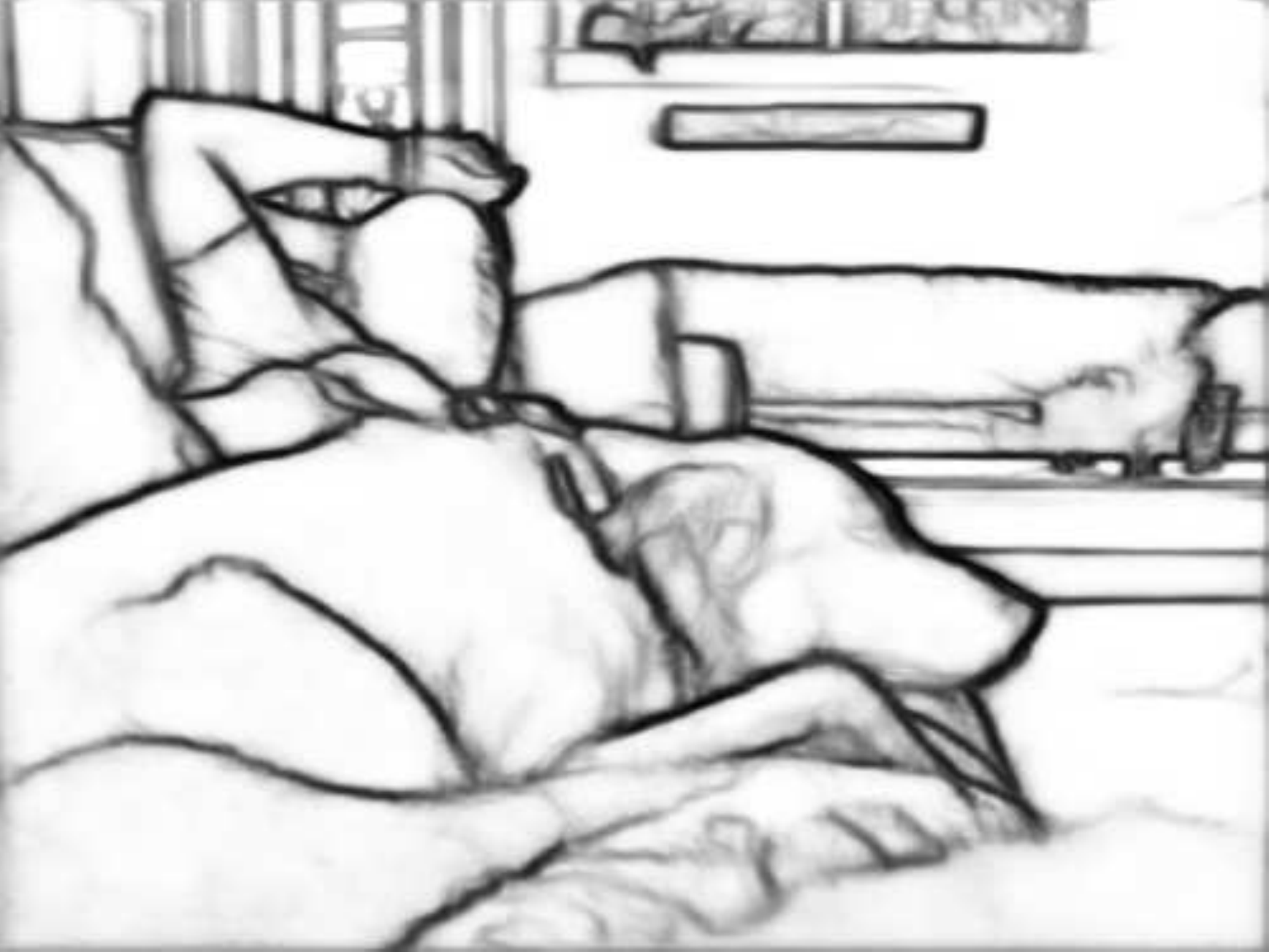} &
		\includegraphics[width=0.19\linewidth]{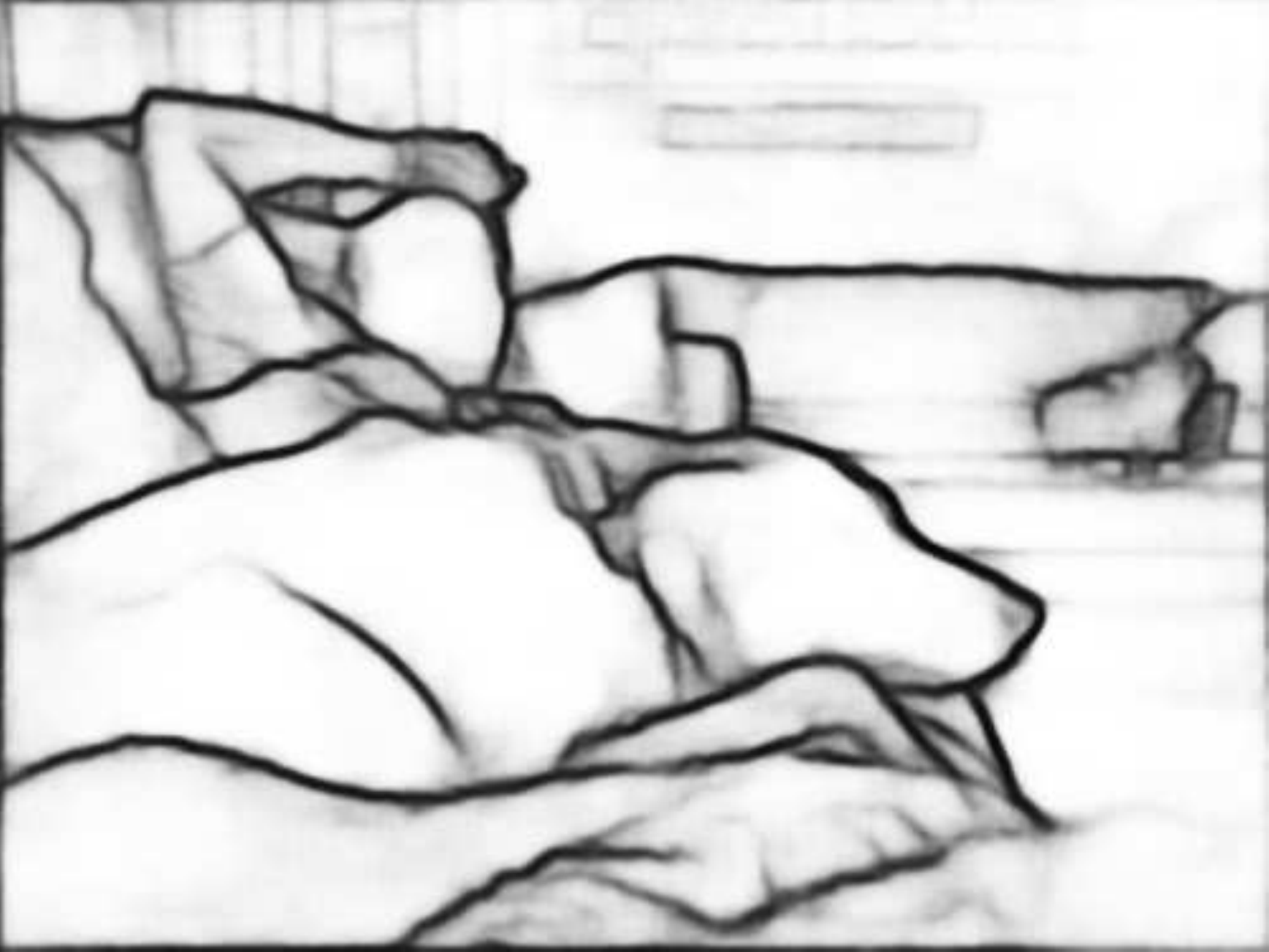} &
		\includegraphics[width=0.19\linewidth]{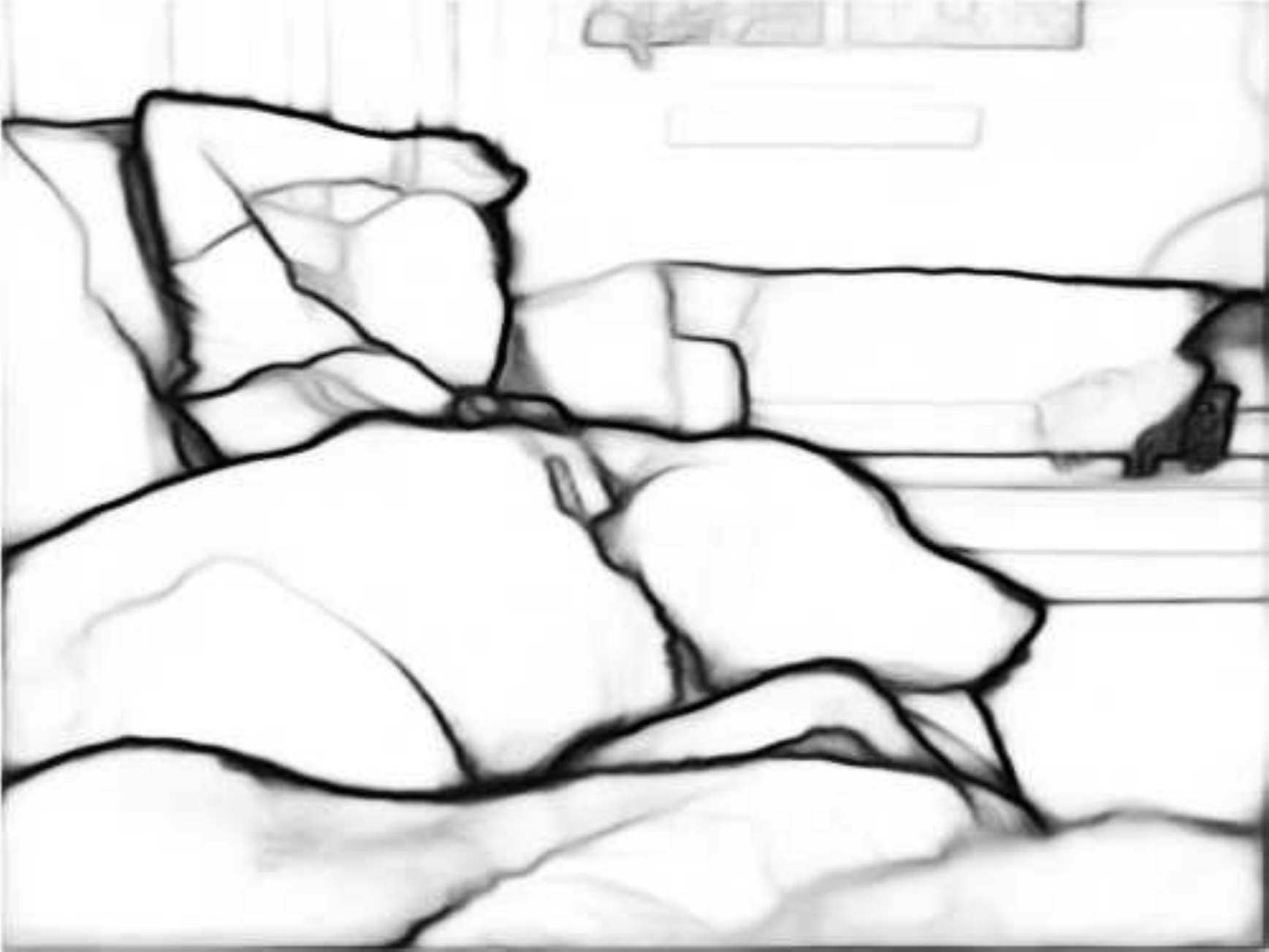} \\
		\includegraphics[width=0.19\linewidth]{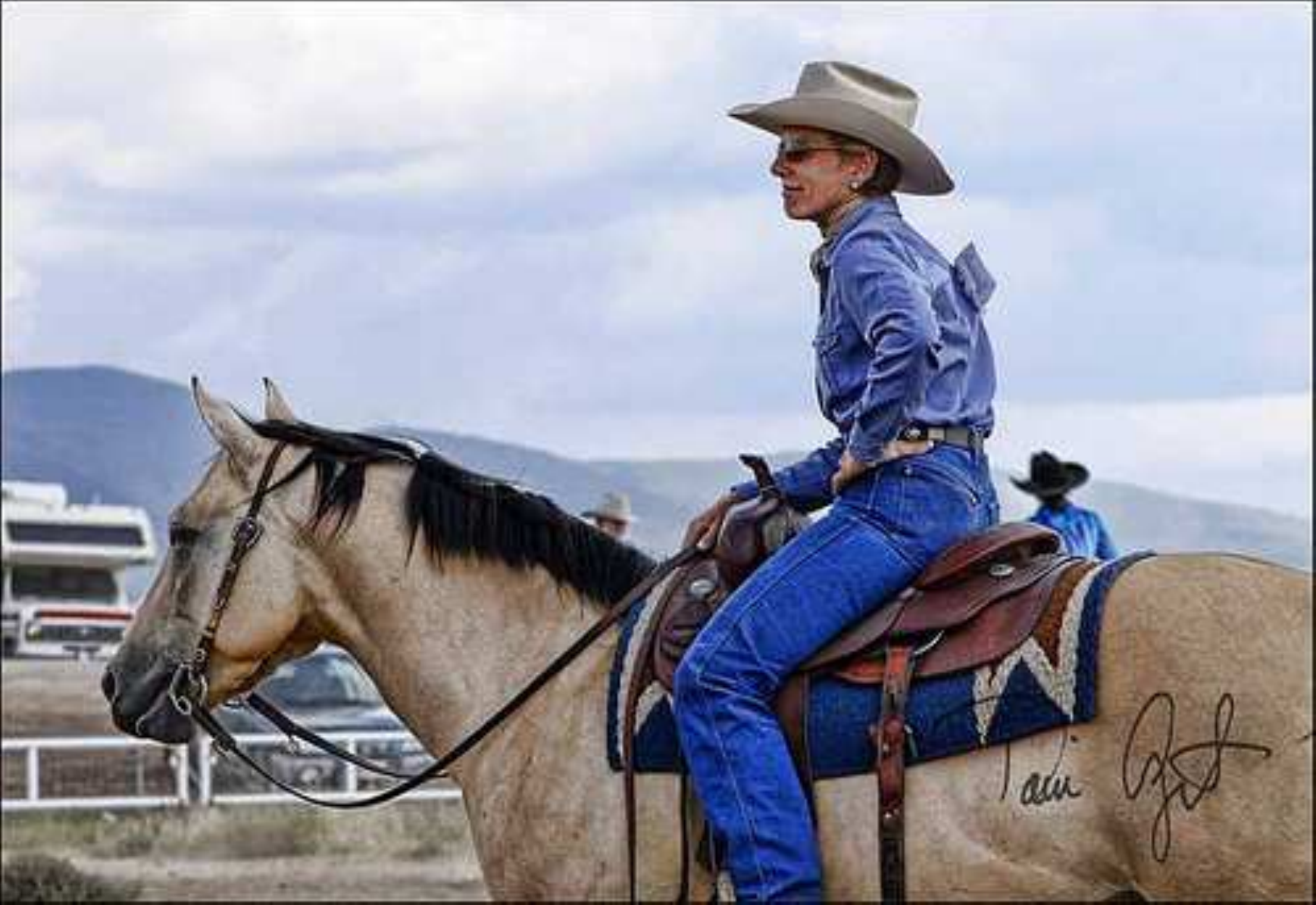} &
		\includegraphics[width=0.19\linewidth]{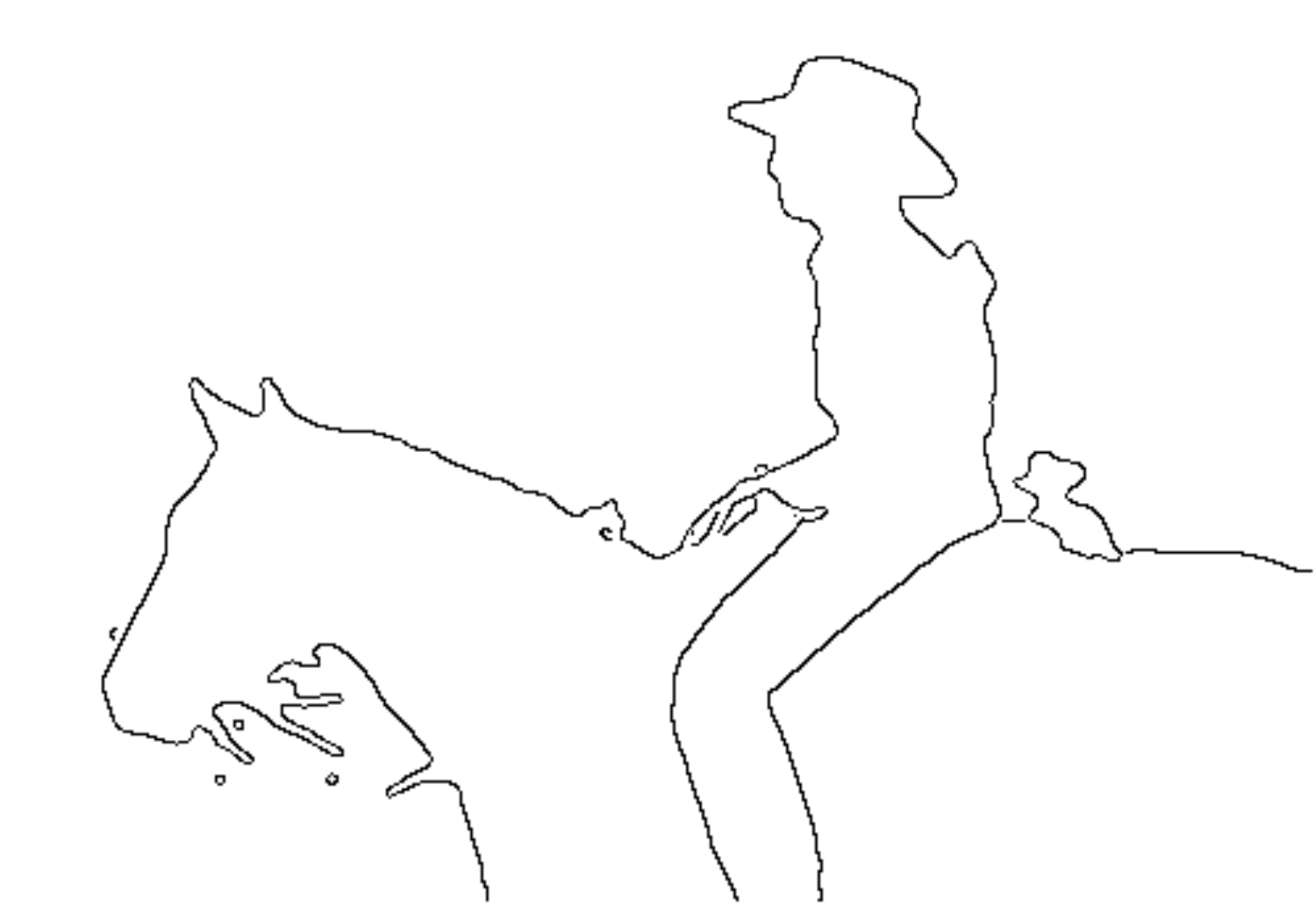} &
		\includegraphics[width=0.19\linewidth]{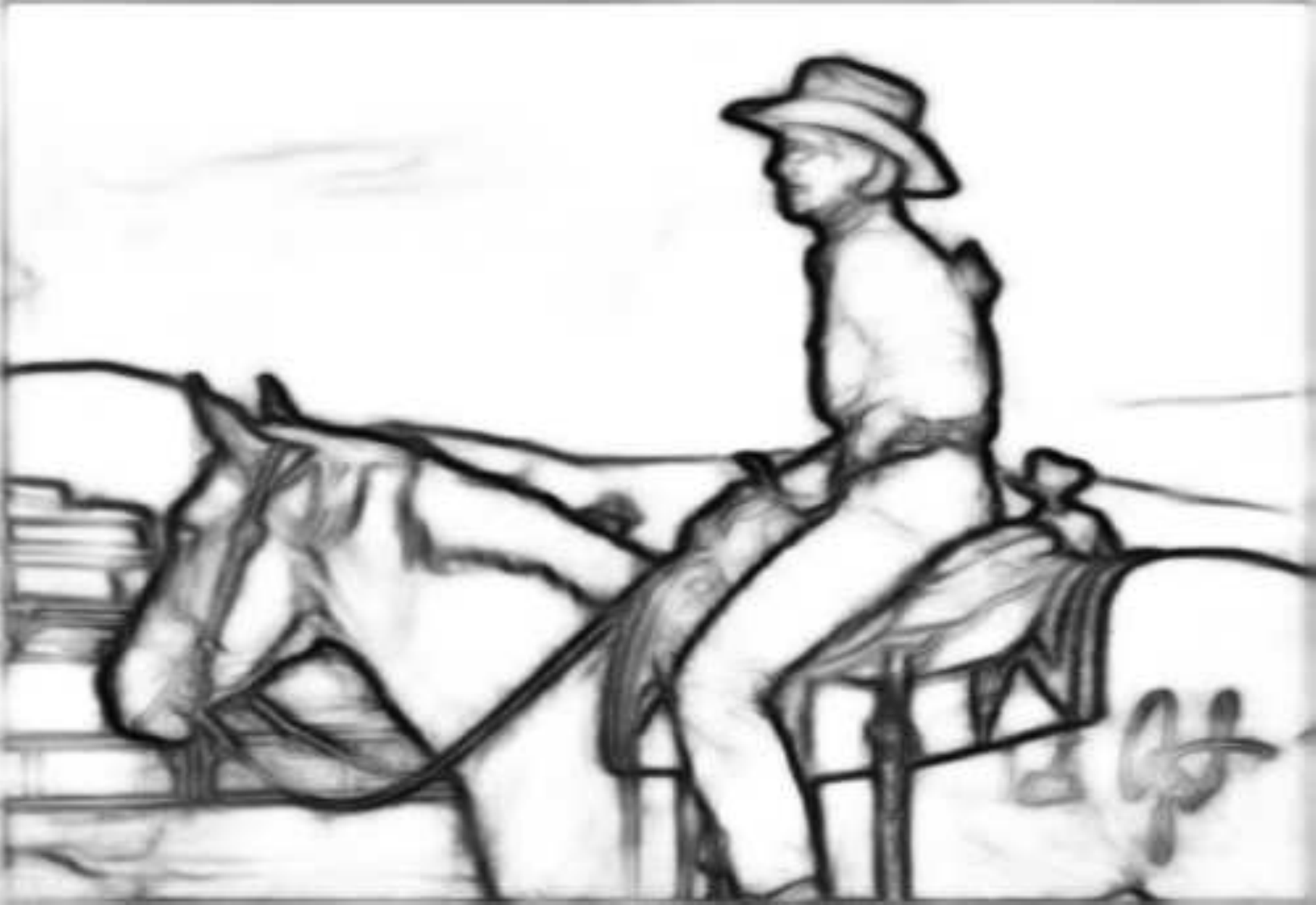} &
		\includegraphics[width=0.19\linewidth]{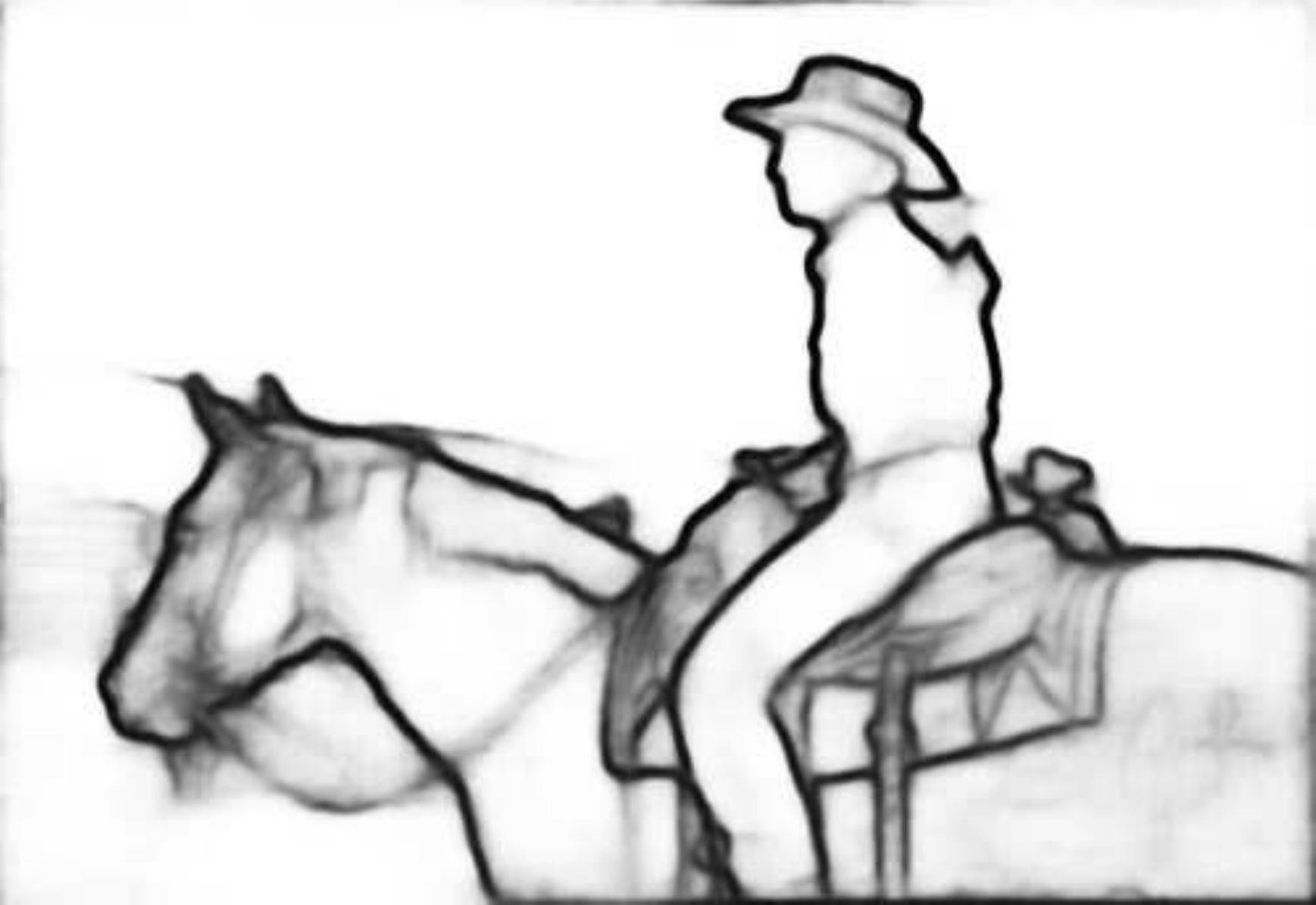} &
		\includegraphics[width=0.19\linewidth]{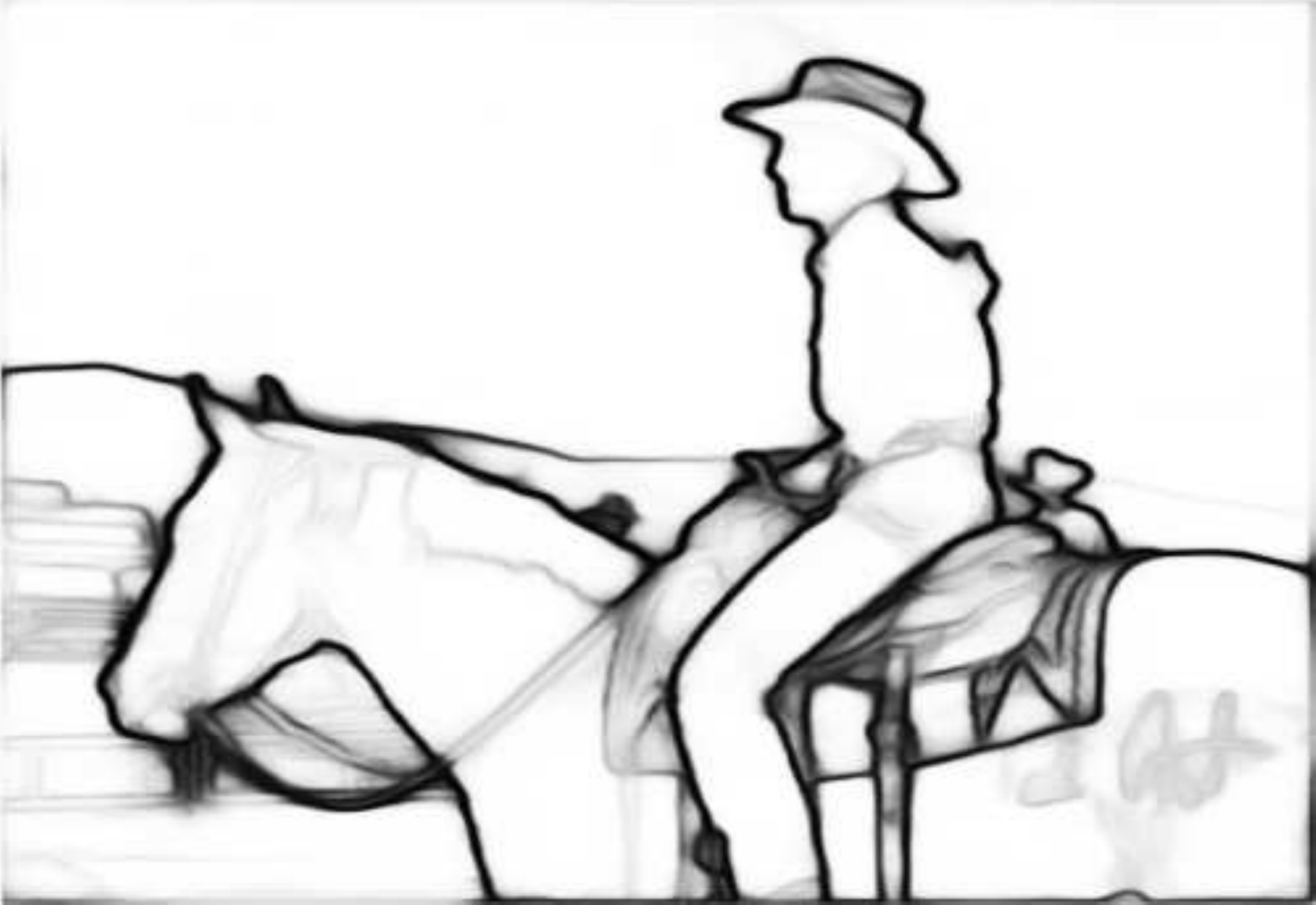} \\
		\includegraphics[width=0.19\linewidth]{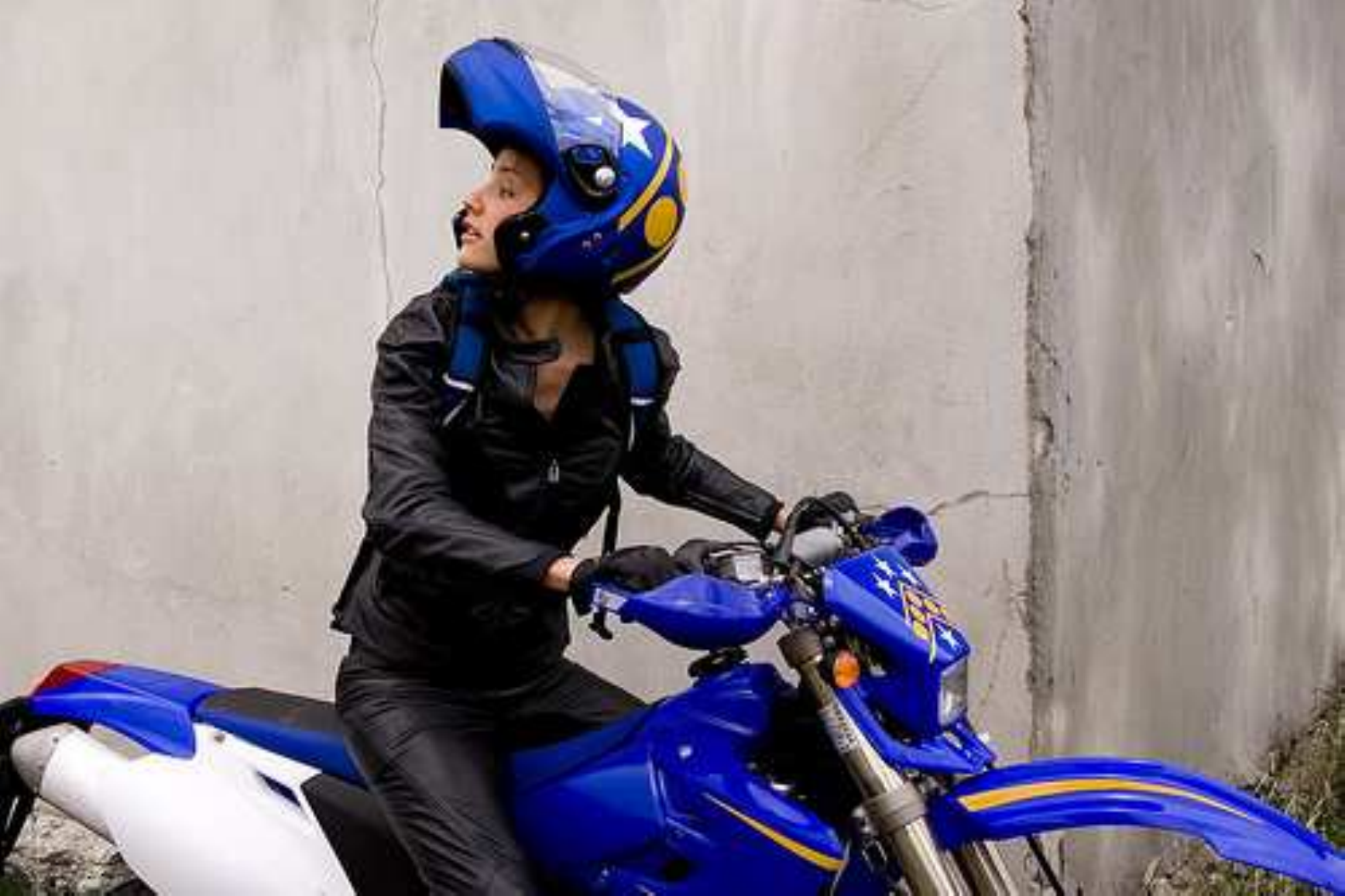} &
		\includegraphics[width=0.19\linewidth]{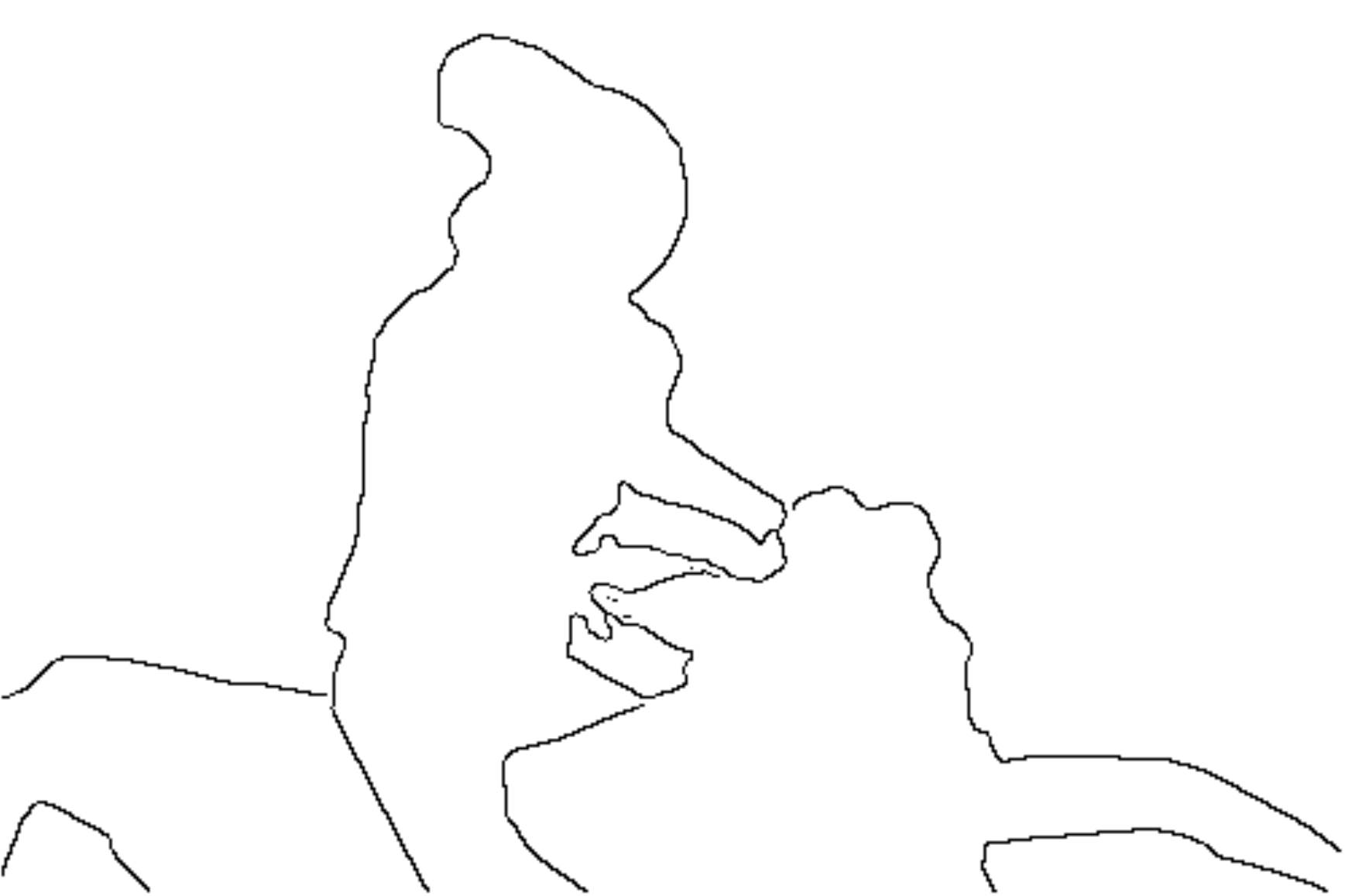} &
		\includegraphics[width=0.19\linewidth]{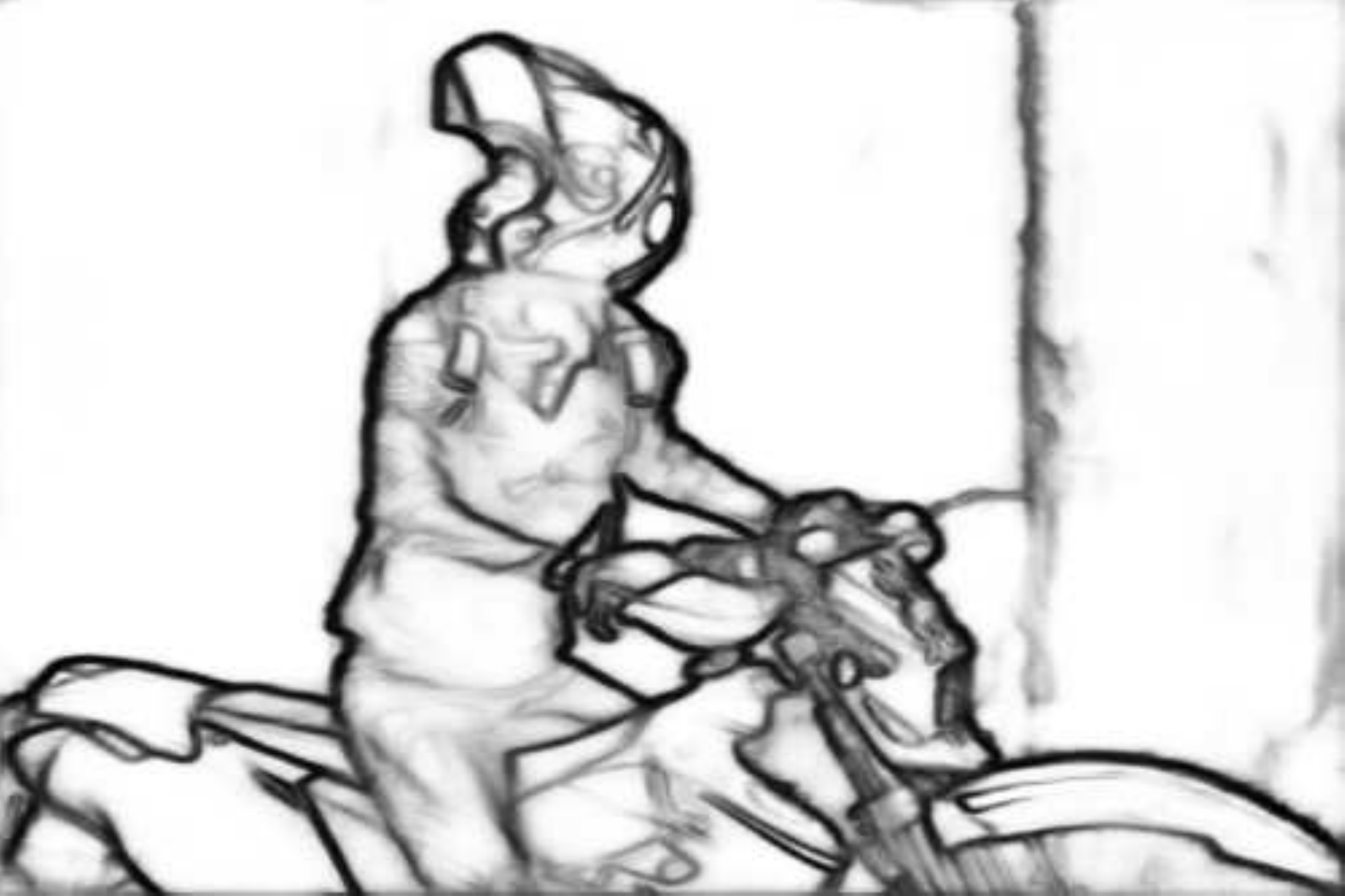} &
		\includegraphics[width=0.19\linewidth]{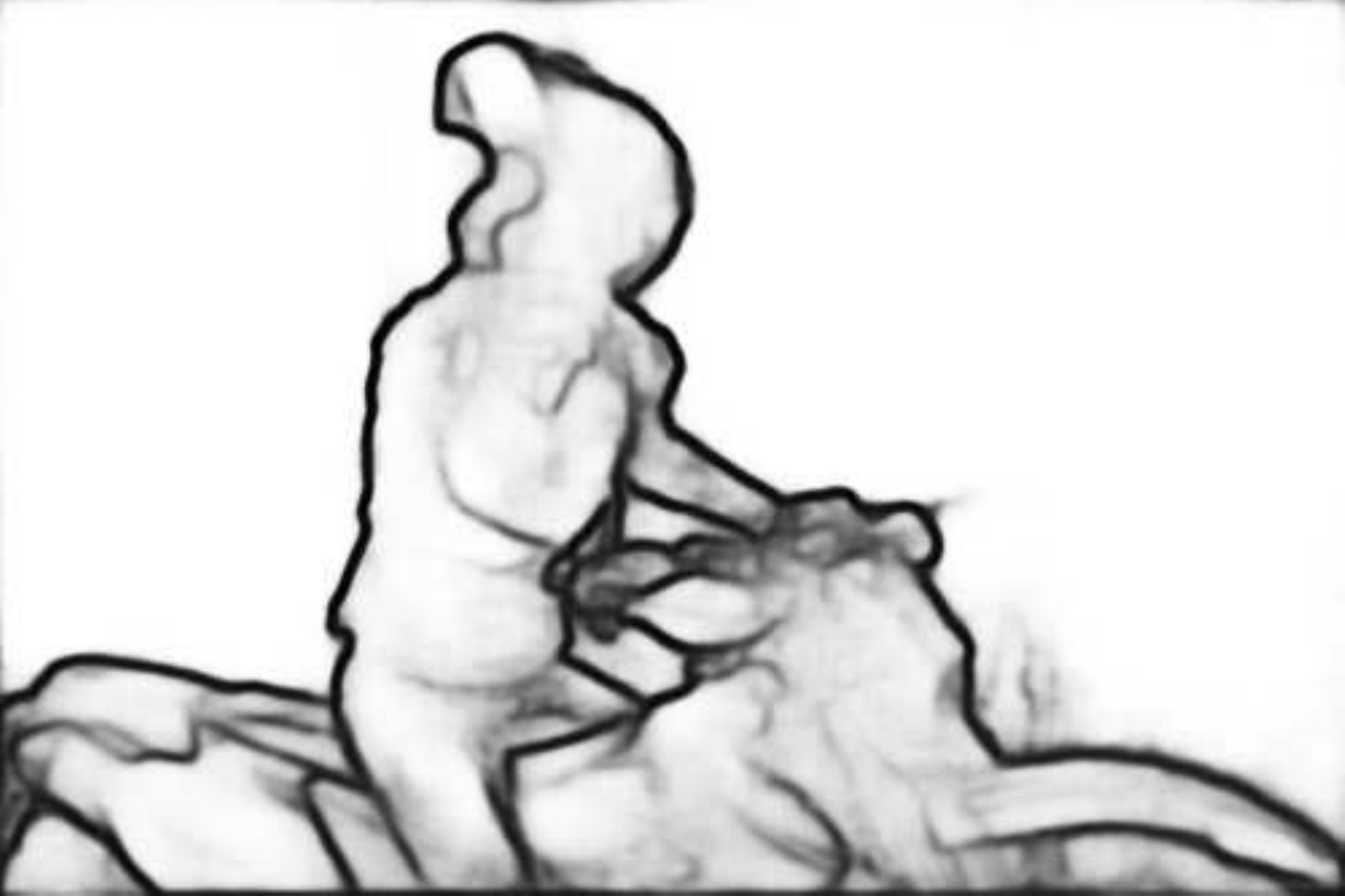} &
		\includegraphics[width=0.19\linewidth]{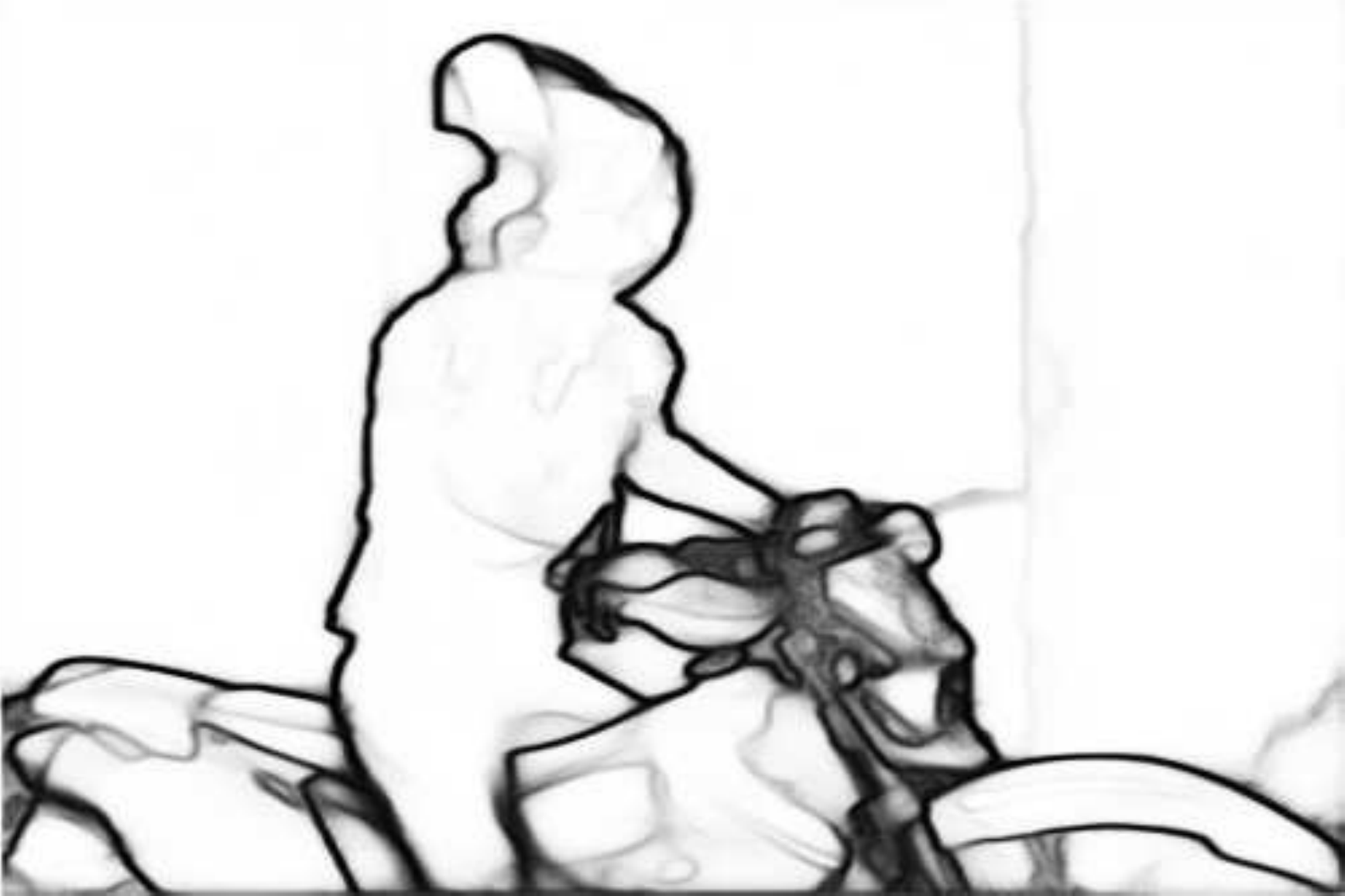} \\
		\includegraphics[width=0.19\linewidth]{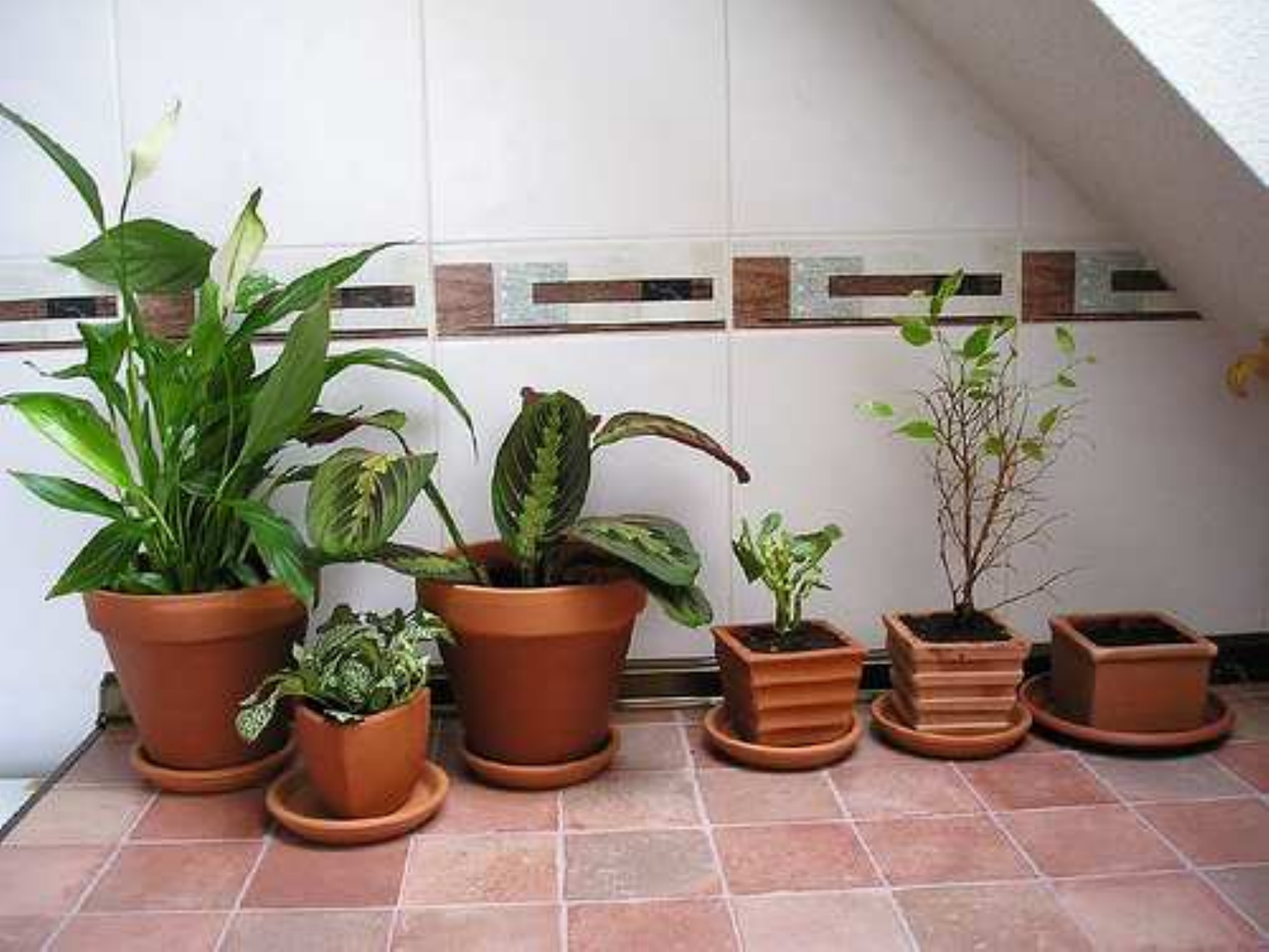} &
		\includegraphics[width=0.19\linewidth]{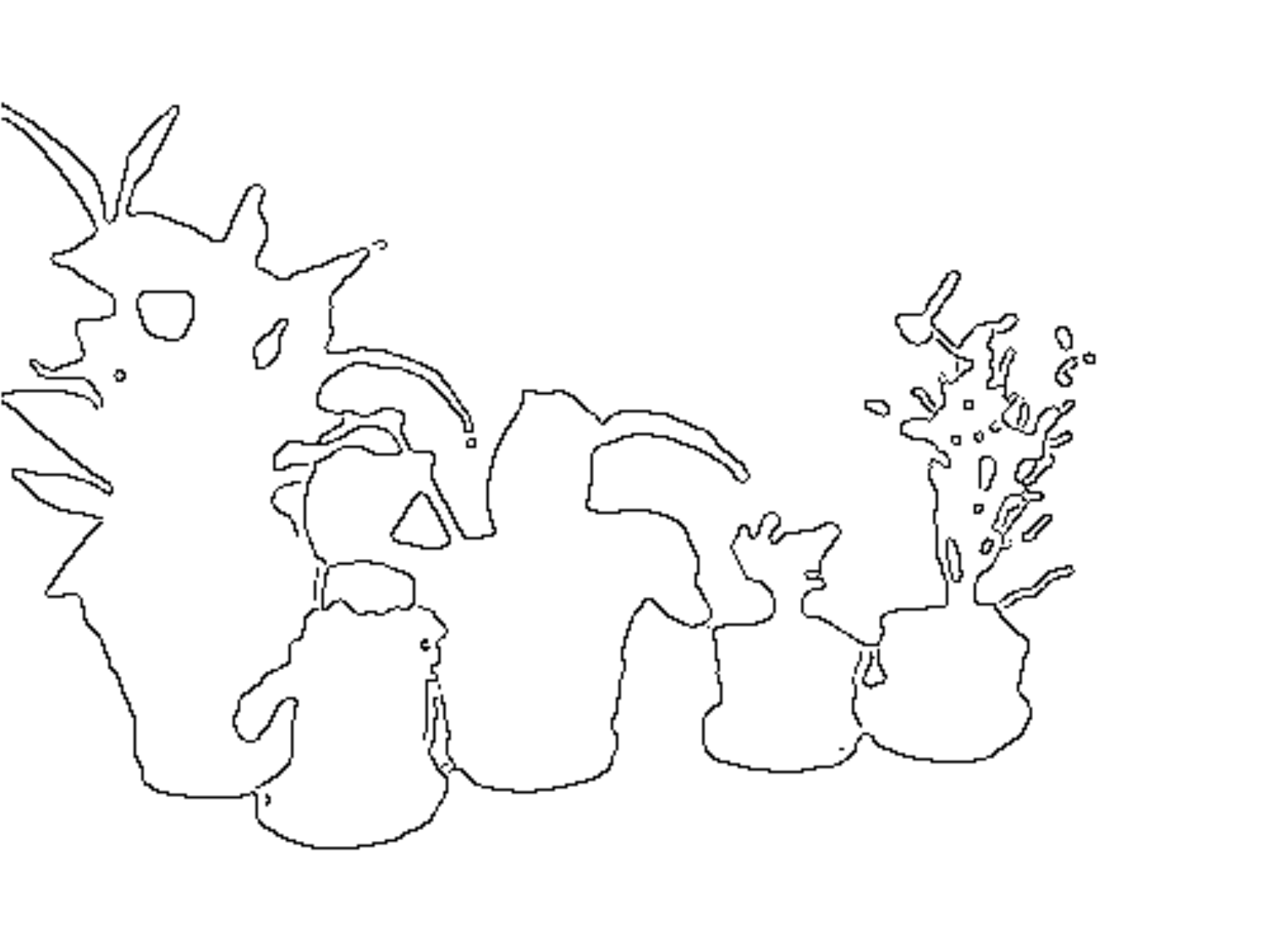} &
		\includegraphics[width=0.19\linewidth]{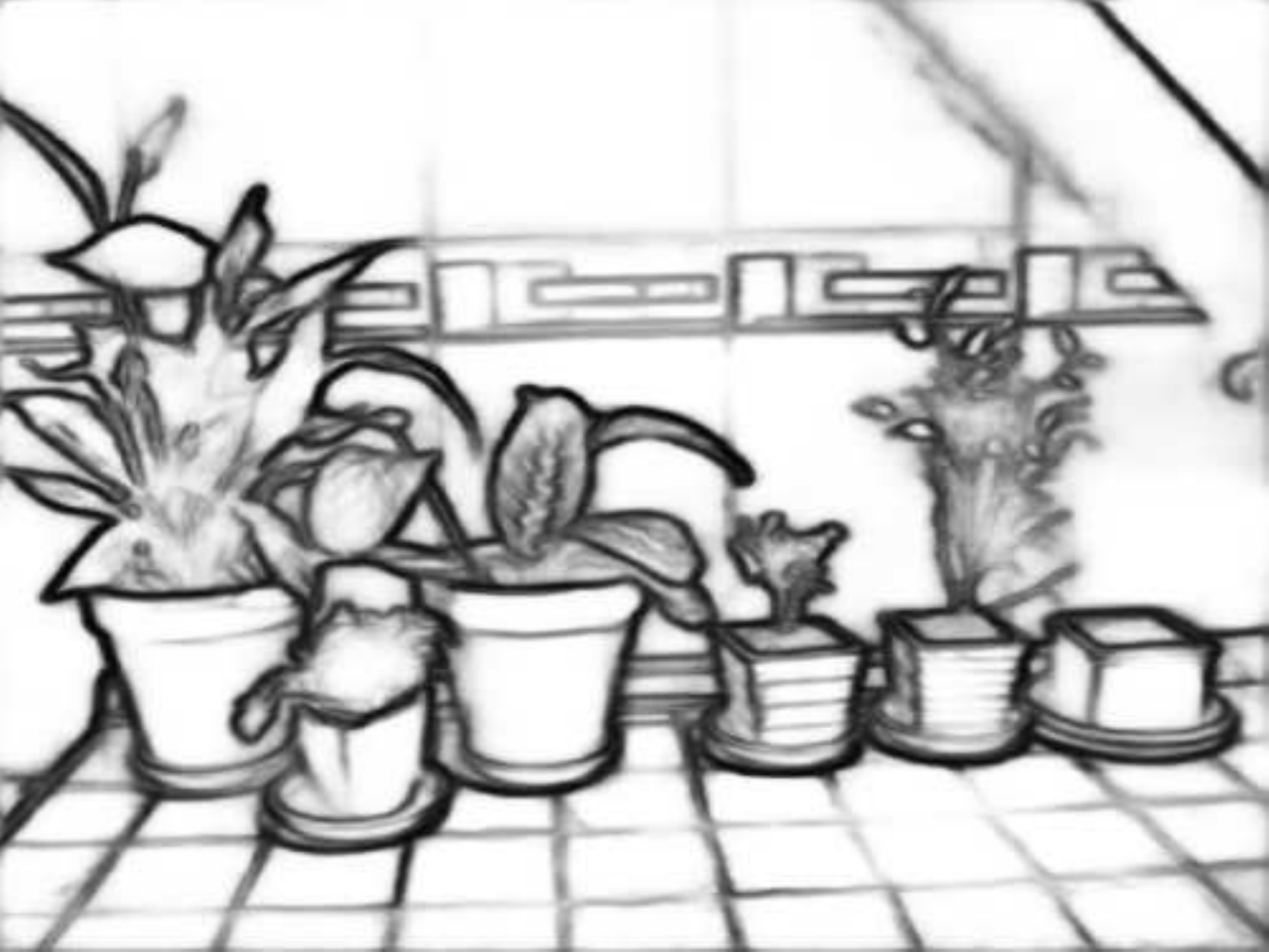} &
		\includegraphics[width=0.19\linewidth]{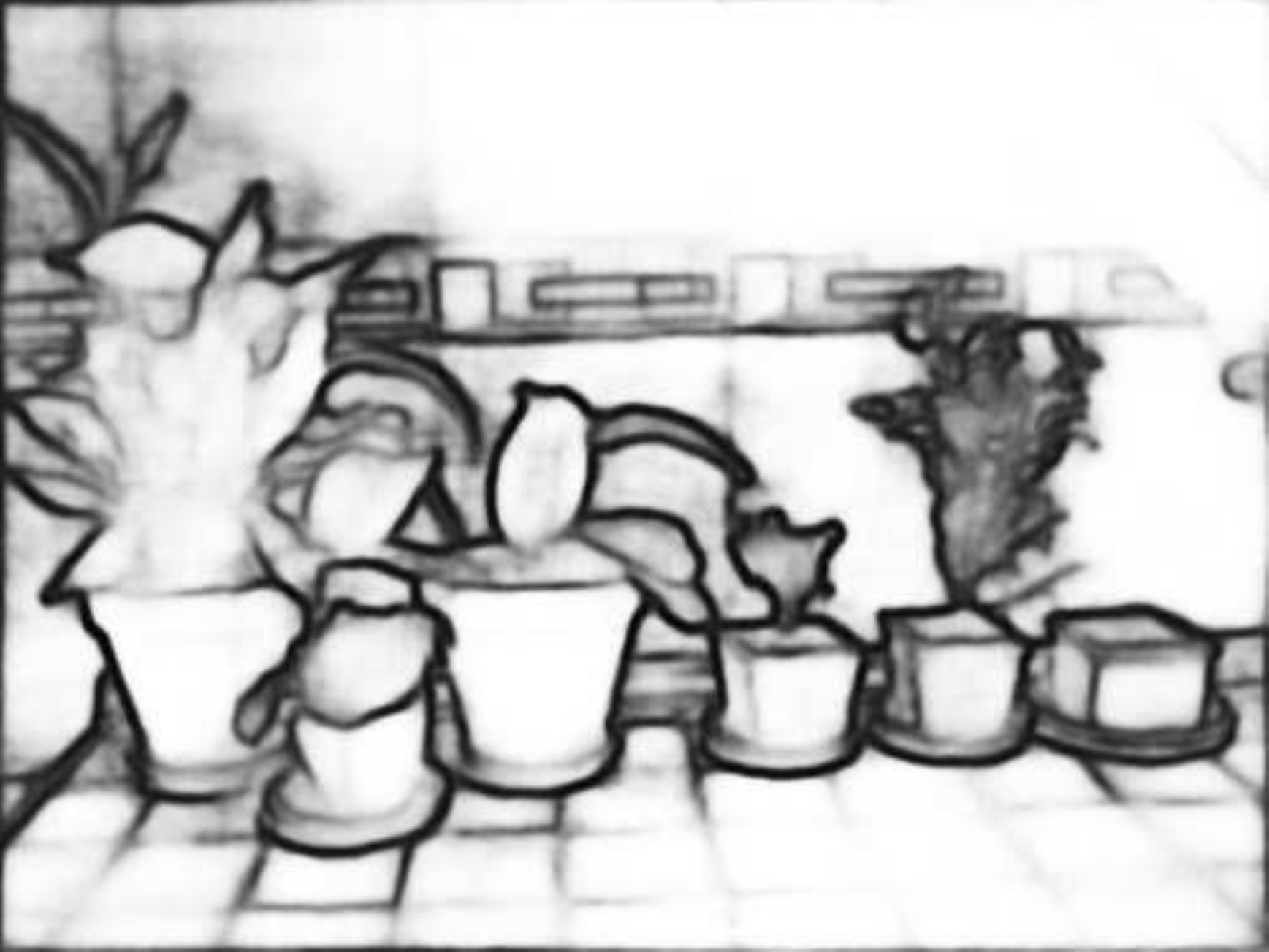} &
		\includegraphics[width=0.19\linewidth]{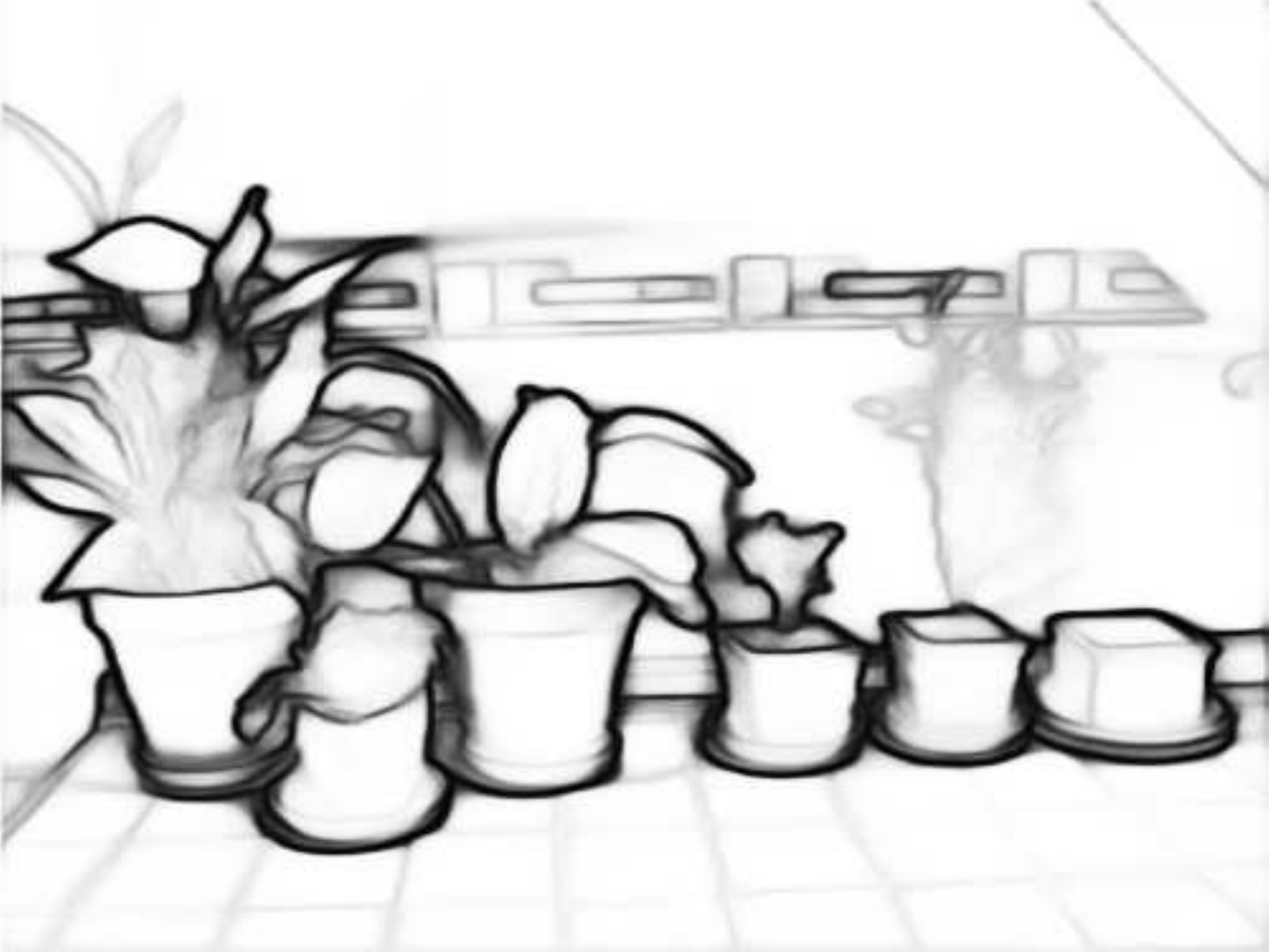} \\
		\includegraphics[width=0.19\linewidth]{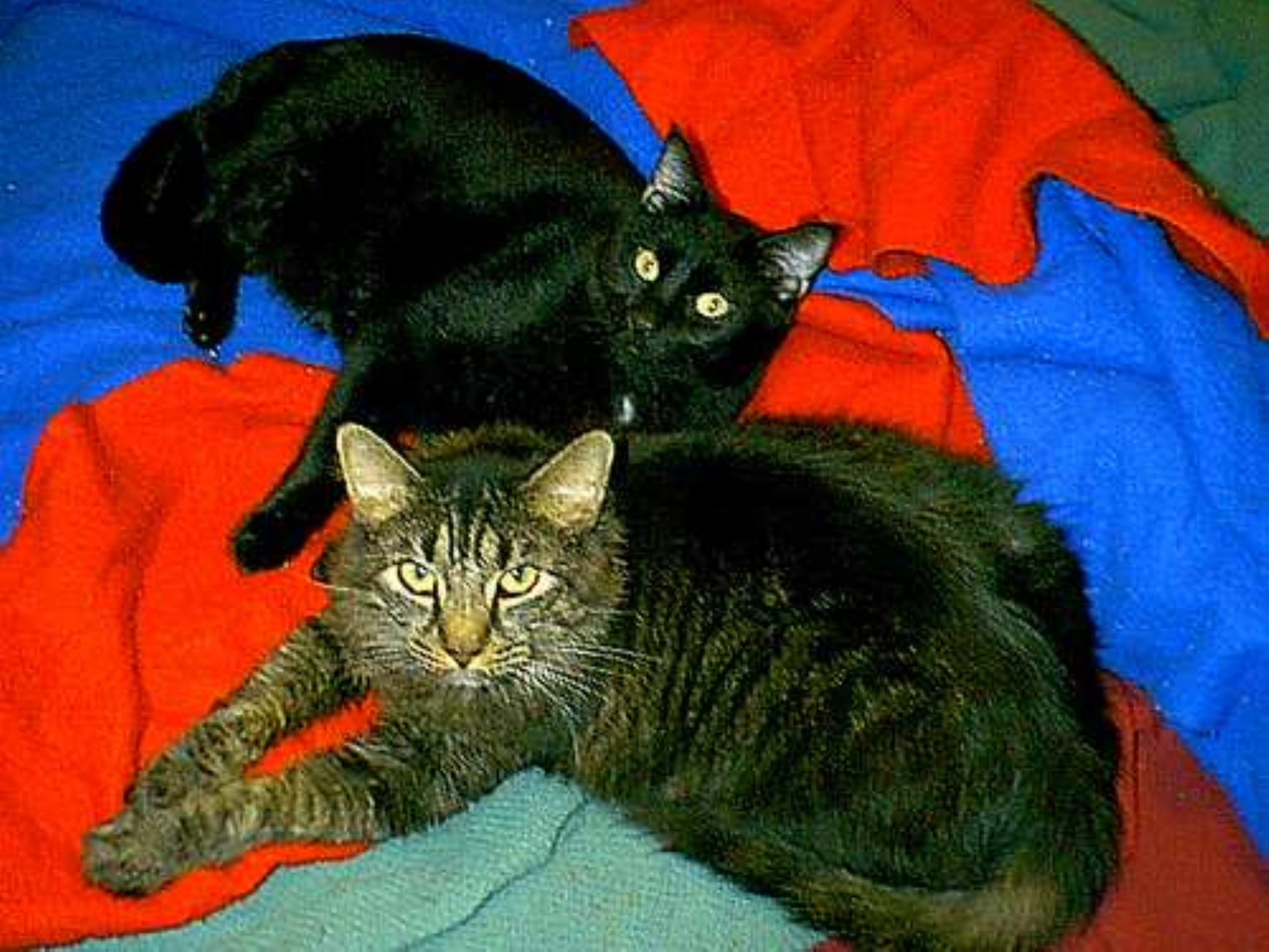} &
		\includegraphics[width=0.19\linewidth]{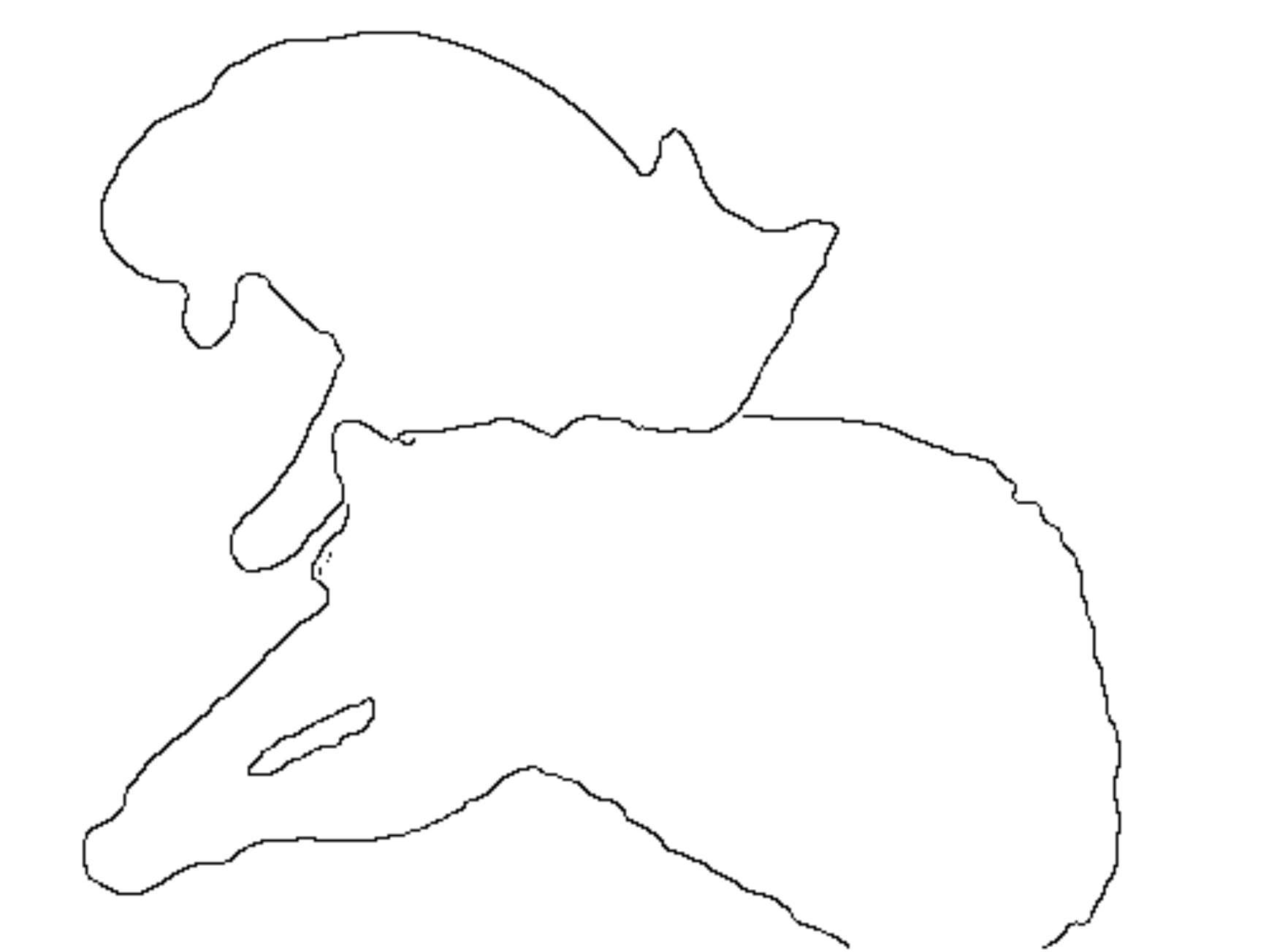} &
		\includegraphics[width=0.19\linewidth]{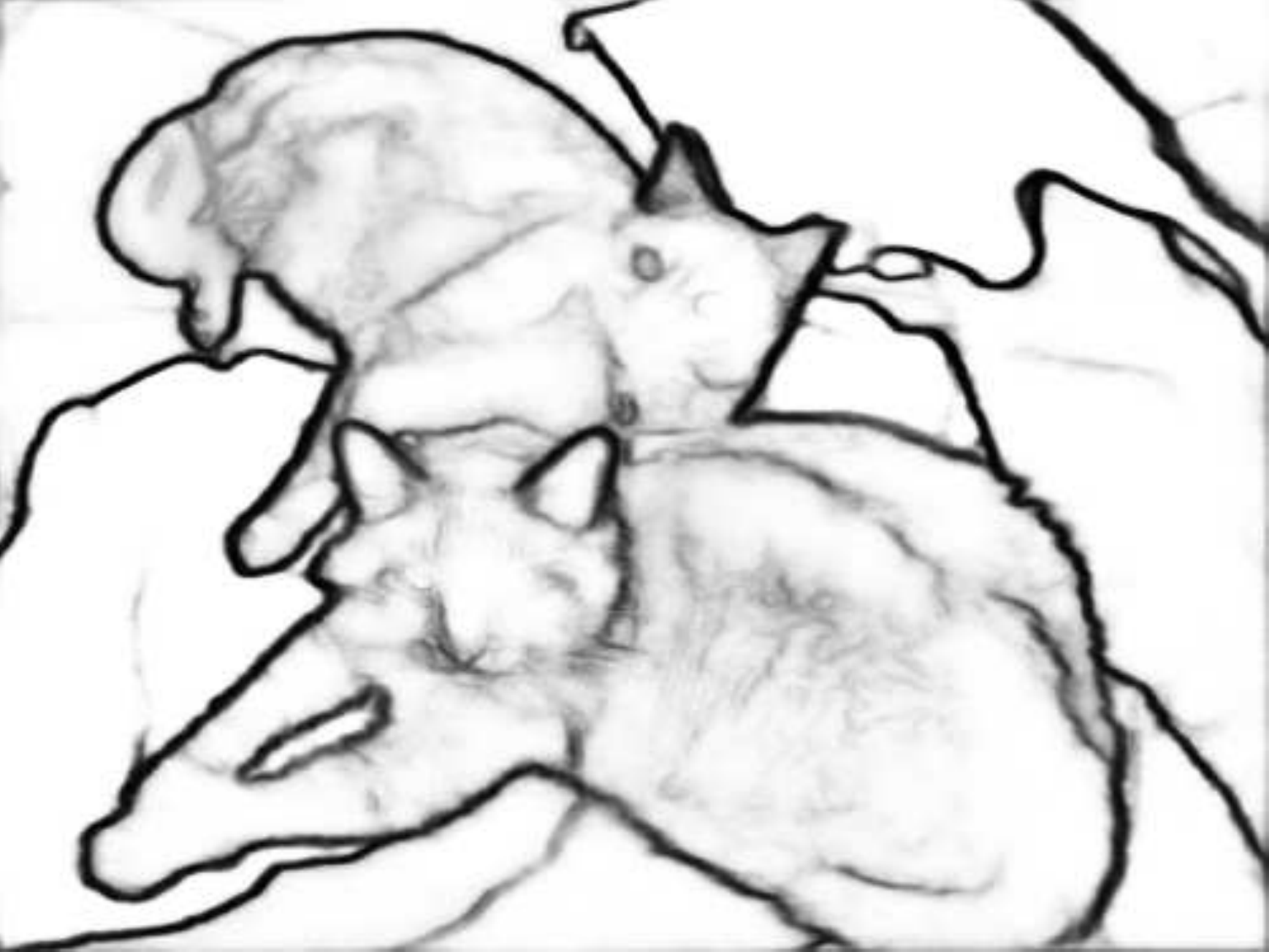} &
		\includegraphics[width=0.19\linewidth]{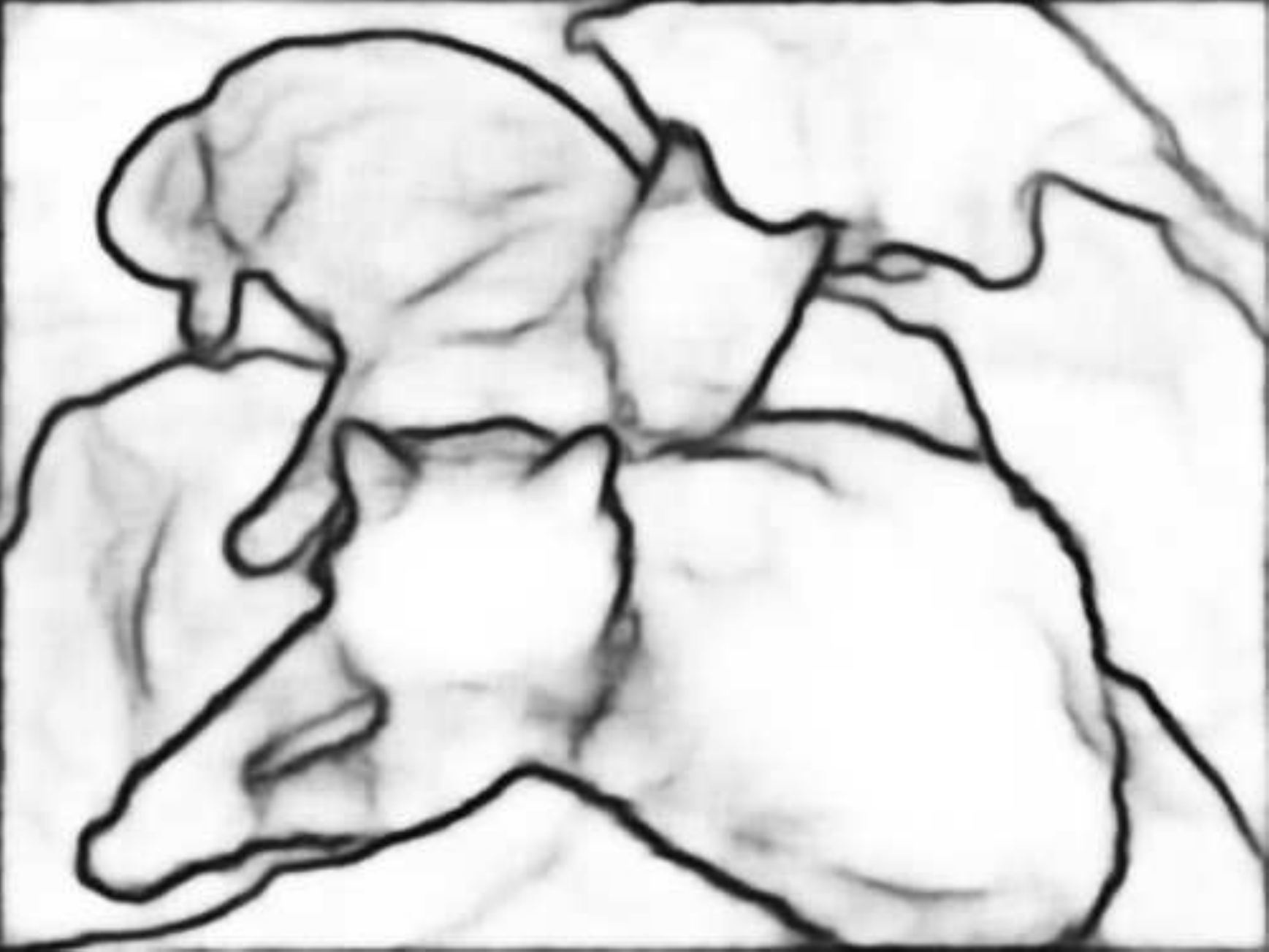} &
		\includegraphics[width=0.19\linewidth]{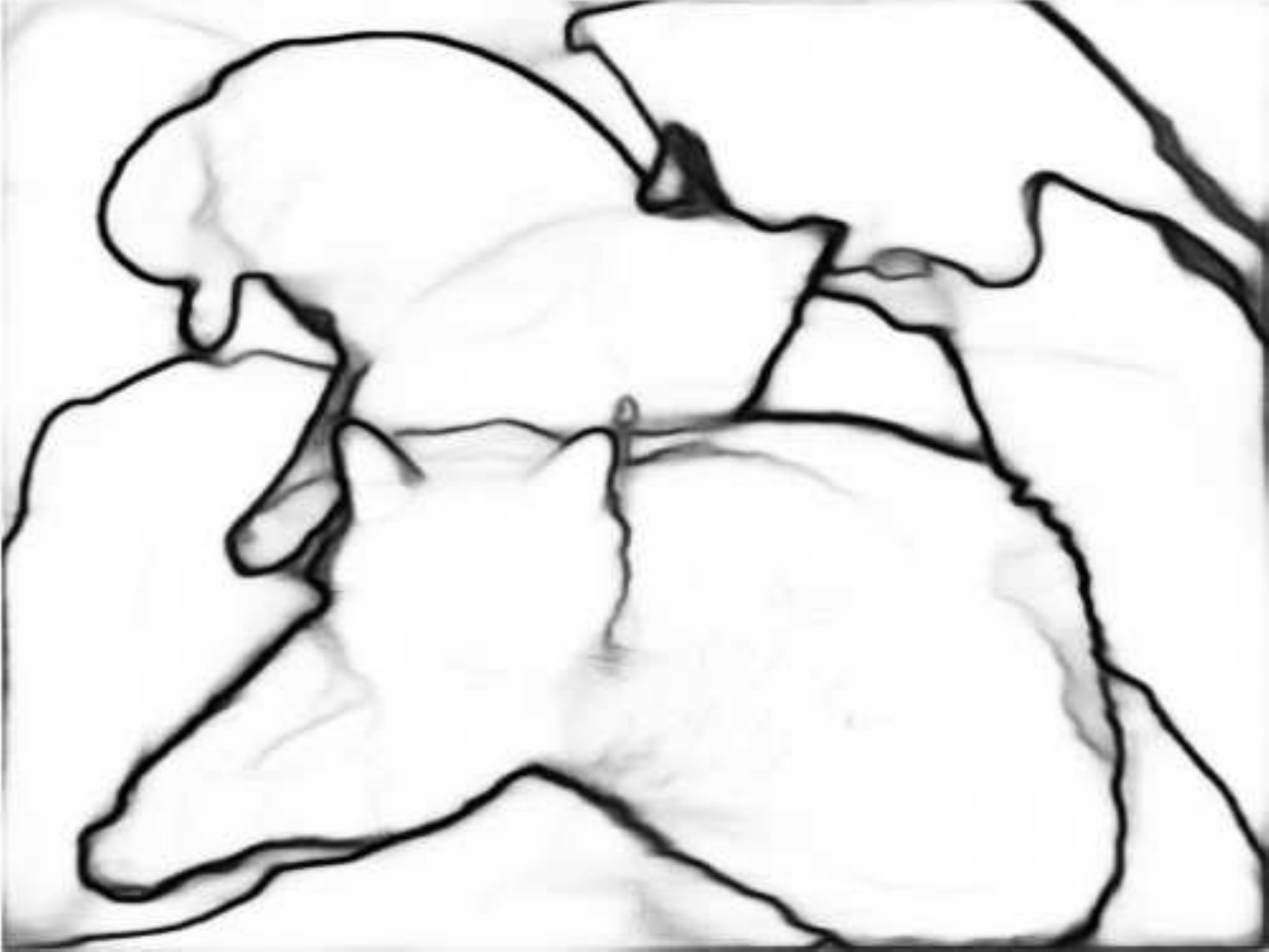} \\
		\includegraphics[width=0.19\linewidth]{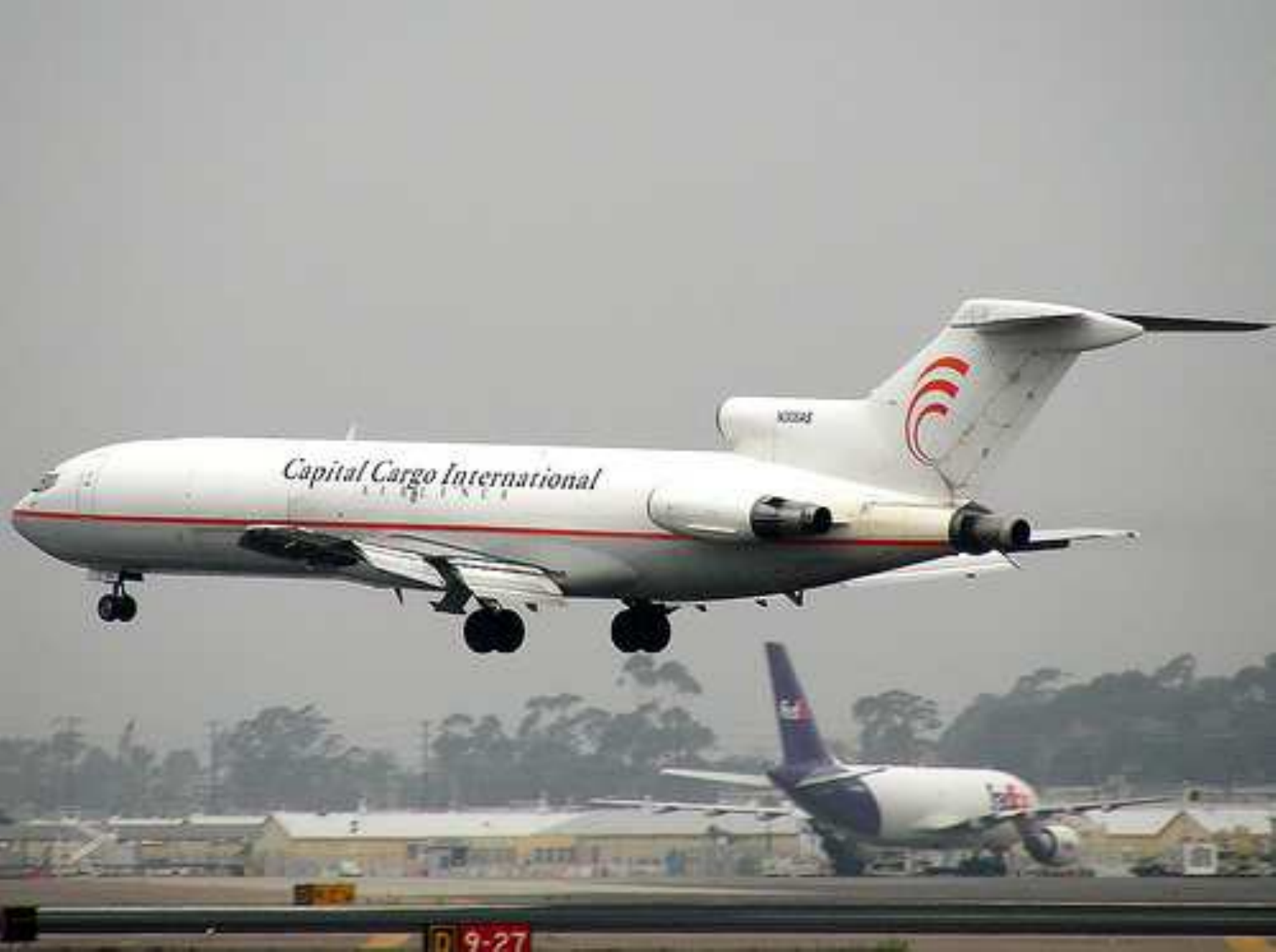} &
		\includegraphics[width=0.19\linewidth]{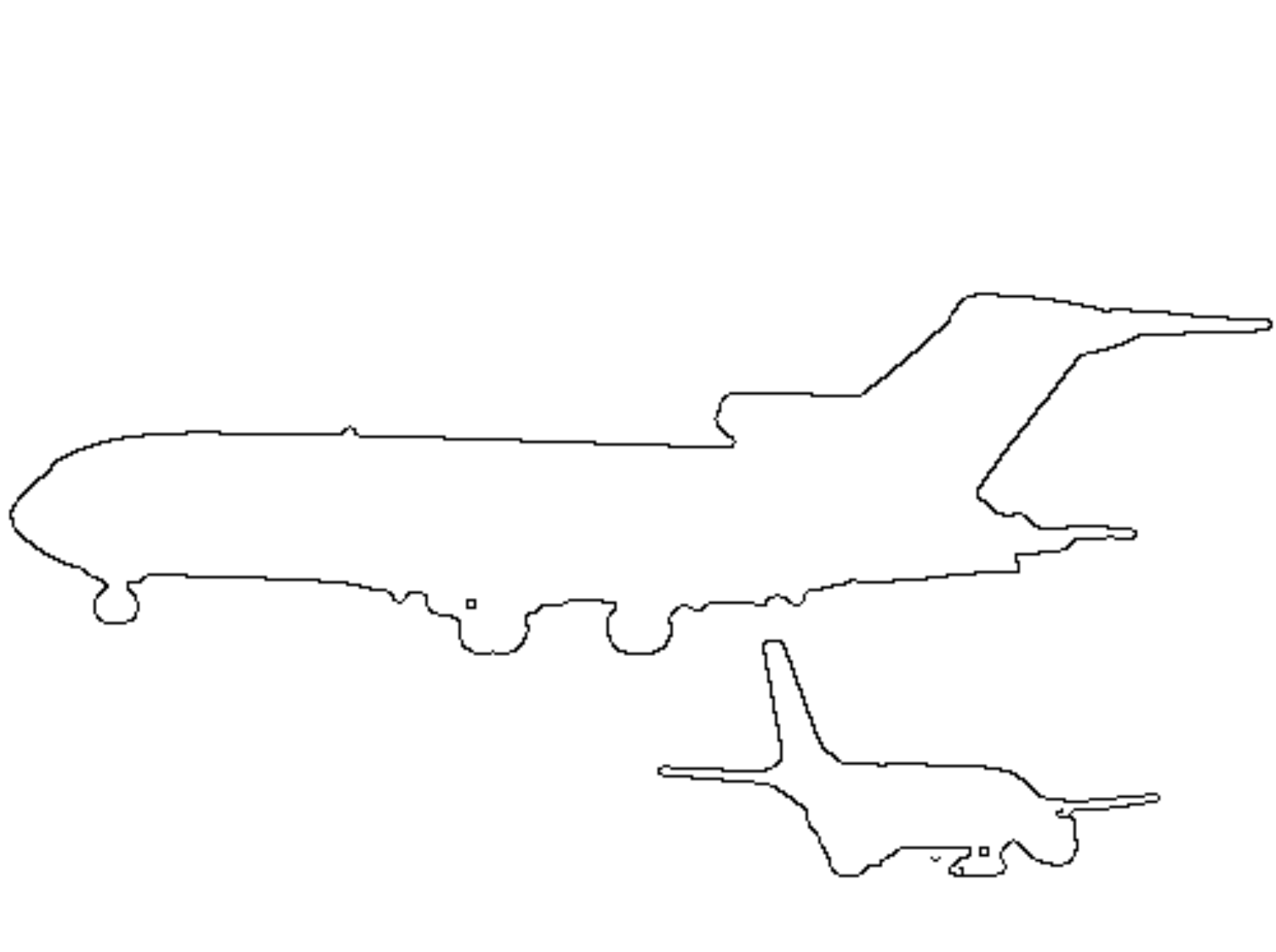} &
		\includegraphics[width=0.19\linewidth]{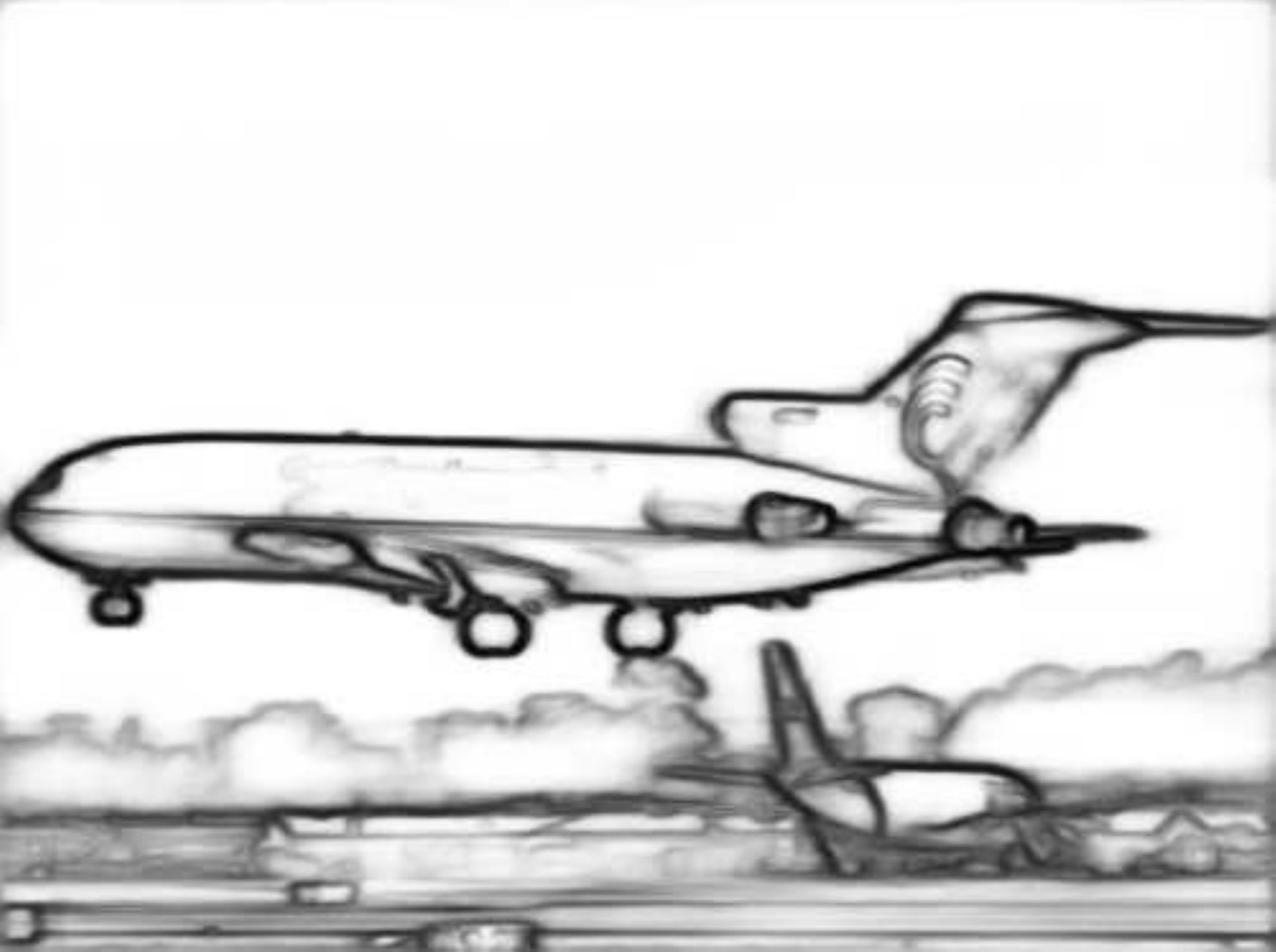} &
		\includegraphics[width=0.19\linewidth]{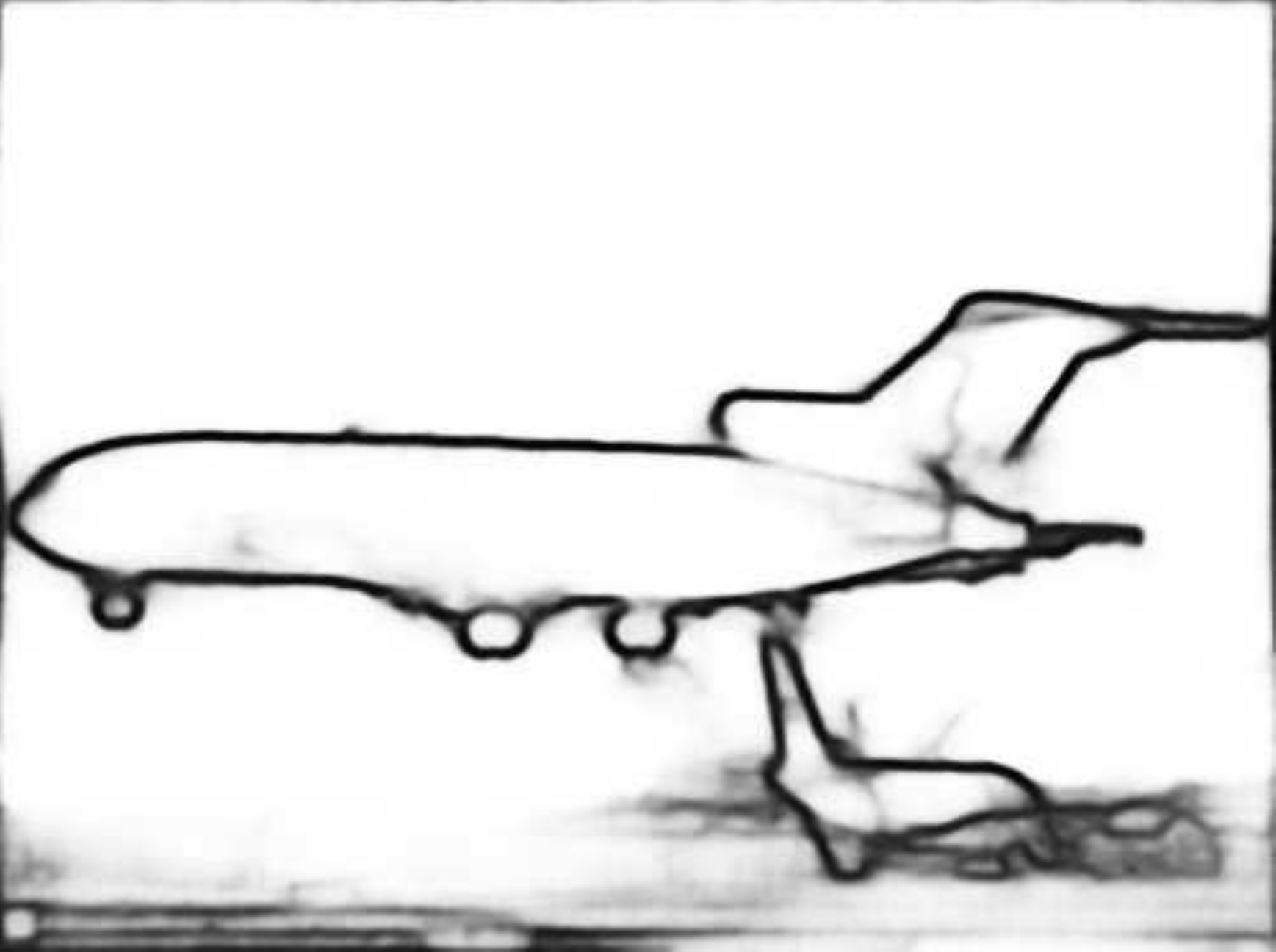} &
		\includegraphics[width=0.19\linewidth]{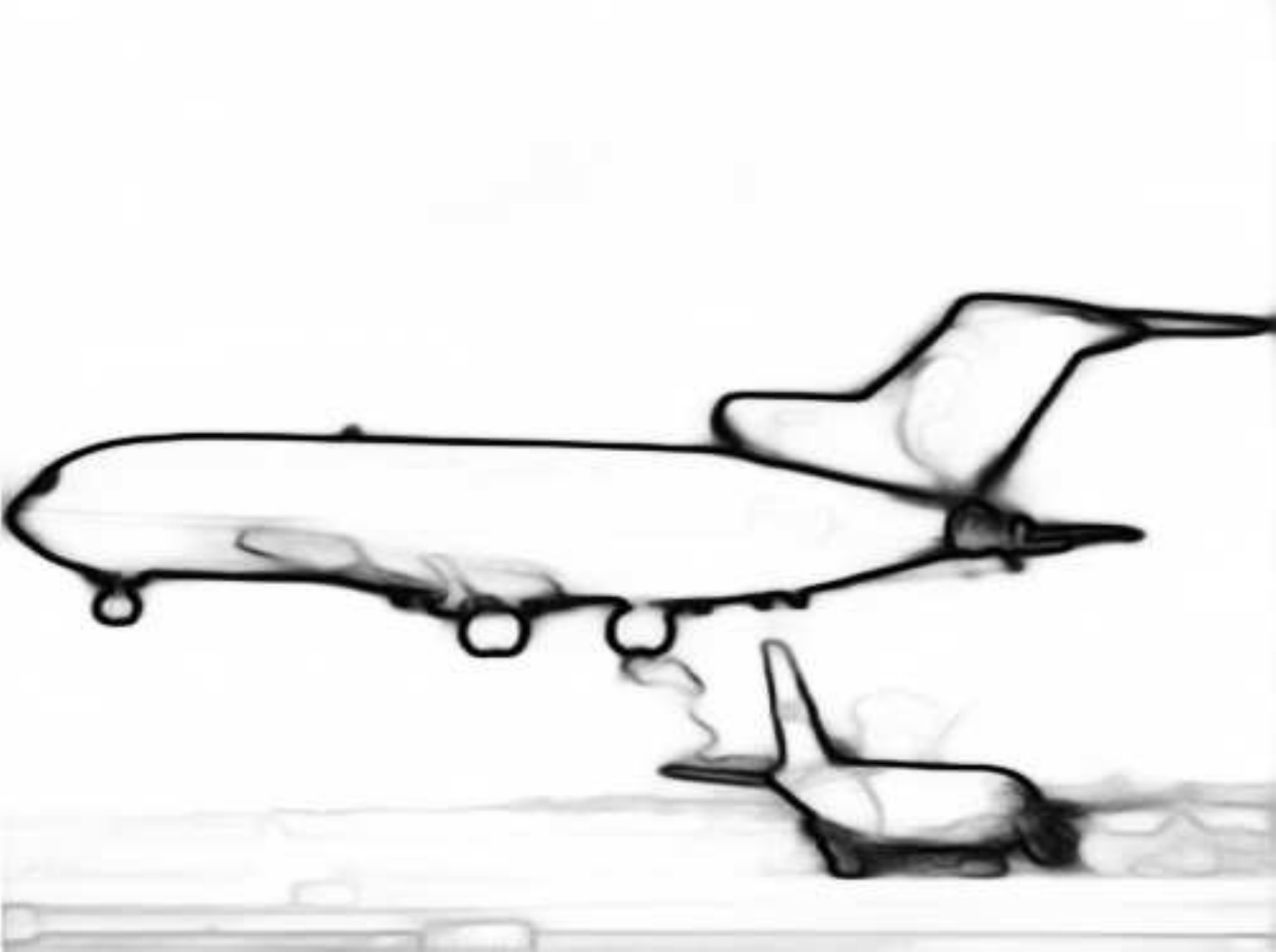} \\
		\includegraphics[width=0.19\linewidth]{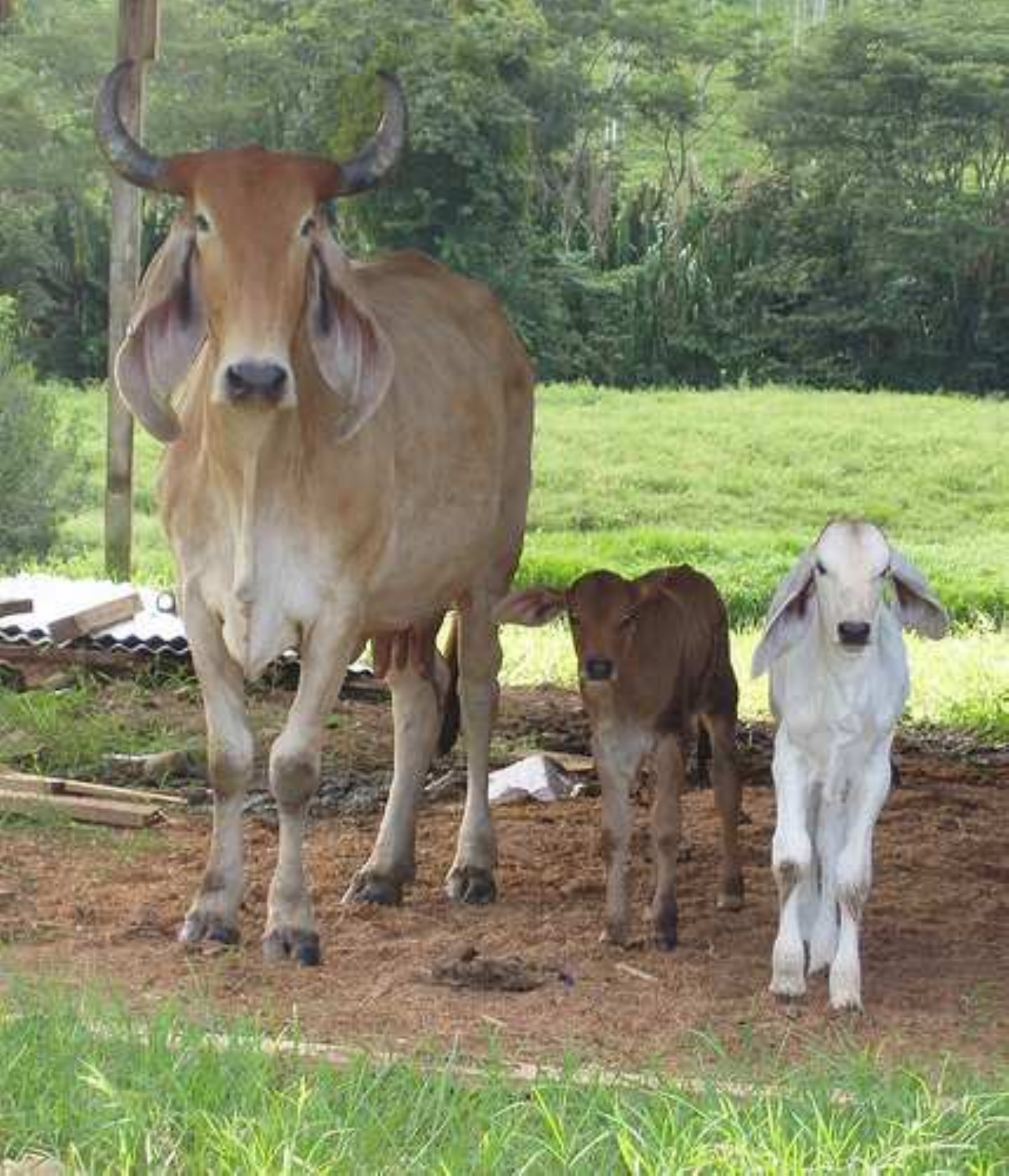} &
		\includegraphics[width=0.19\linewidth]{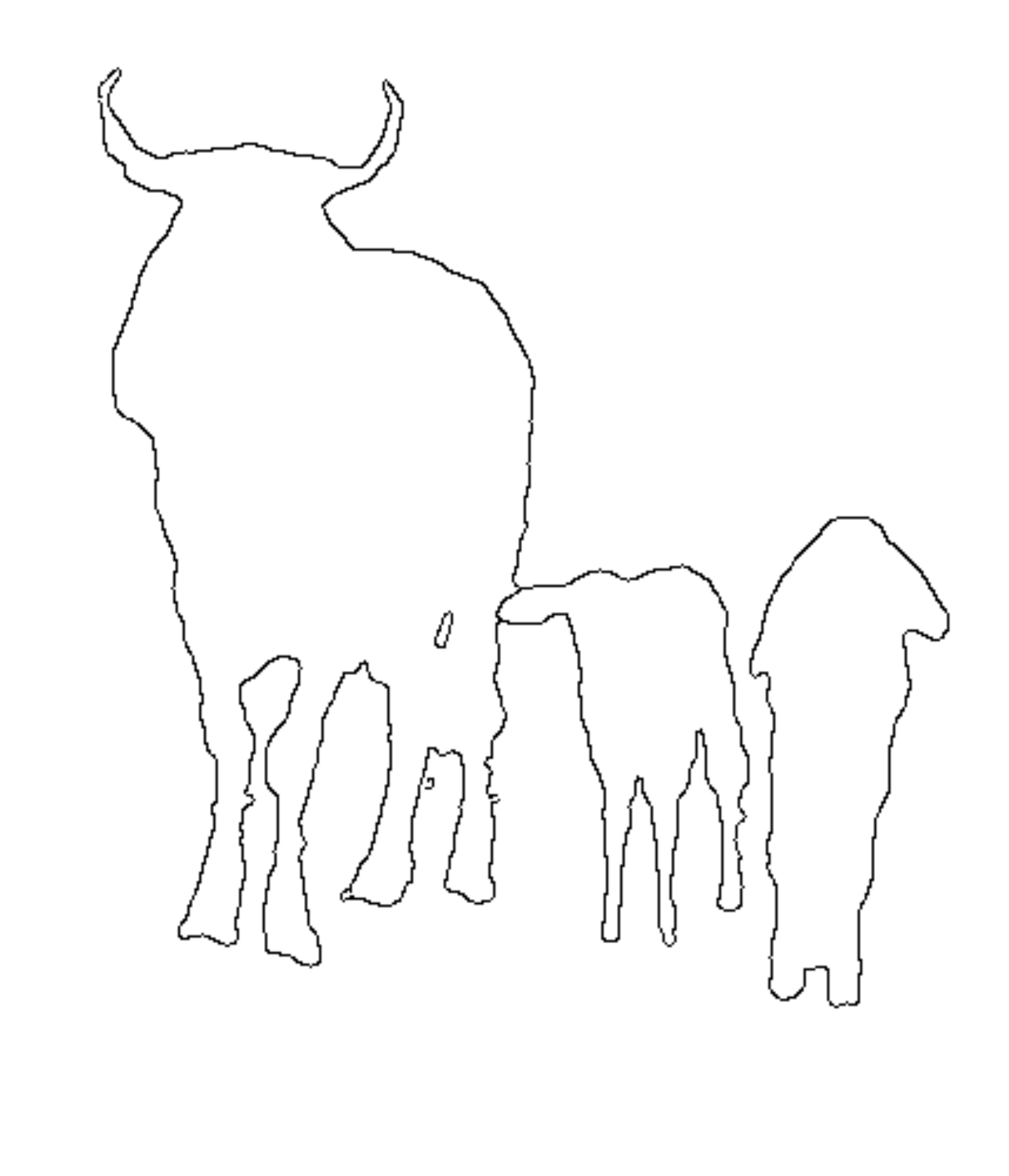} &
		\includegraphics[width=0.19\linewidth]{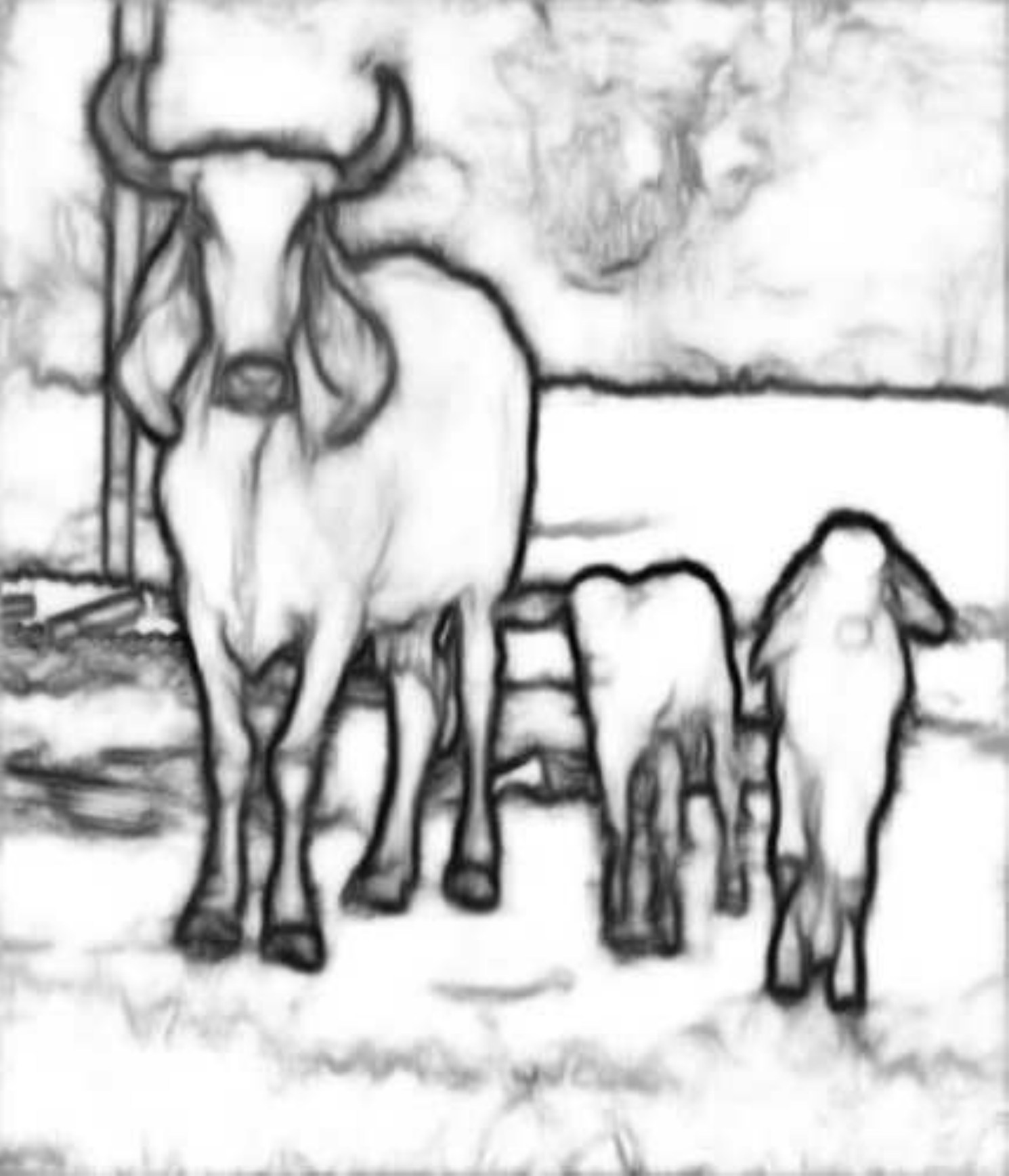} &
		\includegraphics[width=0.19\linewidth]{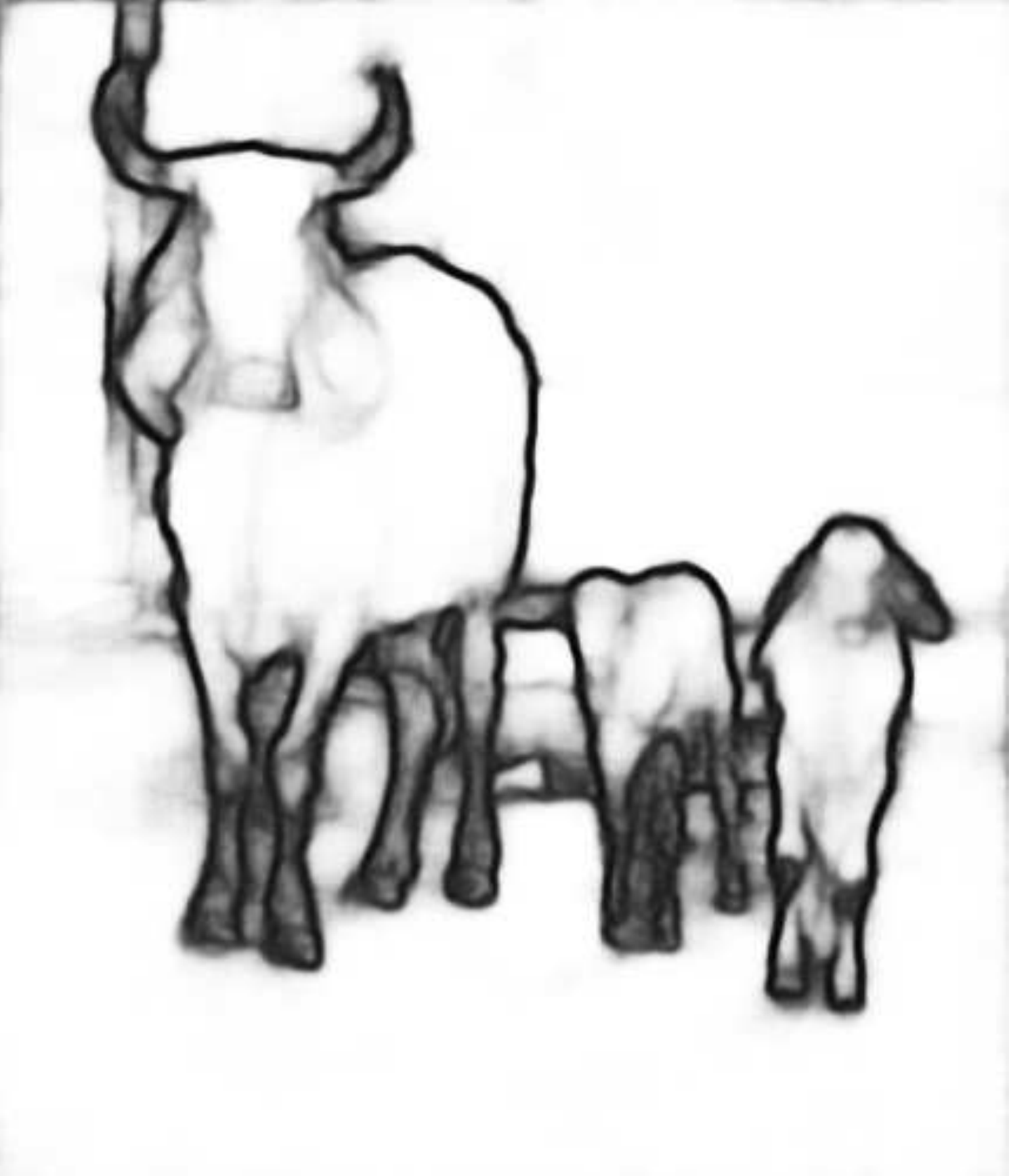} &
		\includegraphics[width=0.19\linewidth]{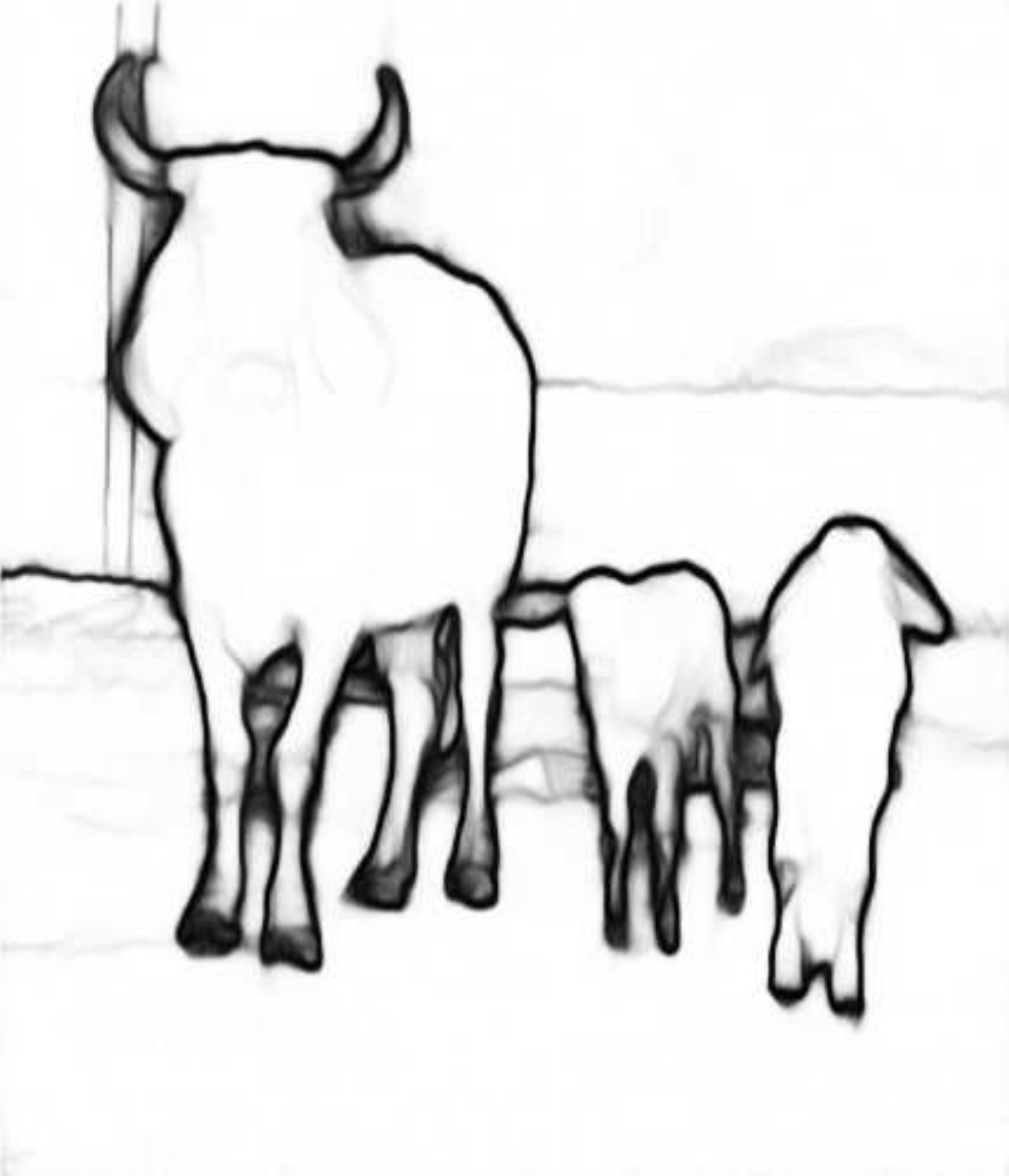} \\
		Raw image & Ground truth & HED-ft & CEDN & TD-CEDN-ft (ours)\\
	\end{tabular}
	\caption{Example results on the PASCAL VOC 2012 validation dataset. }
	\label{Fig:voc-ours-cedn}
\end{figure*}

Fig.~\ref{Fig:voc12} presents the evaluation results on the VOC 2012 validation dataset. Our fine-tuned model achieved the best ODS F-score of 0.588. ``HED-over3" and ``TD-CEDN-over3 (ours)" seem to have a similar performance when they were applied directly on the validation dataset. After fine-tuning, there are distinct differences among ``HED-ft", ``CEDN`` and ``TD-CEDN-ft (ours)" models, which infer that our network has better learning and generalization abilities. Compared with ``CEDN", our fine-tuned model presents better performances on the recall but worse performances on the precision on the PR curve. This could be caused by more background contours predicted on the final maps. Fig.~\ref{Fig:voc-ours-cedn} shows several results predicted by ``HED-ft", ``CEDN" and ``TD-CEDN-ft (ours)" models on the validation dataset. Our method not only provides  accurate predictions but also presents a clear and tidy perception on visual effect.

\begin{figure}[!t]
	\centering
	\includegraphics[width=\linewidth]{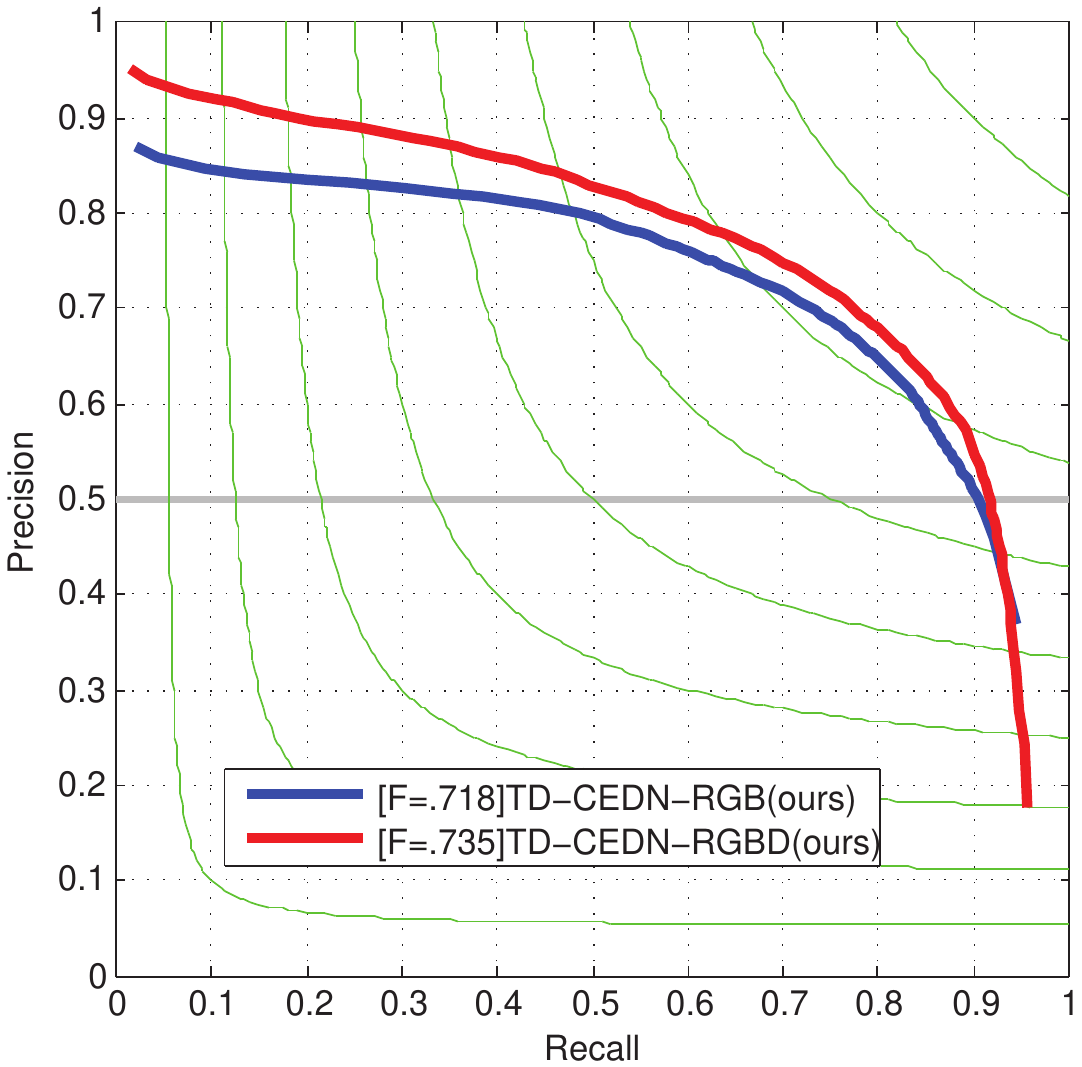}\\
	\centering
	\caption{The PR curve for contour detection on the NYUDv2 testing dataset. Our fine-tuned model achieved the state-of-the-art ODS F-score of 0.735.}
	\label{Fig:nyud}
\end{figure}

\begin{figure*}[!t]
	\small
	\centering
	\renewcommand{\tabcolsep}{2pt}
	\begin{tabular}{cccccc}
		\includegraphics[width=0.16\linewidth]{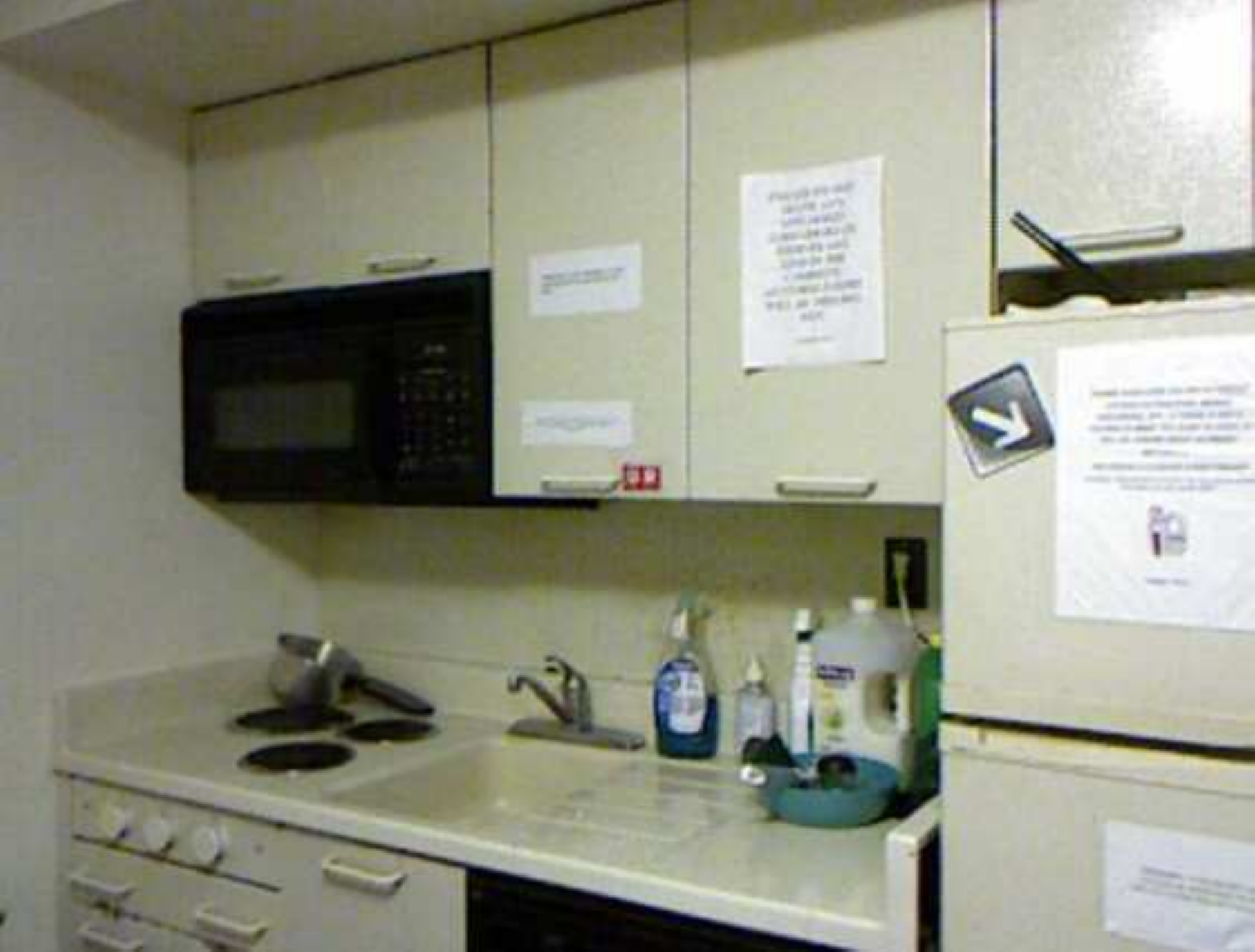} &
		\includegraphics[width=0.16\linewidth]{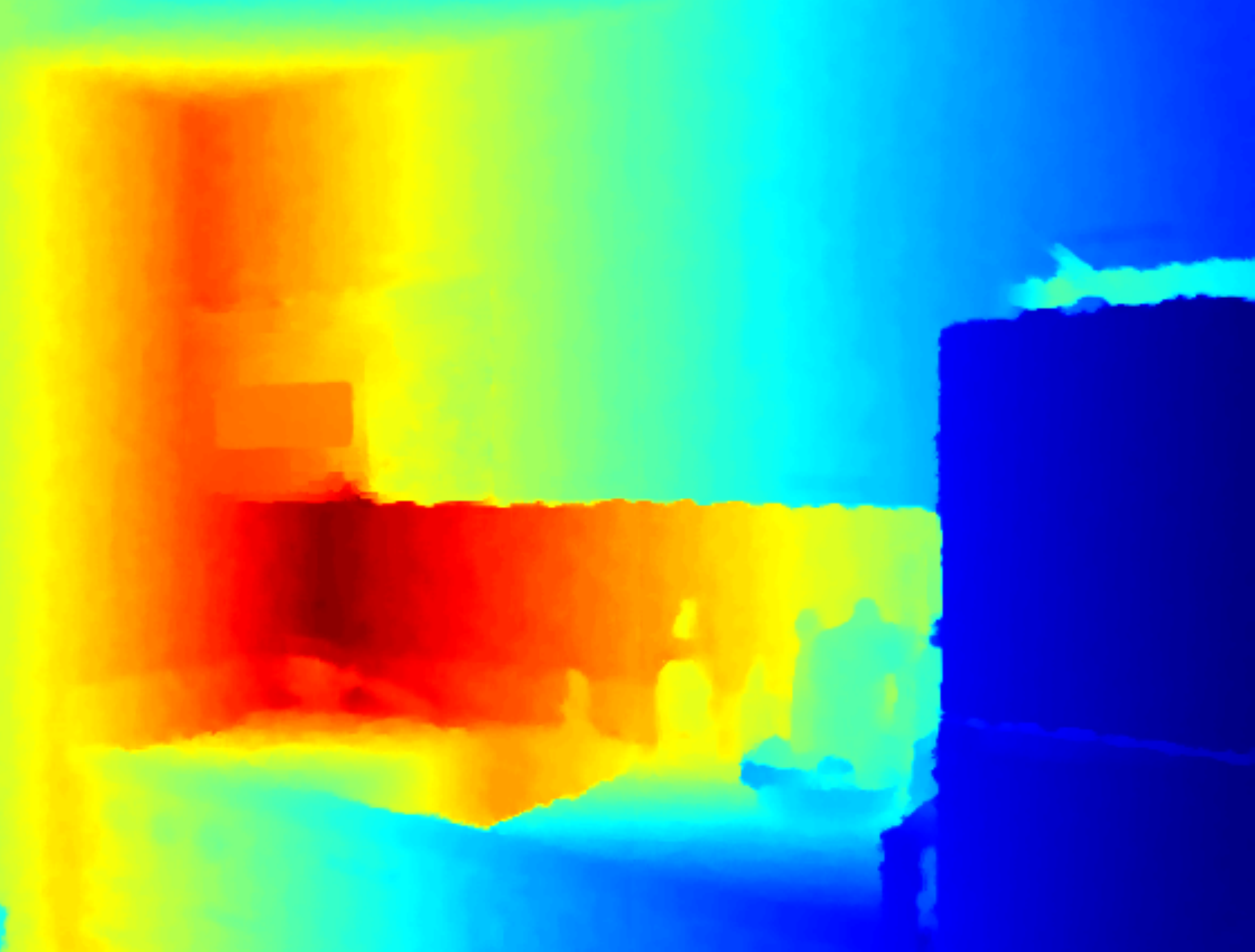} &
		\includegraphics[width=0.16\linewidth]{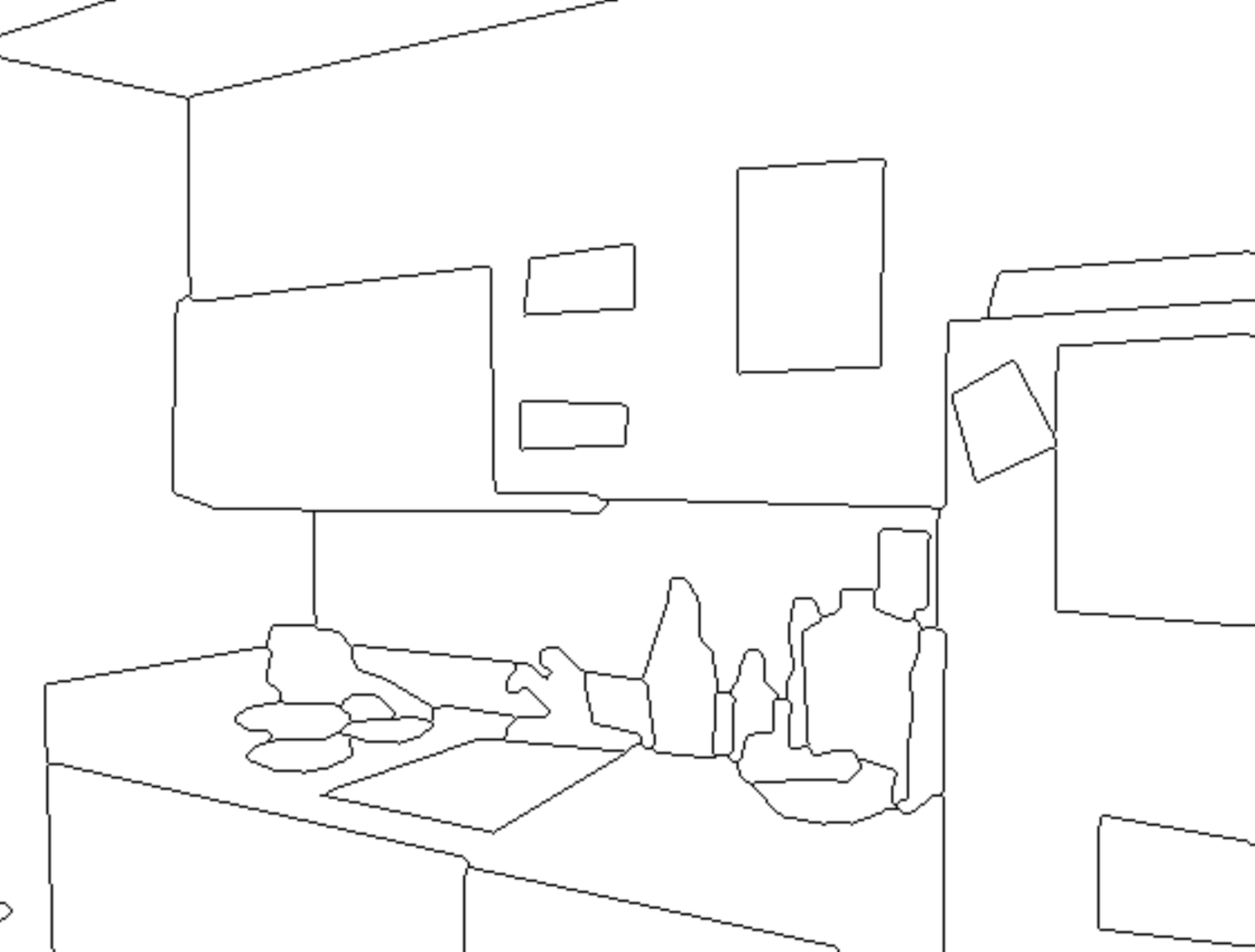} &
		\includegraphics[width=0.16\linewidth]{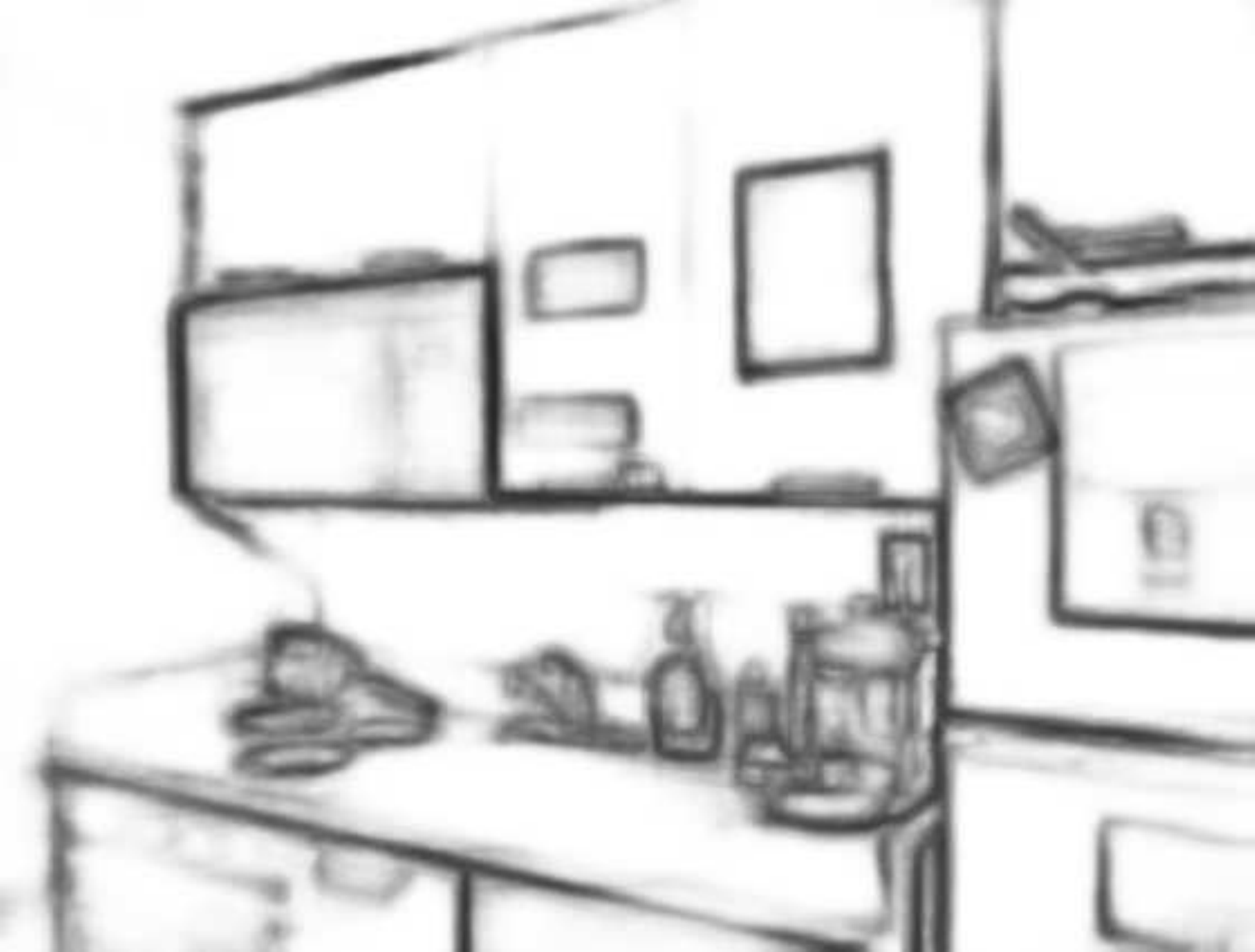} &
		\includegraphics[width=0.16\linewidth]{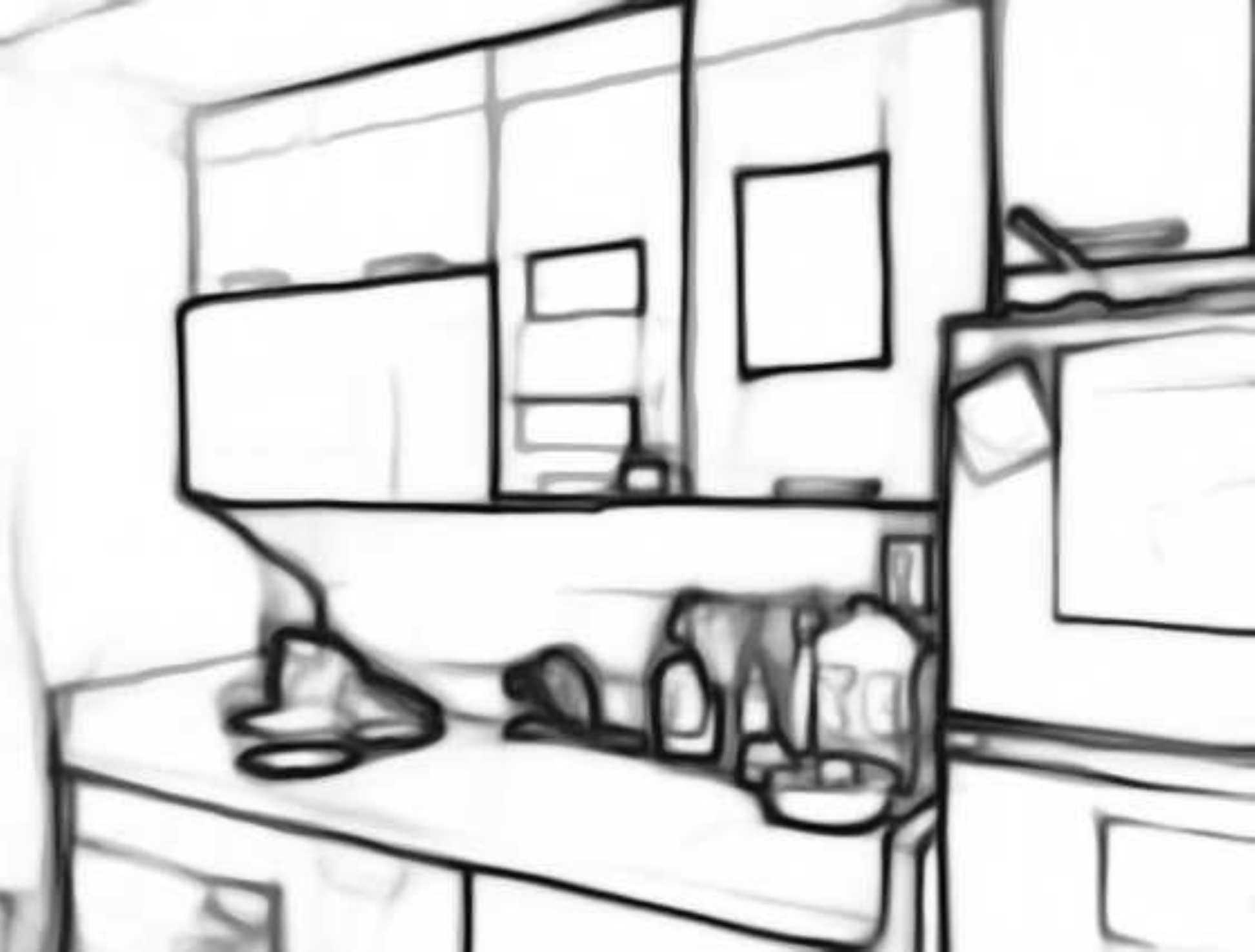} &
		\includegraphics[width=0.16\linewidth]{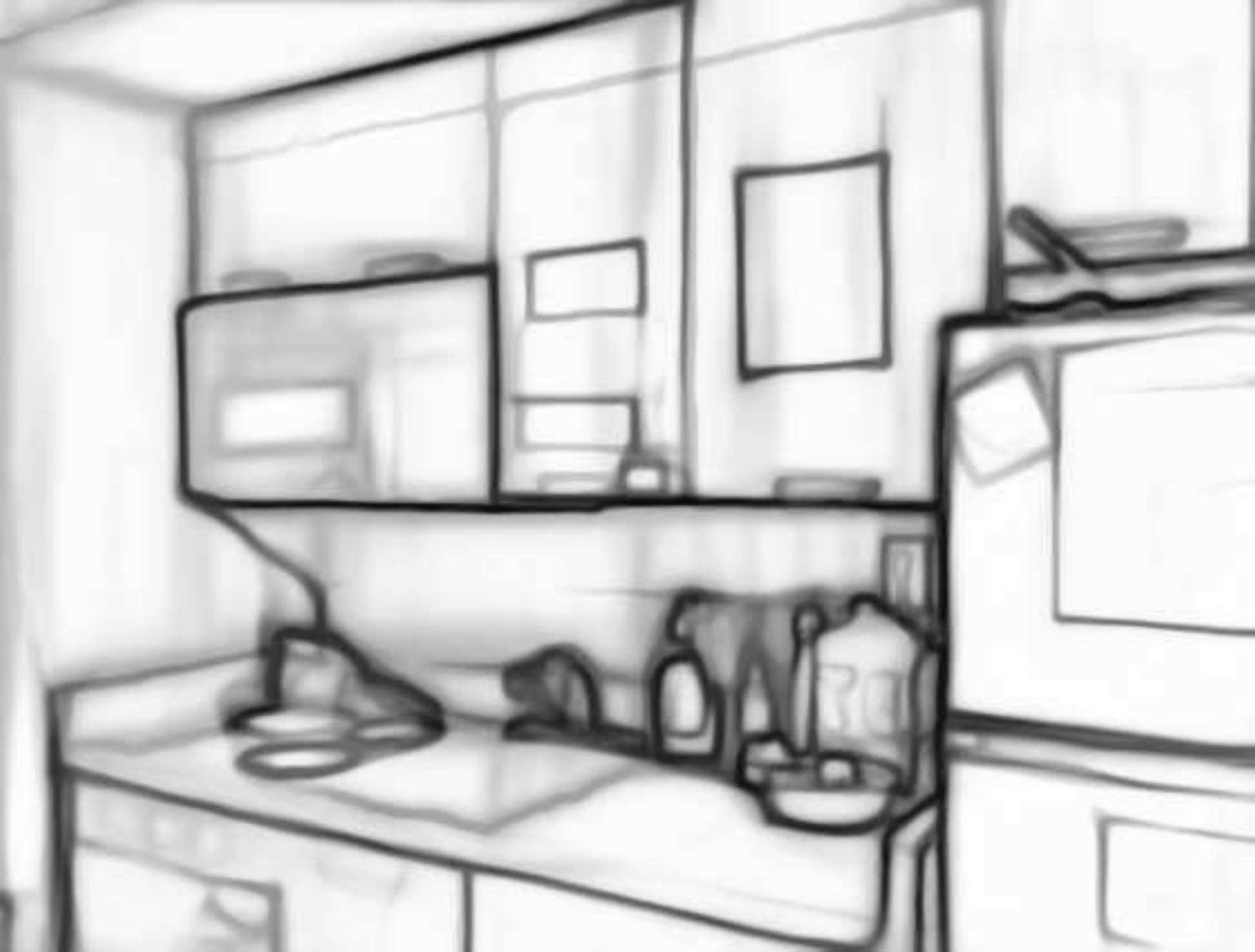} \\
		\includegraphics[width=0.16\linewidth]{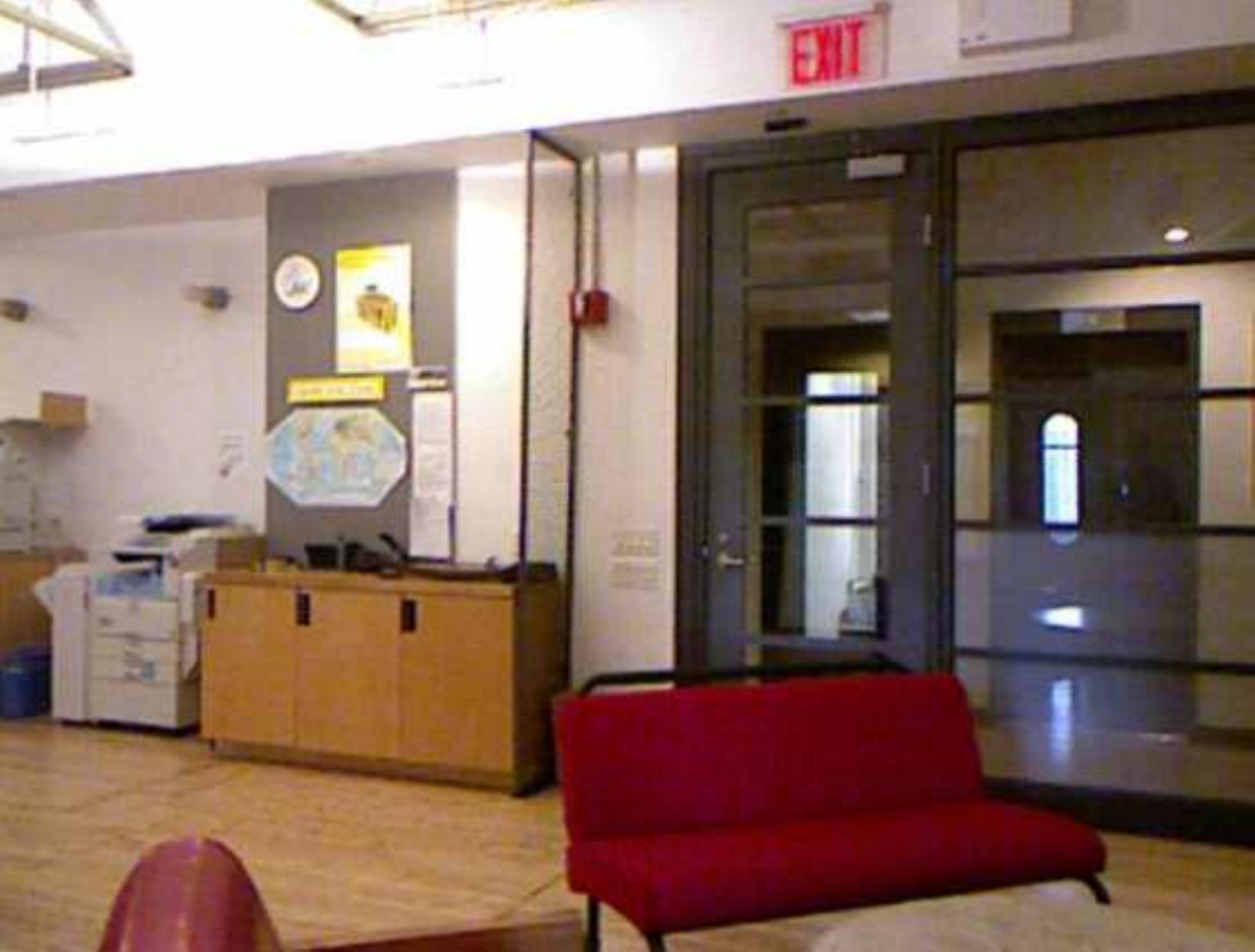} &
		\includegraphics[width=0.16\linewidth]{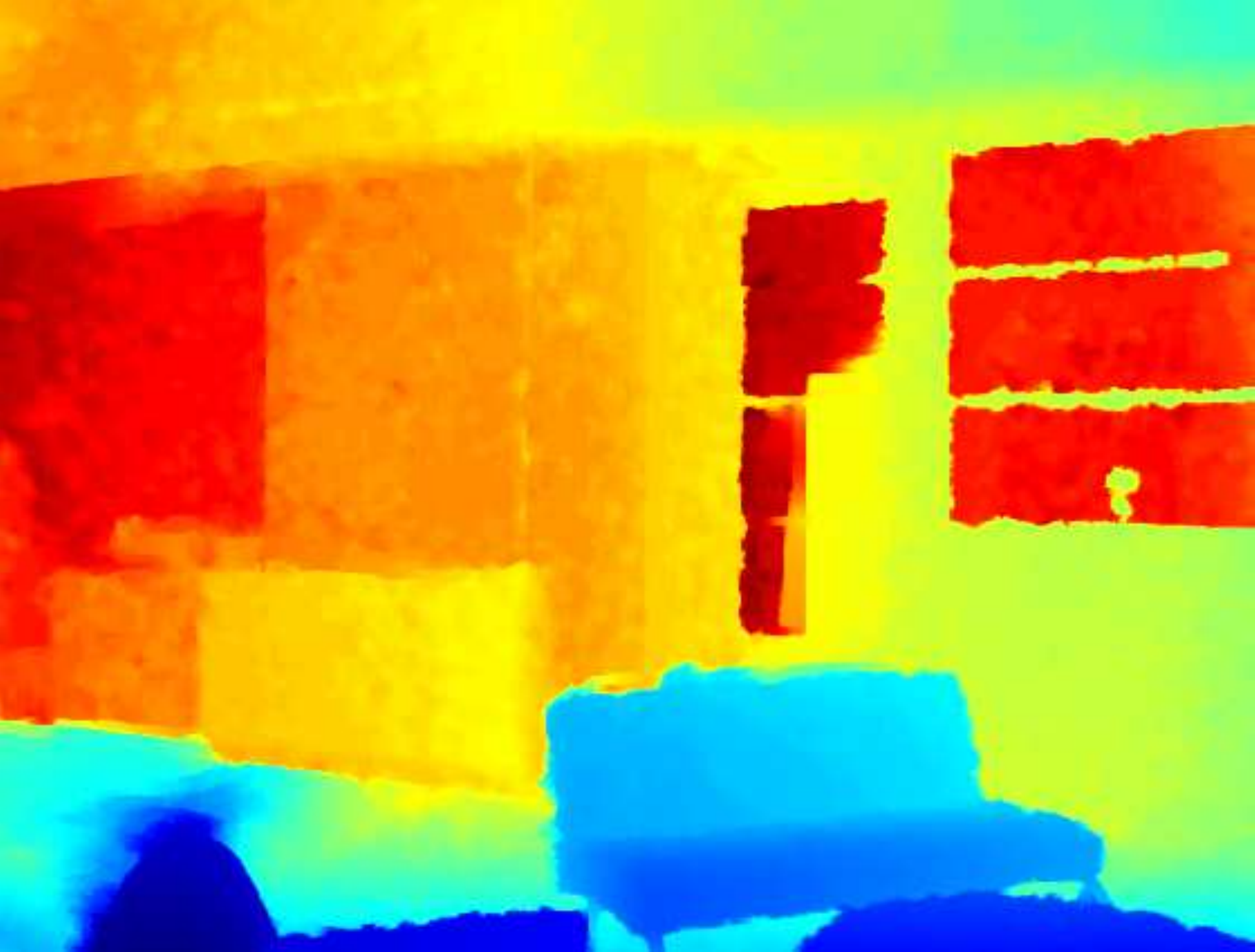} &
		\includegraphics[width=0.16\linewidth]{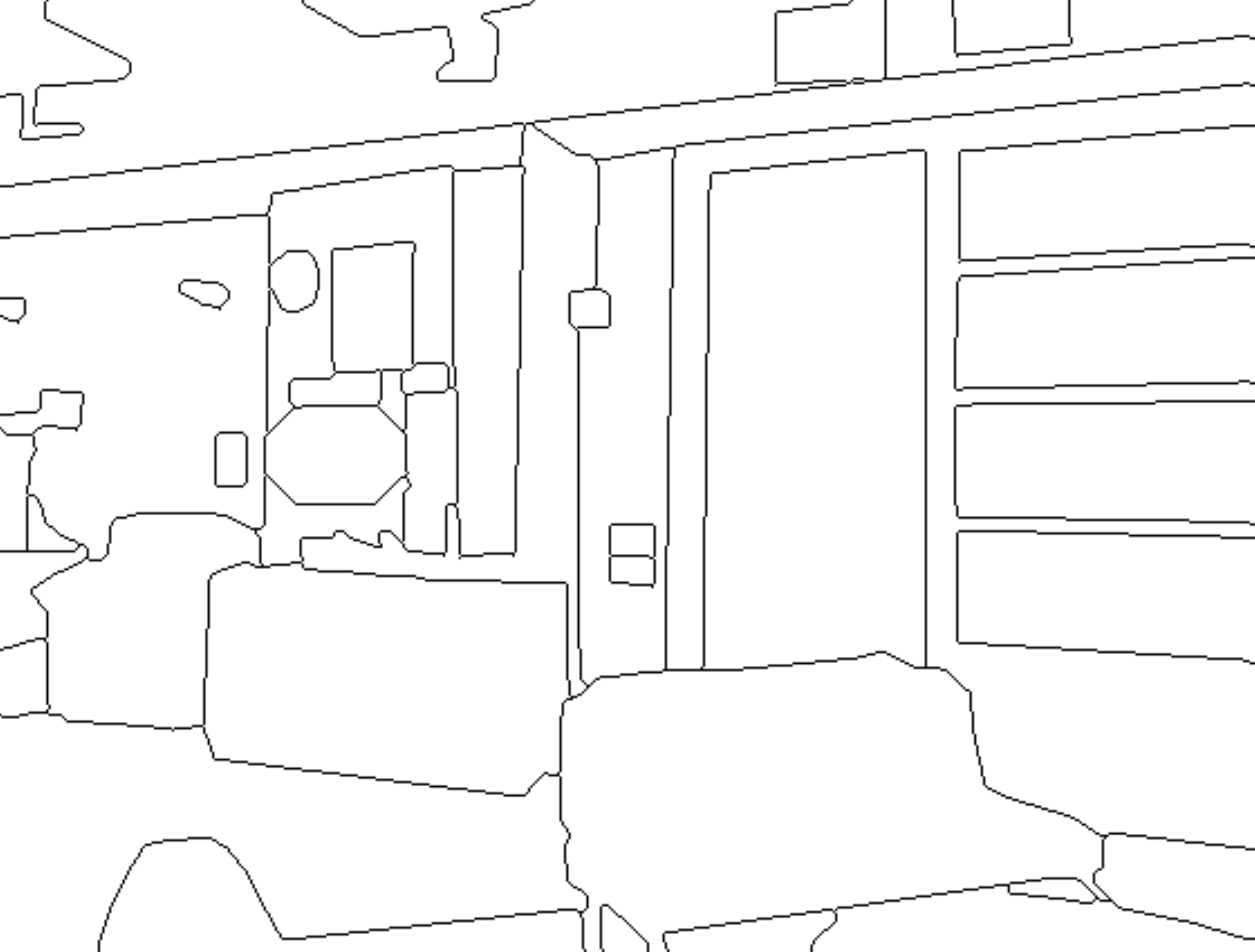} &
		\includegraphics[width=0.16\linewidth]{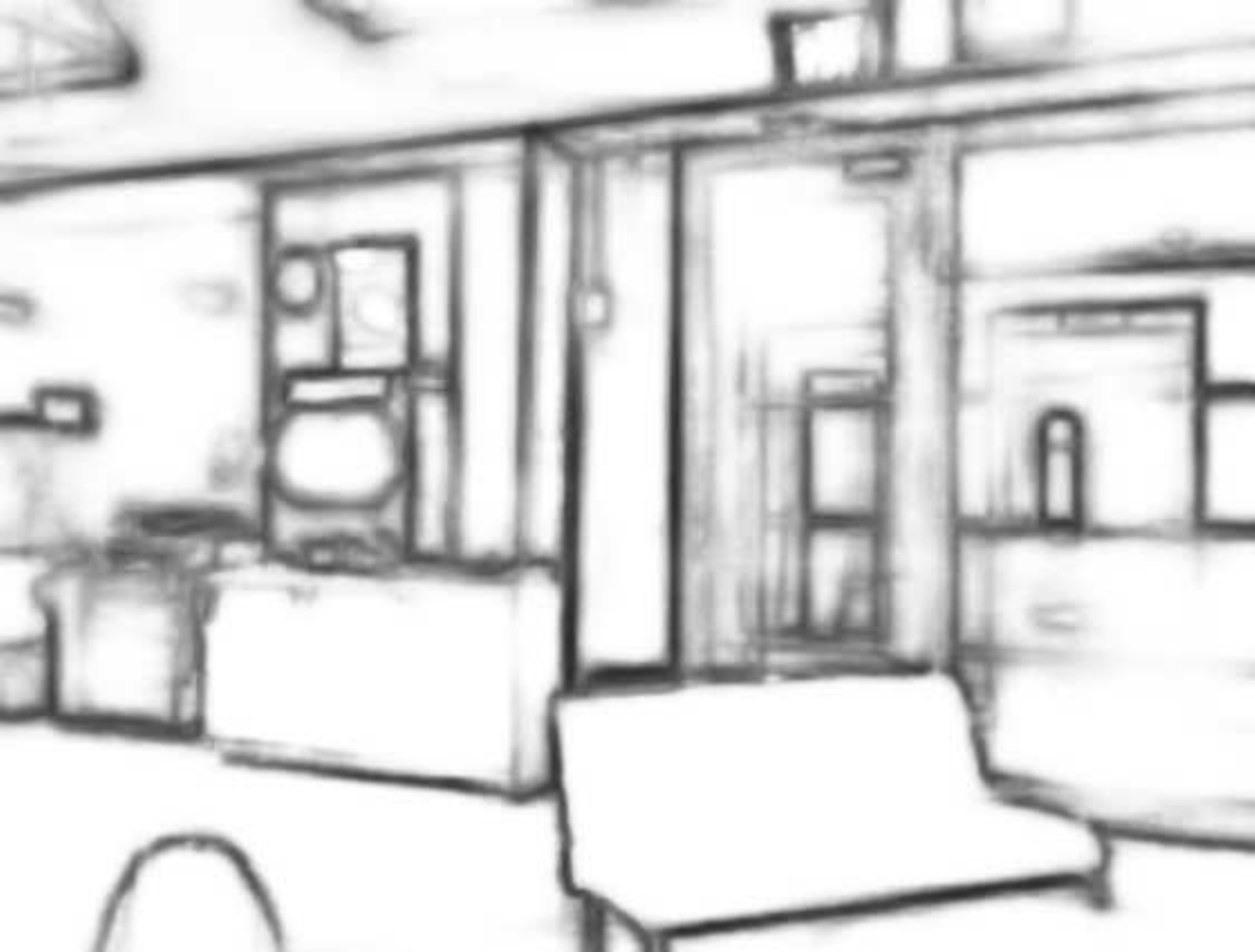} &
		\includegraphics[width=0.16\linewidth]{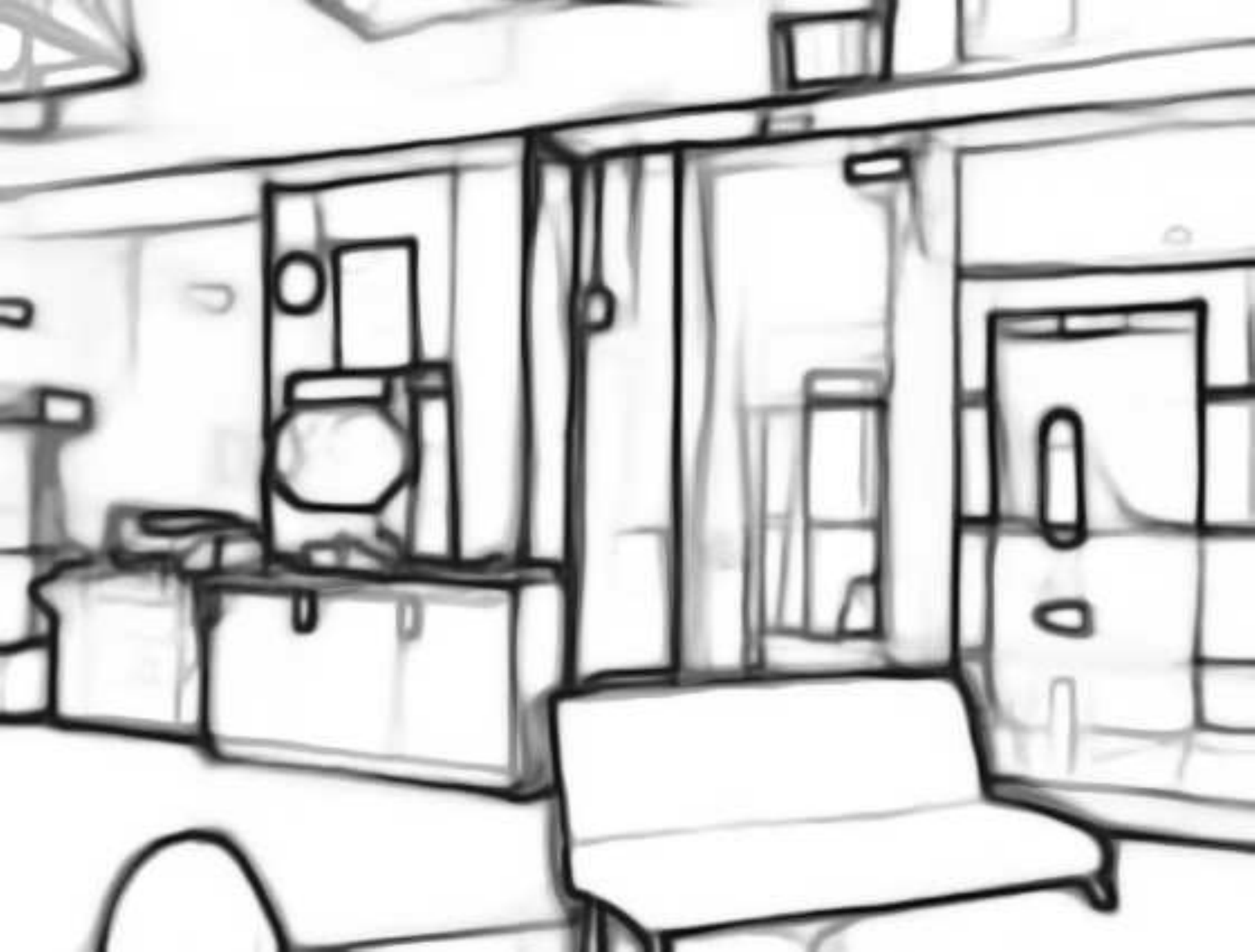} &
		\includegraphics[width=0.16\linewidth]{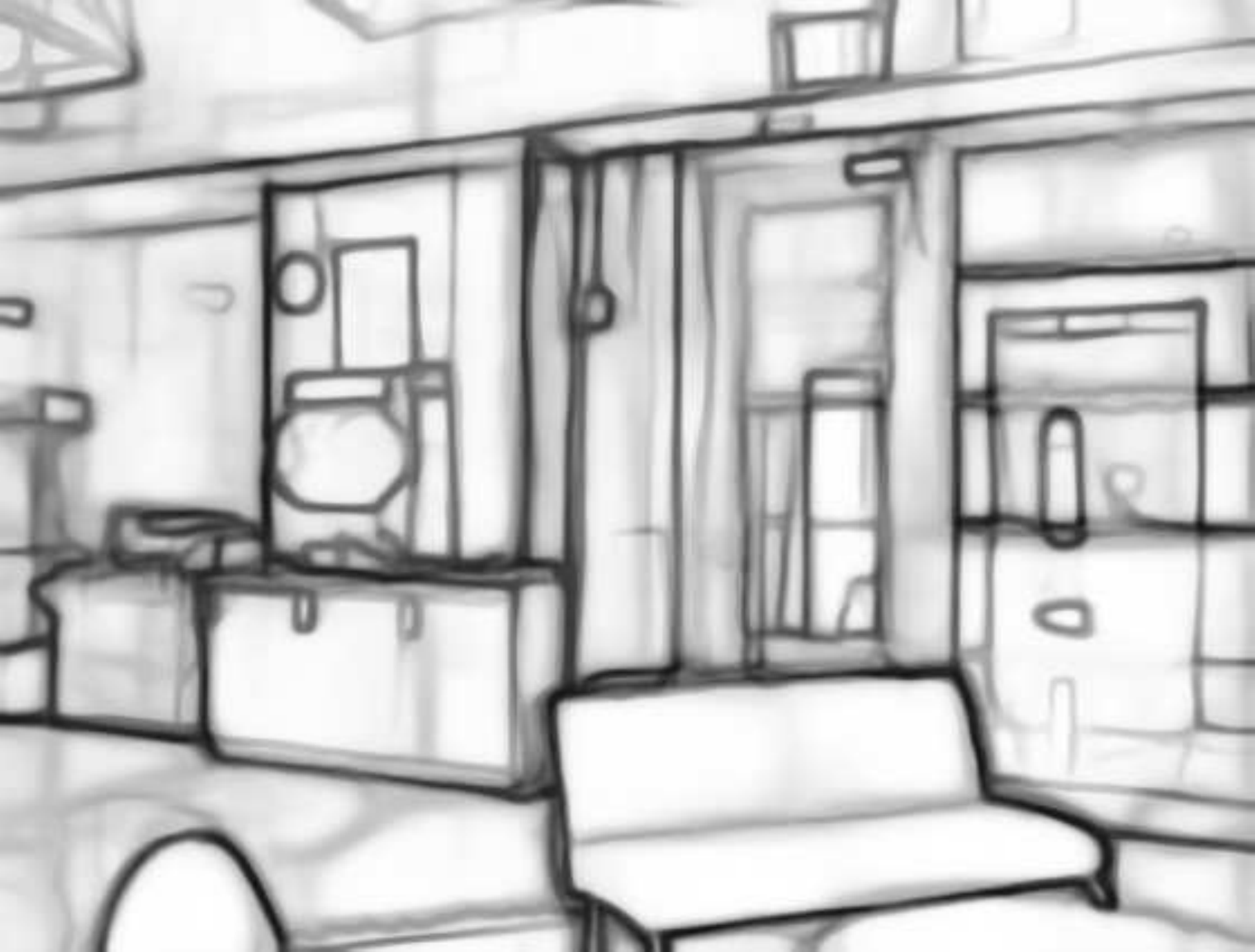} \\
		\includegraphics[width=0.16\linewidth]{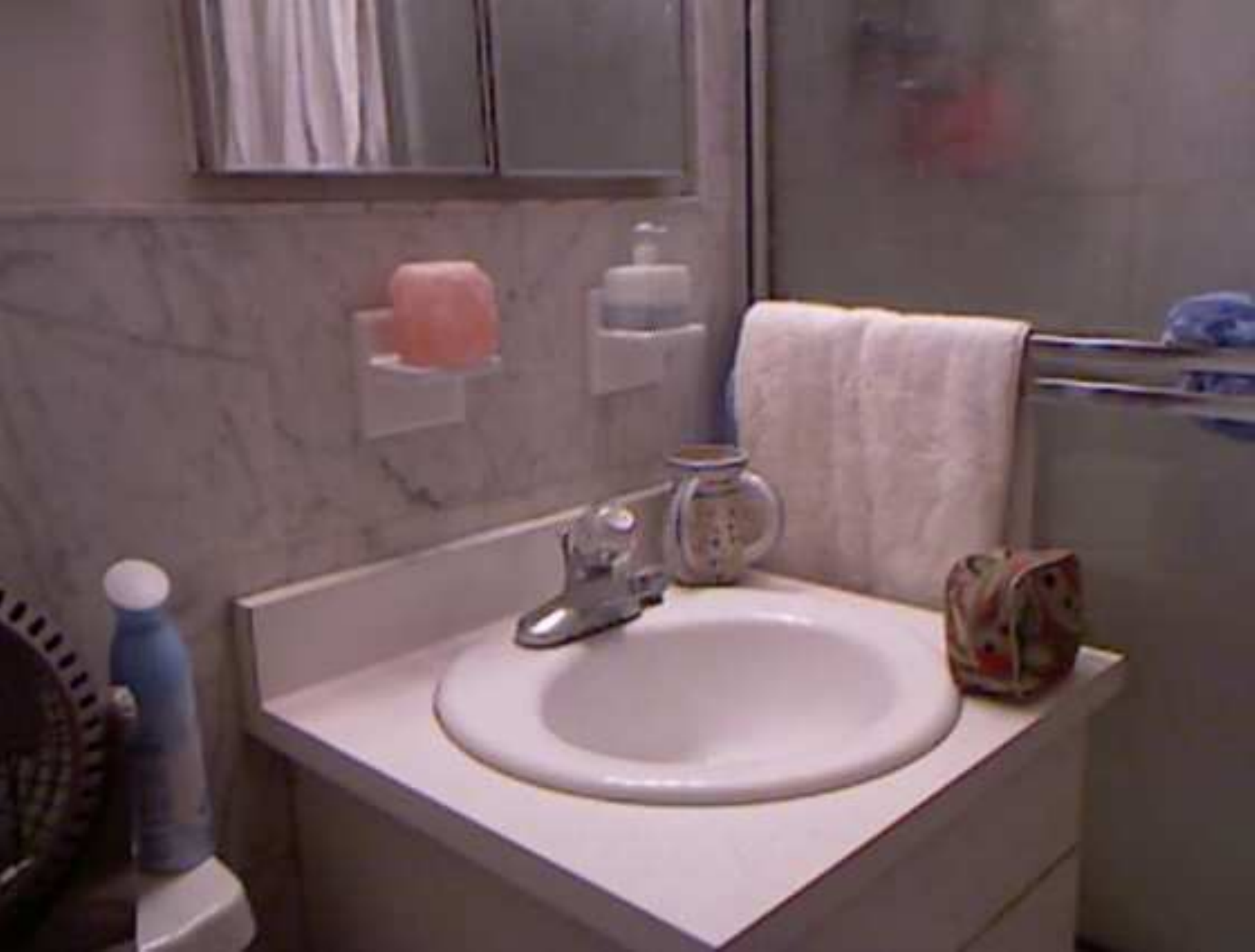} &
		\includegraphics[width=0.16\linewidth]{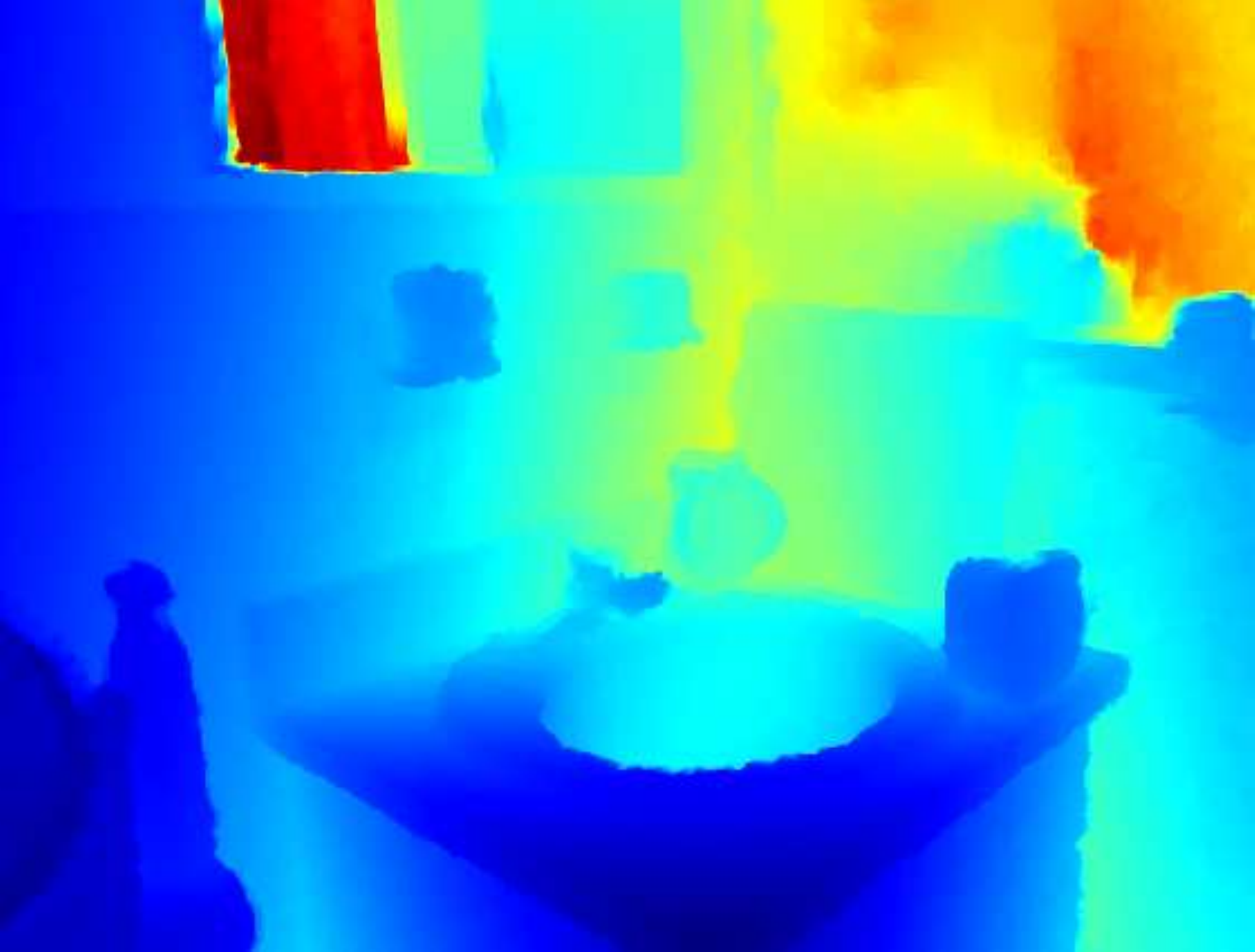} &
		\includegraphics[width=0.16\linewidth]{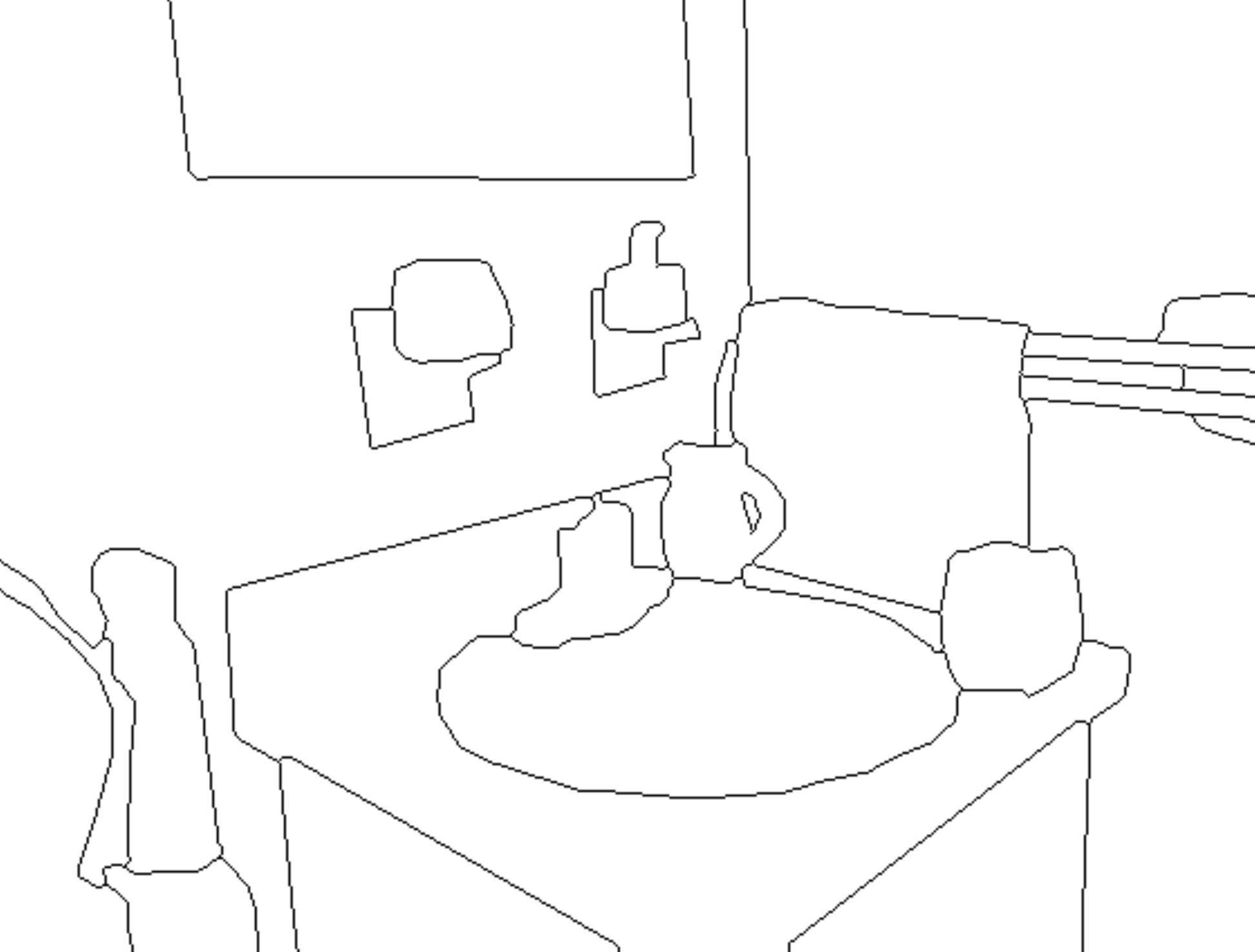} &
		\includegraphics[width=0.16\linewidth]{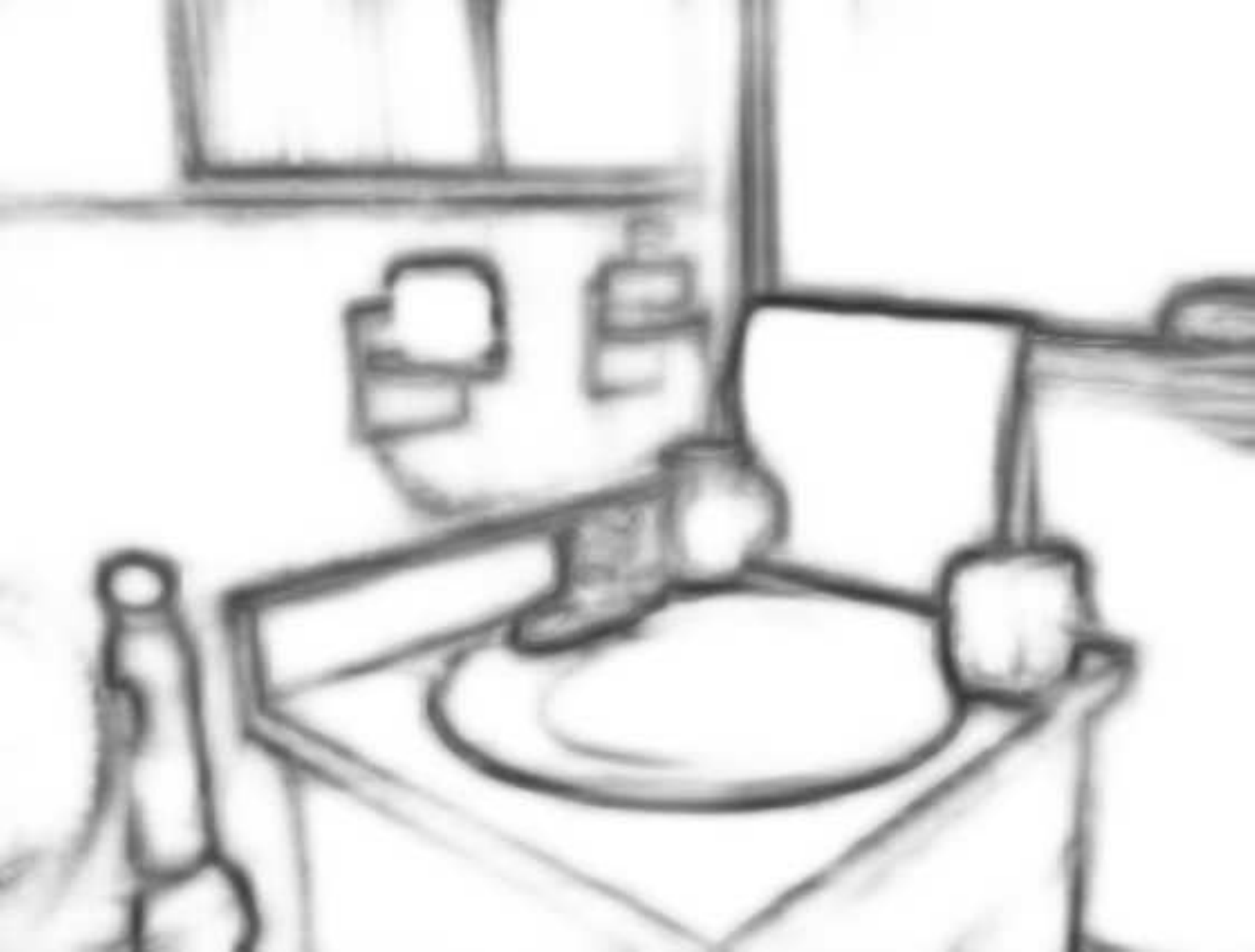} &
		\includegraphics[width=0.16\linewidth]{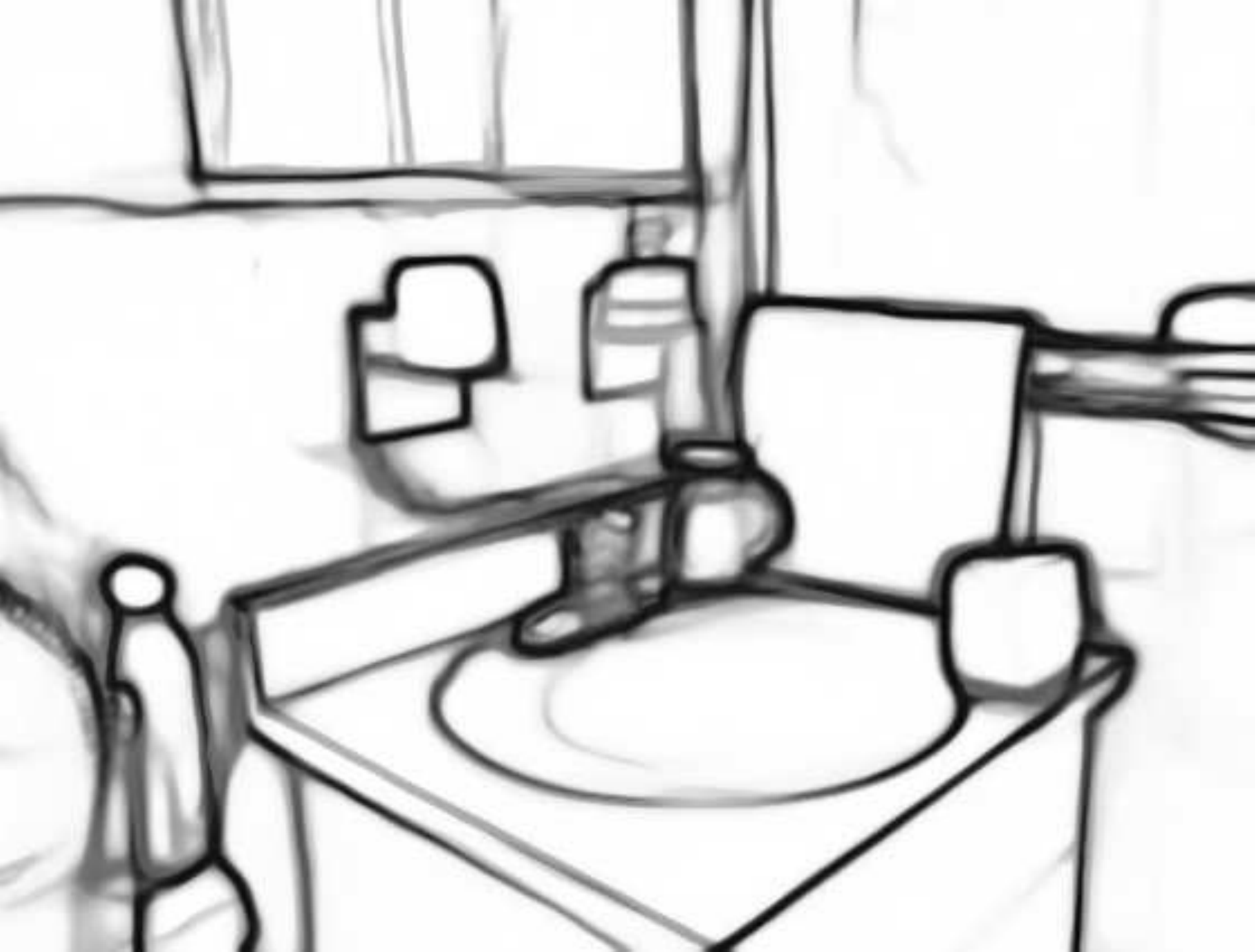} &
		\includegraphics[width=0.16\linewidth]{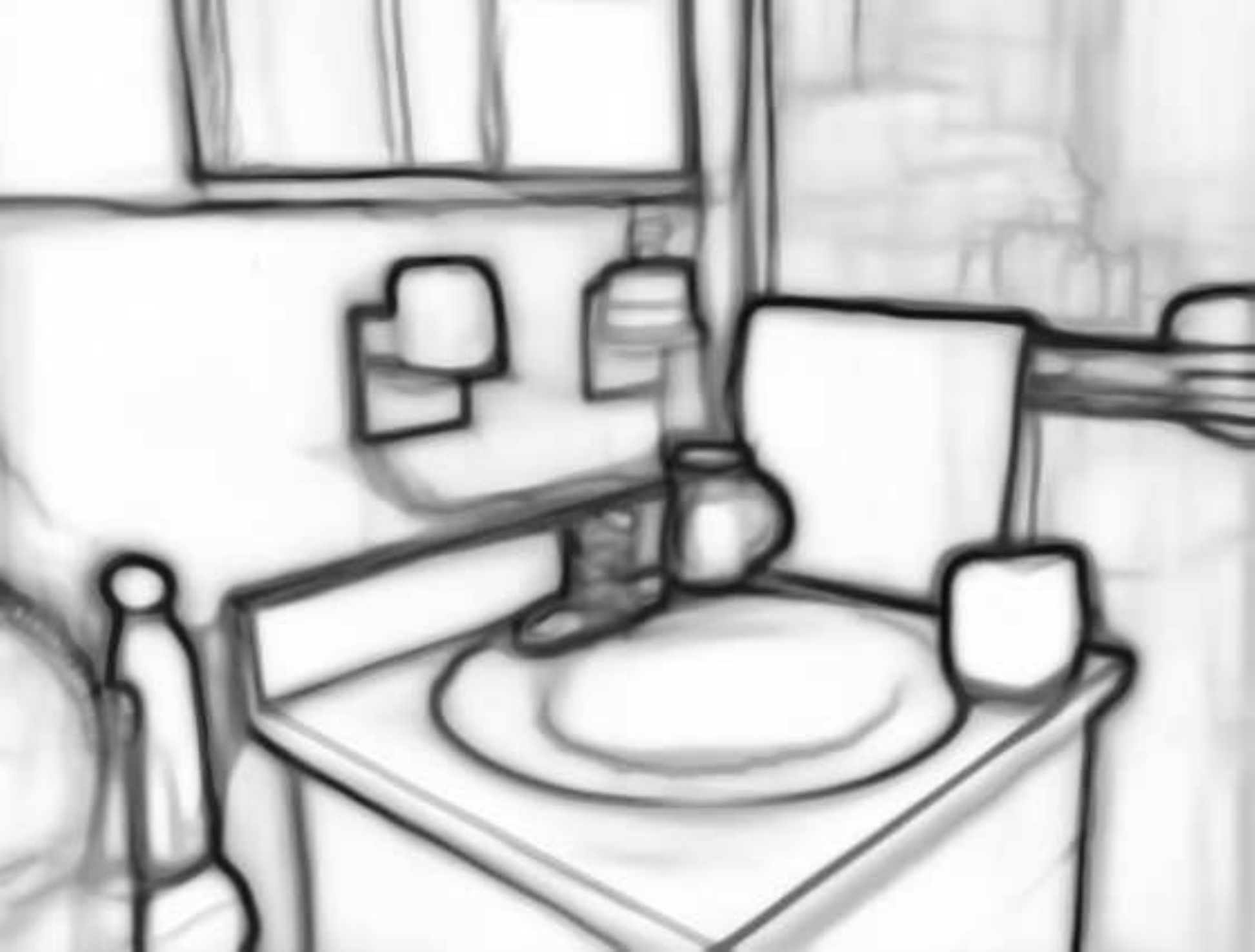} \\
		\includegraphics[width=0.16\linewidth]{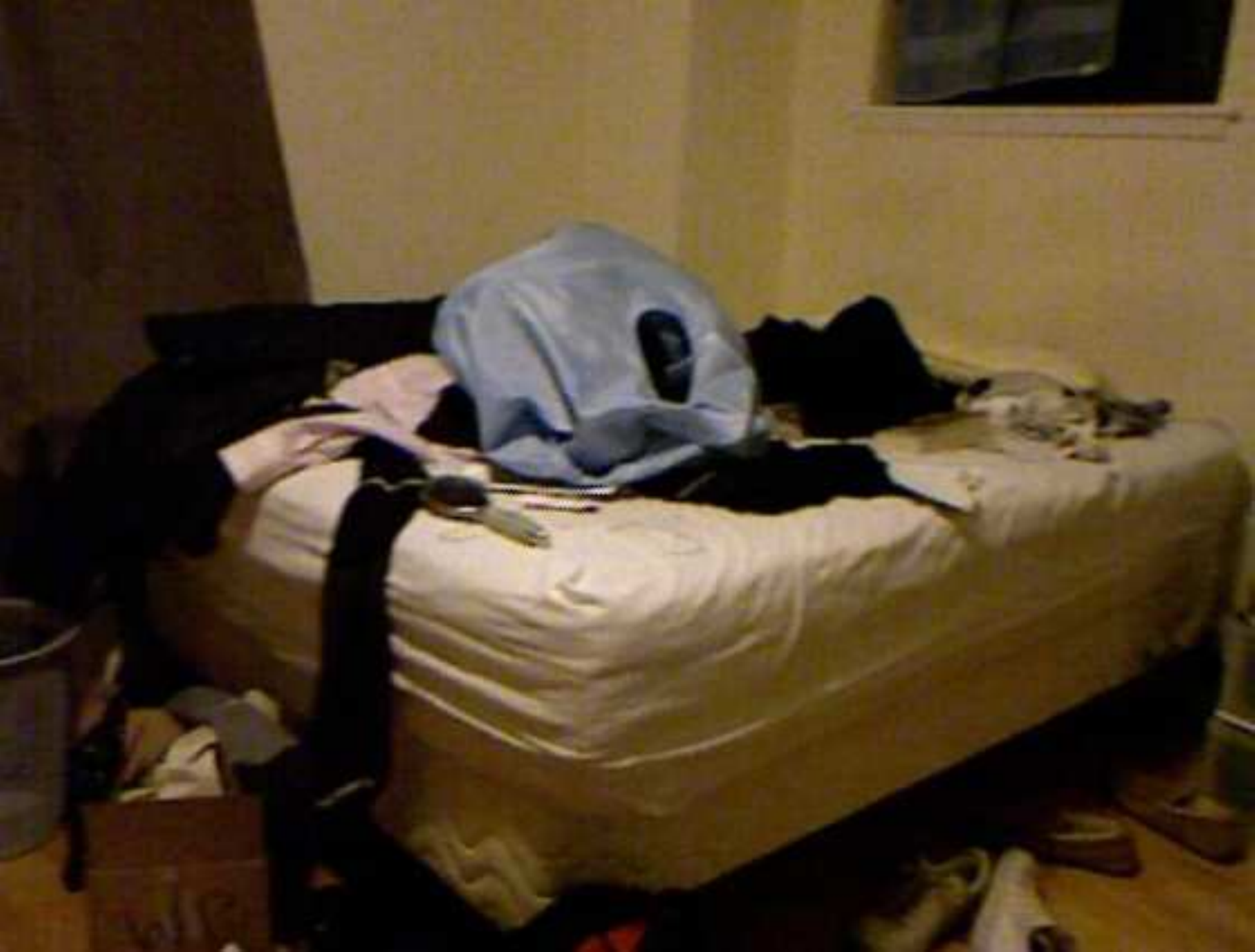} &
		\includegraphics[width=0.16\linewidth]{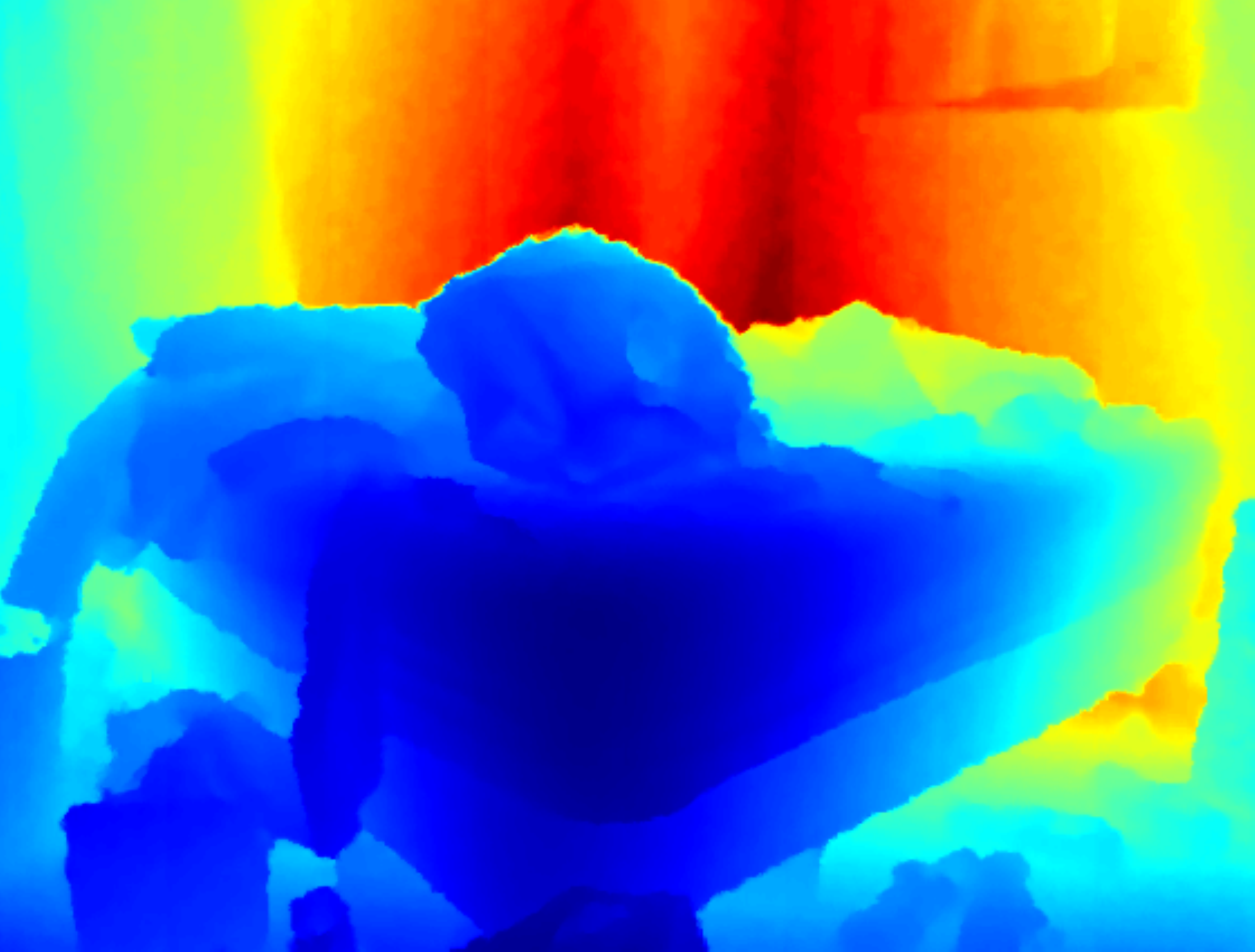} &
		\includegraphics[width=0.16\linewidth]{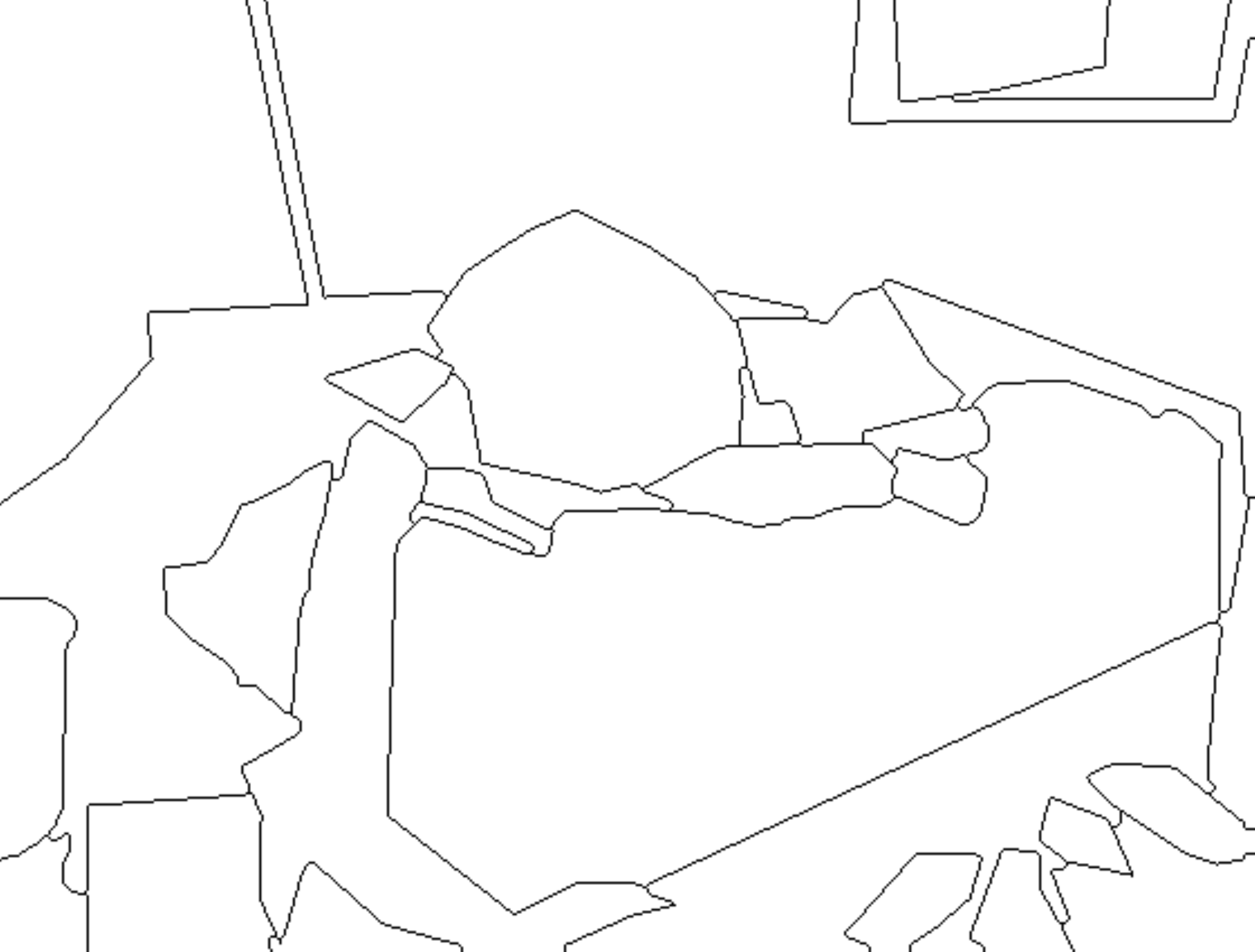} &
		\includegraphics[width=0.16\linewidth]{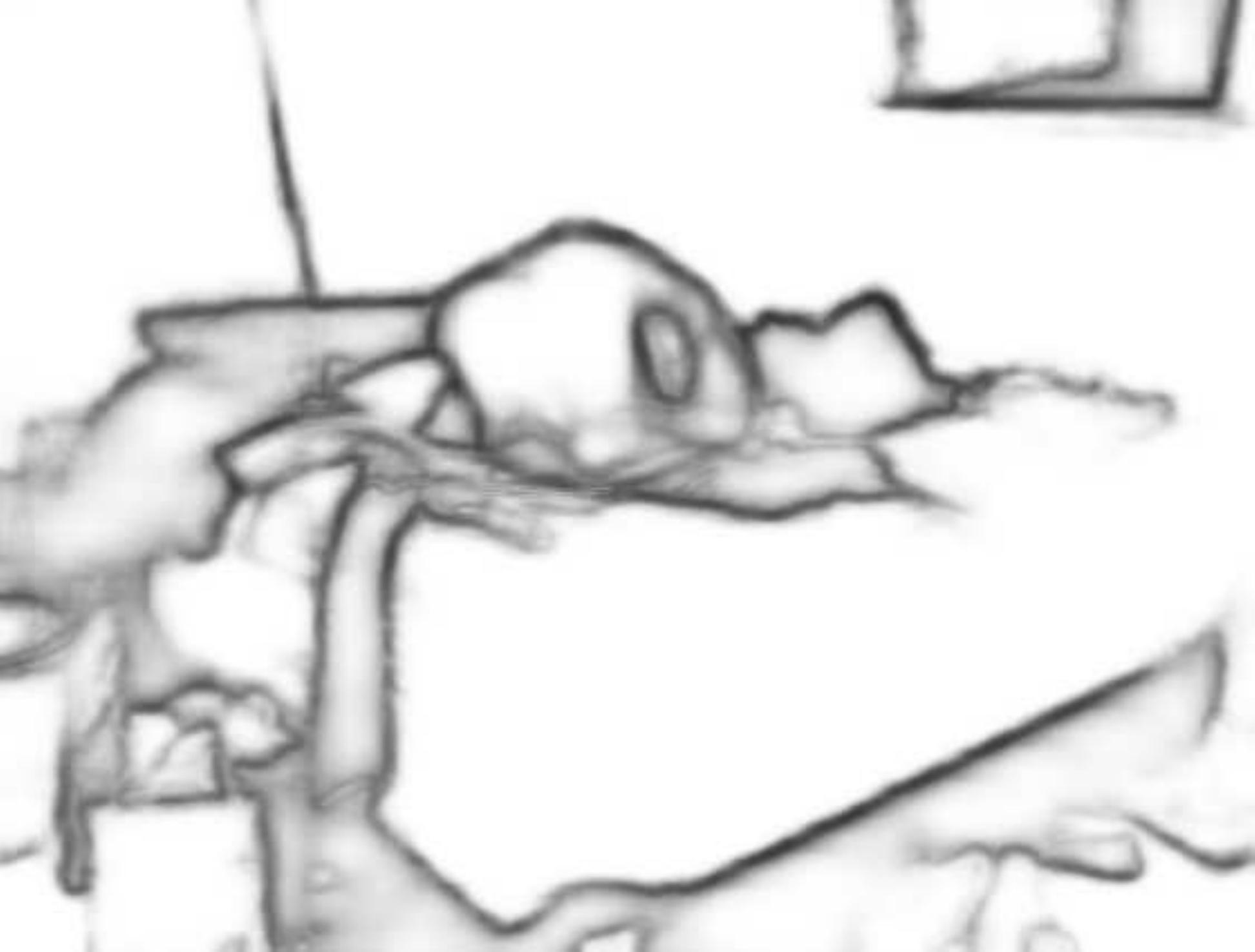} &
		\includegraphics[width=0.16\linewidth]{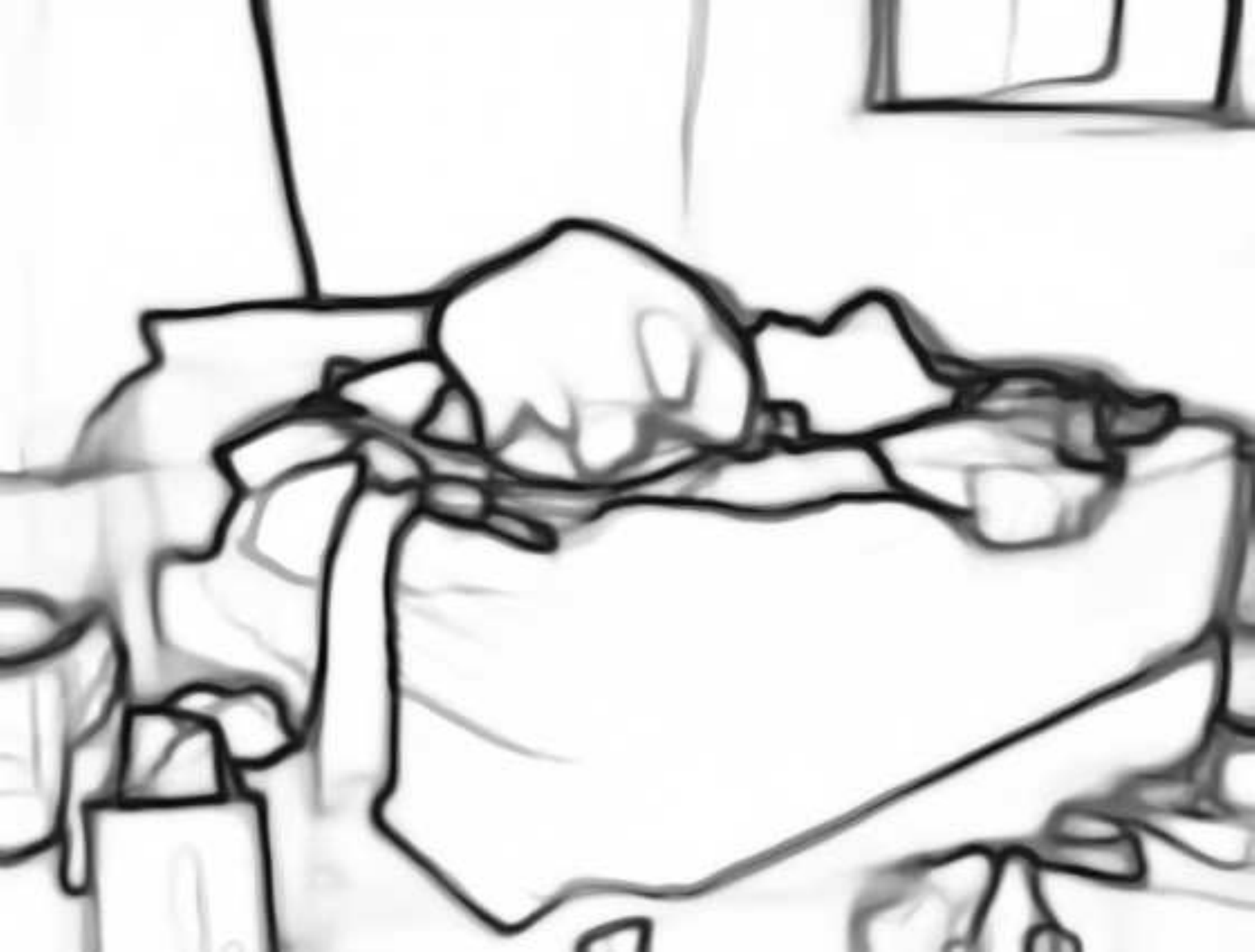} &
		\includegraphics[width=0.16\linewidth]{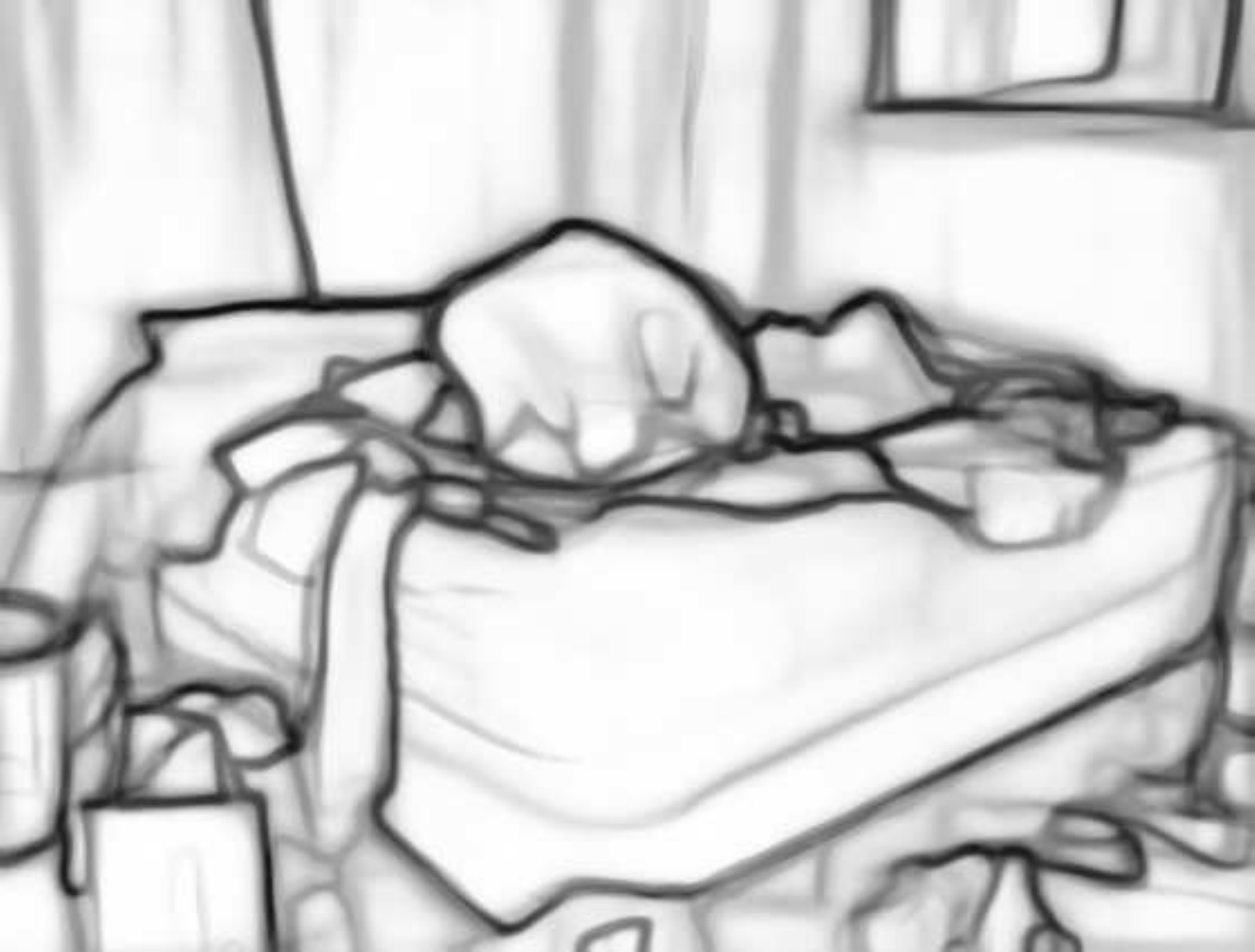} \\
		\includegraphics[width=0.16\linewidth]{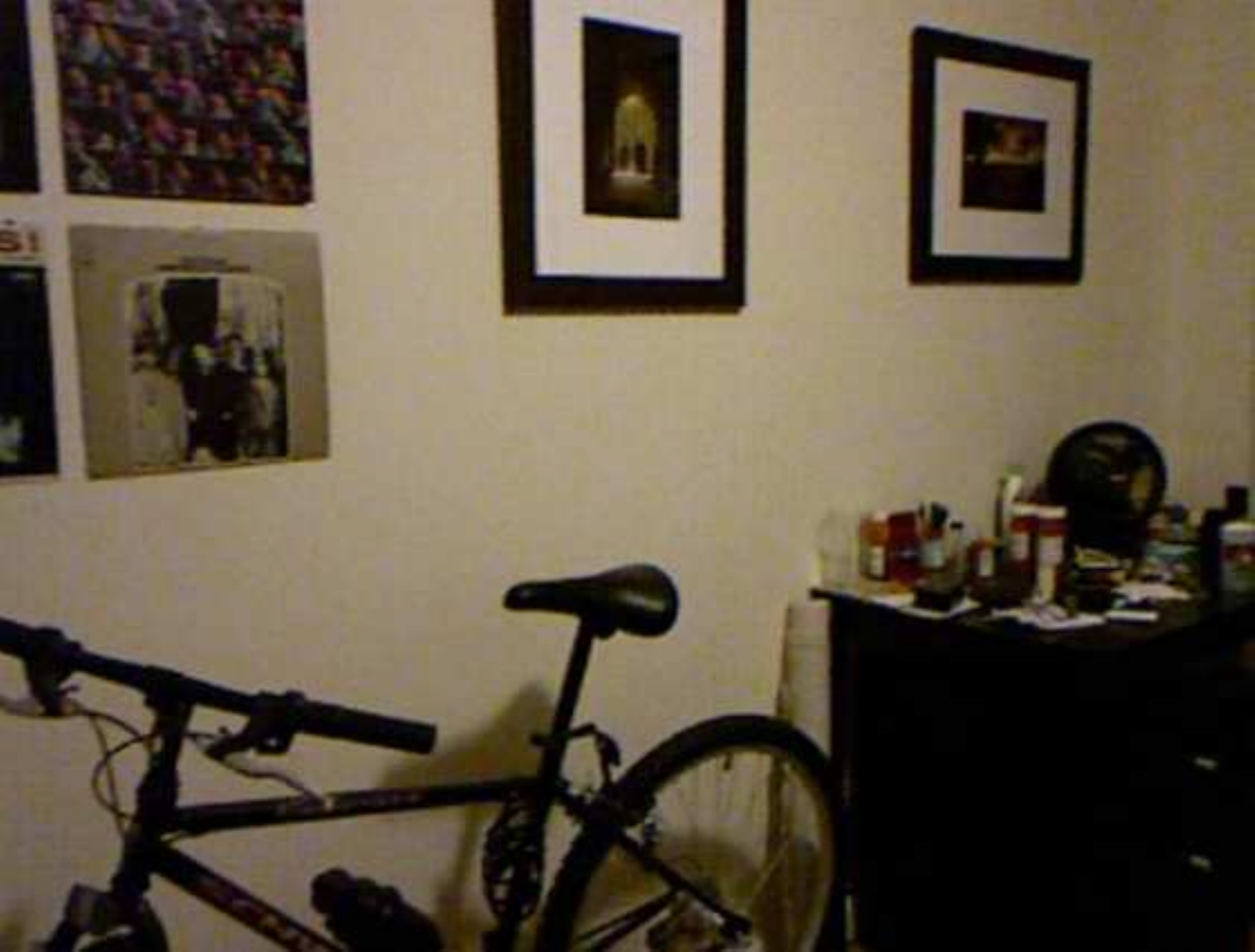} &
		\includegraphics[width=0.16\linewidth]{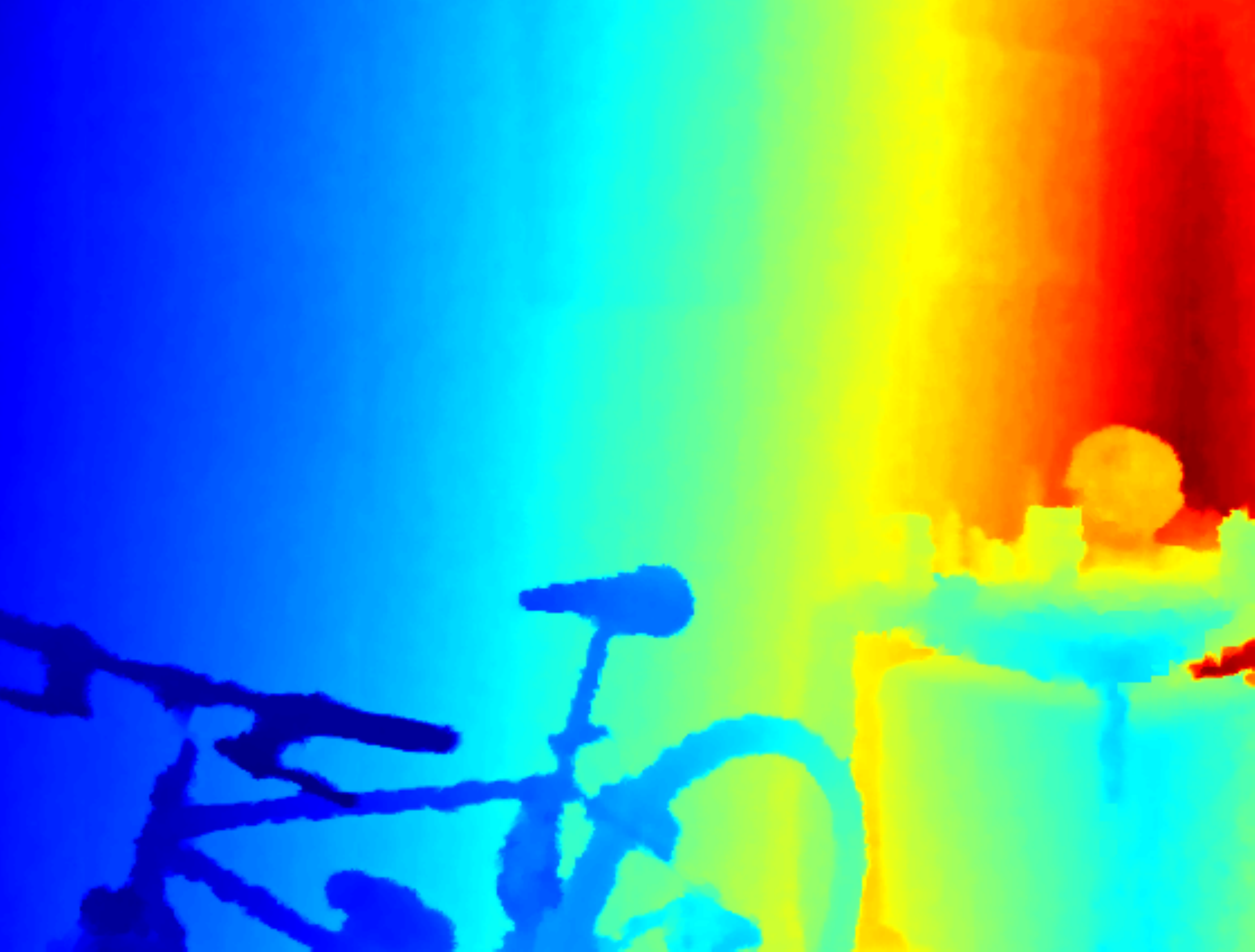} &
		\includegraphics[width=0.16\linewidth]{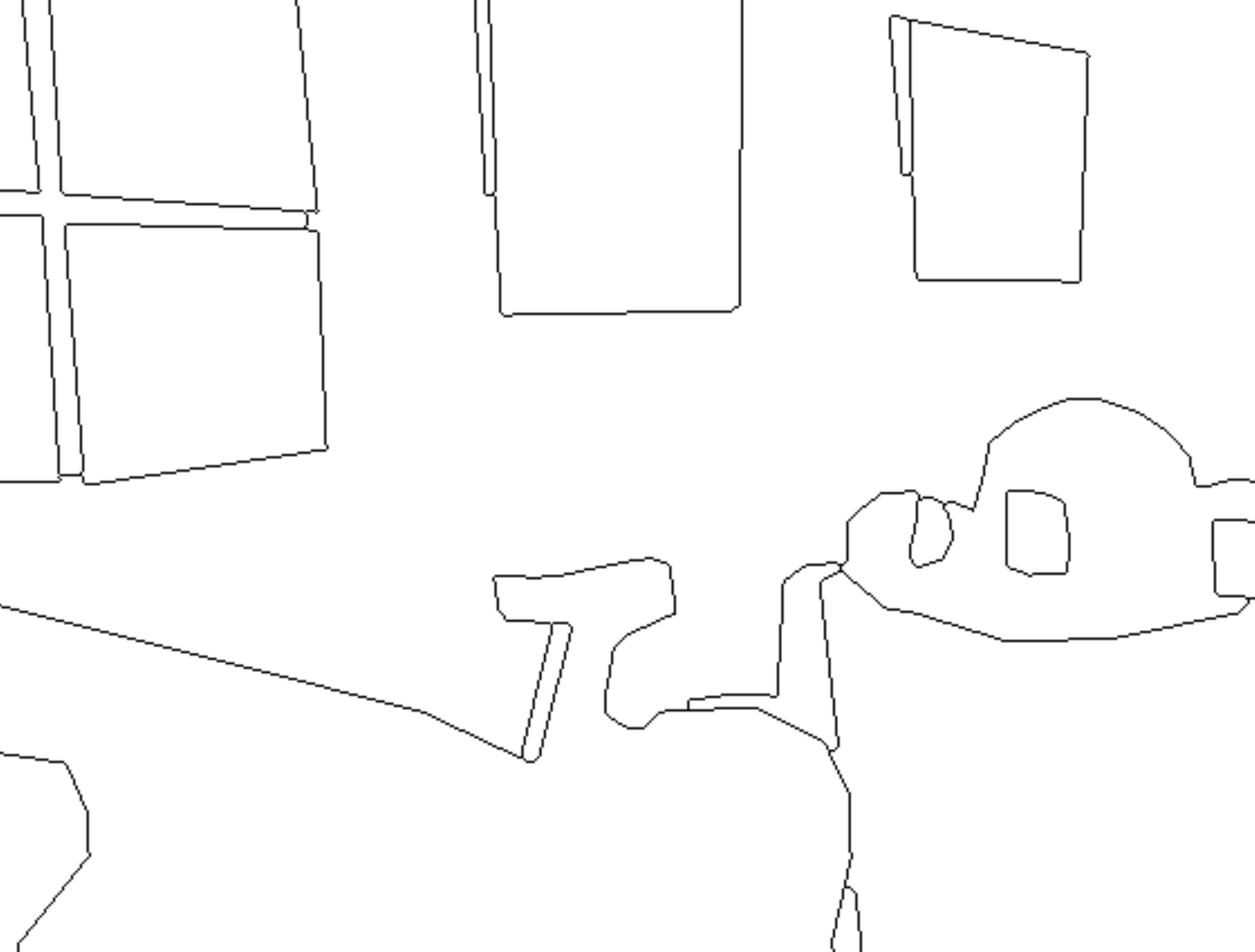} &
		\includegraphics[width=0.16\linewidth]{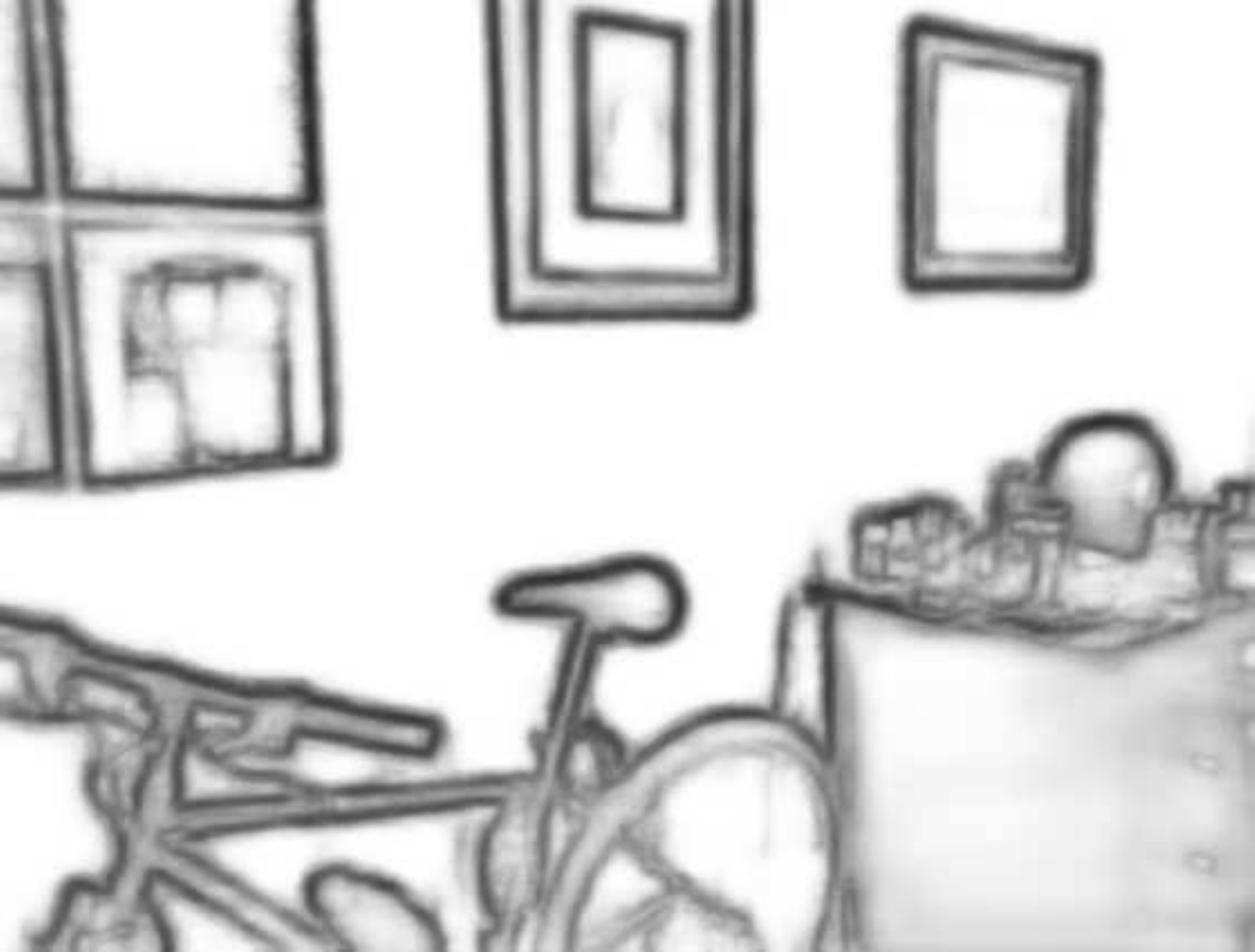} &
		\includegraphics[width=0.16\linewidth]{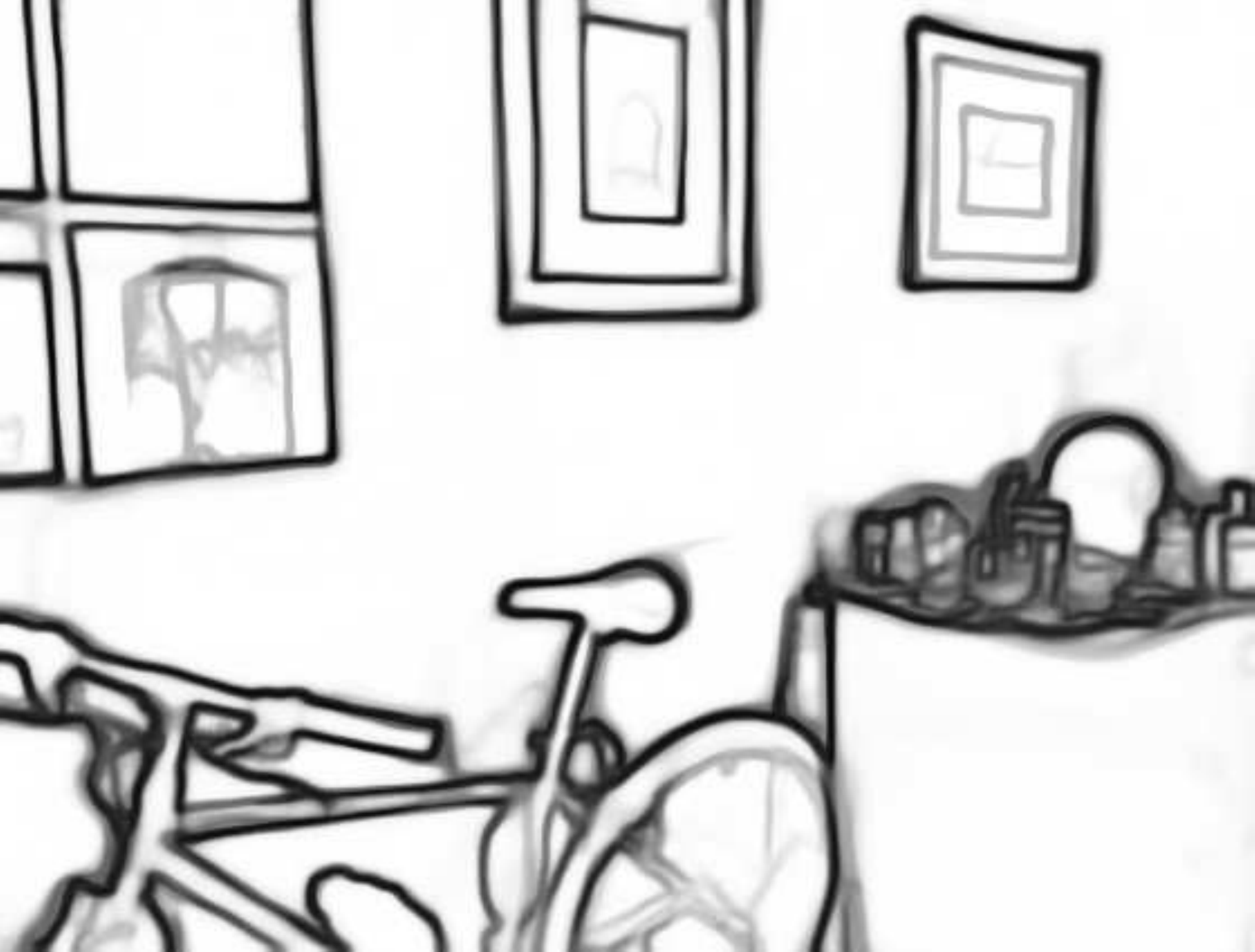} &
		\includegraphics[width=0.16\linewidth]{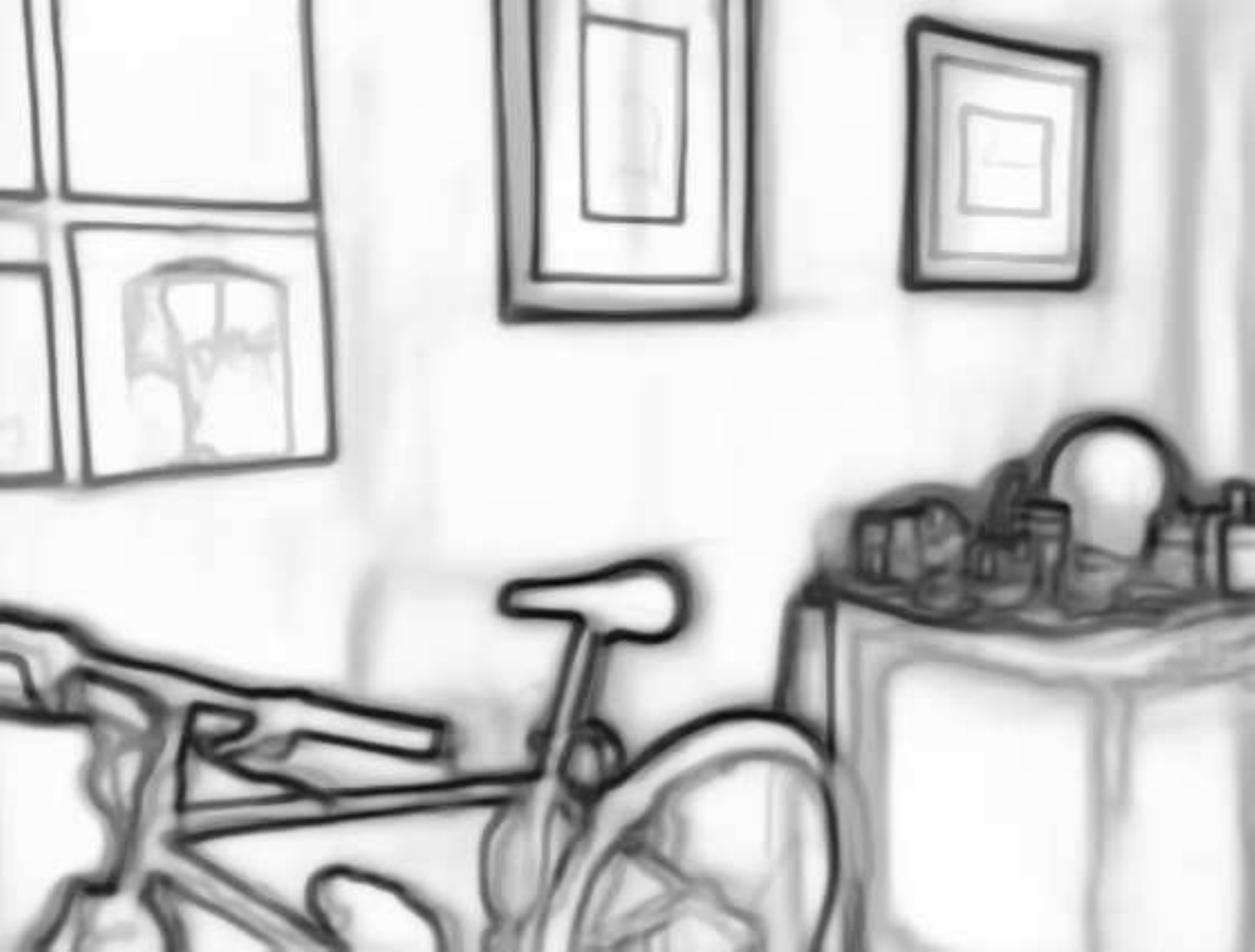} \\
		\includegraphics[width=0.16\linewidth]{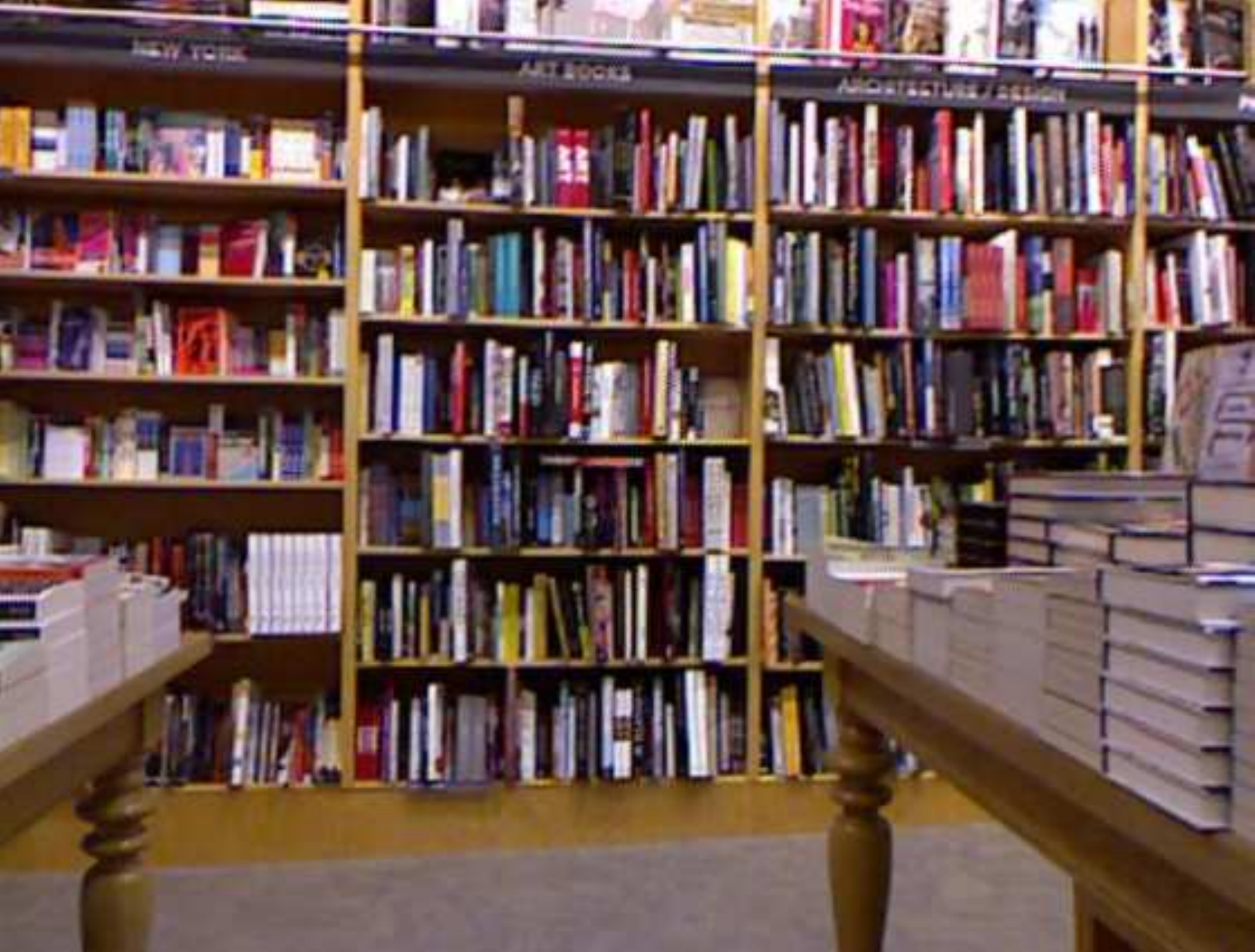} &
		\includegraphics[width=0.16\linewidth]{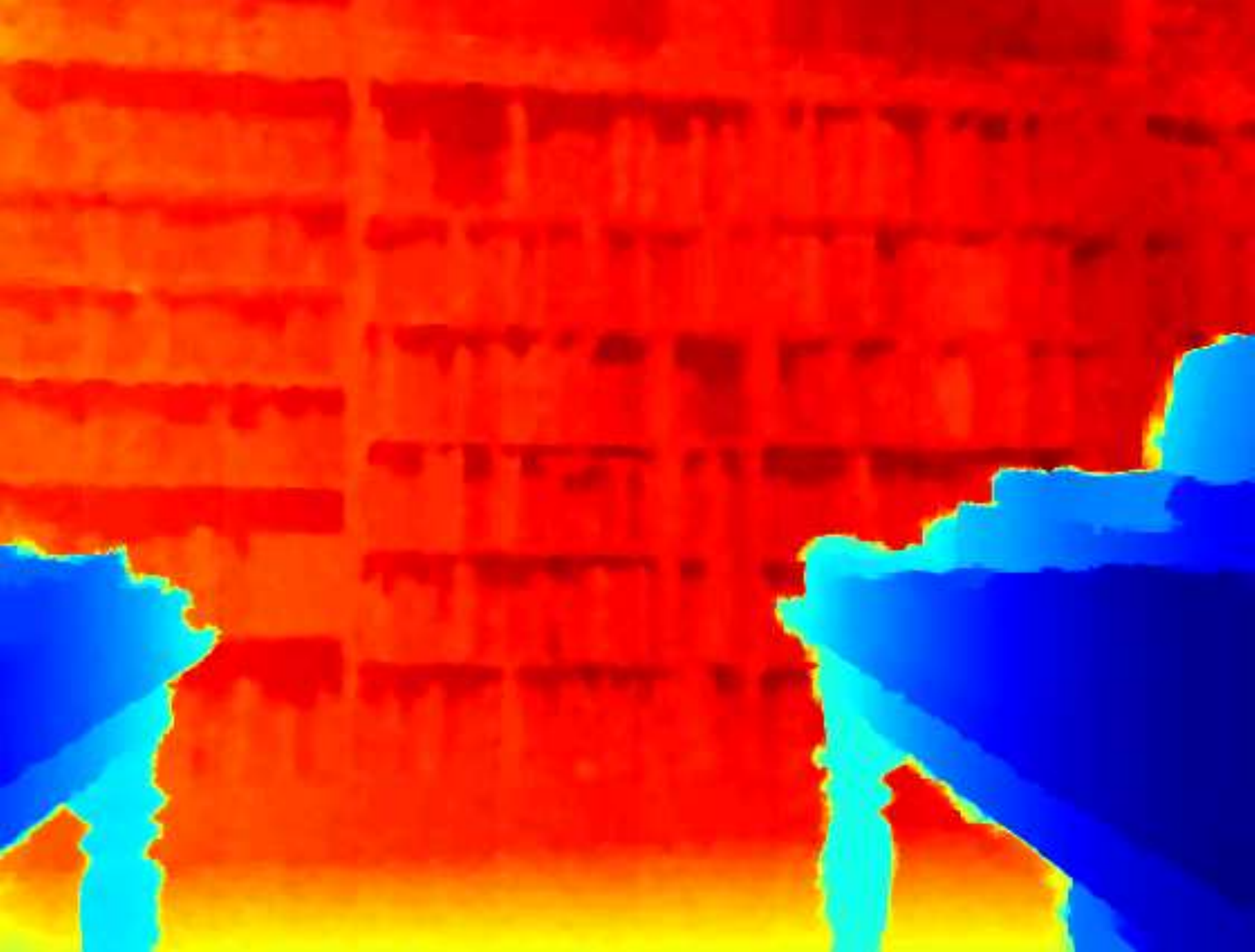} &
		\includegraphics[width=0.16\linewidth]{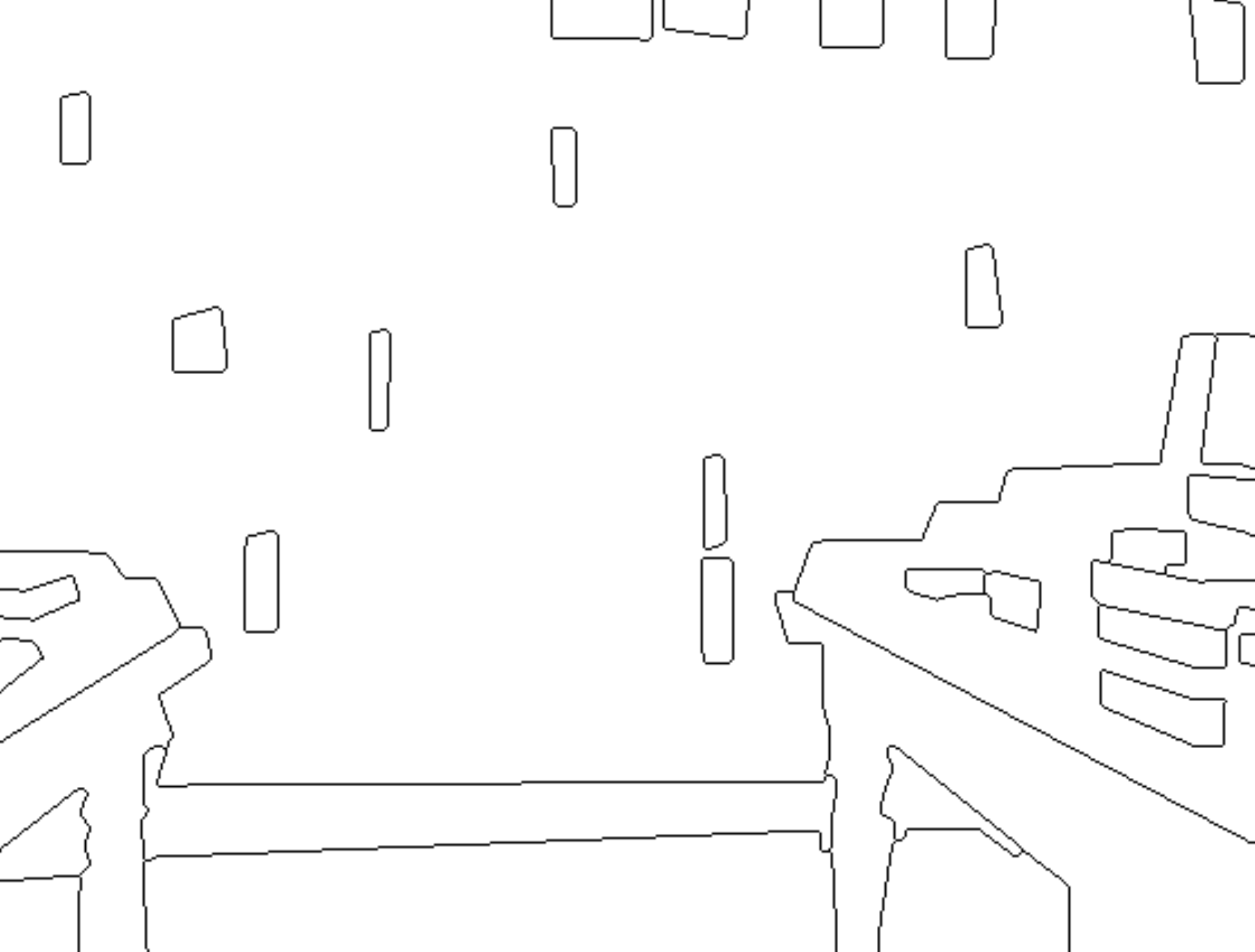} &
		\includegraphics[width=0.16\linewidth]{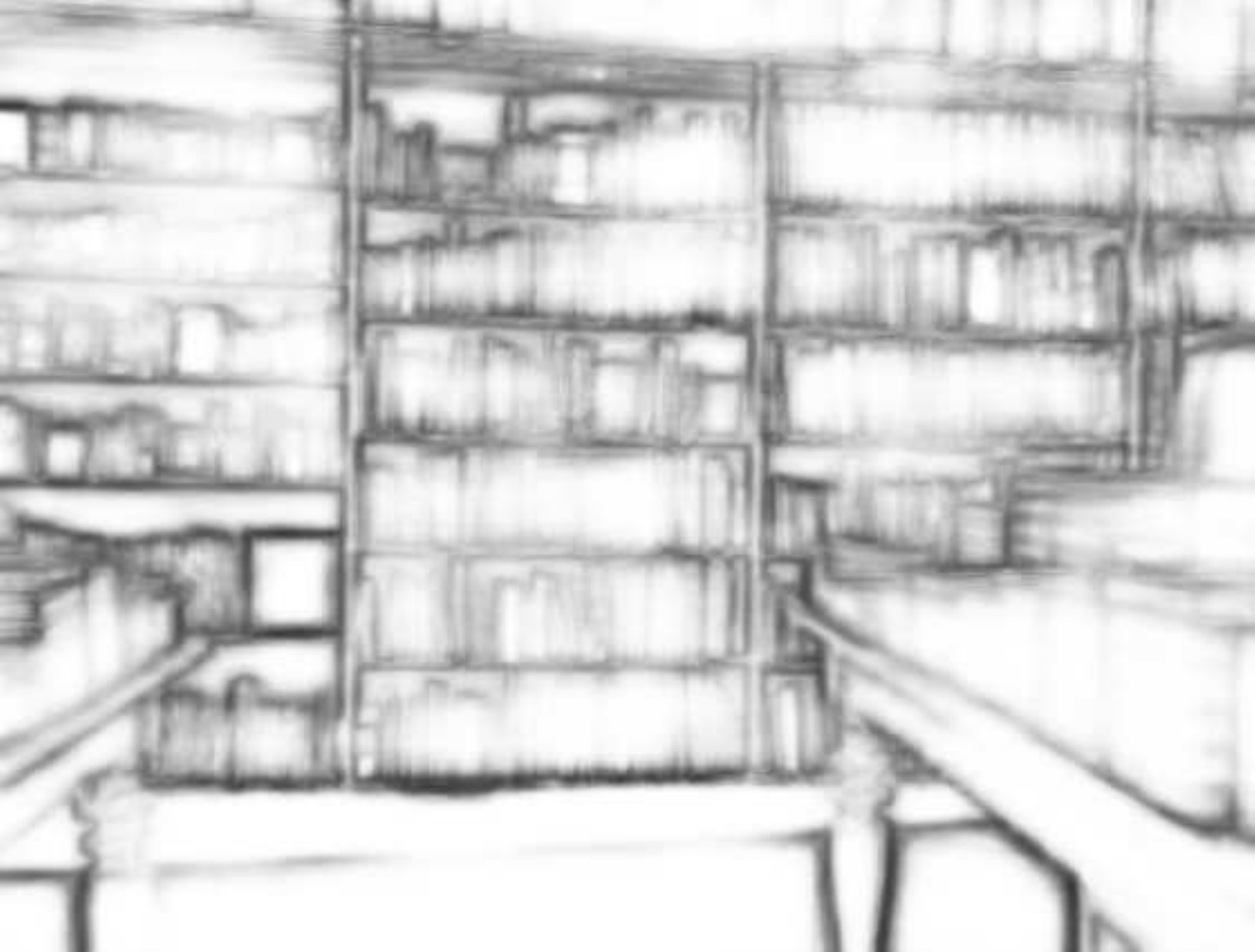} &
		\includegraphics[width=0.16\linewidth]{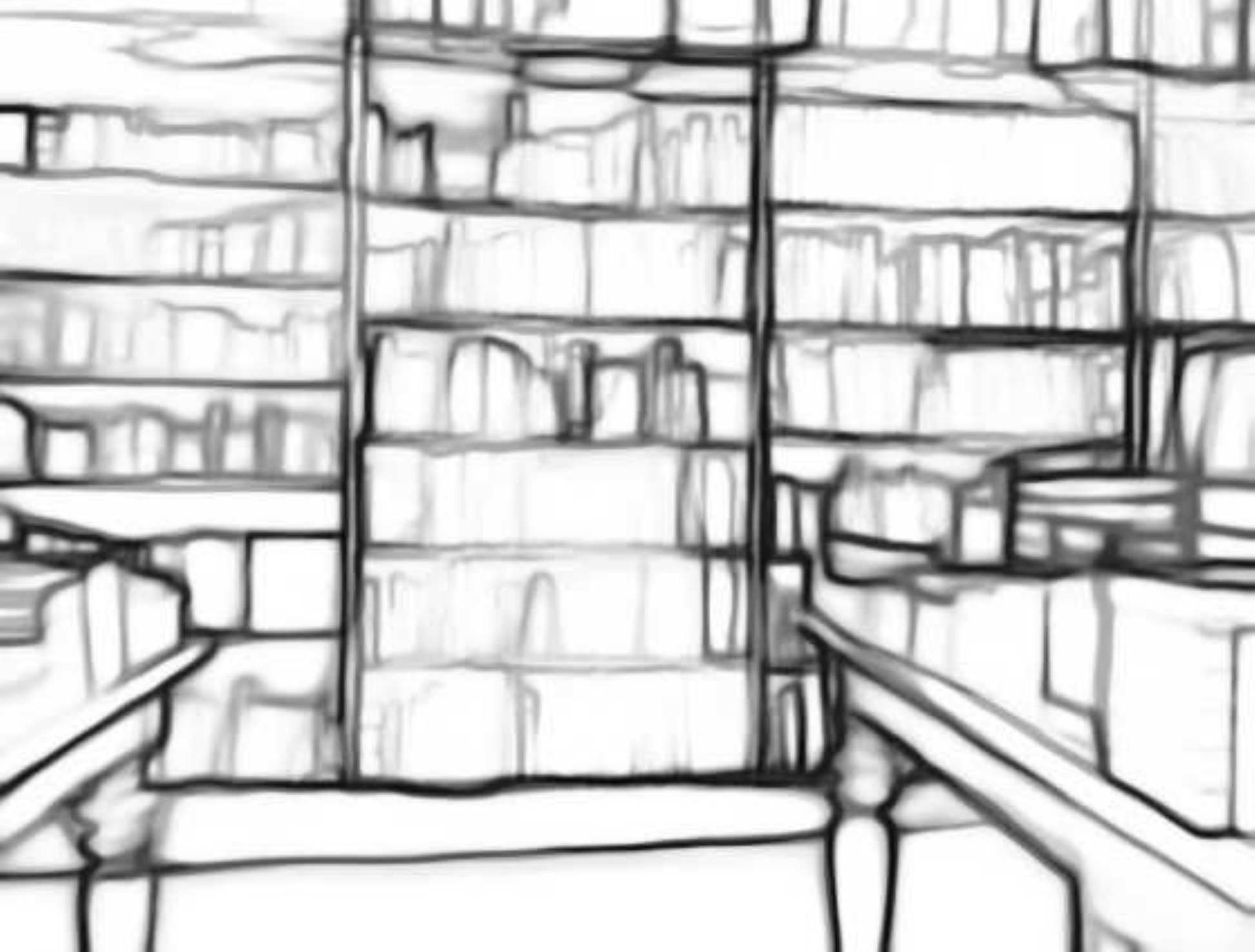} &
		\includegraphics[width=0.16\linewidth]{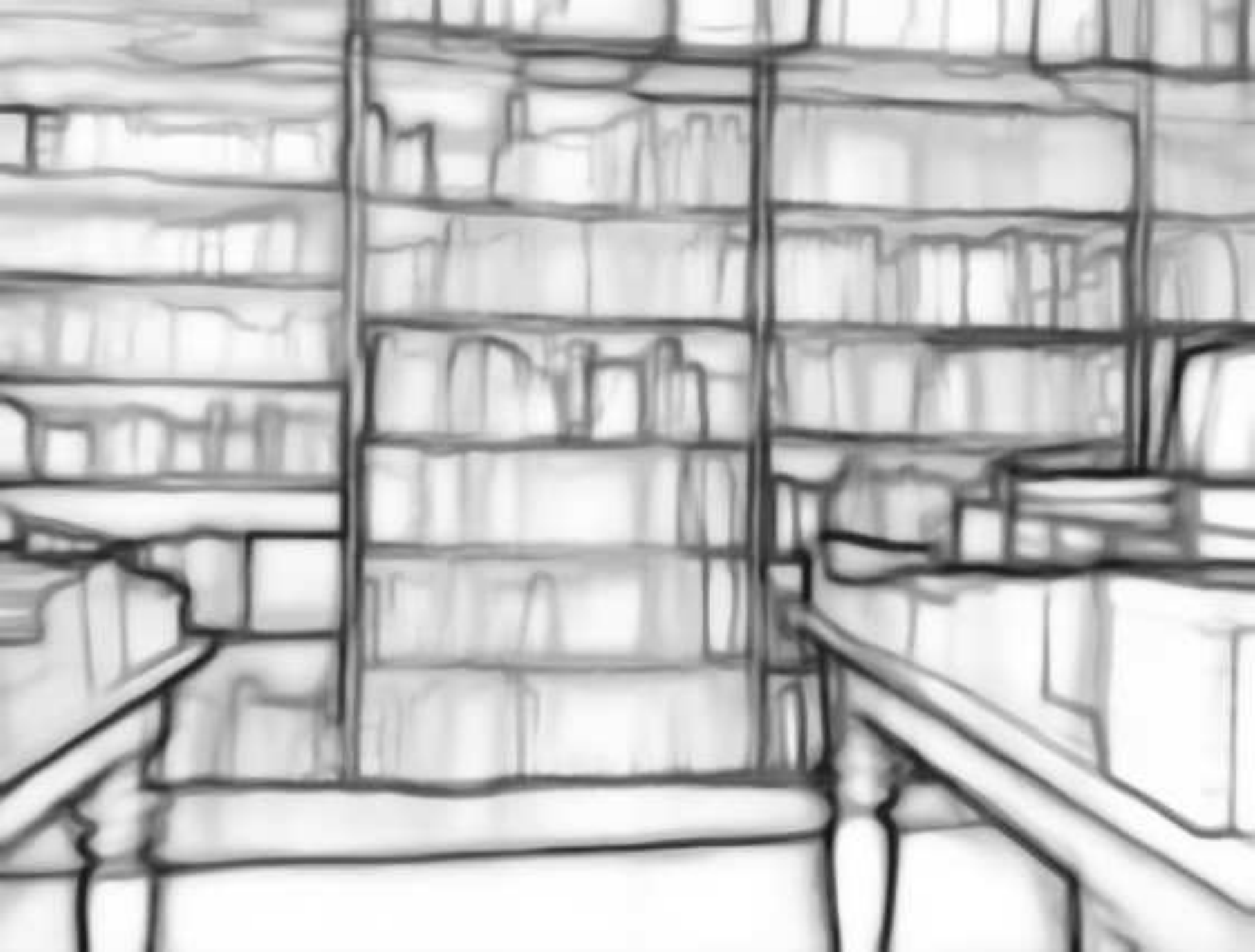} \\
		\includegraphics[width=0.16\linewidth]{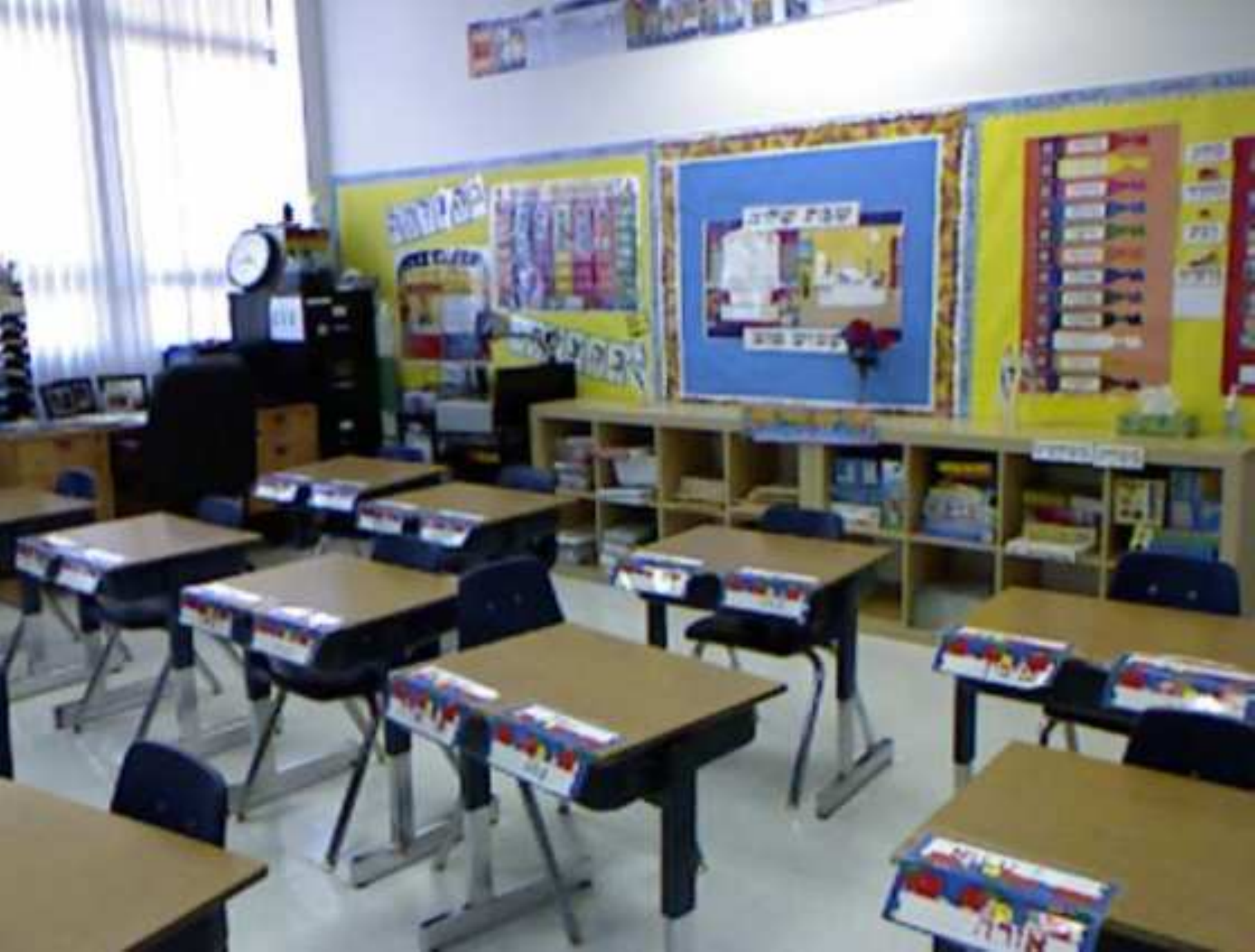} &
		\includegraphics[width=0.16\linewidth]{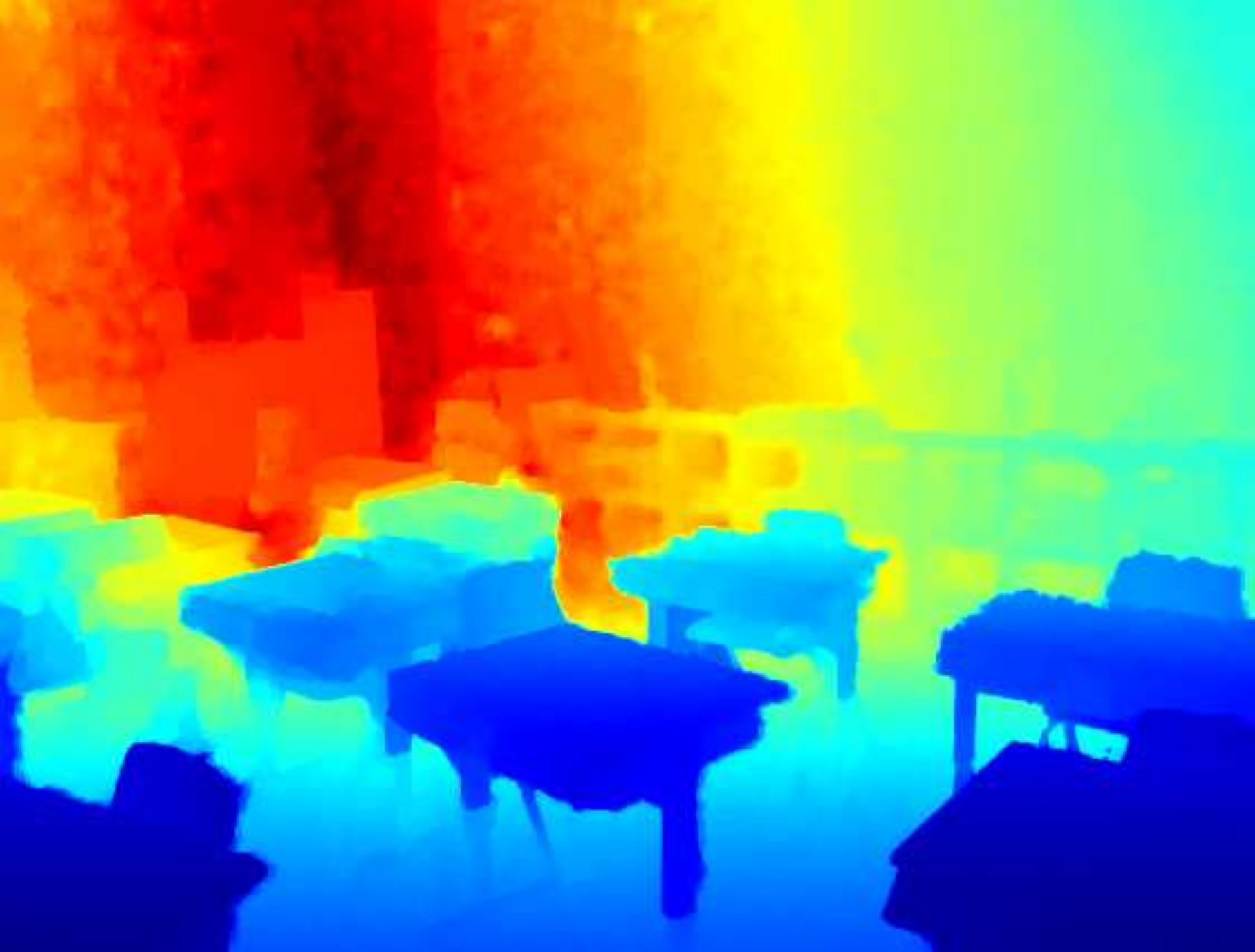} &
		\includegraphics[width=0.16\linewidth]{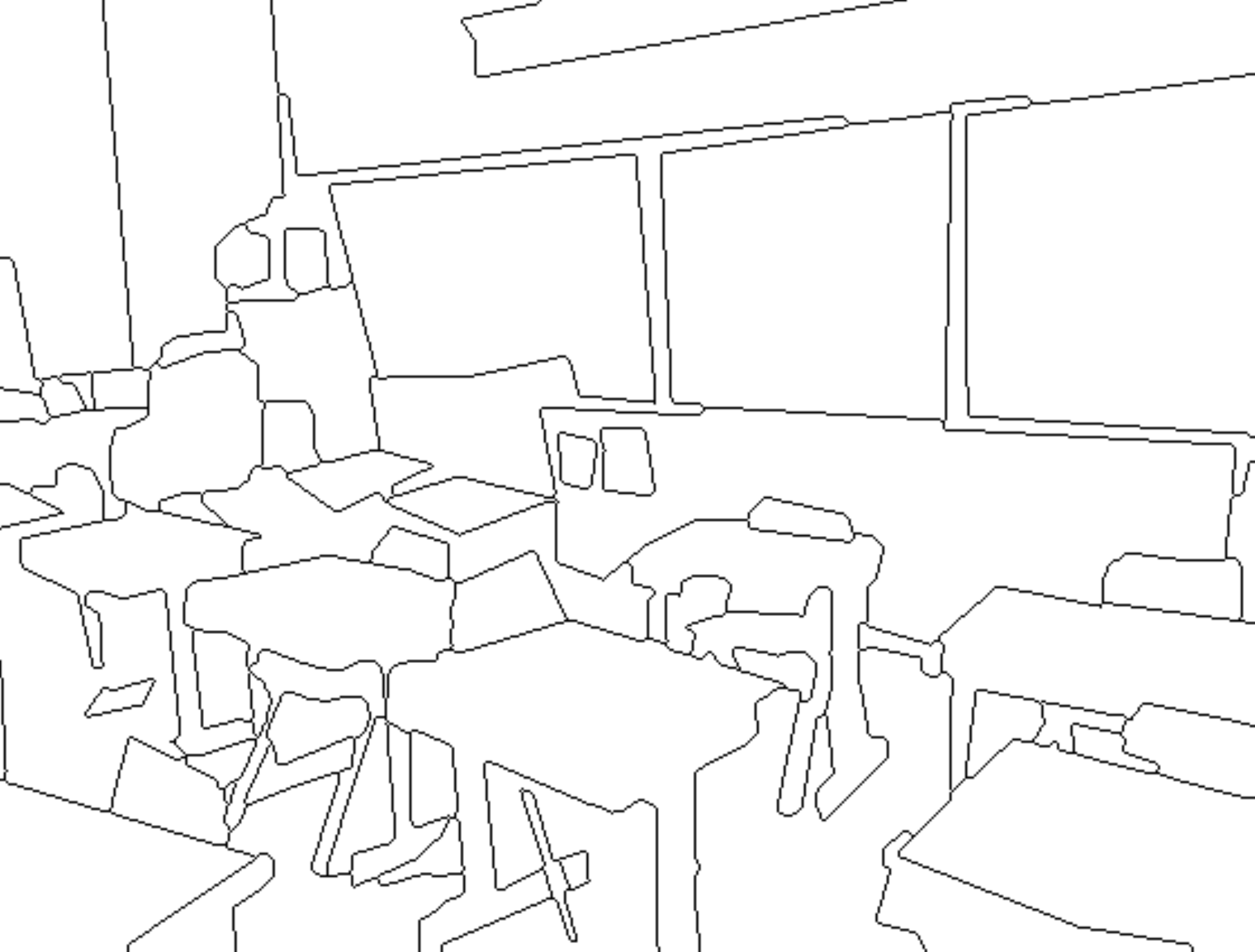} &
		\includegraphics[width=0.16\linewidth]{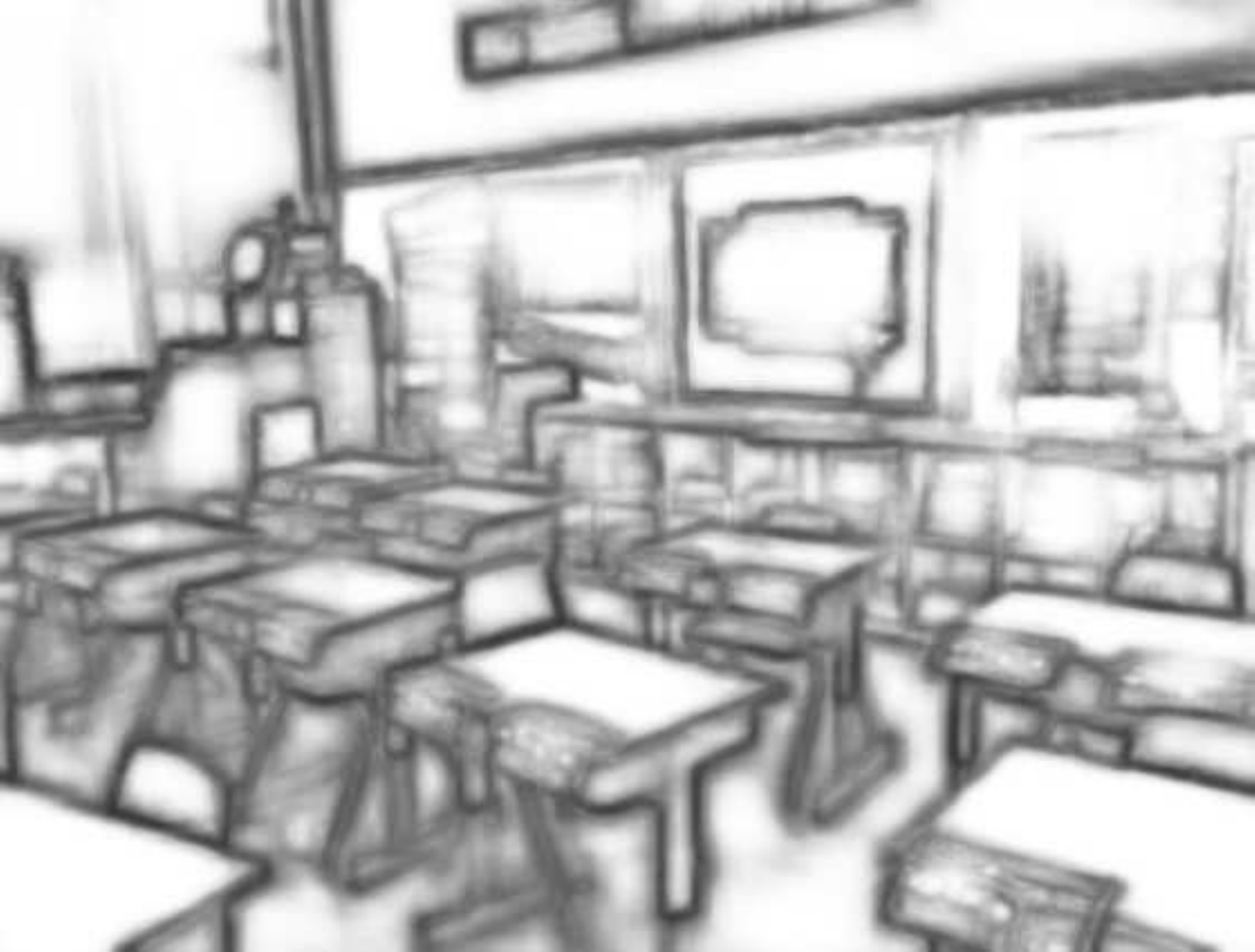} &
		\includegraphics[width=0.16\linewidth]{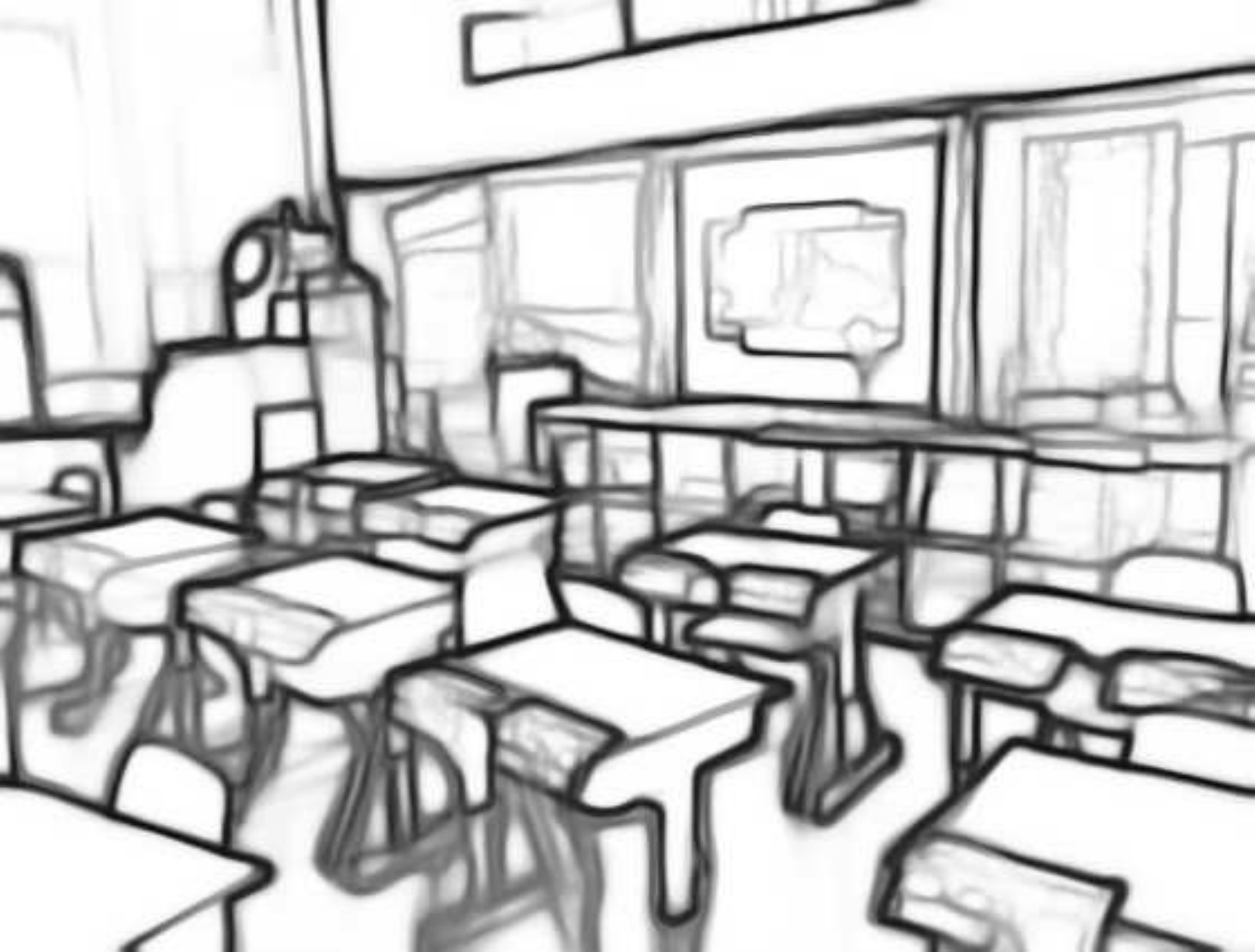} &
		\includegraphics[width=0.16\linewidth]{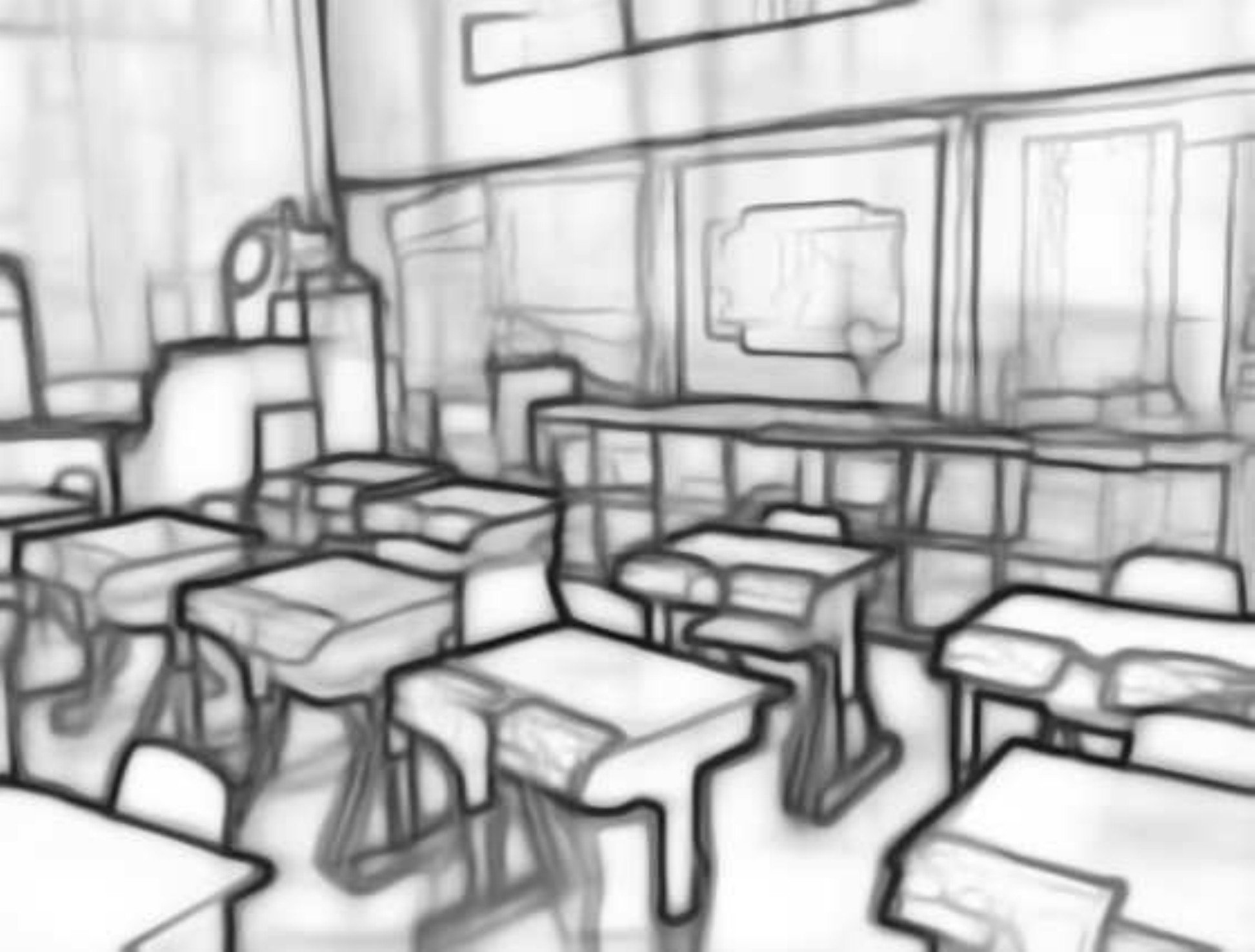} \\
		\includegraphics[width=0.16\linewidth]{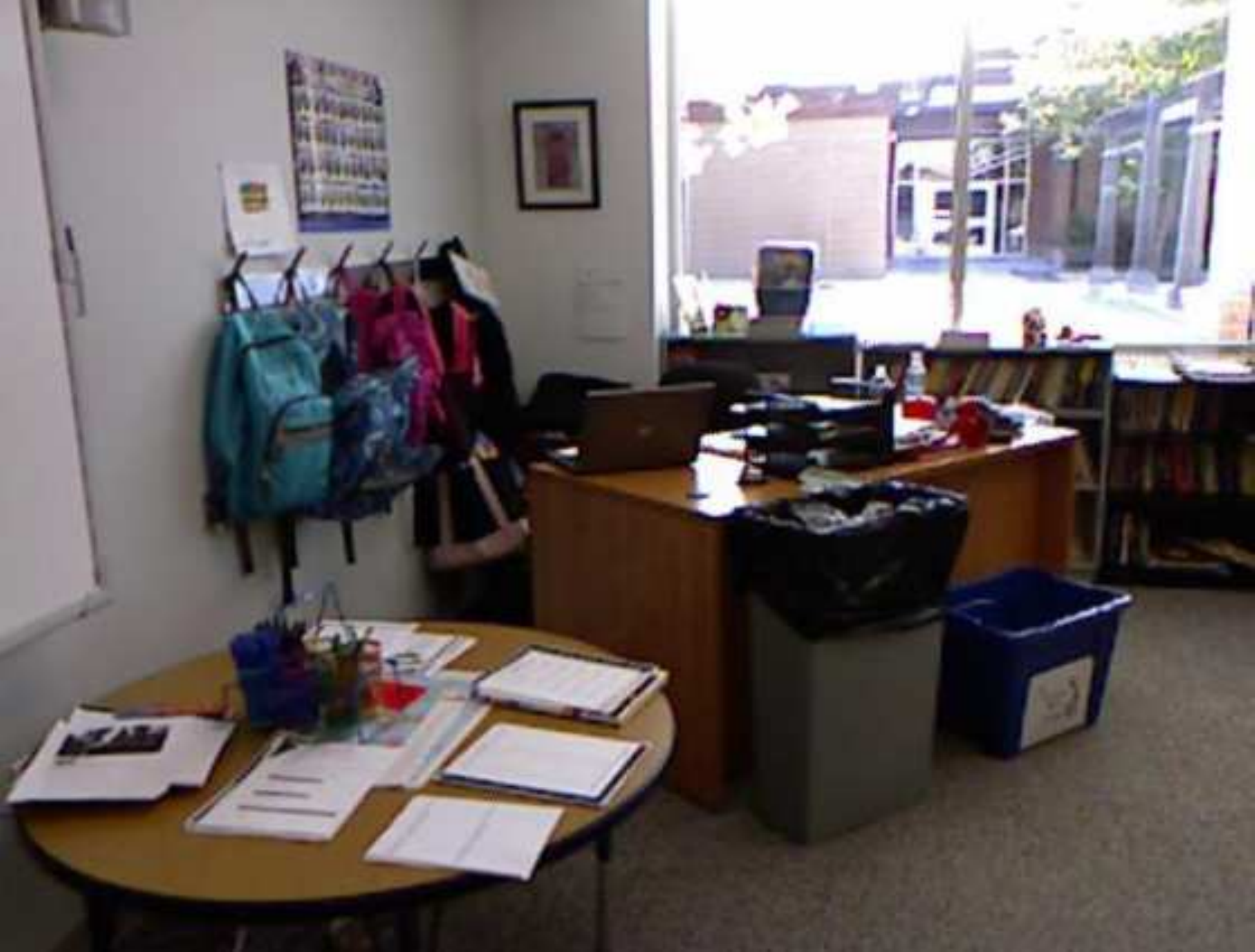} &
		\includegraphics[width=0.16\linewidth]{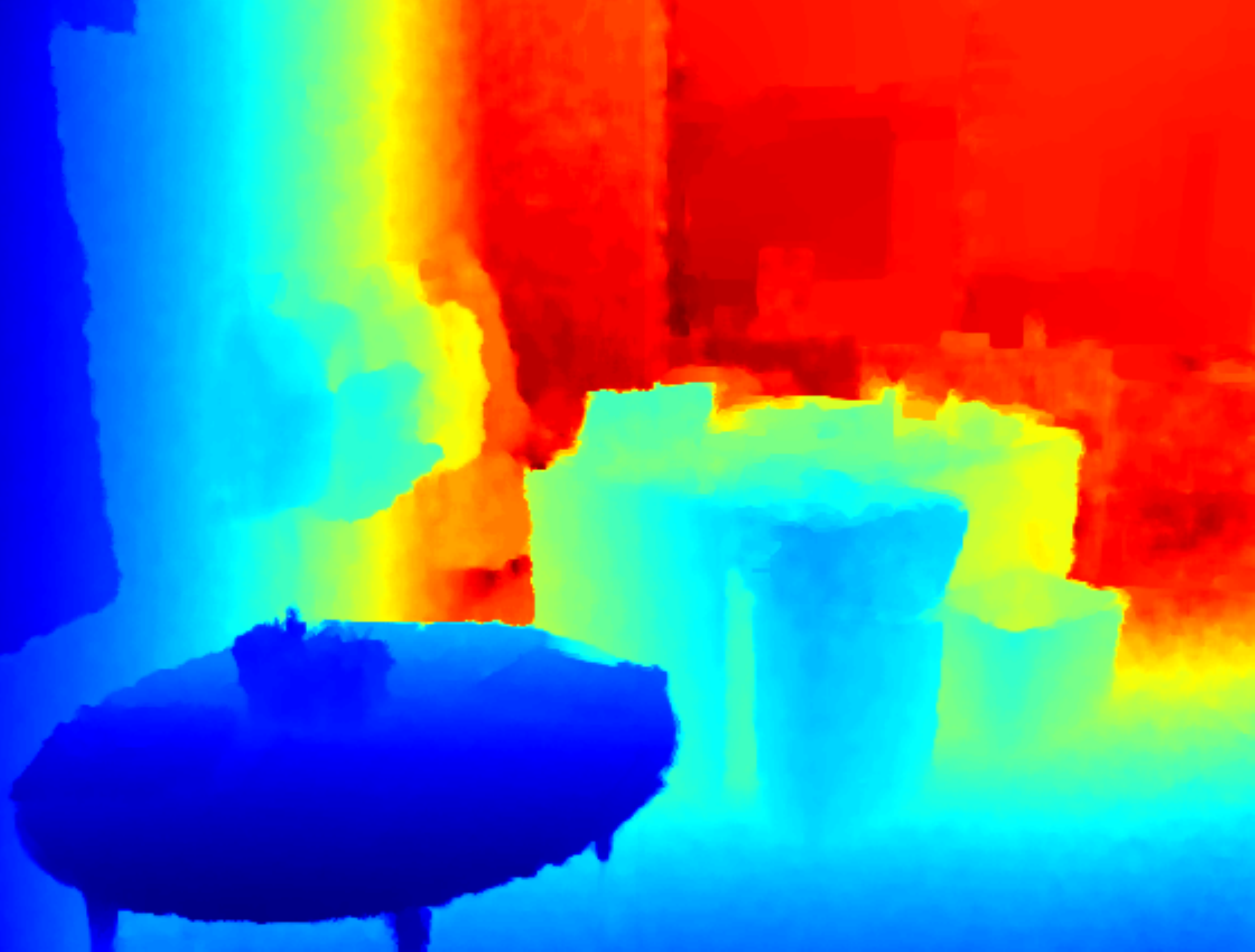} &
		\includegraphics[width=0.16\linewidth]{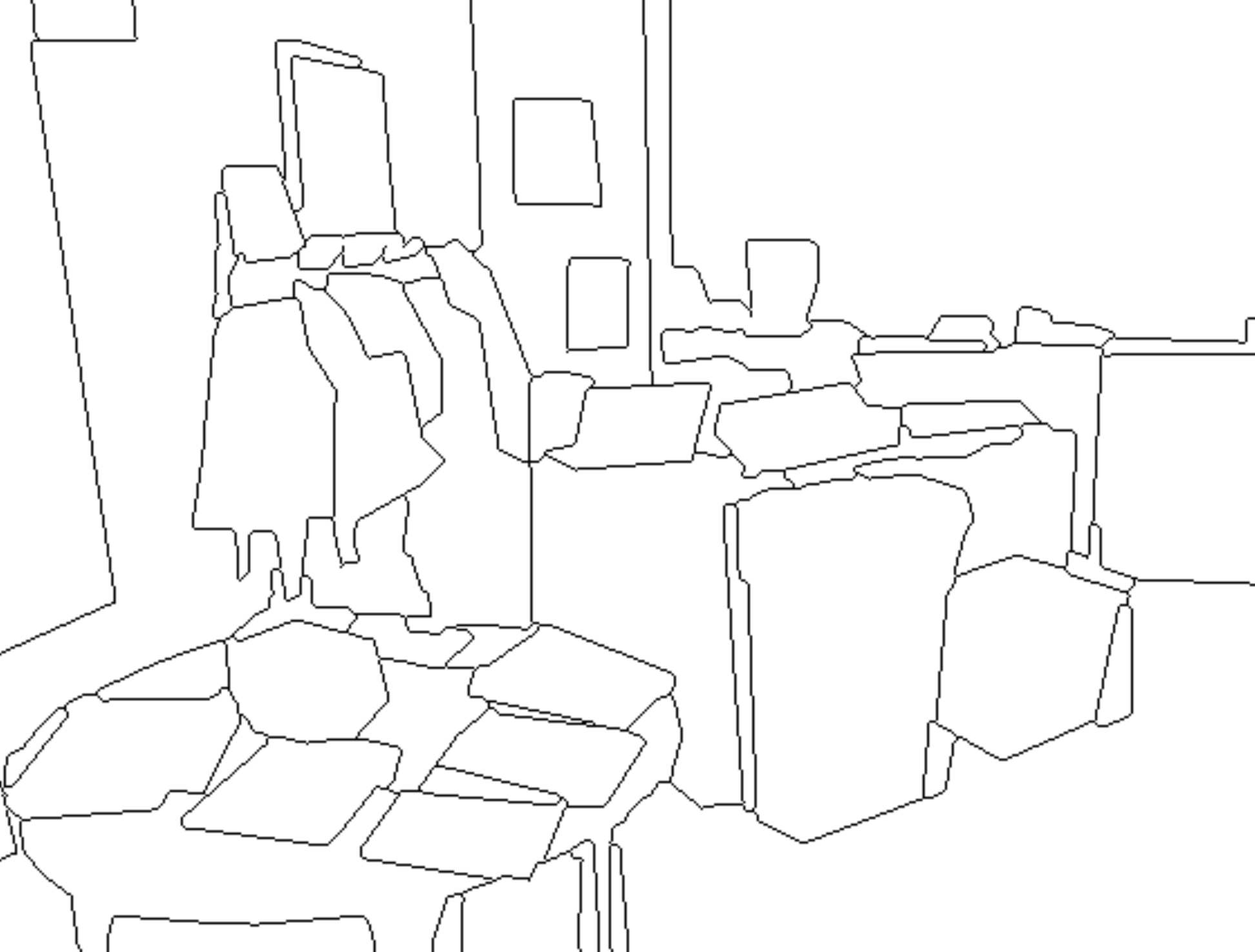} &
		\includegraphics[width=0.16\linewidth]{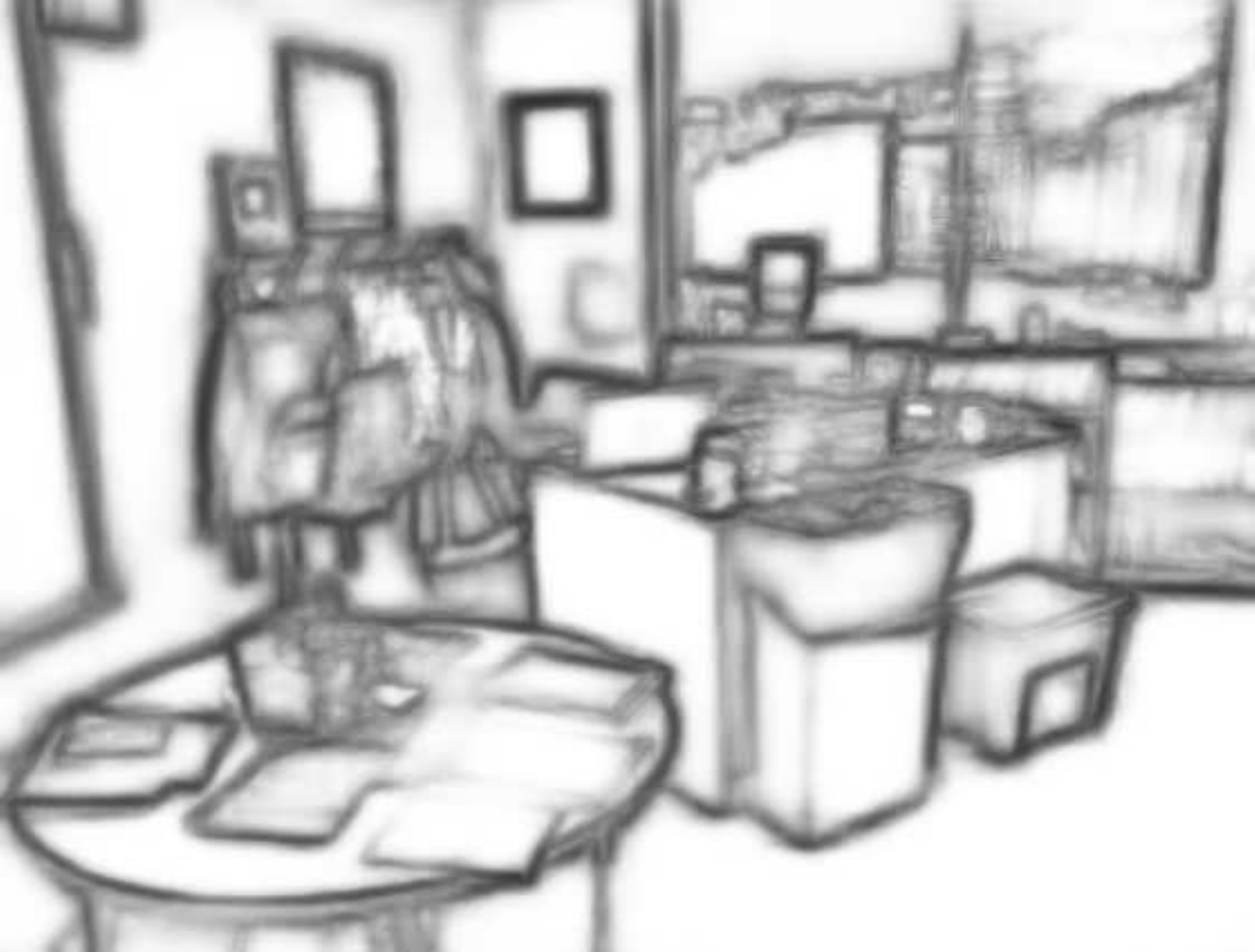} &
		\includegraphics[width=0.16\linewidth]{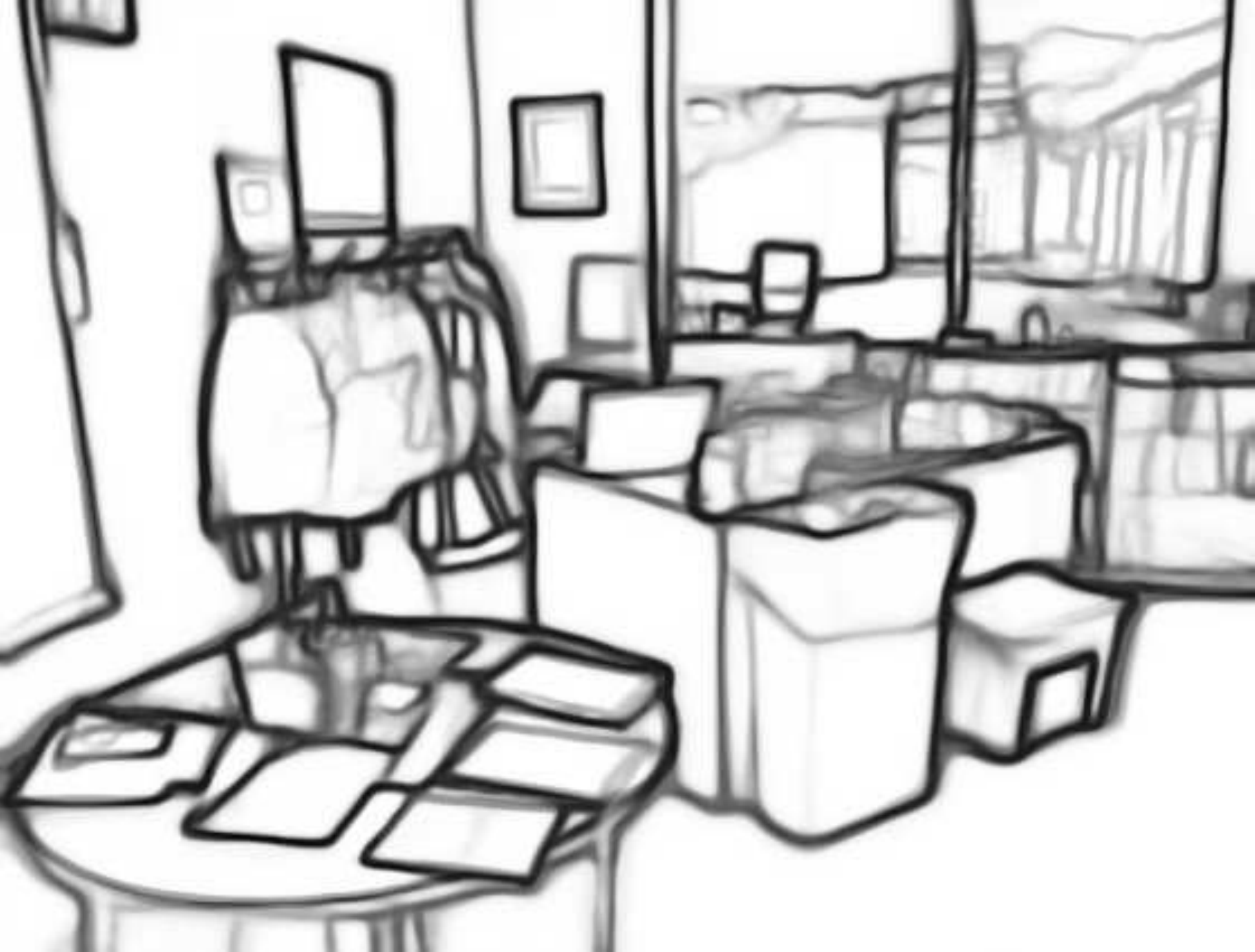} &
		\includegraphics[width=0.16\linewidth]{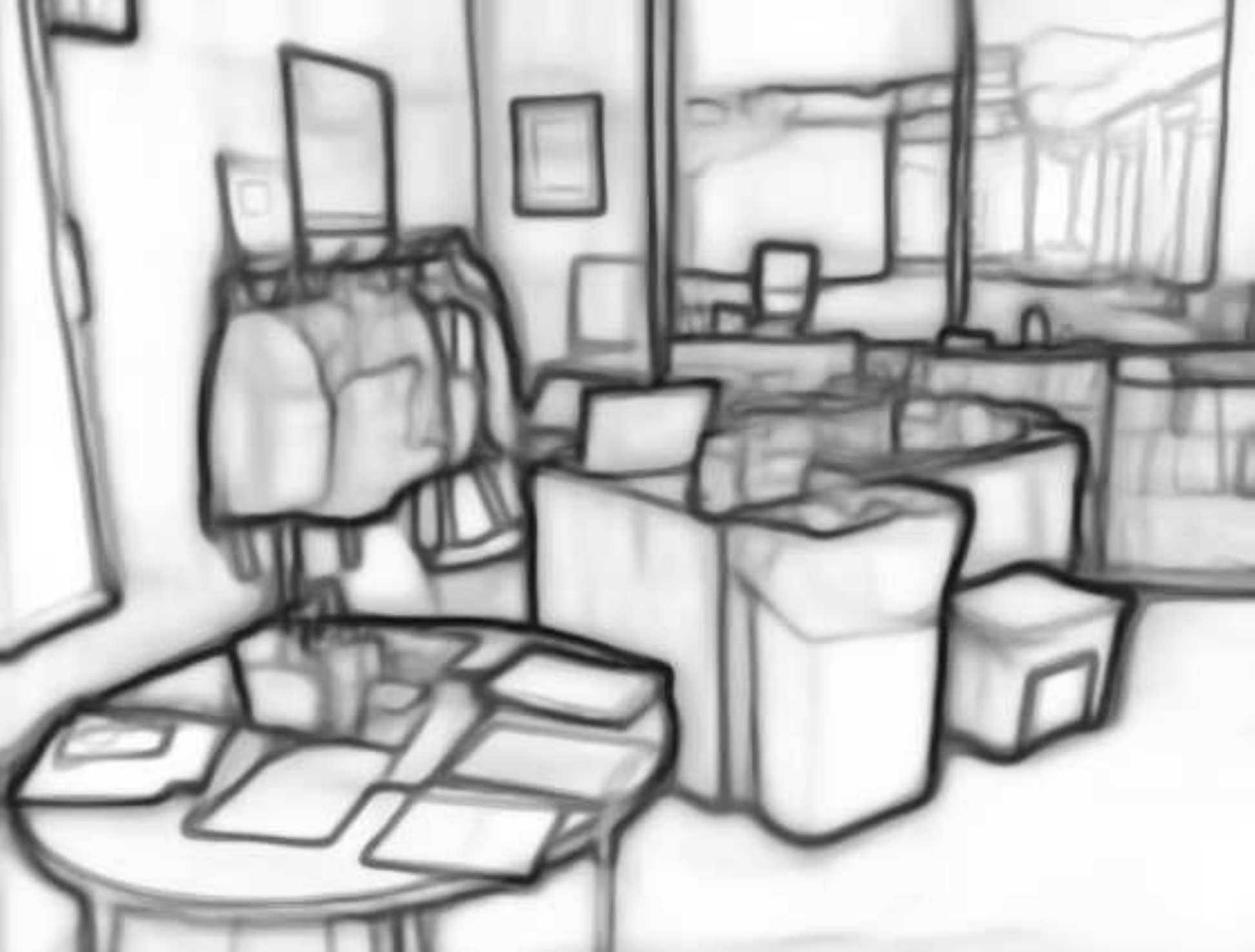} \\
		\includegraphics[width=0.16\linewidth]{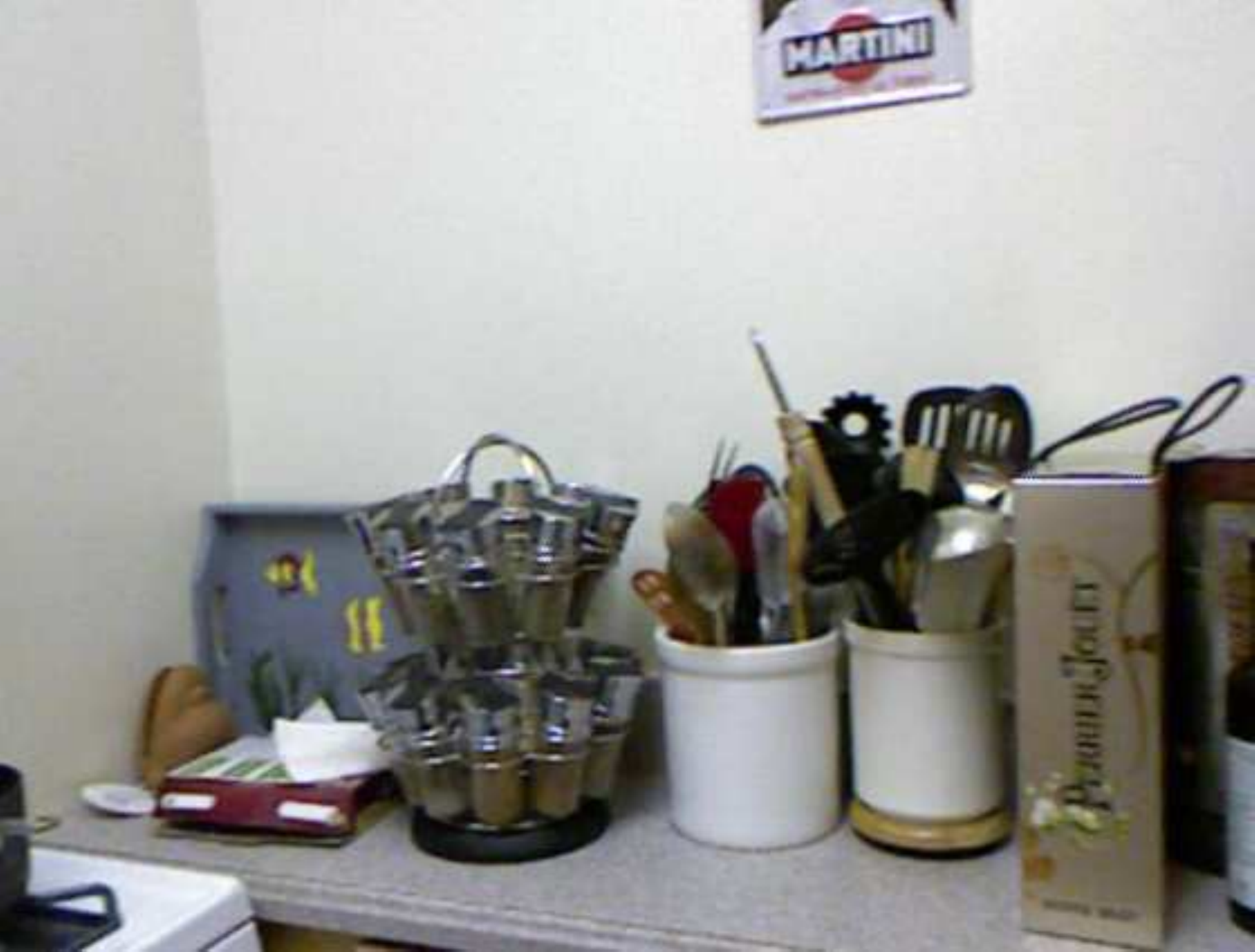} &
		\includegraphics[width=0.16\linewidth]{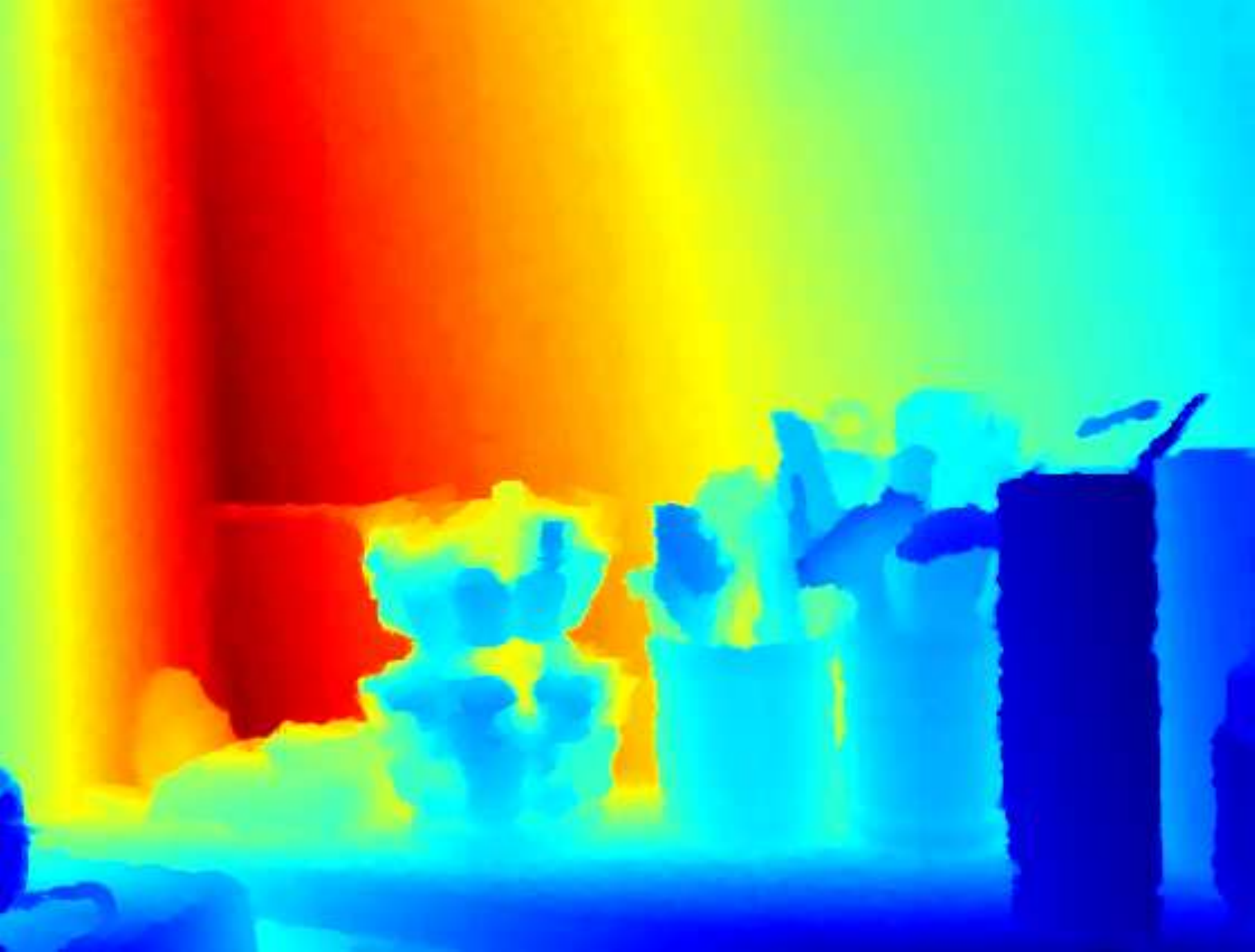} &
		\includegraphics[width=0.16\linewidth]{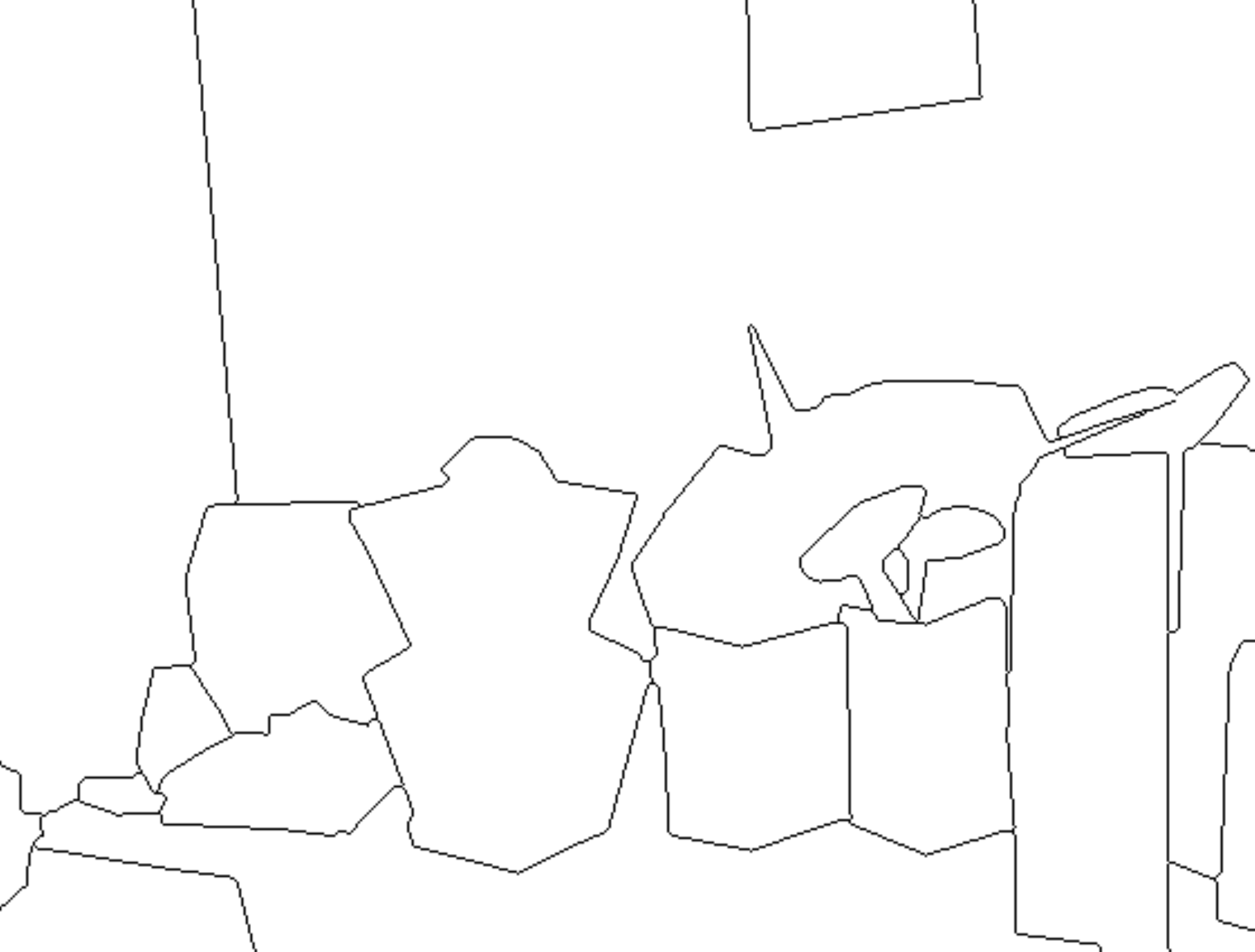} &
		\includegraphics[width=0.16\linewidth]{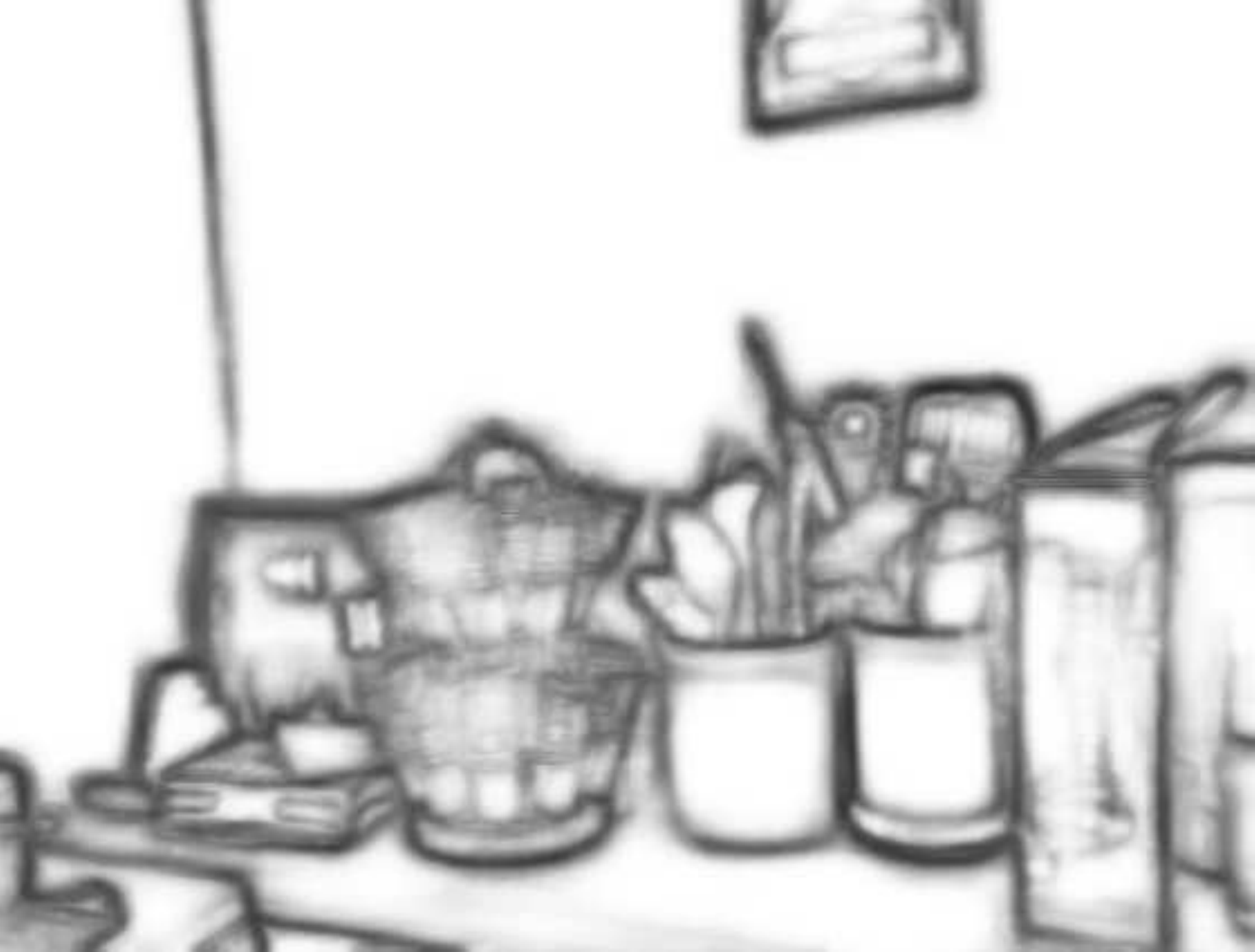} &
		\includegraphics[width=0.16\linewidth]{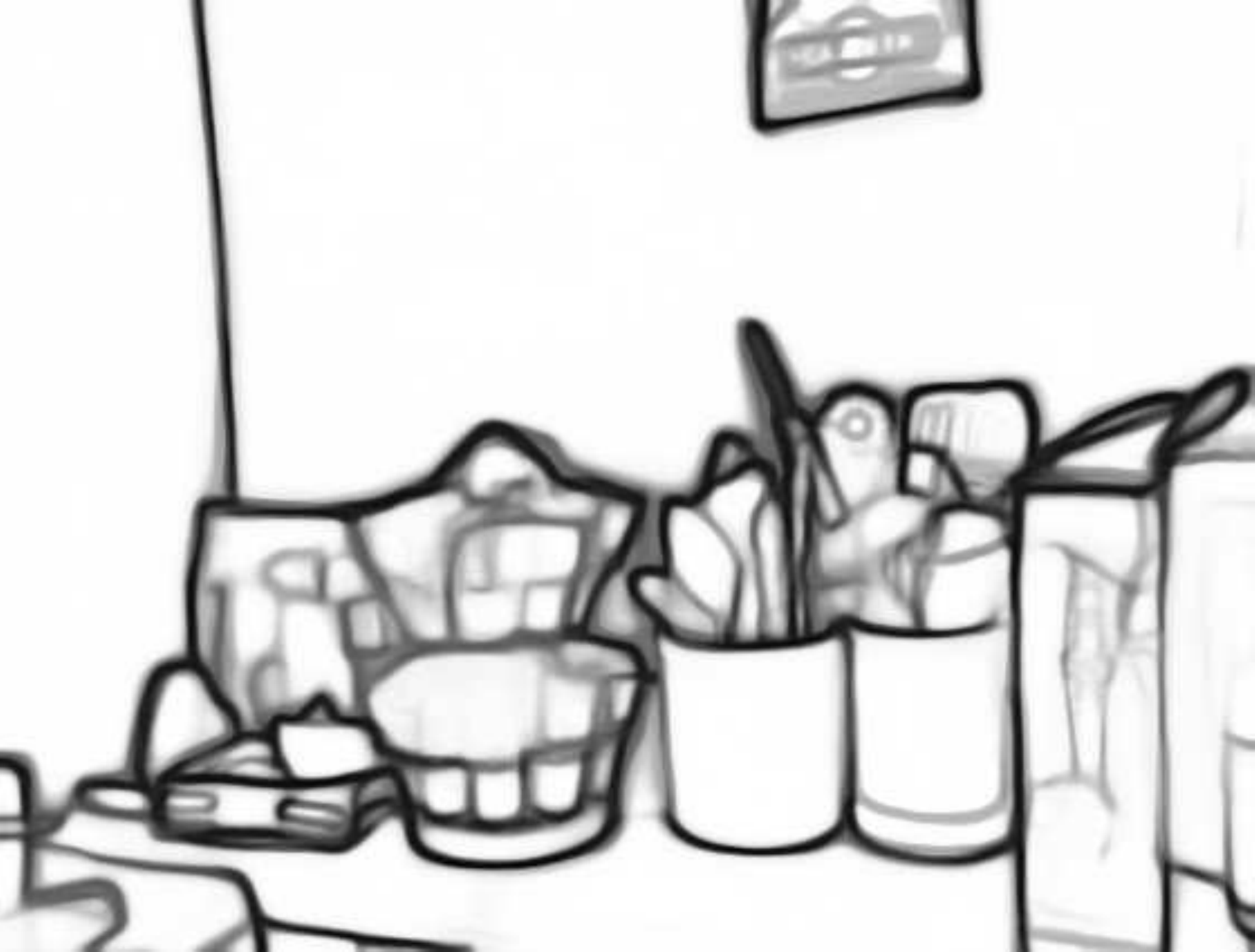} &
		\includegraphics[width=0.16\linewidth]{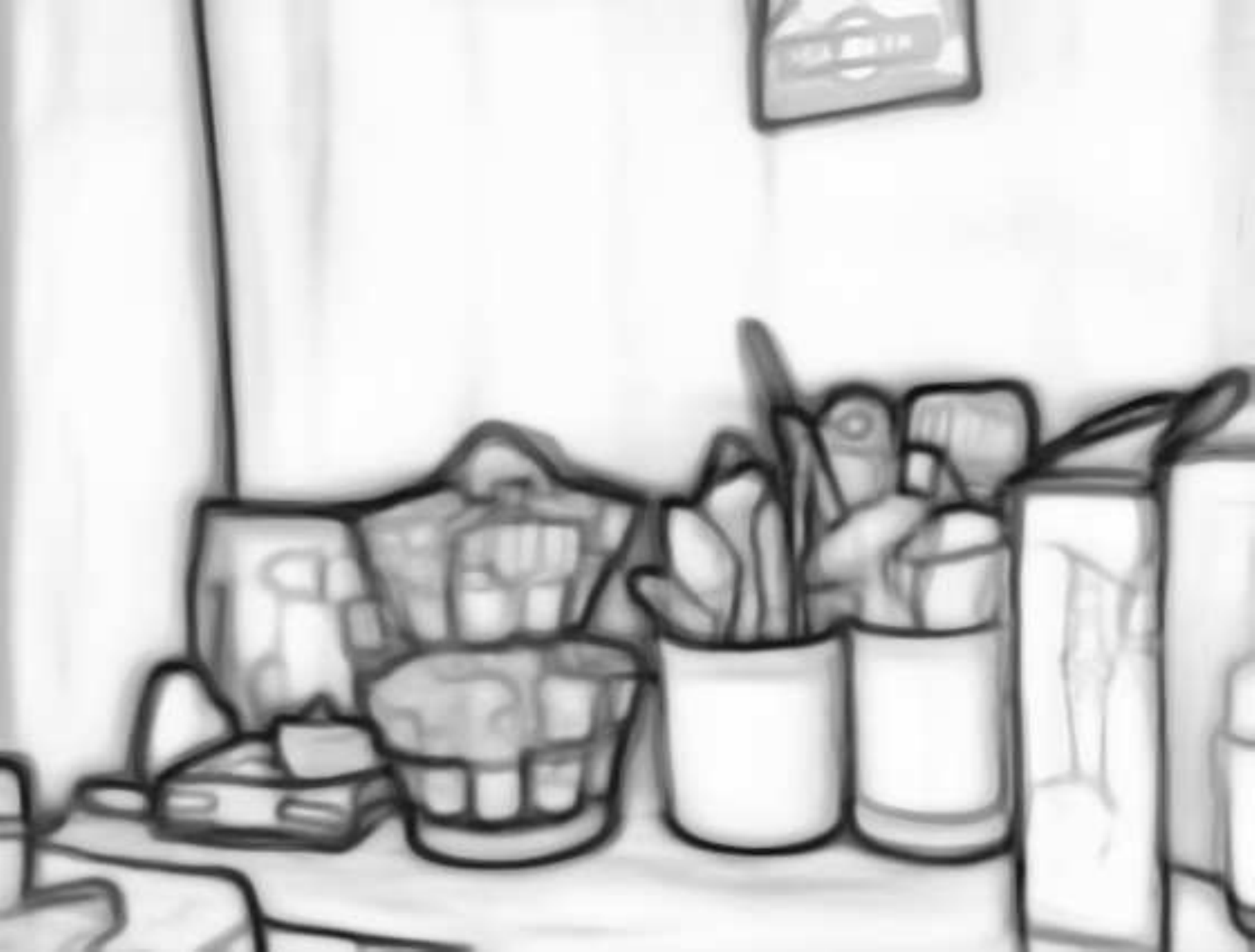} \\
		\includegraphics[width=0.16\linewidth]{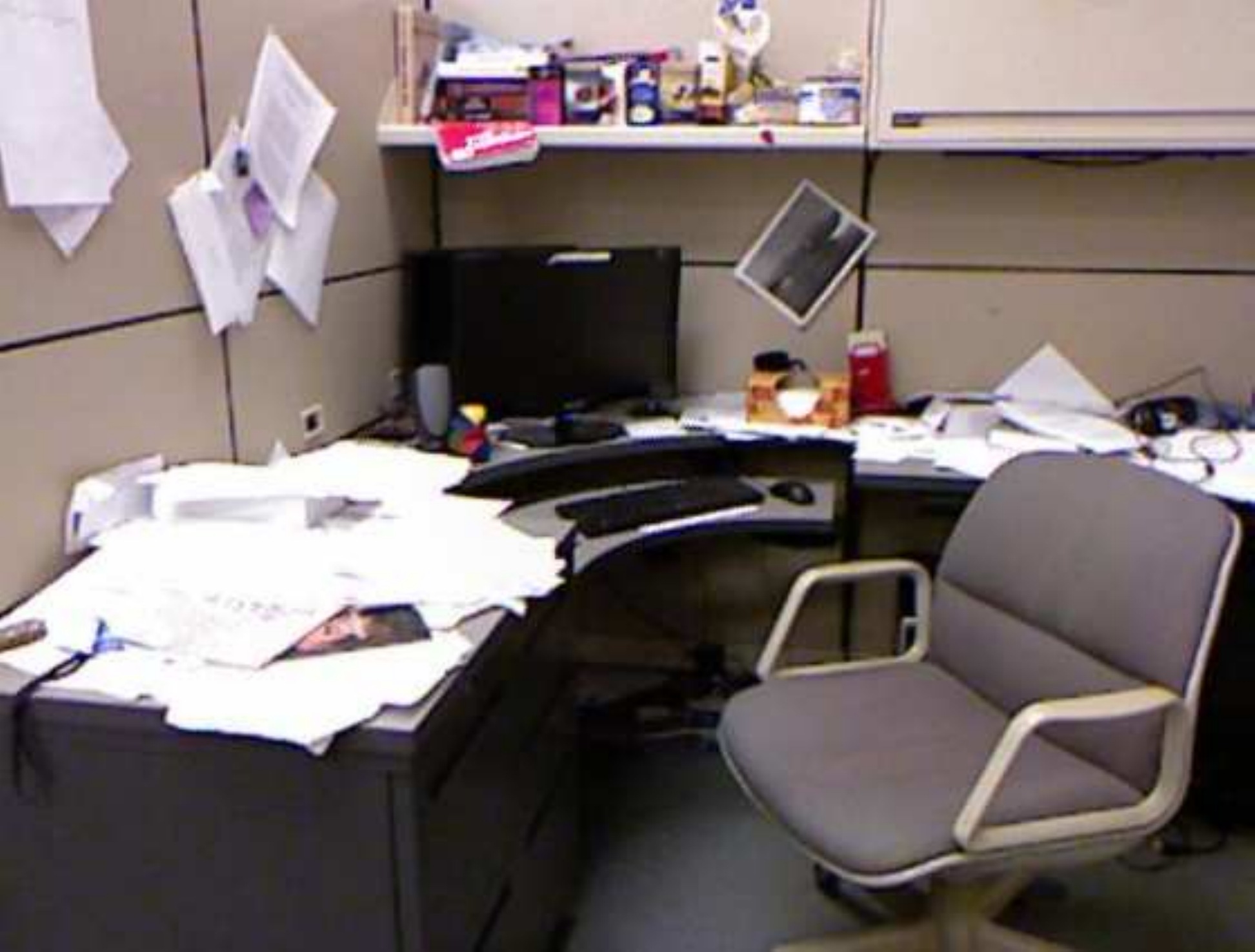} &
		\includegraphics[width=0.16\linewidth]{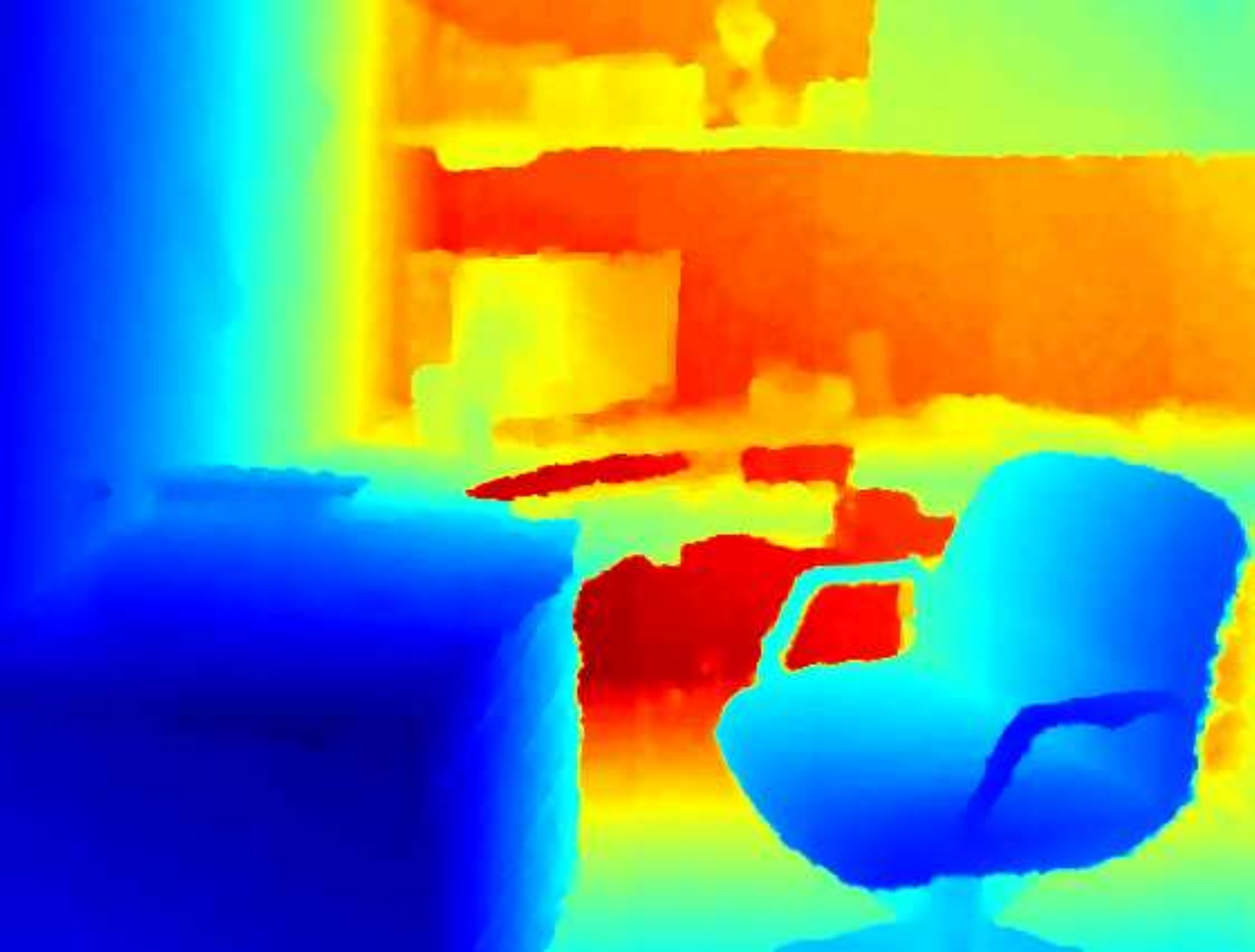} &
		\includegraphics[width=0.16\linewidth]{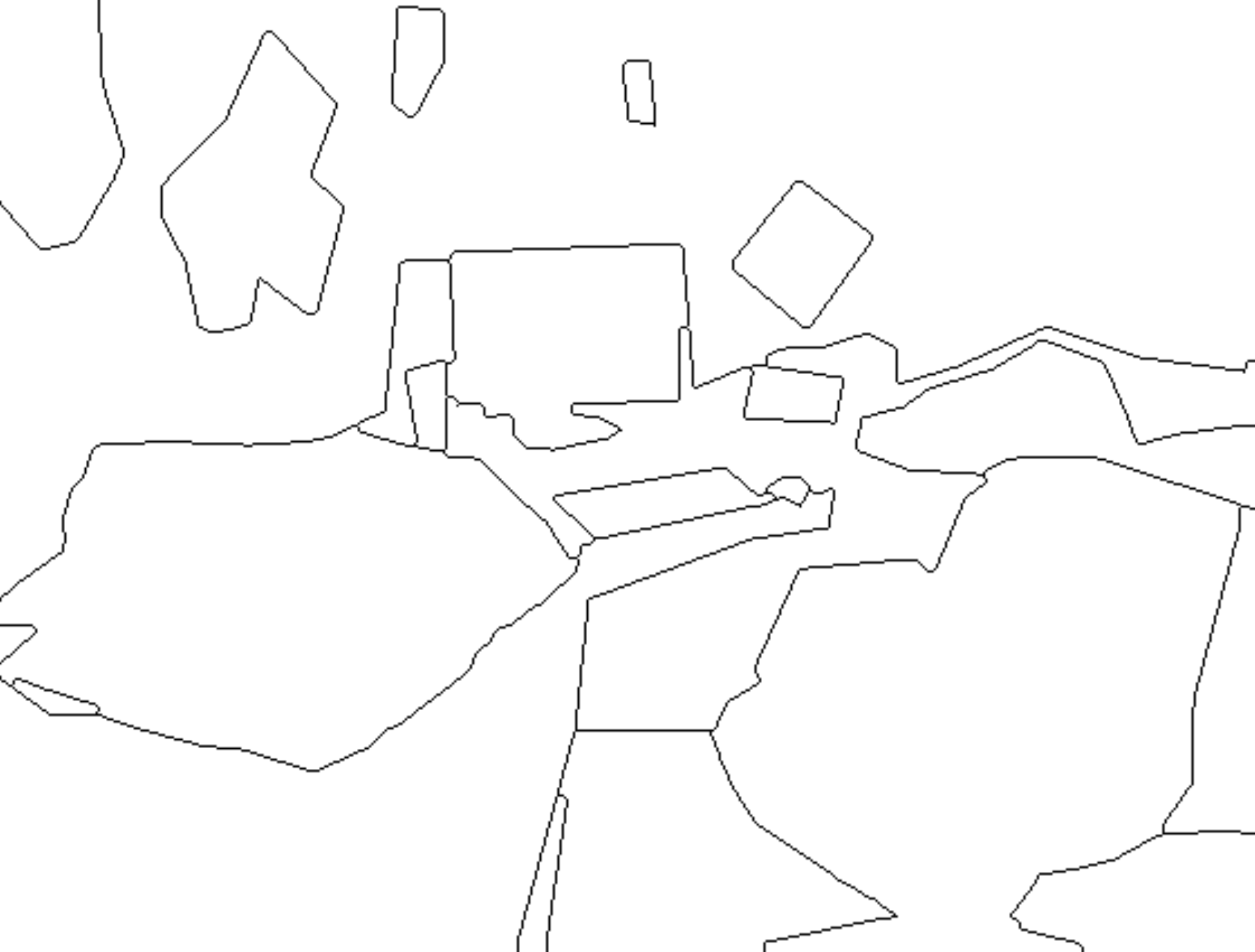} &
		\includegraphics[width=0.16\linewidth]{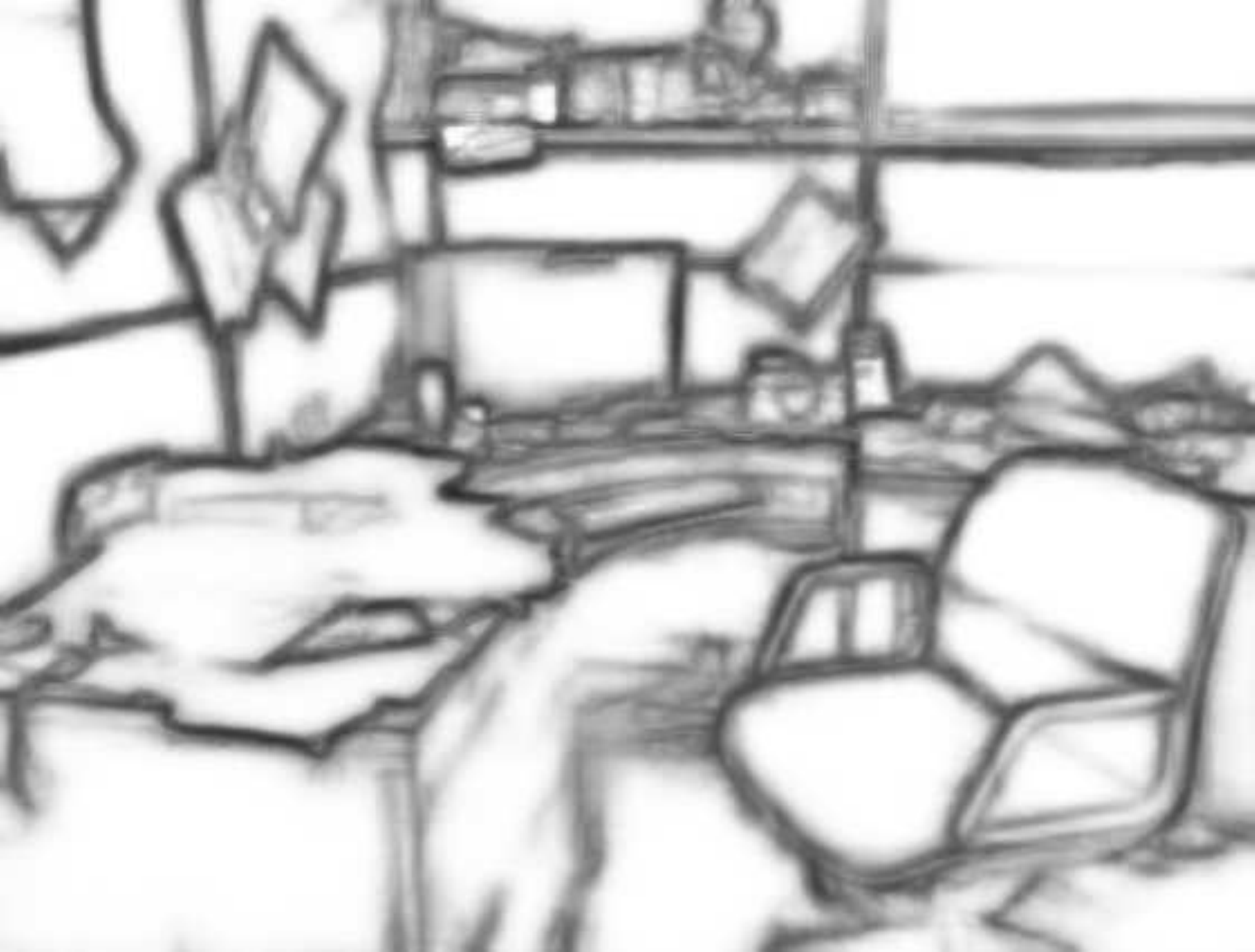} &
		\includegraphics[width=0.16\linewidth]{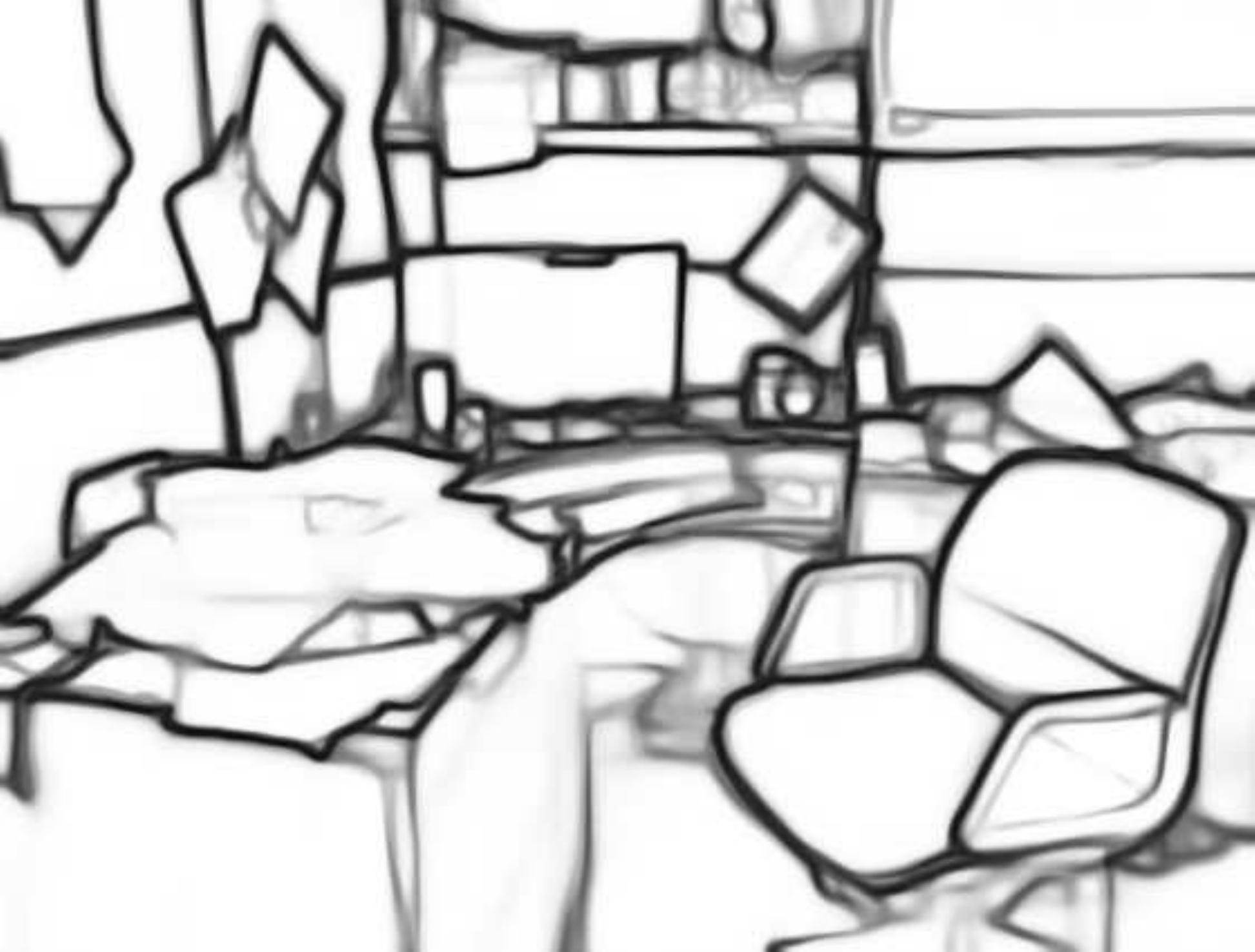} &
		\includegraphics[width=0.16\linewidth]{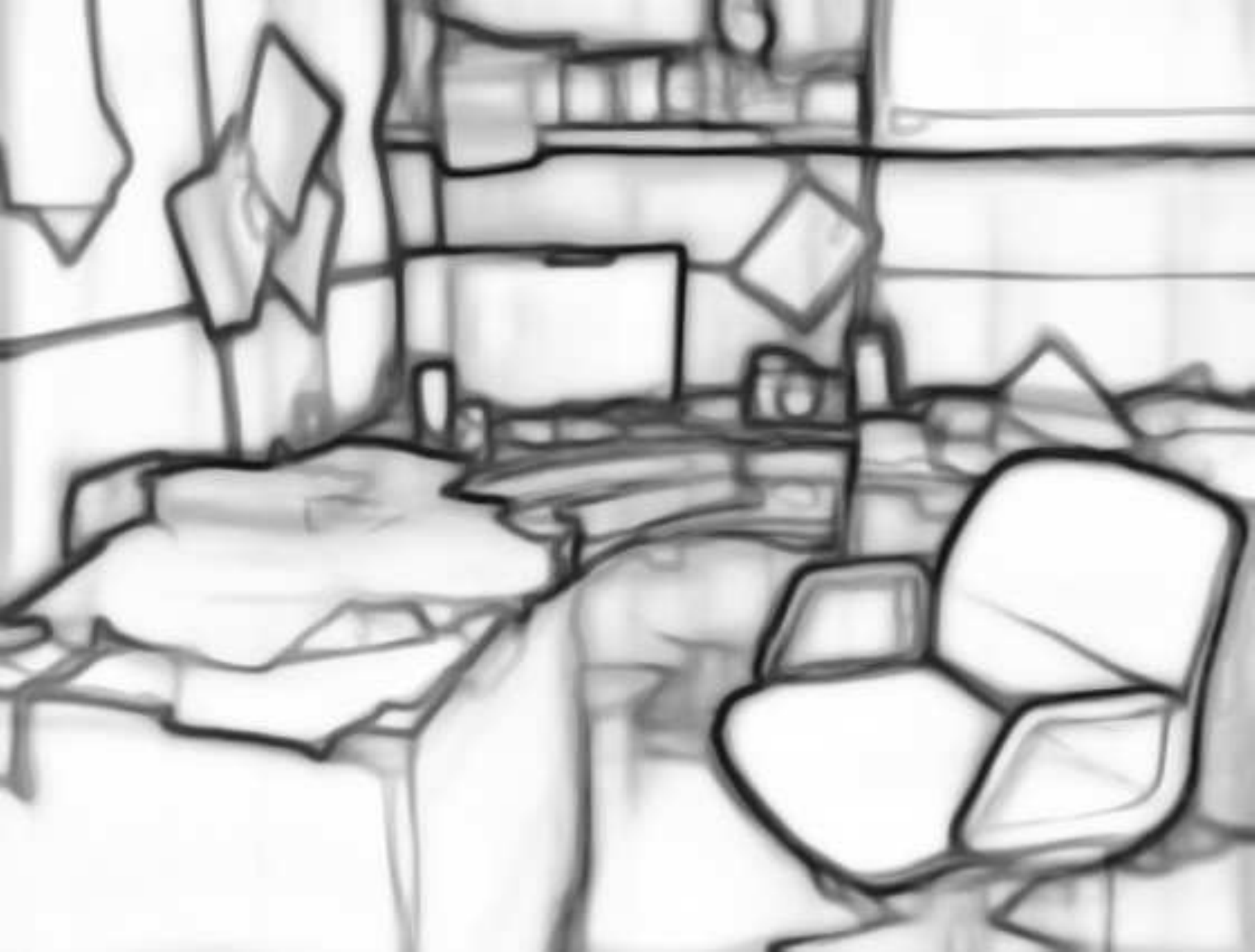} \\
		Raw image & Depth & Ground truth & HED-RGB& TD-CEDN-RGB & TD-CEDN-RGBD\\
		& & & & (ours) & (ours)\\
	\end{tabular}
	\caption{Example results on the NYUD testing dataset. }
	\label{Fig:nyud-ours-hed}
\end{figure*}

\noindent \textbf{NYU Depth:} The NYU Depth dataset (v2)~\cite{silberman2012indoor}, termed as NYUDv2, is composed of 1449 RGB-D images. The dataset is mainly used for  indoor scene segmentation, which is similar to PASCAL VOC 2012 but provides the depth map for each image. The dataset is split into 381 training, 414 validation and 654 testing images. We used the training/testing split proposed by Ren and Bo~\cite{xiaofeng2012discriminatively}. We fine-tuned the model ``TD-CEDN-over3 (ours)" with the NYUD training dataset. The RGB images and depth maps were utilized to train models, respectively. Then, the same fusion method defined in Eq.~(\ref{Eq:fusion}) was applied to average the RGB and depth predictions. Different from HED, we only used the raw depth maps instead of HHA features~\cite{gupta2014learning}.  Fig.~\ref{Fig:nyud} presents the evaluation results on the testing dataset, which indicates the depth information, which has a lower F-score of 0.665, can be applied to improve the performances slightly (0.017 for the F-score). We compared our method with the fine-tuned published model ``HED-RGB". Several example results are listed in Fig.~\ref{Fig:nyud-ours-hed}. Compared the ``HED-RGB" with the ``TD-CEDN-RGB (ours)", it shows a same indication that our method can predict the contours more precisely and clearly, though its published F-scores (the F-score of 0.720 for RGB and the F-score of 0.746 for RGBD) are higher than ours.

\section{Conclusion}
\label{Sec:Conclusion}

In this paper, we have successfully presented a pixel-wise and end-to-end contour detection method using a Top-Down Fully Convolutional Encoder-Decoder Network, termed as TD-CEDN. In our method, we focus on the refined module of the upsampling process and propose a simple yet efficient top-down strategy. The encoder-decoder network with such refined module automatically learns multi-scale and multi-level features to well solve the contour detection issues. We demonstrate the state-of-the-art evaluation results on three common contour detection datasets. Our predictions present the object contours more precisely and clearly on both statistical results and visual effects than the previous networks. Therefore, it's particularly useful for some higher-level tasks. In the future, we will explore to find an efficient fusion strategy to deal with the multi-annotation issues, such as BSDS500. Moreover, we will try to apply our method for some applications, such as generating proposals and instance segmentation.

\section*{Acknowledgment}

This work was partially supported by the National Natural Science Foundation of China (Project No. 41571436), the Hubei Province Science and Technology Support Program, China (Project No. 2015BAA027), the National Natural Science Foundation of China (Project No. 41271431), and the Jiangsu Province Science and Technology Support Program, China (Project No. BE2014866).

\bibliographystyle{IEEEtran}
\bibliography{egbib}

\end{document}